%% file: main.tex
\documentclass{ustthesis}

\usepackage{mathpazo,amsmath,amssymb,epsfig,enumerate,bbm,calc,color,ifthen,capt-of} 
\usepackage{algorithm}
\usepackage[center]{subfigure}
\usepackage{color,graphicx}

\usepackage{hyperref} 
\usepackage[margin=25mm,textheight=247mm,textwidth=145mm]{geometry}
\usepackage{times}
\usepackage{latexsym}
\usepackage{mathrsfs}
\usepackage{amsmath}
\usepackage{amssymb}
\usepackage{booktabs}
\usepackage{graphicx}
\usepackage{multirow}
\usepackage{bibentry} 
\usepackage{hyperref}
\usepackage{ltablex}
\usepackage{algpseudocode}
\usepackage{mathrsfs}
\usepackage{physics}
\usepackage{tikz}
\usepackage{mathdots}
\usepackage{yhmath}
\usepackage{cancel}
\usepackage{color}
\usepackage{siunitx}
\usepackage{array}
\usepackage{tabularx}
\usepackage{extarrows}
\usepackage[numbers]{natbib}
\usepackage{appendix}

\usetikzlibrary{fadings}
\usetikzlibrary{patterns}
\usetikzlibrary{shadows.blur}
\usetikzlibrary{shapes}

\usepackage{pifont}

\usepackage{natbib}

\usepackage[scaled]{helvet} 
\usepackage{courier} 
\usepackage{eulervm} 
\normalfont
\usepackage[T1]{fontenc}

\newcommand{\tab}[1]{\hspace{3mm}}

\usepackage[utf8x]{inputenc}
\usepackage[thaicjk,english]{babel}
\addto\extrasthaicjk{\fontencoding{C90}\selectfont}

\makeatletter
\@namedef{opt@inputenc.sty}{utf8}
\makeatother
\usepackage{CJKutf8}

\title{Effective Transfer Learning for Low-Resource \\ Natural Language Understanding}  

\author{Zihan Liu}     
\degree{\PhD}              
\subject{the Department of Electronic and Computer Engineering}      
\department{Electronic and Computer Engineering}       
\advisor{Prof. Pascale FUNG}     
\depthead{Prof. Andrew Wing On POON}    
\defencedate{2022}{08}{15}

\begin{document}


\maketitle

%



\input{chapter/sec-acknowledgement}

\tableofcontents


\listoffigures


\listoftables


\input{chapter/sec-abstract}




\input{chapter/sec-1-introduction}
\input{chapter/sec-2-background}
\input{chapter/sec-3-crosslingual}
\input{chapter/sec-4-crossdomain}
\input{chapter/sec-5-x2parsing}
\input{chapter/sec-6-conclusion}

\bibliographystyle{plainnat}
\bibliography{main}

\newpage
\input{chapter/sec-publications}

\input{chapter/sec-appendix}



\end{document}

%% file: chapter/sec-acknowledgement.tex
\acknowledgments

First, I would like to thank my supervisor, Prof. Pascale Fung, for supervising me throughout my entire Ph.D journey and giving me a life-changing experience. 
With her careful guidance, I started to learn and get interested in the natural language processing (NLP) field.
Pascale encouraged me to explore novel NLP solutions and inspired me to aim high and conduct impactful research.
She also motivated me to learn how to present the research work and make an interesting speech.
I am extremely lucky and very proud to be a member of the CAiRE research lab and mentored by Pascale. 

Next, I would like to express my appreciation to Prof. Bertram Shi, Prof. Qifeng Chen, Prof. Yangqiu Song, and Prof. Goran Glavaš to be on my thesis supervision committee. I want to thank Tania Leigh Wilmshurst, who proofread my papers and thesis numerous times and gave me insightful and useful advice on academic writing.
I am also fortunate to have a chance to gain research experience in the industry during my internship at NVIDIA. I would like to thank Dr. Bryan Catanzaro for sharing his vision and advice on the research. I am also very grateful for my mentor Dr. Mostofa Patwary for guiding the research direction and providing valuable advice on the research project. I also want to thank Ryan Prenger, Mohammad Shoeybi, Shrimai Prabhumoye, and Wei Ping for the insightful research discussion and brainstorming during this internship.

In the last four years at HKUST, I have an amazing learning and playing experience with my lab friends. I want to express my deep appreciation to Peng Xu, Andrea Madotto, and Genta Indra Winata for providing the guidance at the very beginning of my Ph.D journey. Thanks also to Chien-Sheng (Jason) Wu, who provided valuable advice on conducting research and choosing research topics. 
I would like to thank Genta Indra Winata, with whom I worked closely on multilingual, cross-lingual, cross-domain, and code-switching research, and also to Andrea Madotto and Zhaojiang Lin, with whom I worked closely on dialogue systems.
I want to thank Yan Xu, Wenliang Dai, Tiezheng Yu, Nayeon Lee, Jamin Shin, Etsuko Ishii, and Dan Su for the discussion and consultation on dialogue systems, multimodal systems, abstractive summarization, and question answering. Thanks go to Prof. Nancy Ip, Xiaopu Zhou and Tiffany TW Mak, who provided me a chance to work on Biomedical NLP area, and also to Samuel Cahyawijaya, who leads the Alzheimer's disease detection project.
Thanks also to Yejin Bang, Ziwei Ji, Holy Lovenia, Rita Frieske, Bryan Wilie, Willy Chung, Zeng Min, Leila Khalatbari, and many others for the great discussion and research experience. My graduate life would not be this colorful without them. I wish them all the best, and look forward to meeting them in the near future.

I would like to express my gratitude to my friends, Kai Ren, Jinyuan Xu, Sijun Deng, Yu Lin, and Chengkun Wu, who provided me moral support. I also want to thank Yulu Jin, who motivated me and gave me support during the challenging time. Finally, and most importantly, I would like to send my love to my parents, Xiumei Yan and Zhiyong Liu, and grandparents, Huotu Liu and Rongmei Lin, for their continuous and unconditional love, support, and encouragement. I dedicate this thesis to them.

\endacknowledgments

%% file: chapter/sec-abstract.tex
\begin{abstract}

Natural language understanding (NLU) is the task of semantic decoding of human languages by machines. NLU allows users to interact with machines using natural sentences, and is the fundamental component for any natural language processing (NLP) system.  
Despite the significant achievements on NLU tasks made by machine learning approaches, in particular deep learning, they still rely heavily on large amounts of training data to ensure good performance and fail to generalize well to languages and domains with little training data.  
Obtaining or collecting massive data samples is relatively easy for high-resource languages (e.g., English, Chinese) with significant amounts of textual data on the Internet. However, many other languages
have only a small online footprint (e.g., less than 0.1\% of data resources on the Internet are in Tamil or Urdu). This makes collecting datasets for these low-resource languages much more difficult. Similarly, datasets for low-resource domains (e.g., rare diseases), which have very few data resources and domain experts, are also much more challenging to collect than for high-resource domains (e.g., news). To enable machines to better comprehend natural sentences in low-resource languages and domains, it is necessary to overcome the data scarcity challenge, when very few or even zero training samples are available.


Cross-lingual and cross-domain transfer learning methods have been proposed to  learn task knowledge from large training samples of high-resource languages and domains and transfer it to low-resource languages and domains.
However, previous methods failed to effectively tackle the two main challenges in developing cross-lingual and cross-domain systems, namely, 1) that it is difficult to learn good \textbf{representations} from low-resource target languages (domains); and 2) that it is difficult to transfer the \textbf{task knowledge} from high-resource source languages (domains) to low-resource target languages (domains) due to the discrepancies between languages (domains). How to meet these challenges in a deep learning framework calls for new investigations.

In this thesis, we focus on addressing the aforementioned challenges in a deep learning framework. 
First, we propose to further refine the representations of task-related keywords across languages. We find that the representations for low-resource languages can be easily and greatly improved by focusing on just the keywords. 
Second, we present an Order-Reduced Transformer for the cross-lingual adaptation, and find that modeling partial word orders instead of the whole sequence can improve the robustness of the model against word order differences between languages and task knowledge transfer to low-resource languages.
Third, we propose to leverage different levels of domain-related corpora and additional masking of data in the pre-training for the cross-domain adaptation, and discover that more challenging pre-training can better address the domain discrepancy issue in the task knowledge transfer.
Finally, we introduce a coarse-to-fine framework, Coach, and a cross-lingual and cross-domain parsing framework, X2Parser. Coach decomposes the representation learning process into a coarse-grained and a fine-grained feature learning, and X2Parser simplifies the hierarchical task structures into flattened ones. We observe that simplifying task structures makes the representation learning more effective for low-resource languages and domains.

In all, we tackle the data scarcity issue in NLU by improving the low-resource representation learning and enhancing model robustness on topologically distant languages and domains in the task knowledge transfer. Experiments show that our models can effectively adapt to low-resource target languages and domains, and significantly outperform previous state-of-the-art models.

\end{abstract}

%% file: chapter/sec-1-introduction.tex
\chapter{Introduction}\label{sec-introduction}

\section{Motivation of Low-Resource Research}
Natural language understanding (NLU) is the task of semantic decoding of human languages by machines, and it allows users to easily interact with them using natural sentences. 
Equipping machines with a language understanding ability is the fundamental and indispensable procedure for constructing natural language processing (NLP) systems, such as reading comprehension, information extraction, and text mining~\cite{allen1988natural,devlin2019bert,dong2019unified,goo2018slot,guo2014joint,liu2019roberta,wang2018glue}.
However, NLU tasks (e.g., semantic parsing, sentiment analysis, and named entity recognition) can be challenging, since they require models to be aware of the whole context, understand the vagueness of natural languages, and even be robust to noisy inputs (e.g., grammar mistakes)~\cite{mikheev1999named,mullen2004sentiment,pang2004sentimental,pradhan2004shallow,tang2001using}.
To overcome these difficulties and the ambiguous nature of human languages, various deep learning approaches have been proposed. These have achieved significant progress and reached the state-of-the-art performance, sometimes even comparable to that of humans, in NLU tasks~\cite{devlin2019bert,goo2018slot,lample2016neural,liu2019roberta,medhat2014sentiment,wang2018glue}.

Despite the remarkable progress in NLU methods, they generally rely on large numbers of training samples to achieve close-to-human performance, and their language understanding quality will drastically decline when only a few samples are available for training~\cite{bapna2017towards,conneau2018xnli,gururangan2020don,hedderich2021survey,jia2019cross,kann2019towards,kim2017cross,kozhevnikov2013cross,lauscher2020zero,li2021mtop,li2013active,litschko2021evaluating,liu2020attention,liu2020coach,liu2020cross,liu2021crossner,liu2021importance,liu2021preserving,madotto2020language,pfeiffer2020mad,shah2019robust,yu2021adaptsum}. 
Obtaining or collecting massive data samples is relatively easy for high-resource languages, such as English or Chinese, which are spoken by a large majority of people and make up most resources on the Internet. 
However, around 7,000 languages are spoken around the world, and only a few hundred languages are represented on the websites. Of these, a substantial number are low-resource languages. Most of them have much less than 0.1\% of data resources on the Internet, even those spoken by many people, such as Urdu or Tamil~\footnote{\url{https://w3techs.com/technologies/overview/content_language}}.
This makes curating datasets for these low-resource languages, especially in a massive scale, let alone a great many languages and dialects that lack a keyboard support~\cite{soria2018dldp} or do not have a widespread written tradition~\cite{paterson2014keyboard}, much more difficult than for high-resource languages~\cite{gaikwad2021cross,gu2018universal,magueresse2020low,ranathunga2021neural,tapo2020neural}.
Similarly, datasets for low-resource domains are also much more challenging to collect than high-resource domains. Low-resource domains, such as rare diseases, have very limited data resources and few domain experts. 
In addition, a large number of low-resource domains (e.g., geophysics, cancers) are missing from existing labeled NLU datasets in high-resource languages, not to mention in middle- and low-resources languages.
The knowledge in these domains is known by a relatively small number of people, and gathering it to annotate datasets is particularly difficult and laborious.
Therefore, in order to enable machines to better comprehend low-resource languages, as well as natural sentences in low-resource domains, it is essential to develop models to tackle data scarcity, where very few or even zero training samples are available.

When humans start to learn a new task, their previously learned relevant knowledge will certainly be beneficial and expedite the learning process.
For example, a person who has the skill of sketching can quickly learn landscape painting, since both are related to drawing.
Inspired by this observe, we expect it to be much easier (i.e., many fewer training samples are needed) for a model to learn a task for low-resource languages when it has already learned how to perform this task for a high-resource language.
The motivations of this approach are two-fold: 1) The knowledge of how to perform the NLU task can be well learned from numerous data samples in high-resource languages. 2) Existing techniques have the capability to capture the inherent similarities across languages (e.g., grammar, syntax), which makes the transfer of task knowledge from high-resource languages to low-resource languages possible.
Similarly, learning a task for low-resource domains will also substantially benefit from prior task knowledge of high-resource domains. 
This is because knowledge of how to comprehend natural languages and extract information from high-resource domains can be effectively incorporated with the small-scale low-resource domain information provided by domain experts or unlabeled domain corpora. This enables efficient task knowledge transfer.

Therefore, in order to address low-resource issues, cross-lingual/domain transfer learning methods have been proposed. These methods first initialize models with the task knowledge learned from high-resource languages/domains. Then, when the model starts to learn the same task in low-resource languages/domains, the knowledge used for initialization will be effectively utilized and significantly facilitate the learning process. In this thesis, we focus on advancing cross-lingual and cross-domain transfer learning approaches to construct more scalable, generalizable, and applicable NLU models for low-resource languages and domains.


\section{Cross-Lingual/Domain Transfer Learning}

\begin{figure}[!t]
  \centering
  \includegraphics[width=.99\linewidth]{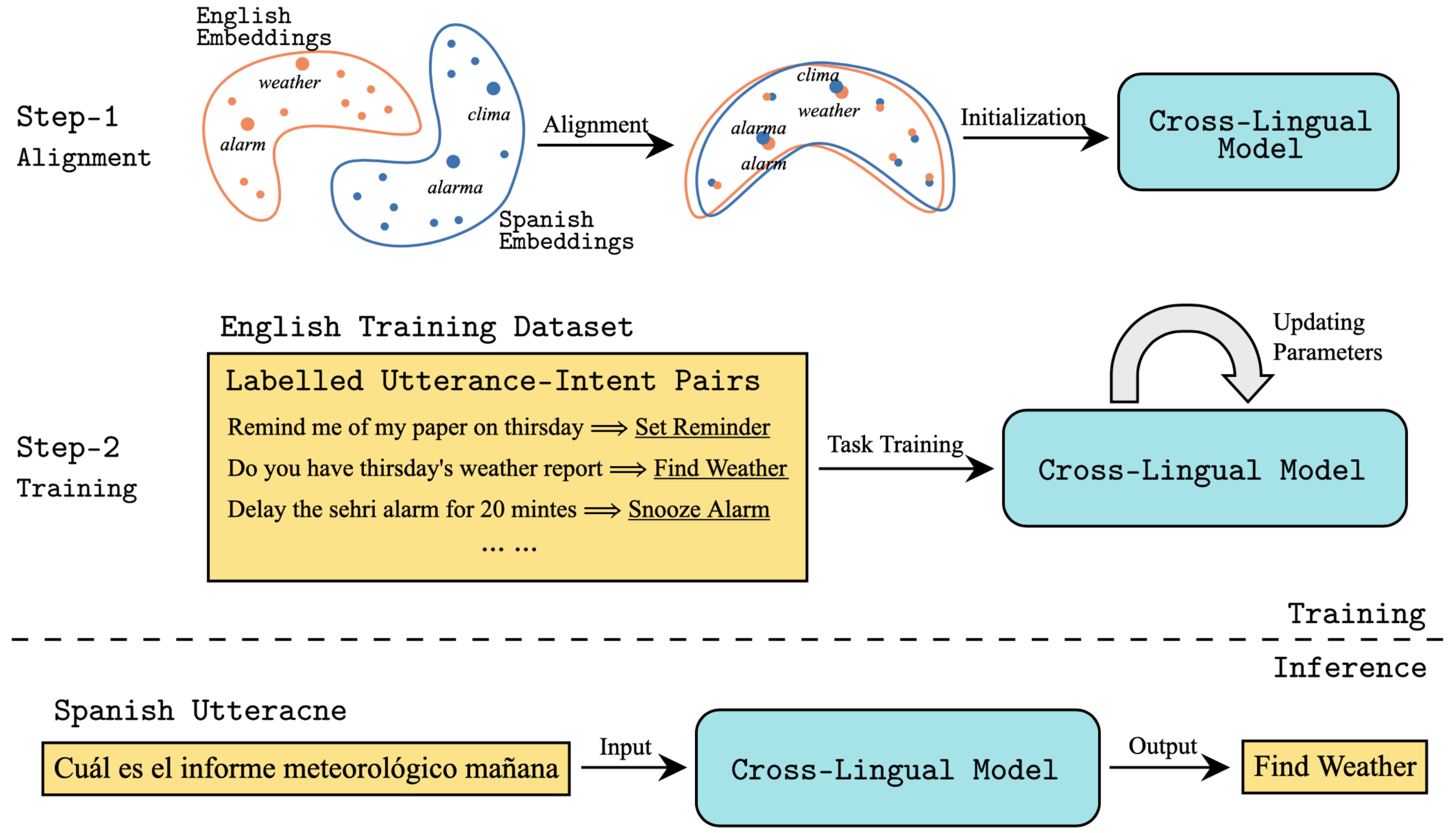}
  \caption{An illustration of cross-lingual intent detection task.}
  \label{fig:cross-lingual-task}
\end{figure}

\paragraph{Cross-Lingual Transfer}
Figure~\ref{fig:cross-lingual-task} provides an illustration of the cross-lingual transfer task, where we use the intent detection (i.e., the model detects the intent type of the input utterance) as an example. 
The model construction consists of two steps.
Step 1, the first and foremost step of building cross-lingual models, is to align the representations across languages~\cite{artetxe2018robust,chen2018unsupervised,conneau2020unsupervised,devlin2019bert,huang2019unicoder,lample2018word,lample2019cross,schuster2019crosslingual}. This step leverages unlabeled multilingual corpora for learning representations for each language separately, and then it learns to align the representations of semantically similar words between low-resource and high-resource languages into a similar vector space. Alternatively, the cross-lingual alignment can be achieved by simultaneously training on the multilingual corpora in a self-supervised fashion.
The intuition behind Step 1 is to equip models with multilingual capability, which is to understand natural sentences from various languages, including low-resource languages. The levels of the multilingual skills will determine how well the models can transfer the task-relevant knowledge learned from high-resource languages to low-resource languages. 
In Step 2, large numbers of task-related training samples from high-resource source languages (e.g., English) are used to train the models to perform the task (e.g., intent detection). 
Thanks to the multilingual comprehension learned in Step 1, the task knowledge can be effectively transferred between languages. This allows the models to perform the same task in low-resource target languages (we use Spanish as an example) in the inference stage.
Moreover, a few training samples in low-resource languages can certainly be incorporated in the training stage of Step 2, so as to make models better perform the task in target languages.

\begin{figure}[!t]
  \centering
  \includegraphics[width=.99\linewidth]{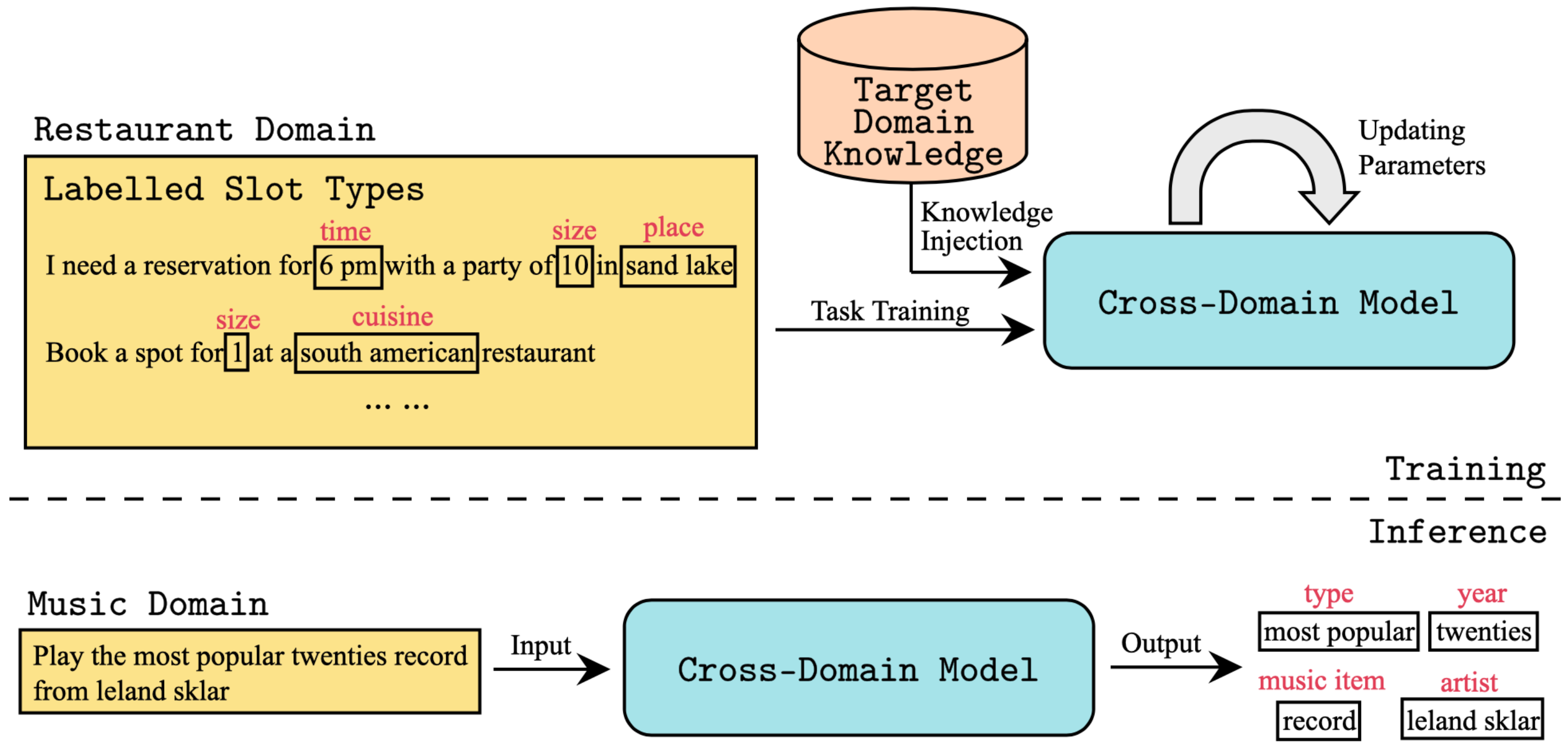}
  \caption{An illustration of cross-domain slot filling task.}
  \label{fig:cross-domain-task}
\end{figure}

\paragraph{Cross-Domain Transfer}
Figure~\ref{fig:cross-domain-task} provides an illustration of the cross-domain transfer task, where we use the slot filling (i.e., the model needs to fill in the slot type for each entity) as an example. 
Similar to cross-lingual transfer learning, one of the most important steps for building cross-domain models is to learn how to perform the task using extensive training samples from high-resource source domains (we use the restaurant domain as an example). 
After this training stage, it could still be difficult for the models to transfer the task knowledge from source domains to target domains (we use the music domain as an example) due to the potentially large domain discrepancies and the lack of language understanding ability in target domains. To mitigate this issue and facilitate the task knowledge transfer, relevant target domain knowledge, which usually contains unlabeled domain-relevant corpora, will be injected into the models~\cite{gururangan2020don,jia2020multi,jia2019cross,shah2019robust,zhou2020sentix}. 
The motivation for the knowledge injection is to allow the models to better comprehend the natural languages in the target domains. Hence, the models can better perform the same NLU task in low-resource target domains by effectively leveraging the task knowledge learned from high-resource source domains.
Similar to cross-lingual transfer learning, a few data samples in low-resource domains can be also integrated with those of high-resource domains in the training, in order to further boost the quality of cross-domain transfer.

\section{Research Problems}

The following are the major challenges to developing cross-lingual and cross-domain adaptation models in low-resource scenarios:

\begin{itemize}
    \item [1.] \textbf{Improving low-resource representation learning:}
    Improving low-resource representation learning allows us to build more robust and scalable domain and language-transferable models. 
    Enormous unlabeled data corpora have been used to learn multilingual representations in a self-supervised fashion, and these representations generally serve as a good initialization for cross-lingual models~\cite{artetxe2017learning,conneau2020unsupervised,devlin2019bert,huang2019unicoder,lample2018word}.
    However, the representations of semantically similar words might not be close enough across languages due to the inherent vagueness in language~\cite{artetxe2018robust,joulin2018loss,lample2018word,schuster2019crosslingual}, and the data scarcity issue further aggravates this issue and hinders the low-resource language adaptation~\cite{conneau2020unsupervised,pires2019multilingual,schuster2019cross,wu2019beto}.
    Another line of work is to continue training the multilingual representations so as to improve the quality of cross-lingual alignments~\cite{cao2019multilingual,faruqui2014improving,schuster2019crosslingual}. 
    However, these methods will inevitably incorporate data biases towards certain domains and could negatively affect the original multilingual capability for partial words.
    
    For learning representations in low-resource domains, one intuitive approach is to leverage a relatively large unlabeled domain corpus for language modeling or domain-adaptive pre-training~\cite{beltagy2019scibert,gururangan2020don,jia2020multi,lee2020biobert}.
    However, such methods are not guaranteed to be effective when the size of the unlabeled target domain corpora is small and very few labeled training samples are available. Moreover, there still remains a challenge in terms of how to select suitable and relevant target domain text and how to inject the knowledge into the model.
    Another approach is to incorporate the commonly shared task knowledge from various source domains and private domain-specific representations for target domains~\cite{bapna2017towards,jia2020multi,shah2019robust,wu2019transferable}. However, the learned representations lack generalizability to all target domains, especially when they are not close to the source domains and have specialized labels.
    
    \item [2.] \textbf{Task knowledge transfer:}
    The inherent discrepancies between languages (e.g., lexicons, language structures) can greatly decrease the quality of task knowledge transfer to low-resource target languages, and it is very challenging to adapt cross-lingual NLU models to target languages that are topologically distant from the source language~\cite{conneau2018xnli,schuster2019cross}. 
    One way to address this challenge is to leverage large-scale data corpora across numerous languages and improve the quality of multilingual representation alignment~\cite{conneau2020unsupervised,devlin2019bert,joulin2018loss,lample2018word}. However, such methods, although generally improving the quality of cross-lingual adaptation on target languages, do not focus on addressing the performance drop caused by the differences in language structures.
    
    Likewise, large domain discrepancies will also hinder the task knowledge transfer from source domains to target domains. A number of previous studies have focused on transferring NLU models to distant target domains based on domain descriptions or potential domain relations~\cite{hedderich2021survey,jia2019cross,shah2019robust}.
    However, these methods cannot generalize well to all kinds of low-resource target domains, and domain relations cannot always be easily extracted.
    Generalizing the learned task knowledge to low-resource target domains that are distant from the source domains has been a long-standing and challenging research problem~\cite{gururangan2020don,jia2019cross,wu2019transferable}.
    
\end{itemize}

In this thesis, we address the major challenges mentioned above.
First, we propose a cross-lingual embedding refinement, a transferable latent variable model, and a label regularization method for the cross-lingual adaptation. These methods further refine the representations of task-related keywords across languages. We find that the representations for low-resource languages can be easily and greatly improved by focusing on just the keywords, and experiments show that our methods significantly outperform previous state-of-the-art models.
Second, we present an Order-Reduced Transformer for the cross-lingual adaptation, and find that modeling partial word orders instead of the whole sequence can improve the task knowledge transfer to low-resource languages and the robustness of the model against the word order differences between languages. We also fit models with various amounts of word order information, and observe that encoding excessive or insufficient word orders into the model results in inferior quality of the task knowledge transfer.
Third, we propose to leverage different levels of domain-related corpora and additional masking of data in the pre-training for the cross-domain adaptation. We discover that focusing on a factional corpus containing domain-specialized entities and utilizing a more challenging manner of pre-training can better address the domain discrepancy issue in the task knowledge transfer.
Finally, we introduce a coarse-to-fine framework, Coach, and a cross-lingual and cross-domain parsing framework, X2Parser. Coach decomposes the representation learning process into a coarse-grained and a fine-grained feature learning, and X2Parser simplifies the hierarchical task structures into flattened ones. We observe that simplifying task structures makes the representation learning more effective for low-resource languages and domains, and learning flattened representations (X2Parser) is also more efficient for long and complex queries.

In sum, we focus on building transferable language and domain adaptation models by \textit{improving low-resource representation learning} and \textit{tackling the challenge of task knowledge transfer}.
We first demonstrate how to enhance the robustness of representation learning for low-resource languages. Then, we show that simplifying task structures makes the representation learning more effective for low-resource languages and domains.
Additionally, we find that modeling partial word orders instead of the whole sequence can improve the task knowledge transfer to low-resource languages. Finally, we discover that leveraging a more challenging pre-training can better address the domain discrepancy issue in the task knowledge transfer.

\section{Thesis Outline}
The contents of this thesis are organized around learning to transfer in low-resource scenarios, and our experiments focus on adapting models to low-resource languages and domains. The rest of the thesis is divided into five chapters and organized as follows:

\begin{itemize}
    \item Chapter 2 introduces the background and related work on representation learning for low-resource scenarios, NLU models, and sequence modeling networks.
    
    \item Chapter 3 presents our proposed cross-lingual methods, which focus on learning robust representations for low-resource target languages, and introduces an Order-Reduced Transformer to tackle the issues raised by inherent language discrepancies during the task knowledge transfer.
    
    \item Chapter 4 presents our proposed cross-domain model, Coach, which effectively learns coarse-grained and fine-grained representations separately, and introduces a collected dataset for evaluating the cross-domain NER models, as well as a continual pre-training method for transferring task knowledge to topologically distant low-resource domains.
    
    \item Chapter 5 introduces a cross-lingual and cross-domain parsing framework, X2Parser, which decomposes the hierarchical task representations and enables a more efficient low-resource representation learning.
    
    \item Chapter 6 summarizes this thesis and the significance of the low-resource transfer approaches, and discusses possible future research directions.
    
\end{itemize}

\newpage

%% file: chapter/sec-2-background.tex
\chapter{Background and Preliminaries}\label{sec-background}

\section{Overview}
In this chapter, we first provide a literature review and background knowledge on the sequence models, which are the fundamental network architectures in our work.
Then, we present monolingual and multilingual representation learning methods in natural language processing (NLP). Next, we introduce natural language understanding (NLU) tasks and the corresponding models. 
Lastly, we provide contemporary studies on low-resource language and domain adaptation.


\section{Sequence Models}
Sequence modeling is one of the fundamental techniques for allowing machines to comprehend human languages due to the sequential nature of natural languages~\cite{bai2018empirical,chung2014empirical,hochreiter1997long,lin2020caire,lin2021xpersona,schuster1997bidirectional,sherstinsky2020fundamentals,su2019generalizing,winata2020lightweight,winata2020meta}. Common sequence models are recurrent neural networks (RNNs) (e.g., long short-term memory~\cite{hochreiter1997long} (LSTM) and gated recurrent units~\cite{chung2014empirical} (GRU)), and Transformer models~\cite{vaswani2017attention}.
These network architectures have a strong capability to model sequential information and capture context dependencies~\cite{devlin2019bert,kim2014convolutional,lample2016neural,lee2019team,lin2019learning,lin2020variational,liu2019incorporating,madotto2020learning,madotto2020language,wang2018glue,winata2019hierarchical}.

\subsection{Recurrent Neural Networks} 
\begin{figure}[!t]
  \centering
  \includegraphics[width=.92\linewidth]{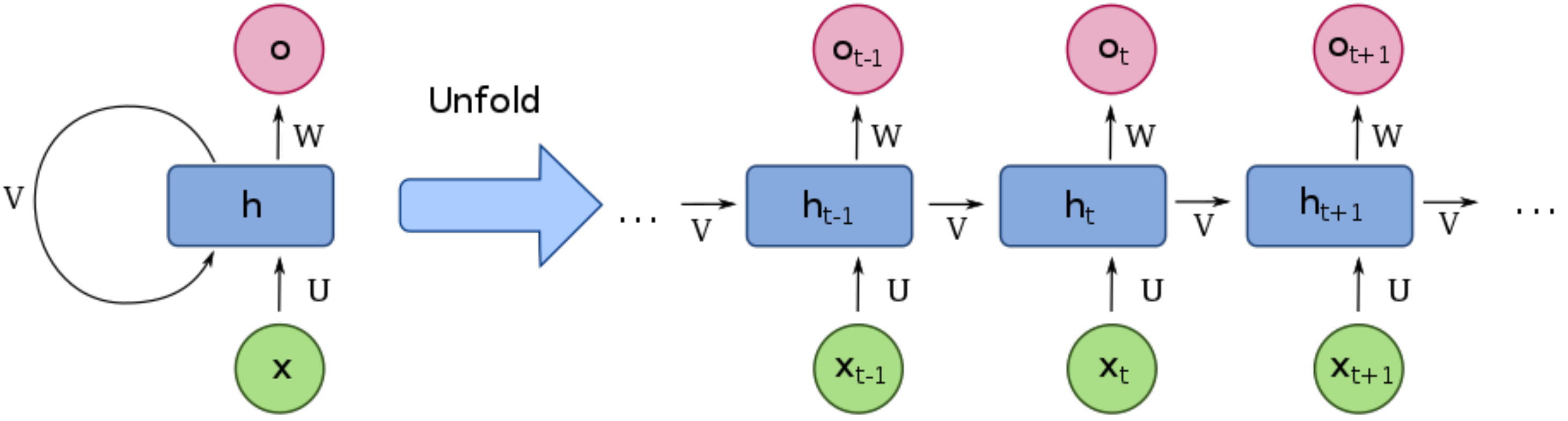}
  \caption{An illustration of recurrent neural networks. This figure is taken from Wikipedia Commons.}
  \label{fig:rnn}
\end{figure}

As shown in Figure~\ref{fig:rnn}, the RNN~\cite{schuster1997bidirectional} is a class of artificial neural networks where connections between nodes form a directed graph along a temporal sequence. Hence, an RNN is able to exhibit dynamic temporal behavior. 
Different from feed-forward neural networks, an RNN uses the internal state to process input sequences, and feeds the hidden layer output at time $t-1$ back to the same hidden layer at time $t$ via recurrent connections. The information stored in the hidden layer can be viewed as a summary of input sequences up to the current time.

The hidden state at time t can be expressed as:
\begin{equation}
    h_t = \sigma(U x_t + V h_{t-1} + b),
\end{equation}
where $\sigma$ is a non-linear and differentiable activation function, and $U$, $V$, and $b$ are trainable parameters (weight matrices). In the training process, network parameters are optimized based on loss functions and updated using the back-propagation through time (BPTT) method considering the influence of past states through recurrent connections. The trainable parameters are updated after each training sample or mini-batch.

Despite the nature of modeling temporal sequences, RNN architectures still have several issues: gradient vanishing and exploding problems, as well as a lack of long-term dependencies of words~\cite{pascanu2013difficulty}. Therefore, more advanced recurrent neural networks are proposed to solve these issues~\cite{chung2014empirical,gers2000recurrent,graves2005framewise,hochreiter1997long,koutnik2014clockwork,li2018independently,yao2015depth}, and the most commonly used among them are LSTM~\cite{hochreiter1997long} and GRU~\cite{chung2014empirical}.

\begin{figure}[!t]
  \centering
  \includegraphics[width=.52\linewidth]{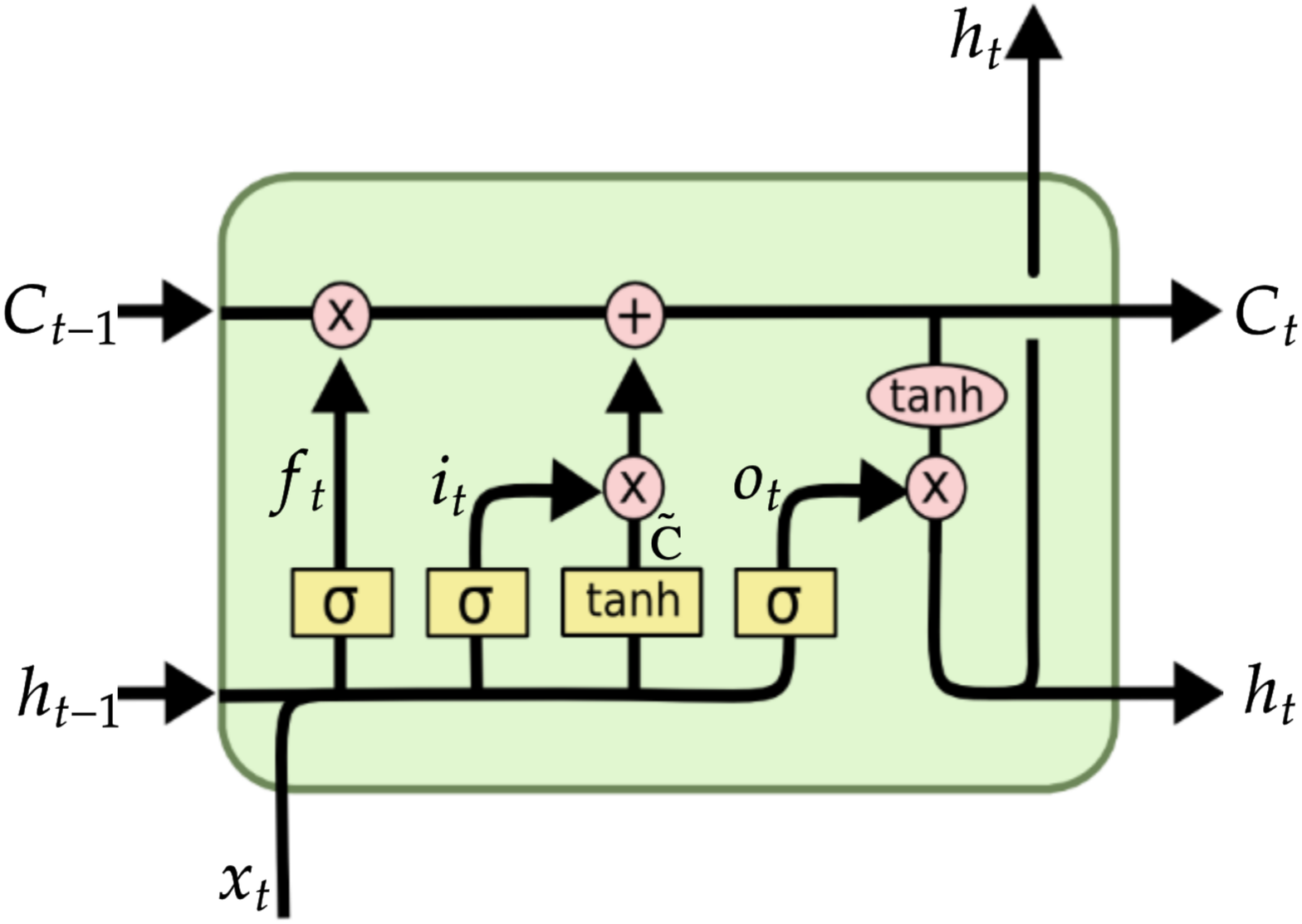}
  \caption{The architecture of a long short-term memory (LSTM) cell. This figure is taken from \url{https://colah.github.io}.}
  \label{fig:lstm_architecture}
\end{figure}

\subsubsection{Long Short-Term Memory}
LSTM is a special kind of RNN, which is explicitly designed to avoid the long-term dependency problem. As shown in Figure~\ref{fig:lstm_architecture}, LSTM consists of an input gate ($i_t$), forget gate ($f_t$), and output gate ($o_t$). 

The forget gate can remove the information from the previous states that is no longer useful for the new inputs:
\begin{equation}
    f_t = \sigma (W_f \cdot [h_{t-1}, x_t] + b_f),
\end{equation}
where $\sigma$ denotes the sigmoid function, $W_f$ and $b_f$ are trainable parameters, and $h_{t-1}$ contains the information from the previous states. Hence, $f_t$ tends to keep the relevant and useful information for the current inputs $x_t$.
The input gate decides which values the model will update. Then, $\tilde{C}$ is added as a scale to measure how much the model wants to update each state:
\begin{equation}
    i_t = \sigma (W_i \cdot [h_{t-1}, x_t] + b_i),
\end{equation}
\begin{equation}
    \tilde{C}_t = tanh (W_C \cdot [h_{t-1}, x_t] + b_C),
\end{equation}
where $W_i$, $b_i$, $W_C$, and $b_C$ are trainable parameters.
Next, the new state $C_t$ will be generated based on the previous state ($C_{t-1}$), the forget gate, and the input gate:
\begin{equation}
    C_t = f_t * C_{t-1} + i_t * \tilde{C}.
\end{equation}
Finally, the output of the LSTM ($h_t$) will be based on the output gate and the new state:
\begin{equation}
    o_t = \sigma (W_o \cdot [h_{t-1}, x_t] + b_o),
\end{equation}
\begin{equation}
    h_t = o_t * tanh (C_t),
\end{equation}
where $W_o$ and $b_o$ are trainable parameters.

\begin{figure}[!t]
  \centering
  \includegraphics[width=.5\linewidth]{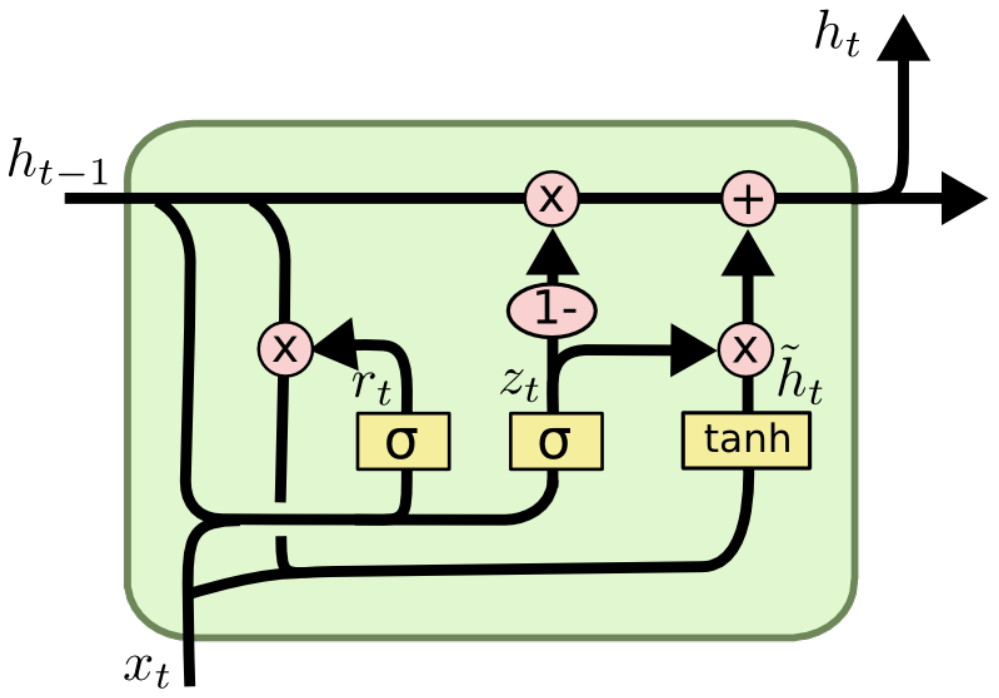}
  \caption{The architecture of a Gated Recurrent Units (GRU) cell. This figure is taken from \url{https://colah.github.io}.}
  \label{fig:gru_architecture}
\end{figure}

\subsubsection{Gated Recurrent Units}
GRU~\cite{chung2014empirical} is a variant of LSTM, which combines the forget and input gates into a single update gate, and also merges the cell state and hidden state. As shown in Figure~\ref{fig:gru_architecture}, GRU consists of a reset gate ($r_t$) and an update gate ($z_t$), which is simpler than the standard LSTM model.
The forwarding process can be formulated as:
\begin{equation}
    z_t = \sigma (W_z \cdot [h_{t-1}, x_t] + b_z),
\end{equation}
\begin{equation}
    r_t = \sigma (W_r \cdot [h_{t-1}, x_t] + b_r),
\end{equation}
\begin{equation}
    \tilde{h}_t = tanh (W_h \cdot [r_t * h_{t-1}, x_t] + b_h),
\end{equation}
\begin{equation}
    h_t = (1-z_t)*h_{t-1} + z_t * \tilde{h}_t,
\end{equation}
where $W_z$, $b_z$, $W_r$, $b_r$, $W_h$, and $b_h$ are trainable parameters, 
$h_{t-1}$ and $h_{t}$ are the previous and present hidden state, and $x_t$ is the current input. The reset gate controls how much of the previous state that the model needs to remember, and the update gate balances the ratio between the previous state and the new state.

\begin{figure}[!t]
  \centering
  \includegraphics[width=.6\linewidth]{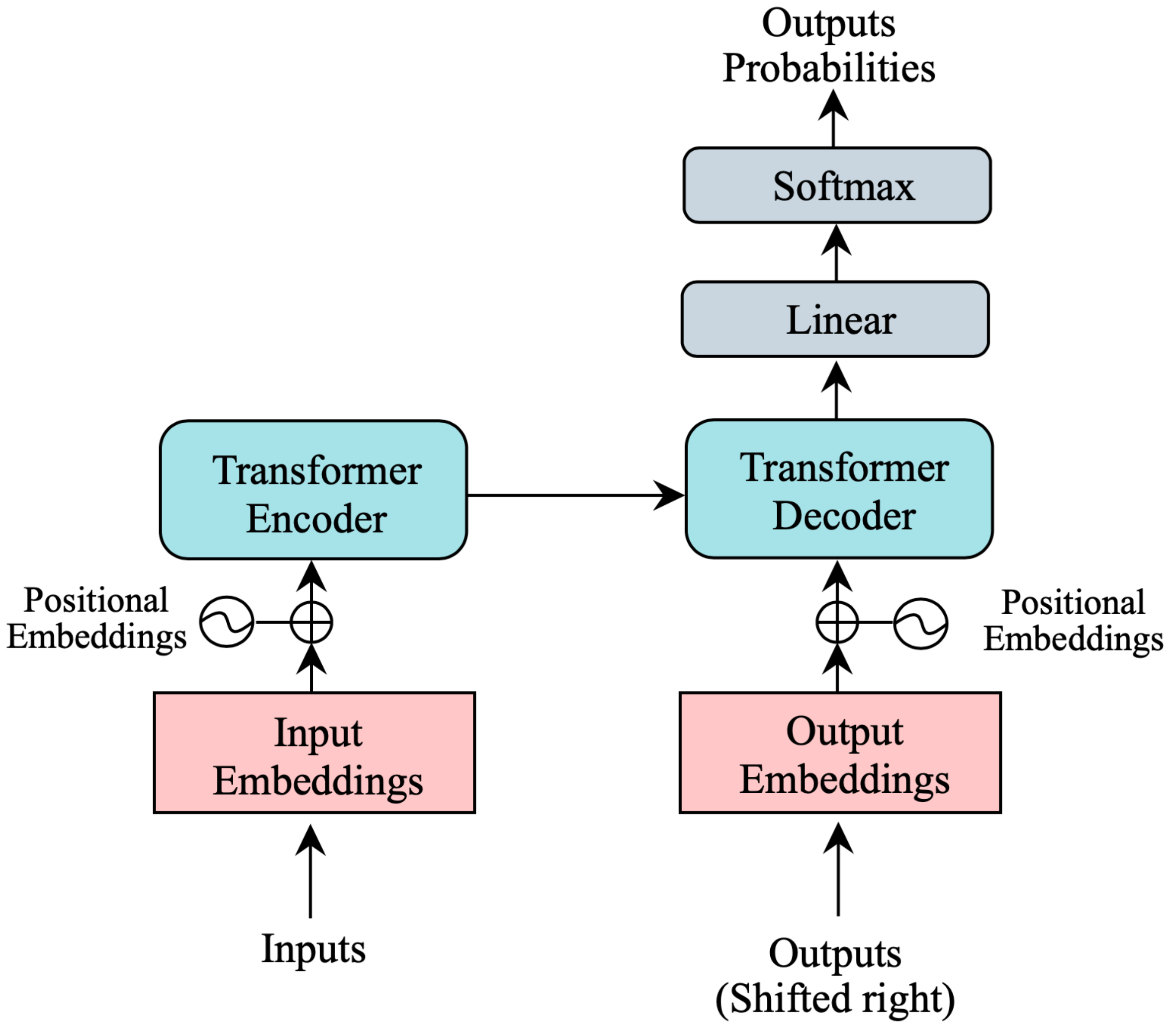}
  \caption{The architecture of Transformer. This figure is adapted from~\citet{vaswani2017attention}.}
  \label{fig:transformer_architecture}
\end{figure}

\subsection{Transformers}
Recently, Transformer~\cite{vaswani2017attention} was proposed to replace the RNN-based architectures (e.g., LSTM, GRU)~\cite{devlin2019bert,rothman2021transformers,wolf2020transformers}. As shown in Figure~\ref{fig:transformer_architecture}, Transformer consists of an encoder, a decoder, input and output embeddings, and positional embeddings.

\subsubsection{Embedding Layers}
Given the input sequence $X=\{x_1, x_2, ..., x_n\} \in R^{n \times 1}$, we define the embedding matrix $E$, which maps each token ($x_i, i \in [1,n]$) to a vector of dimension $d \times 1$.

Differently from RNN-based models, Transformer does not use any recurrent states, and thus by construction it is not able to model temporal dependency in the input sequence. To cope with this issue, Transformer~\cite{vaswani2017attention} uses a sinusoidal positional embedding ($PE$). $PE$ is made of sine and cosine functions of different frequencies:
\begin{equation}
    \begin{split}
    PE_{(pos,2i)} = sin(pos/10000^{2i/d}), \\
    PE_{(pos,2i+1)} = cos(pos/10000^{2i/d}),
    \end{split}
\end{equation}
where  $pos$ is the position in the sequence, $i$ is the $i^{th}$ position in the embedding dimension, and $d$ denotes the size of the dimension (same as the embedding matrix $E$).
Therefore, given the input sequence X, its embedded representations are defined as
\begin{equation}
    H = E(X) + PE(X), \label{eq:embedded}
\end{equation}
where $H \in R^{n \times d}$ is the resulting embeddings of the input.

\begin{figure}[!t]
  \centering
  \includegraphics[width=.8\linewidth]{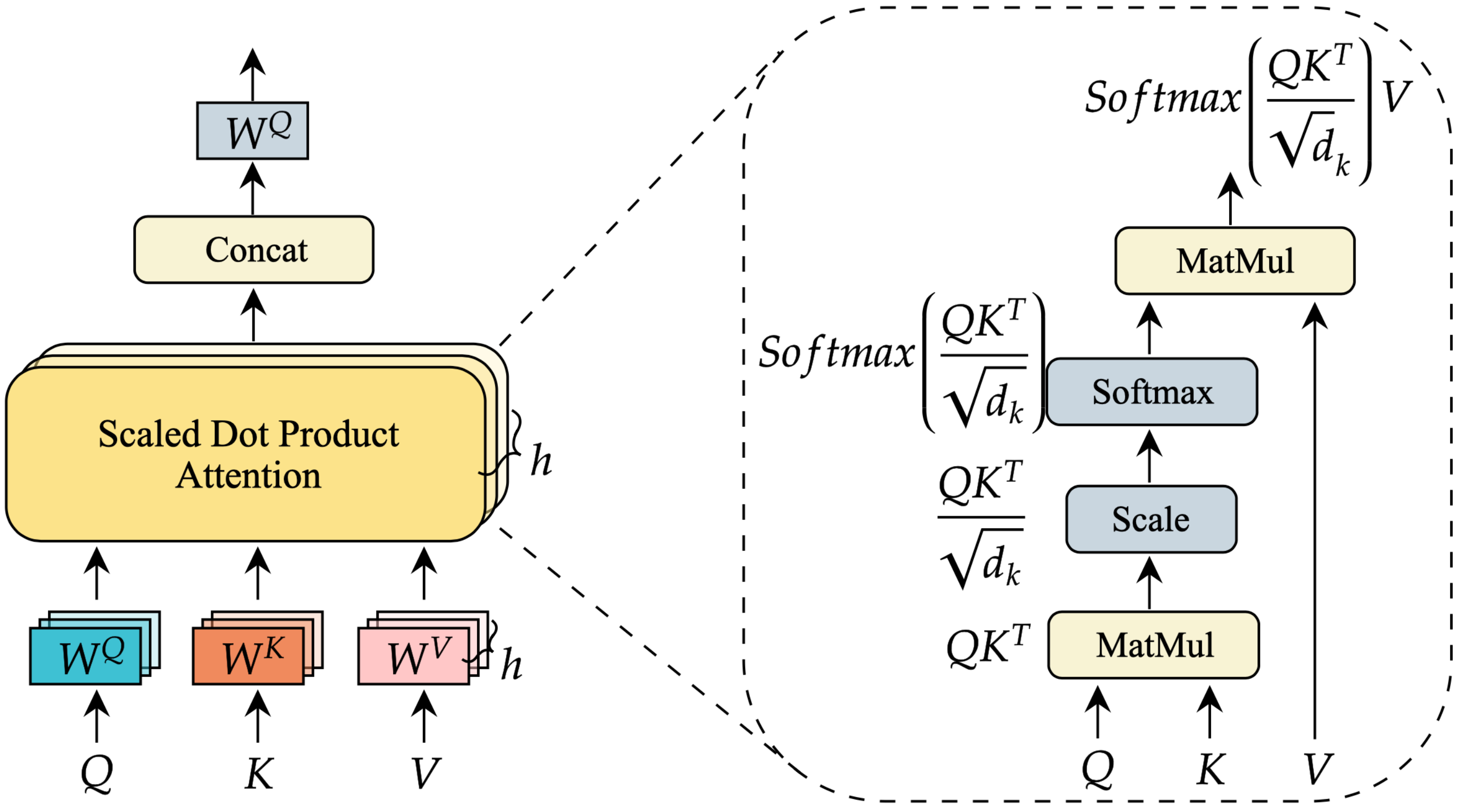}
  \caption{Illustration of multi-head attention (left) and its component scaled dot product attention (right). This figure is adapted from~\citet{vaswani2017attention}.}
  \label{fig:multi_head_attention}
\end{figure}

\subsubsection{Multi-Head Attention}
Figure~\ref{fig:multi_head_attention} provides an illustration of an essential module used in Transformer called multi-head attention. A major component of multi-head attention is the scaled dot product attention, which is formulated as:
\begin{equation}
    Attention (Q,K,V) = Softmax(\frac{QK^T}{\sqrt{d_k}}) V,
\end{equation}
where $Q \in R^{d_q}$, $K \in R^{d_k}$, and $V \in R^{d_v}$ are the input vectors, and normally we set $d_q = d_k$.

Instead of using a single-head attention, we can apply multiple $h$ heads that learn the linear projections of the query ($Q$), key ($K$), and value ($V$), which is beneficial in learning representation subspaces at different positions:
\begin{equation}
    MultiHead (Q,K,V) = Concat (head_1, head_2, ..., head_h) W^O,
\end{equation}
\begin{equation}
    head_i = Attention(QW^Q_i, KW^K_i, VW^V_i),
\end{equation}
where we assign $Q=K=V=Z$, and the projection matrices $W^Q_i \in R^{d \times d_q}$, $W^K_i \in R^{d \times d_k}$, $W^V_i \in R^{d \times d_v}$, and $W^O_i \in R^{hd_v \times d}$.

\begin{figure}[!t]
  \centering
  \includegraphics[width=.3\linewidth]{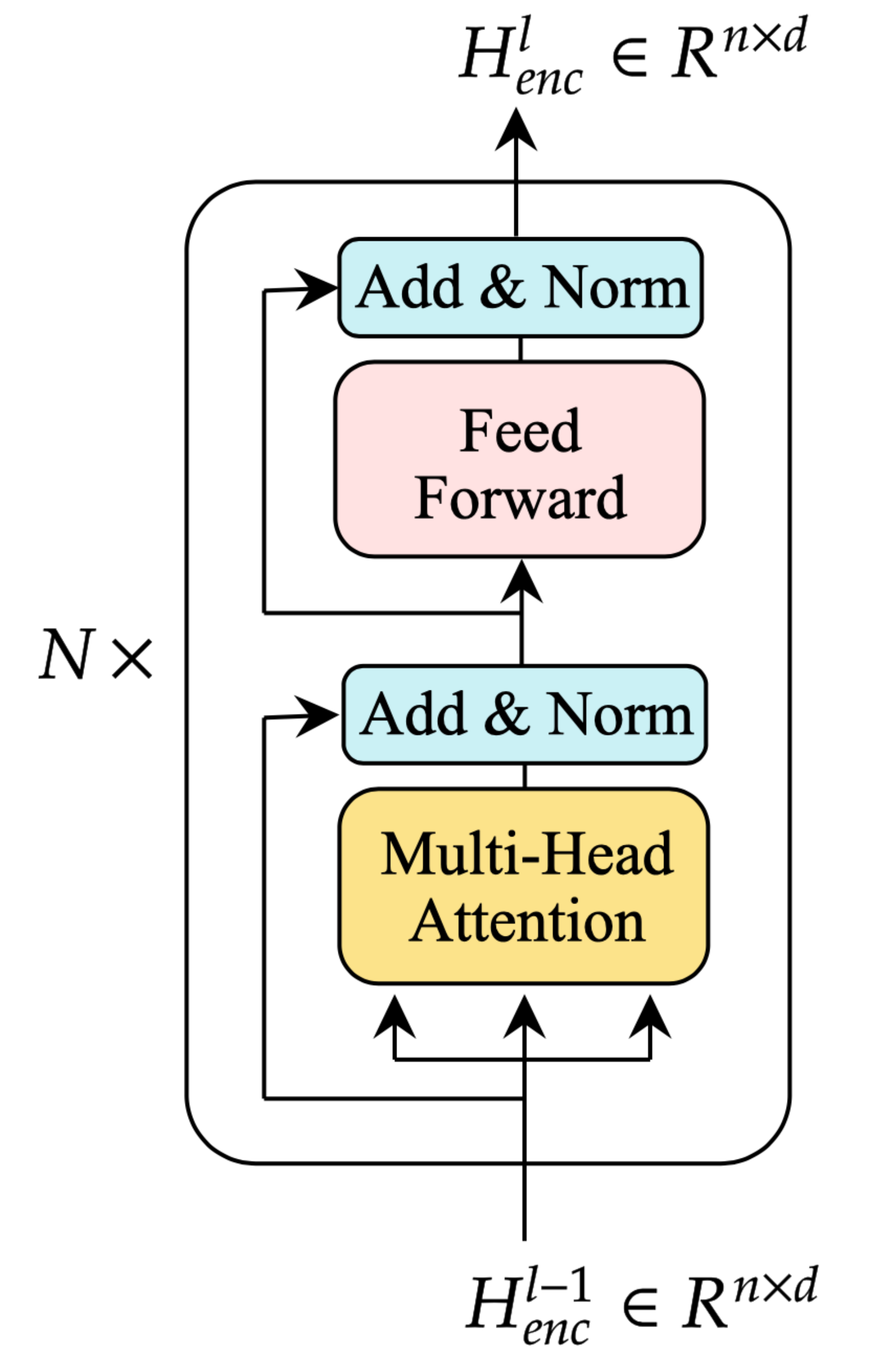}
  \caption{The structure of the Transformer encoder. This figure is adapted from~\citet{vaswani2017attention}.}
  \label{fig:trs_encoder}
\end{figure}

\subsubsection{Encoder}
Figure~\ref{fig:trs_encoder} shows the structure of the Transformer encoder, which is made of a stack of ($N$) encoder layers. Each encoder layer consists of a multi-head attention module, a feed forward ($FF$) layer, and two layer normalization modules. $H_{enc}^{l}$ denotes the outputs from the $l^{th}$ layer, and the inputs of the first layer ($H_{enc}^0$) are the embedded representations $H_{enc}$ from Eq~\ref{eq:embedded}. Each layer returns a transformed version of the embedded representation and feeds the outputs to the next layer:
\begin{equation}
    H_{enc}' = LayerNorm (H_{enc}^{l-1} + MultiHead(H_{enc}^{l-1},H_{enc}^{l-1},H_{enc}^{l-1})),
\end{equation}
\begin{equation}
    H_{enc}^{l} = LayerNorm (H_{enc}' + FF(H_{enc}')),
\end{equation}
\begin{equation}
    FF(x) = max(0, W_1 x + b_1) W_2 + b_2,
\end{equation}
where $FF$ is a feed forward neural network with two layers, and ReLU activation.
Therefore, the forward function of the Transformer encoder can be formulated as:
\begin{equation}
    H_{enc}^N = TRS_{enc} (E(X) + PE(X)), 
\end{equation}
where $TRS_{enc}$ denotes the Transformer encoder with N layers, and $E$ and $PE$ are the embedding matrix and the positional embedding matrix, respectively.

\begin{figure}[!t]
  \centering
  \includegraphics[width=.32\linewidth]{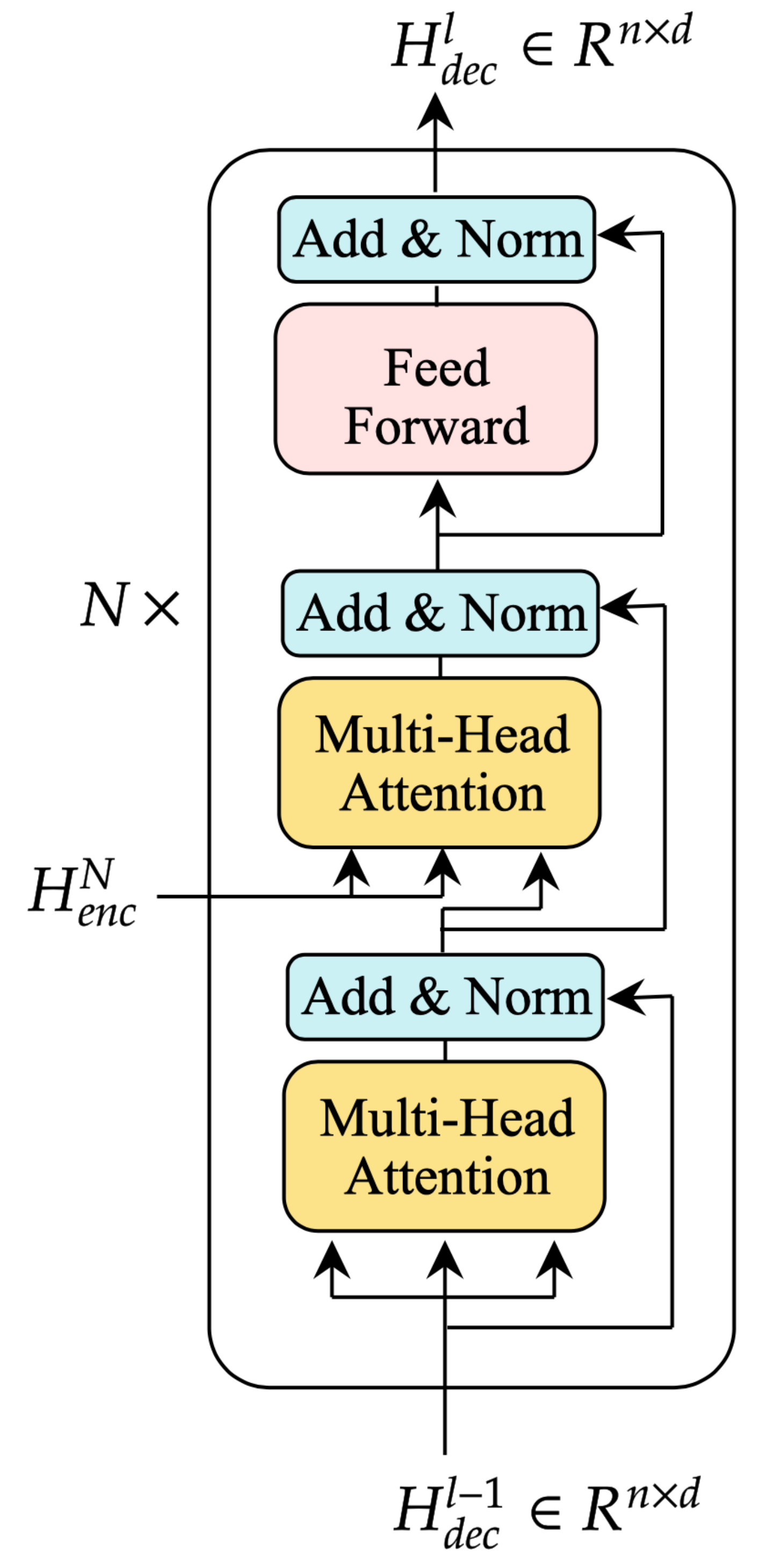}
  \caption{The structure of the Transformer decoder. This figure is adapted from~\citet{vaswani2017attention}.}
  \label{fig:trs_decoder}
\end{figure}

\subsubsection{Decoder}
Figure~\ref{fig:trs_decoder} provides the structure of the Transformer decoder, which is made of a stack of ($N$) decoder layers. Each decoder layer consists of two multi-head attention modules, a feed forward ($FF$) neural network, and their corresponding layer normalization. 
$H_{dec}^{l}$ denotes the outputs from the $l^{th}$ layer, and the inputs of the first layer ($H_{dec}^0$) is the embedded representations for the previous output tokens. Each layer returns a transformed version of the embedded representation:
\begin{equation}
    H'_{dec} = LayerNorm (H_{dec}^{l-1} + MultiHead (H_{dec}^{l-1}, H_{dec}^{l-1}, H_{dec}^{l-1})),
\end{equation}
\begin{equation}
    H''_{dec} = LayerNorm (H'_{dec} + MultiHead (H_{enc}^N, H_{enc}^N, H'_{dec})),
\end{equation}
\begin{equation}
    H_{dec}^{l} = LayerNorm (H''_{dec} + FF (H''_{dec})),
\end{equation}
where $H_{enc}^N$ comes from the outputs of the encoder, and the transformed representation ($H_{dec}^{l}$) will be fed to the next layer.
Therefore, the forward function of the Transformer decoder can be formulated as:
\begin{equation}
    H_{dec}^N = TRS_{dec} (E(Y) + PE(Y)),
\end{equation}
where $TRS_{dec}$ denotes the Transformer decoder with N layers. Note that the number of layers in the decoder can be different from that in the encoder.

\section{Representation Learning in NLP}
Developing methods to represent the meaning of words (or sentences) lays the foundation for understanding and processing natural languages. In this section, we cover the background of learning word representations and contextualized representations (i.e., language model pre-training), and how to extend them into multilingual versions.

\subsection{Word Embeddings}
Learning word representations through embedding is a fundamental technique to capture latent word semantics~\cite{bollegala2016joint,clark2015vector,mikolov2013efficient,mikolov2013distributed,niu2017improved,pennington2014glove,qiu2014learning}. Word2Vec embeddings~\cite{mikolov2013efficient,mikolov2013distributed} were proposed to learn word-level representations by predicting each word's surrounding words. These embeddings have two major training methods: 1) Skip-gram; and 2) continuous bag of words (CBOW)~\cite{mikolov2013efficient}. 
The skip-gram trains the model to predict the context words of a given word, while CBOW uses a reverse technique by training the model to predict the word given its context words.
Both skip-gram and CBOW need large data corpus for training, so as to obtain dense and expressive word embeddings. 
After the training, the Word2Vec embeddings can map words in the discrete space into vector representations in the continuous space, and capture the semantics of words by grouping semantically similar words into a similar vector space.

Another type of effective word embeddings is GloVe~\cite{pennington2014glove}. GloVe combines the advantages of the two major model families in learning word representations, global matrix factorization~\cite{deerwester1990indexing,lund1996producing} and local context window methods~\cite{mikolov2013efficient,mikolov2013distributed}. GloVe uses a specific weighted least squares model that trains on global word-word co-occurrence counts and thus makes efficient use of statistics. The model produces a word vector space with a meaningful sub-structure.

Due to the rich structures of languages, there are numerous rare or misspelled words existing in natural language text. These words are generally ignored by word embeddings, since they are not covered by the vocabulary.
Byte pair encoding (BPE) embeddings~\cite{heinzerling2018bpemb,sennrich2016neural} and FastText~\cite{grave2018learning,lund1996producing,mikolov2018advances} were proposed to address this issue. To address the out-of-vocabulary (OOV) issue and better capture the morphology of the language, FastText and BPEmb~\cite{heinzerling2018bpemb} incorporate the representations of subwords. Specifically, if a word is unknown, its representations are formed by summing all vectors of the subwords.

\subsection{Multilingual Word Embeddings} 

Multilingual word embeddings and the cross-lingual transfer have become popular NLP research topics, due to the globalization and the public awareness of the digital language divide\footnote{E.g.,~\url{http://labs.theguardian.com/digital-language-divide/}}~\cite{ammar2016massively,chen2018unsupervised,duong2017multilingual,joulin2018loss,lample2018word,lu2015deep,ruder2019survey}.
Multilingual word embeddings extend word representations into multiple languages, which allows NLP methods or models to be applied for different languages.
The goal of building multilingual word embeddings is to map the multiple monolingual word embeddings into a shared space (i.e., cross-lingual alignment). After mapping, semantically similar words across languages will be close to each other.
Building multilingual word embeddings brings two benefits: First, they enable us to align the semantics of words across languages, which is key to the bilingual lexicon induction, machine translation, and cross-lingual information retrieval tasks. Second, they enable the model to transfer task knowledge learned in one language to other languages~\cite{chen2018unsupervised,lample2018word,ruder2019survey}.

Constructing multilingual word embeddings is mainly based on word-level alignment methods~\cite{chen2018unsupervised,lample2018word,joulin2018loss}, which usually rely on parallel word-aligned data. The existing approaches can be categorized into three types: 1) mapping-based approaches; 2) pseudo-multilingual corpora-based approaches; and 3) joint methods.

\subsubsection{Mapping-based Approaches}
Mapping-based approaches, by far the most prominent category of multilingual word embedding models, aim to learn a mapping from the monolingual embedding spaces to a joint multilingual space. Four types of mapping methods have been proposed: a) regression methods; b) orthogonal methods; c) canonical methods; and d) margin methods. Their details are as follows:

\paragraph{Regression methods}
Regression methods map the embeddings of the source language to the target language space by maximizing their similarity.
\citet{mikolov2013exploiting} observed that words and their translations show similar geometric constellations in monolingual embedding spaces after an appropriate linear transformation is applied, and suggests that it is possible to transform the vector space of a source language s to the vector space of the target language t by training a transformation matrix ($W$):
\begin{equation}
    MSE = \sum_{i=1}^{n} || W x_i^{s} - x_i^{t} ||^2,
\end{equation}
where $x_i^s$ and $x_i^t$ denote the embeddings of the $i^{th}$ parallel source and target words, respectively, and $n$ is the size of the parallel word-aligned data.
Later on, this basic approach~\cite{mikolov2013exploiting} was adopted by many other works~\cite{artetxe2016learning,artetxe2017learning,dinu2014improving,lample2018word,xing2015normalized,zhang2017adversarial,zhang2016ten}.
\citet{artetxe2017learning} greatly alleviated the reliance on large numbers of word pairs by leveraging the structural similarity of embedding spaces for the transformation matrix learning, and \citet{lample2018word} proposed to use the adversarial training approach to learn the transformation matrix without using any parallel word-aligned data.

\paragraph{Orthogonal methods}
Orthogonal methods map the embeddings in the source language to maximize their similarity with the target language embeddings, but constrain the transformation to be orthogonal.
The most common way in which the basic regression method of the previous section has been improved is to constrain the transformation $W$ to be orthogonal, i.e. $W^{T} W = I$, which is motivated by~\citet{xing2015normalized} in order to preserve the length normalization. \citet{artetxe2016learning} leveraged the orthogonality as a means to ensure monolingual invariance. An orthogonality constraint has also been used to regularize the mapping~\cite{zhang2017adversarial,zhang2016ten} and has been theoretically shown to be self-consistent~\cite{smith2017offline}.

\paragraph{Canonical methods} 
Canonical methods map the embeddings of both languages to a new shared space using Canonical Correlation Analysis (CCA), which maximizes their similarity.
\citet{haghighi2008learning} first leveraged canonical methods for learning translation lexicons for the words of different languages. 
\citet{faruqui2014improving} applied CCA to project the word embeddings of two languages into a shared embedding space.
\citet{artetxe2016learning} showed that the canonical method is similar to the orthogonal method with dimension-wise mean centering. Furthermore, \citet{artetxe2018generalizing} showed that regression methods, canonical methods, and orthogonal methods can be seen as instances of a framework that includes optional whitening and de-whitening steps.

\paragraph{Margin methods}
Margin methods map the embeddings of the source language to maximize the margin between correct translations and other candidates.
\citet{lazaridou2015hubness} reduced the hubness by optimizing a max-margin-based ranking loss instead of MSE.
The idea is to assign a higher cosine similarity to word pairs that are translations of each other than random word pairs.
\citet{joulin2018loss} proposed a margin-based method, which replaces cosine similarity with cross-domain similarity local scaling (CSLS).

\subsubsection{Pseudo-multilingual corpora-based approaches}
Instead of using a seed of a bilingual dictionary to learn the mapping matrix, some approaches use it to construct pseudo-multilingual corpora and train the multilingual word embeddings based on the corpora. \citet{xiao2014distributed} proposed to use a seed bilingual dictionary to randomly replace words in a source language corpus with their translations. Then, they trained this model using an MML loss function~\cite{collobert2008unified} by feeding it context windows of the constructed pseudo-bilingual corpus containing source and target language text.
\citet{gouws2015simple} concatenated the source and target language corpus, replacing each word that was part of a translation pair with its translation equivalent under a certain probability, and \citet{ammar2016massively} extended this approach to multiple languages.
Instead of randomly replacing every word in the corpus with its translation, \citet{duong2016learning} replaced each center word with a translation on-the-fly during CBOW~\cite{mikolov2013efficient} training. They jointly learned to predict both the words and their appropriate translations using PanLex as the seed bilingual dictionary.

In practice, pseudo-multilingual corpora-based methods are more expensive as they require training multilingual word embeddings from scratch based on the concatenation of large monolingual corpora. In contrast, mapping-based approaches are much more computationally efficient as they leverage existing monolingual word embeddings, and the transformation matrix can be learned efficiently.

\subsubsection{Joint Methods}
Mapping-based approaches first optimize multiple monolingual losses, and then the cross-lingual regularization loss. Pseudo-multilingual corpora-based approaches optimize monolingual losses and implicitly the cross-lingual regularization loss. 
Different from them, joint methods optimize monolingual and cross-lingual objectives at the same time.
\citet{klementiev2012inducing} used the multi-task learning technique and jointly optimized a source language and target language with a cross-lingual regularization term. This regularization term leverages the parallel data, such as the Europarl corpus~\cite{koehn2005europarl}, to encourage the representations of source and target language words to be similar.
\citet{kovcisky2014learning} simultaneously trained word embeddings and cross-lingual alignments using FastAlign~\cite{dyer2013simple}. They used a word of the source language sentence to predict the corresponding word in the aligned target language sentence.

\subsection{Language Model Pre-Training}
Recently, language model pre-training~\cite{brown2020language,dai2020kungfupanda,dai2020modality,dai2021multimodal,dai2019transformer,devlin2019bert,dong2019unified,lewis2020bart,liu2019roberta,lovenia2021ascend,peters2018deep,radford2019language,shoeybi2019megatron,xu2020emograph,xu2021caire,xu2021retrieval,yang2019xlnet,yu2021vision} has greatly improved the state-of-the-art results on many downstream tasks in natural language processing. In general, pre-trained language models are divided into bi-directional~\cite{devlin2019bert,dai2019transformer,liu2019roberta,peters2018deep}, uni-directional or causal-decoder~\cite{brown2020language,radford2019language,raffel2019exploring,shoeybi2019megatron}, and encoder-decoder generative models~\cite{lewis2020bart,raffel2019exploring}. These language models are generally pre-trained on a super large-scale monolingual corpus (e.g., Wikipedia, BookCorpus~\cite{zhu2015aligning}, CommonCrawl~\cite{wenzek2020ccnet}), so as to enable their powerful language understanding ability.

\begin{figure}[!t]
  \centering
  \includegraphics[width=.7\linewidth]{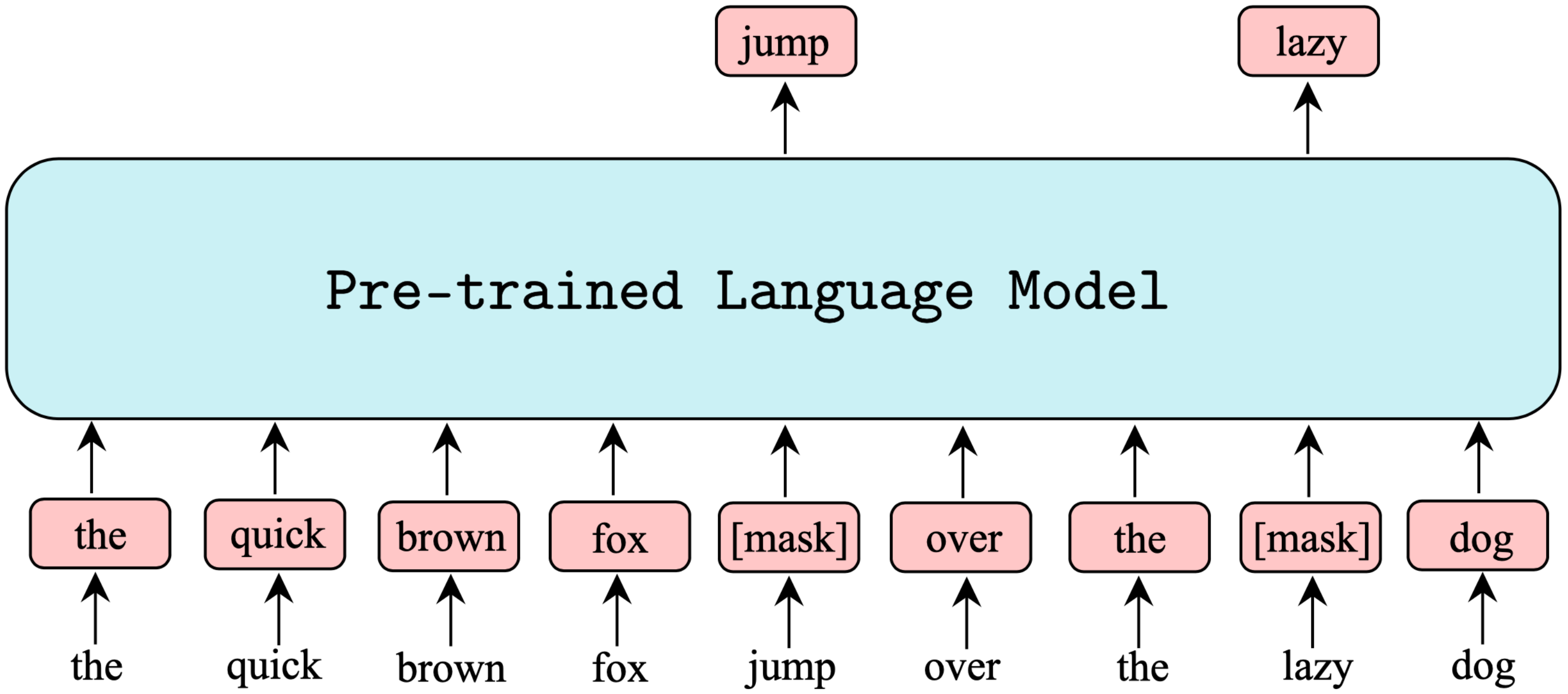}
  \caption{Illustration of masked language modeling (MLM).}
  \label{fig:mlm}
\end{figure}

The bi-directional pre-trained language models are usually trained using the MLM loss proposed by~\citet{devlin2019bert}.
Figure~\ref{fig:mlm} provides an illustration of the MLM. It first randomly masks the tokens in a sentence (illustrated as $[mask]$ in Figure~\ref{fig:mlm}), and then the model is trained to predict the masked tokens based on the context. 
To make the correct prediction and thanks to the massive pre-training corpus, the models learn to comprehend the provided context information even for noisy text. 
As for the uni-directional (casual-decoder) and encoder-decoder generative models, they are usually trained to predict the next token of the current context. The model is trained to learn the likelihood function of the conditional probability $P_\theta (Y_{curr}|Y_{prev})$, which can be formulated as:
\begin{equation}
    P_\theta (Y_{curr}|Y_{prev}) = \prod_{i=1}^{n} P_\theta (y_i | y_0, ..., y_{i-1}),
\end{equation}
where $\theta$ is the trainable parameters, and $\{y_0, ..., y_n\}$ is the sequence with a length of $n$. 
Recent advanced NLU models are usually constructed based on the bi-directional pre-trained language models, while the uni-directional (casual-decoder) and encoder-decoder models are more suitable for natural language generation tasks.

\subsection{Multilingual Language Model Pre-Training}
Monolingual pre-trained language models only allow the input to be a single language, and the task knowledge learned in the source language cannot be transferred to other target languages.
Multilingual language model pre-training extends the monolingual pre-training objective function into a multilingual version~\cite{chi2020cross,chi2021infoxlm,conneau2020unsupervised,devlin2019bert,huang2019unicoder,lample2019cross,liu2020multilingual,liu2021continual,winata2021multilingual,xue2021mt5}. Similar to monolingual pre-trained language models, multilingual pre-trained models can be categorized into two types: 1) bi-directional pre-training; and 2) uni-directional (generative) pre-training. 

\subsubsection{Bi-directional Pre-training}
Bi-directional multilingual pre-training generally relies on the masked language modeling (MLM) objective function proposed in~\citet{devlin2019bert}. Multilingual BERT (M-BERT)~\cite{devlin2019bert} was the first large-scale pre-trained model, and it uses bi-directional pre-training. M-BERT extends BERT and is also pre-trained using the MLM loss function, but it is simultaneously pre-trained on the text corpus of 104 languages.
Surprisingly, M-BERT is shown to possess strong cross-lingual transfer ability, and it achieves remarkable zero-shot cross-lingual performance on multiple downstream tasks, such as named entity recognition, part of speech tagging, dependency parsing, document classification, and natural language inference tasks~\cite{pires2019multilingual,wu2019beto}. Since M-BERT was proposed, more advanced multilingual pre-trained models have been developed by increasing the model sizes~\cite{conneau2020unsupervised,liu2020multilingual,xue2021mt5} and incorporating more objective functions~\cite{chi2020cross,chi2021infoxlm,huang2019unicoder,lample2019cross}.

\citet{lample2019cross} presented a pre-trained language model called XLM and proposed to combine MLM with a translation language modeling (TLM) loss function in the pre-training stage. TLM is similar to MLM, but requires bilingual parallel sentences. It concatenates the parallel source and target sentences, and randomly masks words in both source and target sentences to train the model for prediction. XLM is not only effective at cross-lingual tasks, but can also be a good initialization for machine translation models.
Based on XLM~\cite{lample2019cross}, \citet{huang2019unicoder} added two objective functions: 1) cross-lingual word recovery; and 2) cross-lingual paraphrase classification. They showed that the pre-trained model has more powerful cross-lingual ability with these additional objective functions.
Then, \citet{conneau2020unsupervised} proposed XLM-RoBERTa (XLM-R) which used a larger size of pre-training corpora across 100 languages for the pre-training. They found that both M-BERT and XLM are undertuned and showed that the model's cross-lingual ability and multilingual language understanding can be simply improved by increasing the model capacity. 
Furthermore, \citet{chi2021infoxlm} presented an information-theoretic framework and pre-trained a multilingual language model, InfoXLM, which incorporates a new pre-training objective function based on contrastive learning. Experiments show that InfoXLM further improves upon the cross-lingual transfer performance of XLM-R across multiple downstream tasks, such as cross-lingual natural language inference, cross-lingual sentence retrieval, and cross-lingual question answering.

\subsubsection{Uni-directional Pre-training}
Bi-directional pre-training (e.g., MLM) enables the model to predict the masked tokens in the middle of sentences, while uni-directional pre-training focuses on training the model to predict the next token of the current context. As a result, models trained in a uni-directional fashion are generally applied for the generation tasks, thanks to their powerful generative ability. 
\citet{liu2020multilingual} extended the denoising auto-encoding proposed by~\citet{lewis2020bart} into a multilingual version, and built an mBART model, which was shown to be effective at machine translation tasks, especially in low-resource scenarios. \citet{xue2021mt5} pre-trained mT5 by extending the text-to-text objective function proposed in~\citet{raffel2019exploring}, and further illustrated the cross-lingual ability brought by increasing the size of language models.
Furthermore, \citet{chi2020cross} proposed XNLG which incorporates MLM and denoising autoencoding objective functions. XNLG contains a two-stage pre-training. In stage-1, it follows the pre-training stage in XLM~\cite{lample2019cross}, and uses the pre-trained XLM as the initialization of the encoder and decoder. In stage-2, it uses the denoising autoencoding objective function on a corpus of multiple languages to enable its multilingual generative ability.

\section{Natural Language Understanding}
\subsection{Background}

NLU is the fundamental component of natural language processing systems. It contains a wide variety of tasks, including emotion recognition, semantic parsing, dependency parsing, natural language inference, named entity recognition, part of speech tagging, etc, and the core capability for conducting these tasks is being able to comprehend natural languages and extract task-specific features. In this thesis, we mainly focus on applying our proposed models to the semantic parsing, named entity recognition (NER), and part of speech (POS) tagging tasks. We choose these tasks since they are essential in NLU, and the corresponding models can be used or applied in many NLP systems. We describe these three tasks below.

\subsection{Semantic Parsing}
Semantic parsing is the task of converting a natural language utterance into a logical form: a machine-understandable representation of its meaning.
Semantic parsing plays an essential role in personal assistant and has recently become an interesting research topic. In general, the output of a semantic parser is a data structure that represents the underlying meaning of the given user query, for example, ``show me all important meetings''.

A number of semantic parsing datasets have been developed in recent years~\cite{banarescu2013abstract,coucke2018snips,gupta2018semantic,hemphill1990atis,lawrence2018improving,wang2015building,zelle1996learning}.
One of the main applications is task-oriented semantic parsing, where the task can be cast into a combination of detecting the intent of the query and extracting specific entity values (slot filling)~\cite{coucke2018snips,goo2018slot,hemphill1990atis,liu2016attention}. Sequence-labeling-based models have been proposed to parse the queries~\cite{goo2018slot,liu2016attention,haihong2019novel,wang2018bi,zhang2019joint}, and these use a linear classifier to predict the intent and a sequence labeling structure to assign one label to each token in the query.
Moreover, compositional semantic parsing models are needed for complex queries where one token could have multiple intents or entity values~\cite{berant2014semantic,furrer2020compositional,gupta2018semantic,pasupat2015compositional,shaw2020compositional}.
To tackle these challenges, pre-trained generative models~\cite{lewis2020bart,radford2019language,raffel2019exploring} based on the sequence to sequence framework~\cite{sutskever2014sequence} have been combined with pointer networks~\cite{vinyals2015pointer} to generate new output tokens and also to learn to copy tokens from the input, so as to produce machine-understandable logical forms~\cite{chen2020low,herzig2020span,oren2020improving,pasupat2019span,rongali2020don}. 


\subsection{Named Entity Recognition}
NER is the task of locating and classifying named entities mentioned in unstructured text into pre-defined categories such as persons, organizations, and locations. NER acts as an important pre-processing step for a variety of downstream applications such as information retrieval, question answering, machine translation, etc. It is generally cast as a sequence labeling task, where each token in a sentence has a corresponding label. In recent years, many NER datasets have been introduced~\cite{derczynski2016broad,jia2019cross,lu2018visual,nedellec2013overview,pradhan2012conll,sang2002introduction,sang2003introduction,strauss2016results,wang2020comprehensive}, and deep-learning-based NER models have become dominant and achieved state-of-the-art results~\cite{huang2015bidirectional,lample2016neural,li2020survey,ma2016end,zheng2017joint}.

\citet{huang2015bidirectional} first incorporated a bi-directional LSTM and conditional random field (CRF) for the sequence tagging tasks (including NER), which produced state-of-the-art results. Following~\citet{huang2015bidirectional}, the architecture BiLSTM-CRF has been used as a basic sequence encoder for the NER task~\cite{chiu2016named,lample2016neural,li2017leveraging,lin2017multi,ma2016end,wei2016disease,yang2016multi,zhai2017neural,zheng2017joint}. \citet{yang2016multi} employed deep GRUs~\cite{chung2014empirical} on both character and word levels to encode morphology and context information.
\citet{vzukov2018named} leveraged multiple independent BiLSTM units across the same input and injected diversity among the BiLSTM units by using an inter-model regularization term. Moreover, \citet{katiyar2018nested} introduced a modification to BiLSTM-based sequence labeling models to cope with the nested NER task, and \citet{ju2018neural} proposed to detect nested entities by dynamically stacking flat NER layers until no outer entities are extracted.
Furthermore, pre-trained contextualized models~\cite{devlin2019bert,liu2019roberta,yang2019xlnet} have been used to replace the BiLSTM encoder, and significantly improve the entity tagging performance~\cite{li2020dice,li2020unified,luo2020hierarchical,xia2019multi}.

\subsection{Part of Speech Tagging}
POS tagging is the process of marking up a word in a text or corpus as corresponding to a particular part of speech based on its definition and context.
An accurate POS tagger is essential for many NLP tasks, such as NER, sentiment analysis, dependency parsing, etc.
Recently, deep learning models, such as convolutional neural network (CNN)-based and RNN-based architectures, has been used successfully at solving the POS tagging task~\cite{collobert2008unified,dowlagar2021pre,gui2017part,meftah2018neural,plank2016multilingual,senthil2020bi,wang2015part,zupon2020analysis}.
\citet{simov2001hybrid} provided a hybrid approach to POS tagging in Bulgarian by combining an RNN with a rule-based approach, and \citet{mohnot2014hybrid} incorporated a hidden Markov model with rule-based methods for POS tagging in Hindi. \citet{gui2017part} leveraged adversarial networks to improve the POS tagging in the Twitter domain. Moreover, \citet{zupon2020analysis} analyzed the effectiveness of capsule networks for the POS tagging, and found that they performed nearly as well as a more complex LSTM under certain training conditions. Similar to other NLU tasks, leveraging a pre-trained language model~\cite{devlin2019bert,liu2019roberta,yang2019xlnet} as the sequence encoder can significantly improve the POS tagging performance~\cite{delobelle2020robbert,martin2020camembert,nguyen2020bertweet,nguyen2020phobert}.

\section{Low-Resource Transfer Learning}
\subsection{Low-Resource Problems}
Deep neural networks are known for requiring large amounts of training data, and the shortage of data samples in low-resource scenarios limits their performance in such cases~\cite{conneau2018xnli,jia2019cross,schuster2019cross}. 
Compared to high-resource languages (or domains), it is much more difficult to collect labeled task-related data samples or unlabeled corpora for low-resource languages (or domains).
To alleviate the data-hungry dependencies, it is straightforward to leverage the data samples from high-resource scenarios and then transfer the knowledge into the corresponding low-resource scenarios. Such low-resource transfer learning ensures the effectiveness of deep learning models despite a data shortage in the target languages or domains~\cite{conneau2020unsupervised,jia2019cross,pfeiffer2020mad,shah2019robust}.
In this thesis, we focus on the data scarcity issue in low-resource languages and domains, which is the main challenges in low-resource NLP~\cite{hedderich2021survey}.

\subsection{Cross-Lingual Transfer Learning}
Cross-lingual transfer learning approaches were proposed to cope with the data scarcity issue in low-resource languages~\cite{artetxe2019massively,chi2020cross,chi2021infoxlm,conneau2020unsupervised,kim2017cross,pfeiffer2020mad}.
Zero-shot cross-lingual models are usually built based on the training samples of the source language and allowed a direct transfer to target languages without using any of their training samples~\cite{ahmad2019difficulties,artetxe2019massively,chen2018xl,conneau2020unsupervised,schuster2019cross}. Meanwhile, few-shot cross-lingual models can leverage only a few training samples from the target languages which are then combined with the training samples in the source language for training~\cite{hardalov2021few,stappen2020cross,upadhyay2018almost,wiesner2021injecting}.
Since only a few or even zero data samples for the target languages are available, the cross-lingual models are generally required to have a good cross-lingual/multilingual alignment quality after the initialization, so as to successfully transfer the learned knowledge from one language to the others~\cite{artetxe2017learning,conneau2020unsupervised,devlin2019bert,lample2018word}.

Numerous cross-lingual models leveraged multilingual word embeddings~\cite{artetxe2016learning,artetxe2017learning,joulin2018loss,lample2018word} to initialize their embedding layers, and adaptive modules are built on top of the embeddings~\cite{chen2019multi,chen2018xl,kim2017cross,lin2018multi,schuster2019cross}.
\citet{kim2017cross} used common and private BiLSTM to learn language-general and language-specific representations and showed the effectiveness on the POS tagging task. Similarly, \citet{chen2019multi} integrated the language-invariant and language-specific features from multiple source languages for the cross-lingual transfer of the semantic parsing, NER, and sentiment analysis tasks.
A teacher-student framework was used by~\citet{chen2018xl} to transfer the knowledge from the teacher working on the source language to the student working on the target language. Taking this further, \citet{daza2019translate} trained a model in a cross-lingual setting to generate new task-related labels for low-resource languages, and \citet{lin2018multi} introduced a multilingual and multi-task architecture for few-shot cross-lingual sequence labeling tasks.

Recently, multilingual pre-trained models~\cite{conneau2020unsupervised,devlin2019bert,lample2019cross} have shown their strong cross-lingual ability, and simply adding a linear layer on top of them for fine-tuning achieves remarkable and even state-of-the-art cross-lingual performance on multiple downstream NLP tasks, such as semantic parsing, natural language inference, NER, POS tagging, question answering, etc~\cite{pires2019multilingual,wu2019beto}.
Furthermore, several approaches have been proposed to further improve the cross-lingual transfer ability on top of the pre-training framework. 
The first direction is to increase the number of model's parameters, as well as the size of the data corpus used for pre-training. \citet{conneau2020unsupervised} and \citet{xue2021mt5} showed that simply enlarging the model and corpus size can bring significant improvements on numerous downstream NLP tasks.
The second direction is to add more effective objective functions in the pre-training stage. \citet{huang2019unicoder} added cross-lingual word recovery and paraphrase classification based on the MLM in XLM~\cite{lample2019cross}. Taking this further, \citet{chi2021infoxlm} incorporated a contrastive-learning-based objective function for pre-training and further improved the performance on XLM-R~\cite{conneau2020unsupervised} across multiple NLP tasks.
The third direction is to leverage parameter-efficient fine-tuning approaches~\cite{houlsby2019parameter,parovic2022bad,zaken2022bitfit}. \citet{pfeiffer2020mad} proposed an adapter-based framework for multi-task cross-lingual transfer which makes the training process much more efficient. \citet{ansell2021mad} proposed to generate language adapters from language representations based on typological features, which further improved the fine-tuning efficiency.

\subsection{Cross-Domain Transfer Learning}
Cross-domain transfer learning approaches have been proposed to tackle the data scarcity issue in the low-resource domains~\cite{guo2018multi,gururangan2020don,he2020contrastive,jia2019cross,li2013active,nguyen2021dozen,shah2019robust}.
For many NLU tasks such as the semantic parsing and NER tasks, adapting to a new domain usually requires the model to detect new label types that do not exist in source domains. This increases the challenges of the zero-shot domain adaptation tasks, due to the difficulties of recognizing unseen label types~\cite{he2020contrastive,shah2019robust,wu2019transferable,zhao2018zero}. 
A great many domain-adaptive methods were proposed by incorporating the training samples from the source domain and a few training samples from the target domain~\cite{guo2018multi,gururangan2020don,jia2019cross,ye2021crossfit,yu2019sparc,zhang2021simple}.
\citet{jia2019cross} proposed the cross-domain language modeling to allow a fast adaptation from the source domain to target domains, and \citet{jia2020multi} further improved the domain adaptation performance by introducing a multi-cell compositional LSTM framework.
\citet{wu2019transferable} shared the model parameters for all source domains to allow the zero/few-shot adaptation to the entities in new domains, while \citet{shah2019robust} utilized entity descriptions and entity examples in target domains to boost the model's zero/few-shot semantic parsing performance.
\citet{chen2020low} adopted the pointer-generator networks~\cite{see2017get} to improve the domain adaptation performance for the compositional semantic parsing task.

Recently, pre-trained language models have shown their strong language understanding ability and powerful domain adaptation ability~\cite{devlin2019bert,dong2019unified,hung2022multi2woz,liu2019roberta,yang2019xlnet}, and they have become backbone models for the domain adaptation systems~\cite{el2021domain,gururangan2020don,myagmar2019cross,zhou2020sentix,zou2021unsupervised}. 
Taking this further, domain-specific or task-specific pre-training has been used to further improve the domain adaptation performance.
Extensive domain-specific pre-trained models are constructed based on the existing strong pre-trained language models~\cite{beltagy2019scibert,feng2020codebert,lee2020biobert,liu2021ner,su2019vl,sun2019videobert,wu2020tod}.
\citet{lee2020biobert} used large-scale biomedical corpora to continue pre-training BERT~\cite{devlin2019bert} and achieved state-of-the-art performance on multiple text mining tasks in the biomedical domain. Similarly, SciBERT~\cite{beltagy2019scibert} was constructed using a large multi-domain corpus of scientific publications, and VideoBERT~\cite{sun2019videobert} was pre-trained based on the concatenation of the videos and their corresponding captions. In order to quickly adapt to the sentiment domain, \citet{zhou2020sentix} pre-trained a sentiment-aware language model via domain-invariant sentiment knowledge from large-scale review datasets. Moreover, \citet{gururangan2020don} leveraged domain-related and task-related corpora to continue pre-training RoBERTa~\cite{liu2019roberta}, and provided an in-depth analysis for these two approaches. Furthermore, parameter-efficient domain-adaptation has also been explored~\cite{chronopoulou2021efficient,dingliwal2021prompt,hung2022ds}. 
\citet{dingliwal2021prompt} proposed to use prompt-tuning for building automatic speech recognition (ASR) systems and improved the training efficiency.
\citet{hung2022ds} leveraged domain adapters for encoding domain information, which improved the training efficiency.

\newpage

%% file: chapter/sec-3-crosslingual.tex
\chapter{Adaptation to Low-Resource Target Languages}\label{sec-crosslingual}
Adapting cross-lingual models to low-resource target languages where only a few or even zero training samples are available is a challenging yet essential task. 
There are two major challenges in constructing low-resource cross-lingual models. 
The first is how to effectively learn the low-resource representations. Addressing this allows us to build a more scalable and robust language transferable model. Current multilingual representations trained by self-supervised learning generally serve as a good initialization for cross-lingual models. However, the data scarcity issue and semantic vagueness across languages are still the primary obstacles for learning a good low-resource representations.
The second challenge is the inherent discrepancies between source and target languages. Such discrepancies (e.g., lexicons, language structures) can greatly decrease the quality of task knowledge transfer to low-resource target languages. 
In this chapter, we focus on improving the low-resource transfer ability of cross-lingual models and tackling the two abovementioned challenges. In the following sections, we will present two key contributions:
\begin{itemize}
    \item We propose a Gaussian noise injection method, a cross-lingual embedding refinement, a transferable latent variable model, and a label regularization method to improve the representation learning and the adaptation robustness for the low-resource languages. We find that the low-resource representations can be easily and greatly improved by focusing on just the task-related keywords.
    
    
    \item We study how word order differences across languages affect the cross-lingual performance, and present an Order-Reduced Transformer. We find that modeling partial word orders instead of the whole sequence can improve the robustness of the model against the word order differences between languages and the task knowledge transfer to low-resource languages.
    
\end{itemize}

\begin{figure}[t!]
\centering
\includegraphics[width=.6\linewidth]{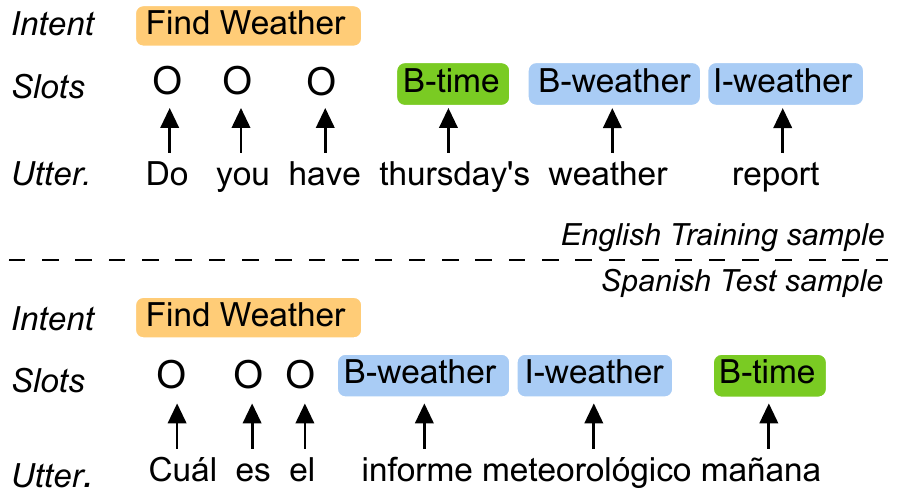}
\caption{Illustration of cross-lingual semantic parsing, where English is the source language and Spanish is the target language.}
\label{fig:semantic-parsing-example}
\end{figure}

\section{Cross-Lingual Parsing Framework}
\label{sec:cross-lingual-semantic-parsing}
In this section, we introduce our methods to enhance the quality of representations learning for low-resource languages. 
Our proposed methods aim to improve the low-resource representations from two perspectives.
First, we focus on improving the model's robustness against the imperfect alignments in cross-lingual representations. Second, we leverage regularization-based methods to further align cross-lingual representations.
We use the task-oriented semantic parsing as a case study to evaluate our proposed methods. Figure~\ref{fig:semantic-parsing-example} provides an example of this task, where the model needs to detect the intent and slot entities in the user utterance~\footnote{The slot tagging usually follows a BIO format (begin-inside-outside) to separate the entity and non-entity tokens. Specifically, there are \{``B-entity\_type'', ``I-entity\_type'', ``O''\}, and ``B'' denotes the beginning token of an entity, ``I'' denotes the inside token of an entity, and ``O'' denotes the non-entity type.}.
Our methods can also be potentially applied for other natural language understanding (NLU) tasks.

\subsection{Model Description}

\subsubsection{Improving Robustness Against Imperfect Alignments}
\begin{figure}[!t]
  \centering
  \includegraphics[width=.6\linewidth]{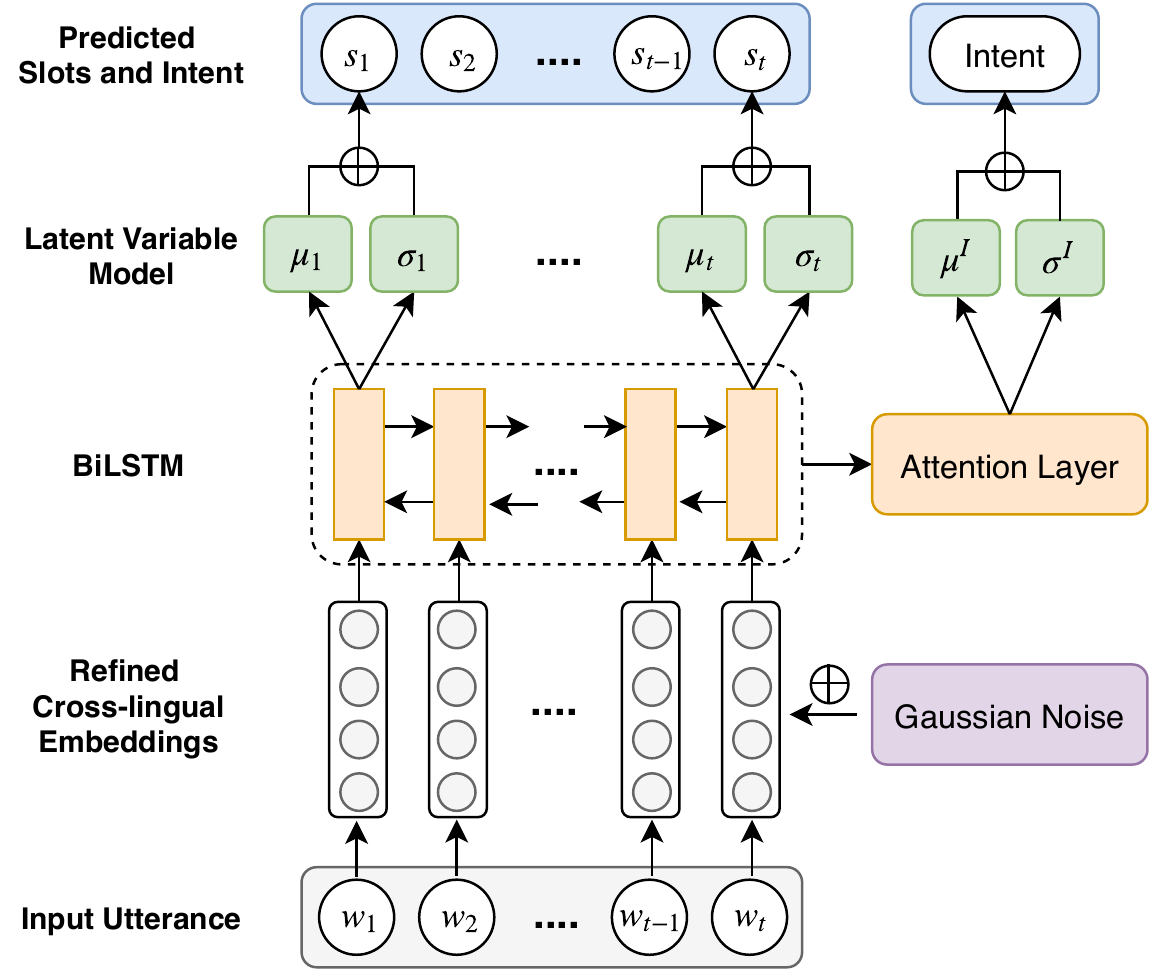}
  \caption{An illustration of the Gaussian noise injection and latent variable model using the refined cross-lingual embeddings.}
  \label{fig:robustness-based-model}
\end{figure}

We proposed three methods to enhance the model's robustness towards imperfect cross-lingual alignments: 1) cross-lingual embeddings refinement; 2) Gaussian noise injection; and 3) latent variable model. Our model architecture is illustrated in Figure~\ref{fig:robustness-based-model}.

There exists imperfect word- and sentence-level representation alignments across languages. To address such variance in the alignment, we turn to probabilistic modeling with latent variables as it has been successfully used in several recent task-oriented dialogues~\cite{wen2017latent,zhao2017learning,zhao2018unsupervised,zhao2018zero, le2018variational}.
However, we notice that naively using latent variables does not help the model improve much in slot filling and intent prediction. We hypothesize that the variance of the cross-lingual word embeddings is too large for the model to learn any meaningful latent variables. Hence, we propose to first refine the cross-lingual embeddings with a few seed word-pairs related to the dialogue domains. We then add Gaussian noise~\cite{zheng2016improving} to further compensate for the imperfect alignment of cross-lingual embeddings.
As a result, a combination of these methods allows us to build a transferable latent variable model that learns a distribution of training language inputs that is invariant to noise in the cross-lingual embeddings. This enables our model to capture the variance of semantically similar sentences across languages.

\begin{figure}[t!]
\centering
\includegraphics[width=.6\linewidth]{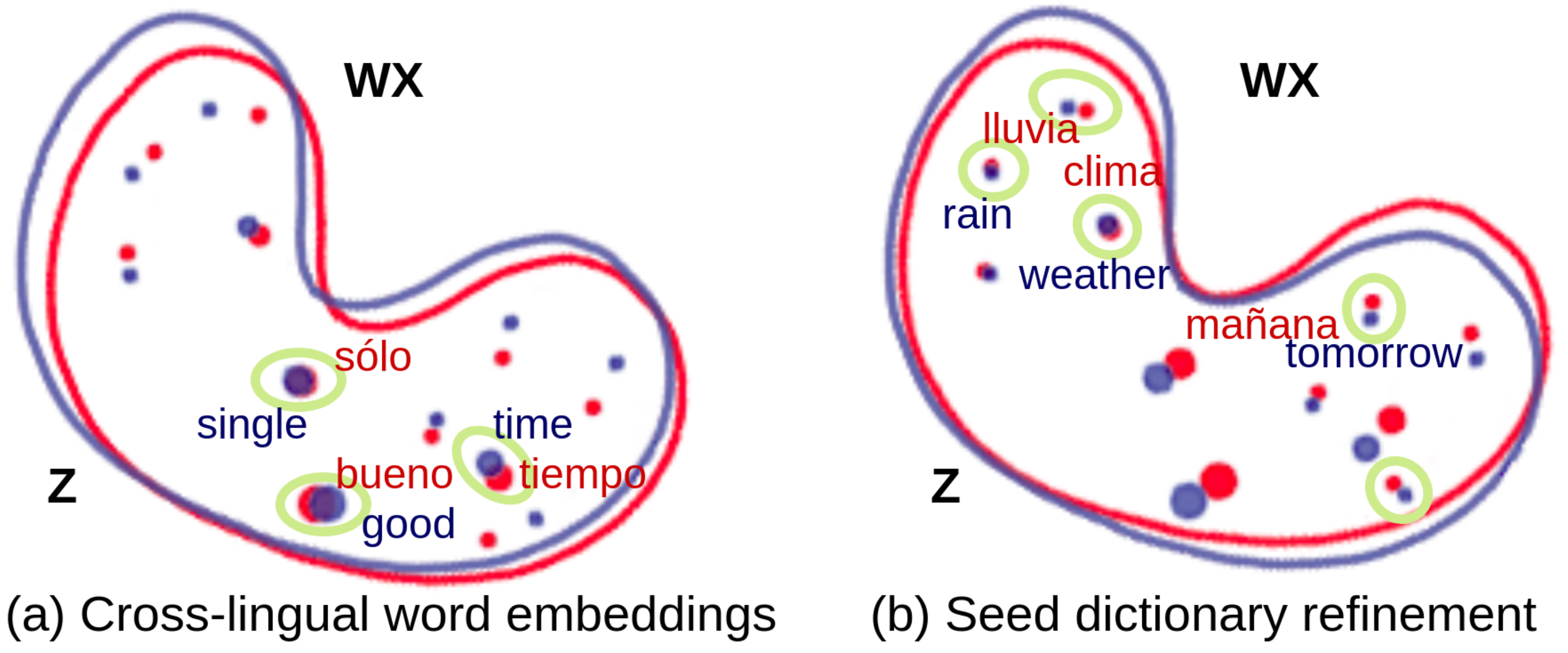}
\caption{(a) The aligned word embeddings. (b) Refined mapping $\mathbf{W}$ with seed dictionary; green circle represents close words in different languages.}
\label{fig:emb-refine}
\end{figure}

\paragraph{Cross-lingual Embedding Refinement}
To further refine the cross-lingual alignments to our task, we draw from the hypothesis that domain-related keywords are more important than others. Hence, as shown in Figure \ref{fig:emb-refine}, we propose to refine the cross-lingual word embeddings~\cite{joulin2018loss}~\footnote{The embeddings are available in https://fasttext.cc} using very few parallel word pairs, which is obtained by selecting 11 English words related to specific domains (weather, alarm, and reminder)~\footnote{We focus on these three domains since they are in the multilingual semantic parsing dataset~\cite{schuster2019cross} on which we evaluate our model. We can focus on other domains based on the task scenario that we are working on.} and translate them using bilingual lexicons. We refine the embeddings by leveraging the framework proposed in~\citet{artetxe2017learning}.

Let $\mathbf{X}$ and $\mathbf{Z}$ be the aligned cross-lingual word embeddings between two languages. $X_{i*}$ and $Z_{j*}$ are the embeddings for the $i^{th}$ source word and $j^{th}$ target word. We denote a binary dictionary matrix $\mathbf{D}$: $D_{ij}=1$ if the $i^{th}$ source language word is aligned with the $j^{th}$ target language word and $ D_{ij}=0 $ otherwise. The goal is to find the optimal mapping matrix $\mathbf{W}^{*}$ by minimizing:
\begin{align}
    \mathbf{W}^{*} = \arg\min_\mathbf{W} \sum_{i,j} D_{ij} ||X_{i*} \mathbf{W} - Z_{j*}||^2. 
    \label{eq:3-1}
\end{align}
Following \citet{artetxe2016learning}, with orthogonal constraints, mean centering, and length normalization, we can maximize the following instead:
\begin{align}
    \mathbf{W}^{*} = \arg\max_{\mathbf{W}} \text{Tr}(\mathbf{X}\mathbf{W} \mathbf{Z}^\text{T} \mathbf{D}^\text{T}).
    \label{eq:3-2}
\end{align}
We iteratively optimize Equation~\ref{eq:3-2} until distances between domain-related seed words are closer than a certain threshold after refinement. Figure~\ref{fig:emb-refine} illustrates better alignment for domain-related keywords (e.g., ``rain'', ``weather'') after refinement.

\paragraph{Gaussian Noise Injection}
To cope with the noise in alignments, we inject Gaussian noise to English embeddings, so the trained model will be more robust to variance. This is a regularization method to improve the generalization ability to the unseen inputs in different languages, particularly languages from different roots such as Thai and Spanish. The final embeddings are $\mathbf{e}^* = [e_1 + N_1, \dots, e_T + N_T]$, where $\mathbf{N} \sim \mathcal{N}(0, 0.1 \mathbf{I})$.

\paragraph{Latent Variable Model}
Given a near-perfect cross-lingual embedding, noise is still caused by the inherent discrepancies between source and target languages. This noise is amplified when combined with imperfect alignment, and makes point estimation vulnerable to the small, but not negligible, differences across languages.
Instead, using latent variables will allow us to model the distribution that captures the variance of semantically similar sentences across different languages. The whole training process is defined as follows:
\begin{equation}
    [h_1...h_t...h_T] = \text{BiLSTM}(\mathbf{e}^*),
\end{equation}
\begin{equation}
    m_t = h_t w_a,~ a_t = \frac{exp(m_t)}{\sum_{j=1}^T exp(m_j)},~ v = \sum_{t=1}^T a_t h_t,
\end{equation}
\begin{equation}
    \left[ \begin{array}{c} { \mu^S_t } \\ { \log \left (\sigma^S_t) ^ { 2 } \right. } \end{array} \right] = \mathbf{W}^S_r h_t, \left[ \begin{array} { c } { \mu^I } \\ { \log \left (\sigma^I) ^ { 2 } \right. } \end{array} \right] = \mathbf{W}^I_{r} v,
\end{equation}
\begin{equation}
    z^S_t \sim q^S_t (z | h_t), ~~ z^I \sim q^I (z | v),
\end{equation}
\begin{equation}
    p^S_t (s_t | z^S_t) = \text{Softmax}(\mathbf{W}^S_g z^S_t),
\end{equation}
\begin{equation}
    p^I (I | z^I) = \text{Softmax}(\mathbf{W}^I_g z^I),
\end{equation}
where attention vector ($v$) is obtained by following~\citet{felbo2017using}, $w_a$ is the weight matrix for the attention layer, $\mathbf{W}_{\{r,g\}}^{\{S,I\}}$ are trainable parameters, superscripts S and I refer to slot prediction and intent detection respectively, subscript ``r'' refers to ``recognition'' for obtaining the mean and variance vectors, while subscript ``g'' refers to ``generation'' for predicting the slots and intents, $q^S_t \sim \mathcal{N}(\mu^S_t, (\sigma^S_t)^2\mathbf{I})$ and $q^I \sim \mathcal{N}(\mu^I, (\sigma^I)^2\mathbf{I})$ are the posterior approximations from which we sample our latent vectors $z^S_t$ and $z^I$, and finally, $p^S_t$ and $p^I$ are the predictions for the slot of the \textit{t-th} token and the intent of the utterance respectively. The objective functions for slot filling and intent prediction are:
\begin{equation}
    \mathcal{L}^I = \mathbb{E}_{z^I} [\log p^I (I|z^I)],
\end{equation}
\begin{equation}
    \mathcal{L}^S_t = \mathbb{E}_{z^S_t} [\log p^S_t (s_t|z^S_t)], ~~~ \mathcal{L}^S = \sum_{t=1}^T L_t^S.
\end{equation}
Hence, the final objective function to be minimized is,
\begin{equation}
    \mathcal{L} = \mathcal{L}^S + \mathcal{L}^I.
\end{equation}
The model prediction is not deterministic since the latent variables $z_t^S$ and $z^I$ are sampled from the Gaussian distributions. Therefore, in the inference time, we use the true mean $\mu_t^S$ and $\mu^I$ to replace $z_t^S$ and $z^I$, respectively, to make the prediction deterministic.

\subsubsection{Improving Alignments Using Regularization-Based Methods}
\begin{figure}[t!]
\centering
\includegraphics[width=.9\linewidth]{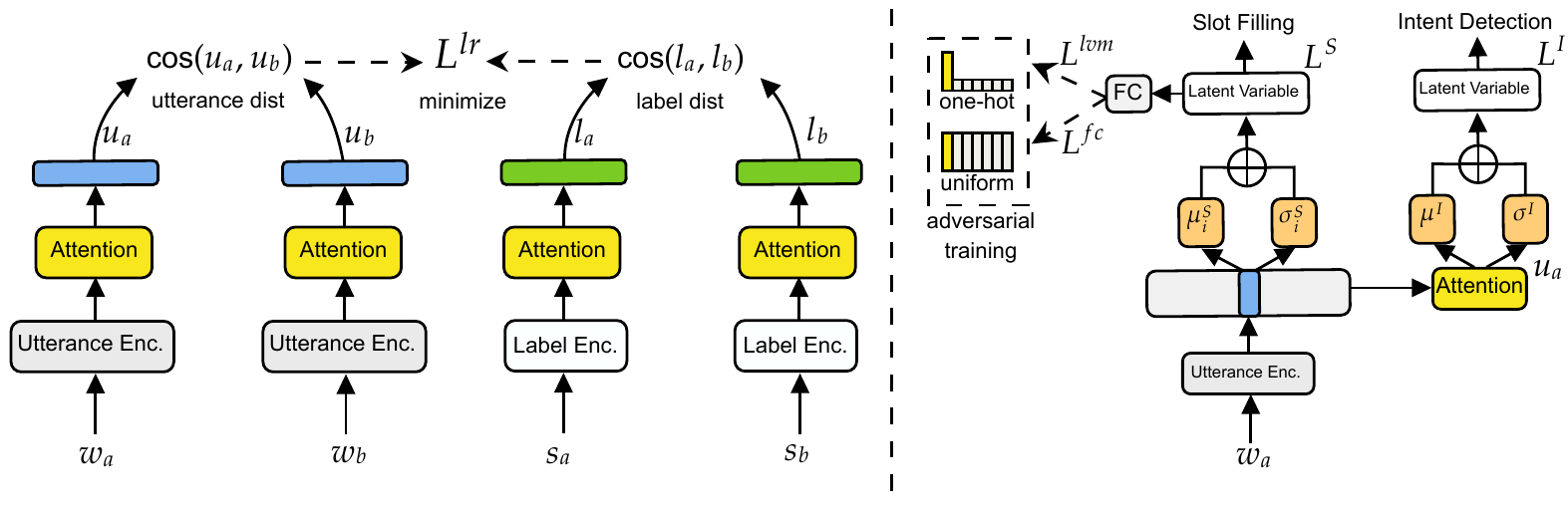}
\caption{Left: Illustration of label regularization (LR). Right: The model architecture with adversarial latent variable model (ALVM), where $FC$ consists of a linear layer and a softmax function.}
\label{fig:regularization-based-model}
\end{figure}

We propose two regularization-based methods to improve the cross-lingual alignment: 1) label regularization; and 2) adversarial latent variable model. Figure~\ref{fig:regularization-based-model} provides an illustration of these two methods.

To improve the quality of cross-lingual alignment, we first propose a label regularization (LR) method, which utilizes the label sequences to regularize the utterance representations. 
We hypothesize that if the label sequences of user utterances are close to each other, these user utterances should have similar meanings. Hence, we regularize the distance of utterance representations based on the corresponding representations of label sequences to further improve the cross-lingual alignments.

Our proposed latent variable model (LVM) generates a Gaussian distribution instead of a feature vector for each token, which improves the adaptation robustness. However, there are no additional constraints on generating distributions, making the latent variables easily entangled for different slot labels. To handle this issue, we leverage adversarial training to regularize the LVM (ALVM). We train a linear layer to fit latent variables to a uniform distribution over slot types. Then, we optimize the latent variables to fool the trained linear layer to output the correct slot type (one-hot vector). In this way, latent variables of different slot types are encouraged to disentangle from each other, leading to a better alignment of cross-lingual representations.

\paragraph{Label Regularization}
Figure~\ref{fig:regularization-based-model} (Left) illustrates an utterance encoder and label encoder, which generate the representations for utterances and labels, respectively. We denote the user utterance as $ \textbf{w} = [w_1, w_2, ..., w_{n}] $, where $n$ is the length of the utterance. Similarly, we represent the slot label sequences as $ \textbf{s} = [s_1, s_2, ..., s_{n}] $. We combine a bidirectional LSTM (BiLSTM)~\cite{hochreiter1997long} and an attention layer~\cite{felbo2017using} to encode and produce the representations for user utterances and slot label sequences. The representation generation process is defined as follows:
\begin{equation}
    [h_1^{w}, h_2^{w}, ..., h_{w}^{w}] = \textnormal{BiLSTM}_{\textnormal{utter}} (\textbf{E}(\textbf{w})),
\end{equation}
\begin{equation}
    [h_1^{s}, h_2^{s}, ..., h_{n}^{s}] = \textnormal{BiLSTM}_{\textnormal{label}} (\textbf{E}(\textbf{s})),
\end{equation}
\begin{equation}
    m_i^{w} = h_i^{w} v^{w},~~ \alpha_i^{w} = \frac{exp(m_i^{w})}{\sum_{t=1}^{n}exp(m_t^{w})},
\end{equation}
\begin{equation}
    m_i^{s} = h_i^{s} v^{s},~~ \alpha_i^{s} = \frac{exp(m_i^{s})}{\sum_{t=1}^{n}exp(m_t^{s})},
\end{equation}
\begin{equation}
    u = \sum_{i=1}^{n} \alpha_i^{w} h_i^{w},~~ l = \sum_{i=1}^{n} \alpha_i^{s} h_i^{s},
\end{equation}
where the superscripts $w$ and $s$ represent utterance and label, respectively, $v$ is a trainable weight vector in the attention layer, $\alpha_i$ is the attention score for each token i, \textbf{E} denotes the embedding layers for utterances and label sequences, and $u$ and $l$ denote the representation of utterance $\textbf{w}$ and slot label $\textbf{s}$, respectively.

In each iteration of the training phase, we randomly select two samples for the label regularization. As illustrated in Figure~\ref{fig:regularization-based-model} (Left),
we first calculate the cosine similarity of two utterance representations $u_a$ and $u_b$, and the cosine similarity of two label representations $l_a$ and $l_b$. Then, we minimize the distance of these two cosine similarities. The objective functions can be described as follows:
\begin{equation}
    \textnormal{cos}(u_a,u_b) = \frac{u_{a} \cdot u_{b}}{||u_{a}||~||u_{b}||},
\end{equation}
\begin{equation}
    \textnormal{cos}(l_a,l_b) = \frac{l_{a} \cdot l_{b}}{||l_{a}||~||l_{b}||},
\end{equation}
\begin{equation}
    L^{lr} = \sum_{a,b} \textnormal{MSE}(\textnormal{cos}(u_a,u_b), \textnormal{cos}(l_a,l_b)), \label{eq:3-19}
\end{equation}
where the superscript $lr$ denotes label regularization, and \textnormal{MSE} represents mean square error. In the zero-shot setting, both samples $u_a$ and $u_b$ come from the source language, while in the few-shot setting, one sample comes from the source language and the other comes from the target language.

Since the features of labels and utterances are in different vector spaces, we choose not to share the parameters of their encoders. During training, it is easy to produce expressive representations for user utterances due to the large number of training samples, but this is difficult for label sequences since the objective function $L^{lr}$ is the only supervision. This supervision is weak at the beginning of the training since utterance representations are not sufficiently expressive, which leads to the label regularization approach being unstable and ineffective. To ensure the representations for slot label sequences are meaningful, we conduct pre-training for the label sequence encoder.

We leverage the large amount of source language training data to pre-train the label sequence encoder. Concretely, we use the model architecture illustrated in Figure~\ref{fig:regularization-based-model} to train the parsing system in the source language, and at the same time, we optimize the label sequence encoder based on the objective function $L^{lr}$ in Equation~\ref{eq:3-19}. The label sequence encoder learns to generate meaningful label sequence representations since the extensive source language training samples ensure the high quality of the utterance encoder. 

\paragraph{Adversarial Latent Variable Model}
Since no constraints are enforced on the latent Gaussian distribution during training, the latent distributions of different slot types are likely to be close to each other. Hence, the distributions for the same slot type in different user utterances or languages might not be clustered well, which could hurt the cross-lingual alignment and prevent the model from distinguishing slot types when adapting to the target language.

To improve the cross-lingual alignment of latent variables, we propose to make the latent variables of different slot types more distinguishable by adding adversarial training to the LVM. As illustrated in Figure~\ref{fig:regularization-based-model} (Right), we train a fully connected layer to fit latent variables into a uniform distribution over slot types. At the same time, the latent variables are regularized to fool the trained fully connected layer by predicting the correct slot type. In this way, the latent variables are trained to be more recognizable. In other words, the generated distributions for different slot types are more likely to repel each other, and for the same slot type are more likely to be close to each other, which leads to a more robust cross-lingual adaptation.
We denote the size of the whole training data as $J$ and the length of data sample $j$ as $|Y_j|$. Note that in the few-shot setting, $J$ includes the number of data samples in the target language.
The process of adversarial training is as follows:
\begin{equation}
    p_{jk} = \mathcal{FC}(z_{jk}^S),
\end{equation}
\begin{equation}
    L^{fc} = \sum_{j=1}^J \sum_{k=1}^{|Y_j|} \textnormal{MSE}(p_{jk}, \mathcal{U}),
\end{equation}
\begin{equation}
    L^{lvm} = \sum_{j=1}^J \sum_{k=1}^{|Y_j|} \textnormal{MSE}(p_{jk}, y_{jk}^{S}),
\end{equation}
where $\mathcal{FC}$ consists of a linear layer and a \textnormal{Softmax} function, and $z_{jk}^{S}$ and $p_{jk}$ are the latent variable and generated distribution, respectively, for the $k^{th}$ token in the $j^{th}$ utterance, 
$\textnormal{MSE}$ represents the mean square error,  $\mathcal{U}$ represents the uniform distribution, and $y_{jk}^{S}$ represents the slot label. The slot label is a one-hot vector where the value for the correct slot type is one and zero otherwise.
We optimize $L^{fc}$ to train only $\mathcal{FC}$ to fit a uniform distribution, and $L^{lvm}$ is optimized to constrain the LVM to generate more distinguishable distributions for slot predictions. Different from the well-known adversarial training~\cite{goodfellow2014generative} where the discriminator is used to distinguish the classes, and the generator is to make the features indistinguishable, in our approach, the $\mathcal{FC}$ layer, acting as the discriminator, is trained to generate a uniform distribution, and the generator is regularized to make latent variables distinguishable by slot types.
The objective functions for the slot filling and intent detection tasks are illustrated as follows:
\begin{equation}
    L^S = \sum_{j=1}^{J} \sum_{k=1}^{|Y_j|} -log(p_{jk}^{S} \cdot (y_{jk}^{S})^\top),
\end{equation}
\begin{equation}
    L^I = \sum_{j=1}^{J} -log(p_{j}^{I} \cdot (y_{j}^I)^\top),
\end{equation}
where $p_{jk}^S$ and $y_{jk}^S$ are the prediction and label, respectively, for the slot of the $k^{th}$ token in the $j^{th}$ utterance, and $p_{j}^{I}$ and $y_{j}^{I}$ are the intent prediction and label, respectively, for the $j^{th}$ utterance.
The optimization for our model is to minimize the following objective function:
\begin{equation}
    L = L^S + L^I + L^{lr} + \alpha L^{fc} + \beta L^{lvm}, \label{eq:3-25}
\end{equation}
where $\alpha$ and $\beta$ are hyper-parameters, $L^{fc}$ only optimizes the parameters in $\mathcal{FC}$, and $L^{lvm}$ optimizes all the model parameters, excluding $\mathcal{FC}$.

\begin{table}[]
\setlength{\tabcolsep}{20pt}
\centering
\resizebox{0.6\textwidth}{!}{
\begin{tabular}{lccc}
\toprule
\textbf{\# Utterance} & \textbf{English}  & \textbf{Spanish}  & \textbf{Thai}  \\ \midrule
Train & 30,521 & 3,617 & 2,156 \\
Validation & 4,181 & 1,983 & 1,235 \\
Test & 8,621 & 3,043 & 1,692 \\ \bottomrule
\end{tabular}
}
\caption{Number of utterances for the multilingual semantic parsing dataset. English is the source language, and Spanish and Thai are the target languages.}
\label{tab:semantic-parsing-data-statistics}
\end{table}

\subsection{Experimental Setup}

\subsubsection{Dataset}
We conduct our experiments on the multilingual semantic parsing dataset~\cite{schuster2019cross}. It contains English, Spanish, and Thai across the weather, reminder, and alarm domains. The dataset includes 12 intent types and 11 slot types, and the data statistics are shown in Table~\ref{tab:semantic-parsing-data-statistics}. We consider English as the source language, and Spanish and Thai are the target languages.

\subsubsection{Training}
We refine the cross-lingual word embeddings proposed in~\citet{joulin2018loss} and freeze them in the training process. We use bi-directional LSTM (BiLSTM) model with hidden dimension size of 250 with a dropout rate of 0.1, and the LVM with both mean and variance in the size of 100. Gaussian noise with zero mean and variance of 0.1 is injected dynamically in different iterations. The label encoder is a 1-layer BiLSTM with a hidden size of 150, and 100-dimensional embeddings for label types. We use the Adam optimizer with a learning rate of 0.001, and a delexicalization (delex.) technique which replaces the tokens that represent numbers, time, and duration with special tokens. For the adversarial training, we realize that the latent variable model is not able to make slot types recognizable if the $\mathcal{FC}$ is too strong. Hence, we decide to first learn a good initialization for $\mathcal{FC}$ by setting both $\alpha$ and $\beta$ parameters in Equation~\ref{eq:3-25} as 1 in the first two training epochs, and then we gradually decrease the value of $\alpha$. We use the accuracy to evaluate the performance of intent prediction and the standard BIO structure to calculate the F1 score for evaluating the performance of slot filling.
We conduct experiments in zero-shot and few-shot settings. In the zero-shot setting, we do not use any training samples in the target language, while in the few-shot setting, we use 1\% and 3\% training samples from the target languages. There are \textbf{36} training samples in Spanish and \textbf{21} training samples in Thai on the 1\% few-shot setting, and \textbf{108} training samples in Spanish and \textbf{64} training samples in Thai on the 3\% few-shot setting. 
In terms of word pairs used for the cross-lingual embeddings refinement, we select \textit{weather, forecast, temperature, rain, hot, cold, remind, forget, alarm, cancel, tomorrow}, which are related to the three dialogue domains (weather, alarm, and reminder). We translate them by leveraging bilingual dictionaries\footnote{https://github.com/facebookresearch/MUSE}. 
The corresponding translations in Spanish and Thai are \textit{clima, pronóstico, temperatura, lluvia, caliente, frío, recordar, olvidar, alarma, cancelar, mañana} and 
\foreignlanguage{thaicjk}{\textit{อากาศ, พยากรณ์, อุณหภูมิ, ฝน, ร้อน, หนาว, เตือน, ลืม, เตือน, ยกเลิก, พรุ่ง }} respectively.

\subsubsection{Baselines}
We compare our model to the following baselines:
\begin{itemize}
    \item \textbf{BiLSTM-CRF}~~ This is the same cross-lingual model structure as in~\citet{schuster2019cross}. It uses the same BiLSTM encoder as our model, and it adds a conditional random field (CRF) on top of the BiLSTM for the sequence labeling.
    \item \textbf{BiLSTM-LVM}~~ We replace the CRF layer in BiLSTM-CRF with our proposed LVM.
    \item \textbf{Zero-shot Parsing}~~ \citet{upadhyay2018almost} used cross-lingual embeddings \cite{bojanowski2017enriching} to conduct zero-shot transfer learning. We implement their proposed model and evaluate on the multilingual semantic parsing dataset~\cite{schuster2019cross}.
    \item \textbf{Multi. CoVe}~~ Multilingual CoVe~\cite{yu2018multilingual} is a bidirectional machine translation system that tends to encode phrases with similar meanings into similar vector spaces across languages. \citet{schuster2019cross} used it for the cross-lingual semantic parsing task.
    \item \textbf{Multi. CoVe w/ auto-encoder}~~ Based on Multilingual CoVe, \citet{schuster2019cross} added an auto-encoder objective so as to produce better-aligned representations for semantically similar sentences across languages.
    \item \textbf{Multilingual BERT (M-BERT)}~~ M-BERT is a single language model pre-trained from monolingual corpora in 104 languages~\cite{devlin2019bert}, and is surprisingly good at cross-lingual model transfer.
    \item \textbf{Translate Train}~~ \citet{schuster2019cross} trained a supervised machine translation system to translate English data into the target language, and then trained the model on the translated dataset.
    \item \textbf{All-shot Settings}~~ We train the BiLSTM-CRF model~\cite{schuster2019cross} on all the target language training samples, and on both the source and target language training datasets.
\end{itemize}

\subsection{Results and Discussion}

\begin{table}[t!]
\centering
\setlength{\tabcolsep}{18pt}
\resizebox{0.99\textwidth}{!}{
\begin{tabular}{lcccccccc}
\toprule
\multicolumn{1}{c}{}     & \multicolumn{4}{c}{\textbf{Spanish}}  & \multicolumn{4}{c}{\textbf{Thai}}      \\ \cmidrule(lr){2-5} \cmidrule(lr){6-9}
\multicolumn{1}{c}{\multirow{1}{*}{\textbf{Model}}} & \multicolumn{2}{c}{\textbf{Intent acc.}} & \multicolumn{2}{c}{\textbf{Slot F1}} & \multicolumn{2}{c}{\textbf{Intent acc.}}   & \multicolumn{2}{c}{\textbf{Slot F1}} \\  \cmidrule(lr){2-3} \cmidrule(lr){4-5} \cmidrule(lr){6-7} \cmidrule(lr){8-9}
\multicolumn{1}{c}{}                       & \textbf{LVM}         & \textbf{CRF}       & \textbf{LVM}       & \textbf{CRF}     & \textbf{LVM}            & \multicolumn{1}{c}{\textbf{CRF}}   & \textbf{LVM}  & \textbf{CRF} \\ \midrule
Vanilla BiLSTM    & 46.36    & 44.13   & 15.64    & 11.32    & 35.12  & \multicolumn{1}{c}{33.57} & 5.82                  & 5.24         \\
\quad+ \textit{noise (N)}   & 72.97  & 66.95  & 46.56  & 20.27  & 40.37          & \multicolumn{1}{c}{37.53} & 10.66    & 6.51         \\
\quad+ \textit{refinement (R)}  & 87.69    & 88.23    & 61.63  & 42.62            & 59.40   & \multicolumn{1}{c}{59.28} & 21.84  & 16.53        \\
\quad+ \textit{noise + refinement}  & 89.21  & 88.79 & 64.04  & 43.98            & 70.81 & \multicolumn{1}{c}{64.48} & 29.54  & 17.46         \\
\quad+ \textit{N + R + delexicalization}      & \textbf{90.20}       & 89.98              & \textbf{65.79}     & 47.70            & \textbf{73.43} & \multicolumn{1}{c}{69.62} & \textbf{32.24}        & 23.11  \\ \midrule 
Zero-shot Parsing                              & \multicolumn{2}{c}{46.64}                & \multicolumn{2}{c}{15.41}            & \multicolumn{2}{c}{35.64}                  & \multicolumn{2}{c}{12.11}            \\
Multi. CoVe w/ auto                            & \multicolumn{2}{c}{53.89}                & \multicolumn{2}{c}{19.25}            & \multicolumn{2}{c}{70.70}                  & \multicolumn{2}{c}{35.62}           \\
Translate Train $^\dagger$        & \multicolumn{2}{c}{\textit{85.39}}   & \multicolumn{2}{c}{\textit{72.87}}  & \multicolumn{2}{c}{\textit{95.85}} & \multicolumn{2}{c}{\textit{55.43}} \\
\bottomrule
\end{tabular}
}
\caption{Zero-shot cross-lingual results, where N refers to the Gaussian noise injection and R refers to the cross-lingual embeddings refinement. $^\dagger$ It is considered as our \textit{upper bound} result due to the usage of machine translation systems.}
\label{tab:robustness-based-alignment-results}
\end{table}

\subsubsection{Improving Robustness Against Imperfect Alignments}
From Table~\ref{tab:robustness-based-alignment-results}, in general, LVM outperforms CRF models. This is because for semantically similar words (e.g., weather and clima), the LVM considers close enough points as being the same distribution, but the CRF is more likely to classify them differently. In addition, we can see that adding only Gaussian noise to the Vanilla BiLSTM improves our prediction performance significantly, which implies the robustness of our model towards the noisy signals which come from the target embedding inputs.

Furthermore, it is clearly visible that cross-lingual embeddings refinement significantly improves the cross-lingual performance, and it is much more effective than Gaussian noise injection and the LVM methods. This shows that low-resource representations can be easily and greatly improved by focusing on just the task-related keywords. 
Interestingly, cross-lingual embedding refinement works better for Spanish than Thai. This is attributed to the quality of alignments in the two languages. Spanish is much more lexically and grammatically similar to English than Thai, so word-level embedding refinement is reasonably good. Jointly incorporating all three methods (Gaussian noise injection, cross-lingual embeddings refinement, and delexicalization) further reduces the noise in the inputs as well as makes the model more robust to noise, which help LVM to more easily approximate the distribution.

\begin{table}[!t]
\centering
\setlength{\tabcolsep}{20pt}
\resizebox{0.6\textwidth}{!}{
\begin{tabular}{lcccc}
\toprule
\multicolumn{1}{c}{\multirow{2}{*}{Model}} & \multicolumn{2}{c}{Spanish}    & \multicolumn{2}{c}{Thai}        \\ \cmidrule(lr){2-3} \cmidrule(lr){4-5}
\multicolumn{1}{c}{}                       & Intent         & Slot           & Intent         & Slot           \\ \midrule
Our model    & \textbf{90.20} & \textbf{65.79} & \textbf{73.43} & \textbf{32.24} \\
\textit{- LVM}  & 85.85   & 61.86   & 66.01    & 25.22          \\
\textit{- LVM + MLP}  & 86.02   & 62.34  & 66.56   & 28.35   \\
\bottomrule
\end{tabular}
}
\caption{Ablation study on LVM. \textit{\{- LVM\}} means removing LVM and \textit{\{- LVM + MLP\}} replaces LVM with a Multi-Layer Perceptron that is of the same size as the LVM.}
\label{tab:lvm-ablation-study-results}
\end{table}

Finally, in Table~\ref{tab:lvm-ablation-study-results}, we ablate the usage of the LVM to see whether the boost of performance comes simply from the increase in the parameter size. By removing or replacing the LVM with the MLP, we can see the clear performance gains by using the LVM.

\begin{table}[t!]
\centering
\setlength{\tabcolsep}{15pt}
\renewcommand{\arraystretch}{1.1}
\resizebox{0.99\textwidth}{!}{
\begin{tabular}{lcccccccc}
\toprule
\multicolumn{1}{l}{\multirow{2}{*}{\textbf{Model}}} & \multicolumn{4}{c}{\textbf{Spanish}} & \multicolumn{4}{c}{\textbf{Thai}} \\ \cmidrule(lr){2-5} \cmidrule(lr){6-9}
\multicolumn{1}{l}{} & \multicolumn{2}{c}{\textbf{Intent Acc.}} & \multicolumn{2}{c}{\textbf{Slot F1}} & \multicolumn{2}{c}{\textbf{Intent Acc.}} & \multicolumn{2}{c}{\textbf{Slot F1}} \\ \bottomrule
\multicolumn{9}{l}{\textit{\textbf{Few-shot settings}}} \\ \midrule
\multicolumn{1}{l}{} & $\tt{1\%}$-$\tt{shot}$ & \multicolumn{1}{c}{$\tt{3\%}$-$\tt{shot}$} & $\tt{1\%}$-$\tt{shot}$ & \multicolumn{1}{c}{$\tt{3\%}$-$\tt{shot}$} & $\tt{1\%}$-$\tt{shot}$ & \multicolumn{1}{c}{$\tt{3\%}$-$\tt{shot}$} & $\tt{1\%}$-$\tt{shot}$ & $\tt{3\%}$-$\tt{shot}$ \\ \midrule
\multicolumn{1}{l}{BiLSTM-CRF} & 93.03 & \multicolumn{1}{c}{93.63} & 75.70 & \multicolumn{1}{c}{82.60} & 81.30 & \multicolumn{1}{c}{87.23} & 52.57 & 66.04 \\
\multicolumn{1}{l}{\hspace{1mm}+ LR} & 93.08 & \multicolumn{1}{c}{95.04} & 77.04 & \multicolumn{1}{c}{84.09} & 84.04 & \multicolumn{1}{c}{89.20} & 57.40 & 67.45 \\  \midrule
\multicolumn{1}{l}{BiLSTM-LVM} & 92.86 & \multicolumn{1}{c}{94.46} & 75.19 & \multicolumn{1}{c}{82.64} & 83.51 & \multicolumn{1}{c}{89.08} & 55.08 & 67.26 \\
\multicolumn{1}{l}{\hspace{1mm}+ LR} & 93.79 & \multicolumn{1}{c}{95.16} & 76.96 & \multicolumn{1}{c}{83.54} & 86.33 & \multicolumn{1}{c}{90.80} & 59.02 & 70.26 \\
\multicolumn{1}{l}{\hspace{1mm}+ ALVM} & 93.78 & \multicolumn{1}{c}{95.27} & 78.35 & \multicolumn{1}{c}{83.69} & 85.40 & \multicolumn{1}{c}{90.70} & 59.75 & 69.38 \\
\multicolumn{1}{l}{\hspace{1mm}+ LR \& ALVM} & 93.82 & \multicolumn{1}{c}{95.20} & 78.46 & \multicolumn{1}{c}{84.19} & 87.43 & \multicolumn{1}{c}{90.96} & 61.44 & 70.88 \\
\multicolumn{1}{l}{\hspace{1mm}+ LR \& ALVM \& delex.} & \textbf{94.71} & \multicolumn{1}{c}{\underline{\textbf{95.62}}} & \textbf{80.82} & \multicolumn{1}{c}{\underline{\textbf{85.18}}} & \textbf{87.67} & \multicolumn{1}{c}{\underline{\textbf{91.61}}} & \textbf{62.01} & \textbf{72.39}  \\ \midrule
\multicolumn{1}{l}{XL-Parsing} & 92.70 & \multicolumn{1}{c}{94.96} & 77.67 & \multicolumn{1}{c}{82.22} & 84.04 & \multicolumn{1}{c}{89.59} & 55.57 & 67.56 \\
\multicolumn{1}{l}{M-BERT} & 92.77 & \multicolumn{1}{c}{95.56} & 80.15 & \multicolumn{1}{c}{84.50} & 83.87 & \multicolumn{1}{c}{89.19} & 58.18 & 67.88 \\  \bottomrule
\multicolumn{9}{l}{\textit{\textbf{Zero-shot settings}}} \\ \midrule
\multicolumn{1}{l}{XL-Parsing} & \multicolumn{2}{c}{90.20} & \multicolumn{2}{c}{65.79} & \multicolumn{2}{c}{73.43} & \multicolumn{2}{c}{32.24} \\
\multicolumn{1}{l}{\hspace{1mm}+ LR} & \multicolumn{2}{c}{91.51} & \multicolumn{2}{c}{71.55} & \multicolumn{2}{c}{74.86} & \multicolumn{2}{c}{32.86} \\
\multicolumn{1}{l}{\hspace{1mm}+ ALVM} & \multicolumn{2}{c}{91.48} & \multicolumn{2}{c}{71.21} & \multicolumn{2}{c}{74.35} & \multicolumn{2}{c}{32.97} \\
\multicolumn{1}{l}{\hspace{1mm}+ LR \& ALVM} & \multicolumn{2}{c}{\textbf{92.31}} & \multicolumn{2}{c}{\textbf{72.49}} & \multicolumn{2}{c}{\textbf{75.77}} & \multicolumn{2}{c}{\textbf{33.28}} \\ \midrule
\multicolumn{1}{l}{M-BERT} & \multicolumn{2}{c}{74.91} & \multicolumn{2}{c}{67.55} & \multicolumn{2}{c}{42.97} & \multicolumn{2}{c}{10.68} \\
\multicolumn{1}{l}{Multi. CoVe} & \multicolumn{2}{c}{53.34} & \multicolumn{2}{c}{22.50} & \multicolumn{2}{c}{66.35} & \multicolumn{2}{c}{32.52} \\
\multicolumn{1}{l}{\hspace{1mm}+ Auto-encoder} & \multicolumn{2}{c}{53.89} & \multicolumn{2}{c}{19.25} & \multicolumn{2}{c}{70.70} & \multicolumn{2}{c}{35.62} \\ \midrule 
\multicolumn{1}{l}{Translate Train} & \multicolumn{2}{c}{\textit{85.39}} & \multicolumn{2}{c}{\textit{72.89}} & \multicolumn{2}{c}{\textit{95.89}} & \multicolumn{2}{c}{\textit{55.43}} \\ \bottomrule
\multicolumn{9}{l}{\textit{\textbf{All-shot settings}}} \\ \midrule
\multicolumn{1}{l}{Target$^\dagger$} & \multicolumn{2}{c}{\textit{96.08}} & \multicolumn{2}{c}{\textit{86.03}} & \multicolumn{2}{c}{\textit{92.73}} & \multicolumn{2}{c}{\textit{85.52}} \\
\multicolumn{1}{l}{Source \& Target$^\ddagger$} & \multicolumn{2}{c}{\textit{98.06}} & \multicolumn{2}{c}{\textit{87.65}} & \multicolumn{2}{c}{\textit{95.58}} & \multicolumn{2}{c}{\textit{88.11}} \\ \bottomrule
\end{tabular}
}
\caption{Zero-shot and few-shot cross-lingual semantic parsing results (averaged over three runs). $^\dagger$denotes supervised training on all the target language training samples. $^\ddagger$denotes supervised training on both the source and target language datasets. XL-Parsing denotes our best model to improve robustness against the imperfect alignments.}
\label{tab:regularization-based-alignment-results}
\end{table}

\subsubsection{Improving Alignments Using Regularization-Based Methods}

The results of our regularization-based alignments are shown in Table~\ref{tab:regularization-based-alignment-results}. The bold numbers denote the best results in the few-shot or zero-shot settings, and the underlined numbers represent that the results are comparable (distances are $\sim$1\%) to the all-shot experiment with all the target language training samples. We comprehensively analyse our proposed methods from the following aspects.

\paragraph{Quantitative Analysis}
We can clearly see consistent improvements made by label regularization and adversarial training. For example, on the 1\% few-shot setting, our model improves on BiLSTM-LVM in terms of accuracy/f1-score by 1.85\%/1.16\% in Spanish, and by 4.16\%/6.93\% in Thai. 
Our model also surpasses a strong baseline, M-BERT, while our model has many fewer parameters compared to M-BERT. For example, on the 1\% few-shot setting, our model improves on M-BERT in terms of accuracy/f1-score by 3.80\%/3.83\% in Thai.
Instead of generating a feature point like CRF, the LVM creates a more robust cross-lingual adaptation by generating a distribution for the intent or each token in the utterance. However, distributions generated by the LVM for the same slot type across languages might not be sufficiently close.
Incorporating adversarial training into the LVM alleviates this problem by regularizing the latent variables and making them more distinguishable. This improves the performance in both intent detection (a sentence-level task) and slot filling (a word-level task) by 0.92\%/3.16\% in Spanish and by 1.89\%/4.67\% in Thai on the 1\% few-shot setting. This proves that both sentence-level and word-level representations are better aligned across languages.

In addition, LR aims to further align the sentence-level representations of target language utterances into a semantically similar space of source language utterances. As a result, there is a 0.93\%/2.82\% improvement in intent detection for Spanish/Thai on the 1\% few-shot setting after we add LR to BiLSTM-LVM. Interestingly, the performance gains are not only on the intent detection but also on the slot filling, with an improvement of 1.77\%/3.94\% in Spanish/Thai.
This is attributed to the fact that utterance representations are produced based on word-level representations from BiLSTM. Therefore, the alignment of word-level representations will be implicitly improved in this process. Furthermore, incorporating LR and ALVM further tackles the inherent difficulties for the cross-lingual adaptation and achieves the state-of-the-art few-shot performance. Notably, by only leveraging 3\% of target language training samples, the results of our best model are on par with the supervised training on all the target language training data.

We observe the remarkable improvements made by LR and ALVM on the our previous model XL-Parsing in the zero-shot setting, and we note that the slot filling performance of our best model in Spanish is on par with the strong baseline Translate Train, which leverages large amounts of bilingual resources. LR improves the adaptation robustness by making the word-level and sentence-level representations of similar utterances distinguishable. In addition, integrating adversarial training with the LVM further increases the robustness by disentangling the latent variables for different slot types. However, the performance boost for slot filling in Thai is limited. We conjecture that the inherent discrepancies in cross-lingual word embeddings and language structures for topologically different languages pairs make the word-level representations between them difficult to align in the zero-shot scenario. We notice that Multilingual CoVe with auto-encoder achieves slightly better performance than our model on the slot filling task in Thai. This is because this baseline leverages large amounts of monolingual and bilingual resources, which greatly benefits the cross-lingual alignment between English and Thai.

\paragraph{Adaptation Ability to Topologically Distant Languages}
Moreover, we observe impressive improvements in Thai, an topologically distant language to English, by utilizing our proposed approaches, especially when the number of target language training samples is small. For example, compared to the BiLSTM-LVM, our best model significantly improves the accuracy by $\sim$4\% and F1-score by $\sim$7\% in intent detection and slot filling, respectively in Thai in the few-shot setting on 1\% data. Additionally, in the same setting, our model surpasses the strong baseline, M-BERT, in terms of accuracy and F1-score by $\sim$4\%.
This illustrates that our approaches provide strong adaptation robustness and are able to tackle the inherent adaptation difficulties to topologically distant languages.

\begin{table}[!t]
\setlength{\tabcolsep}{25pt}
\centering
\resizebox{0.55\textwidth}{!}{
\begin{tabular}{lcc}
\toprule
\multicolumn{1}{c}{\multirow{2}{*}{\textbf{Model}}}      & \multicolumn{2}{c}{\textbf{Thai}}                     \\ \cmidrule(lr){2-3} 
\multicolumn{1}{c}{}                            & \textbf{Intent}           & \textbf{Slot}             \\ \midrule
\multicolumn{3}{l}{\textit{few-shot on 5\% target language training set}}                                         \\ \midrule
\multicolumn{1}{l}{BiLSTM-CRF}                  & 90.05                     & 72.11                     \\
\multicolumn{1}{l}{+ LR}                        & 91.11                     & 73.71                     \\ \midrule
\multicolumn{1}{l}{BiLSTM-LVM}                  & 91.02                     & 73.11                     \\
\multicolumn{1}{l}{+ LR}                        & 91.45                     & 75.18                     \\
\multicolumn{1}{l}{+ ALVM}                 & 91.08                     & 74.67                     \\
\multicolumn{1}{l}{+ LR \& ALVM}           & 91.58                     & 75.87                     \\
\multicolumn{1}{l}{+ LR \& ALVM \& delex.} & \textbf{92.51}   & \textbf{77.03}                     \\ \midrule
\multicolumn{1}{l}{XL-Parsing}  & 91.05  & 73.43   \\ 
\multicolumn{1}{l}{M-BERT}  & 92.02  & 75.52   \\\bottomrule
\end{tabular}
}
\caption{Results of few-shot learning on 5\% Thai training data, which are averaged over three runs. We make the training samples in Thai the same as the 3\% Spanish training samples (\textbf{108}).}
\label{tab:comparison-between-spanish-and-thai}
\end{table}

\paragraph{Comparison Between Spanish and Thai}
To make a fair comparison for the few-shot performance in Spanish and Thai, we increase the training size of Thai to the same as 3\% Spanish training samples, as depicted in Table~\ref{tab:comparison-between-spanish-and-thai}. We can see that there is still a performance gap between the Spanish and Thai (3.11\% in the intent detection task and 8.15\% in the slot filling task). This is because Spanish is grammatically and syntactically closer to English than Thai, leading to a better quality of cross-lingual alignment.

\begin{figure}[!t]
\centering
\subfigure[LVM]{
    \includegraphics[scale=0.35]{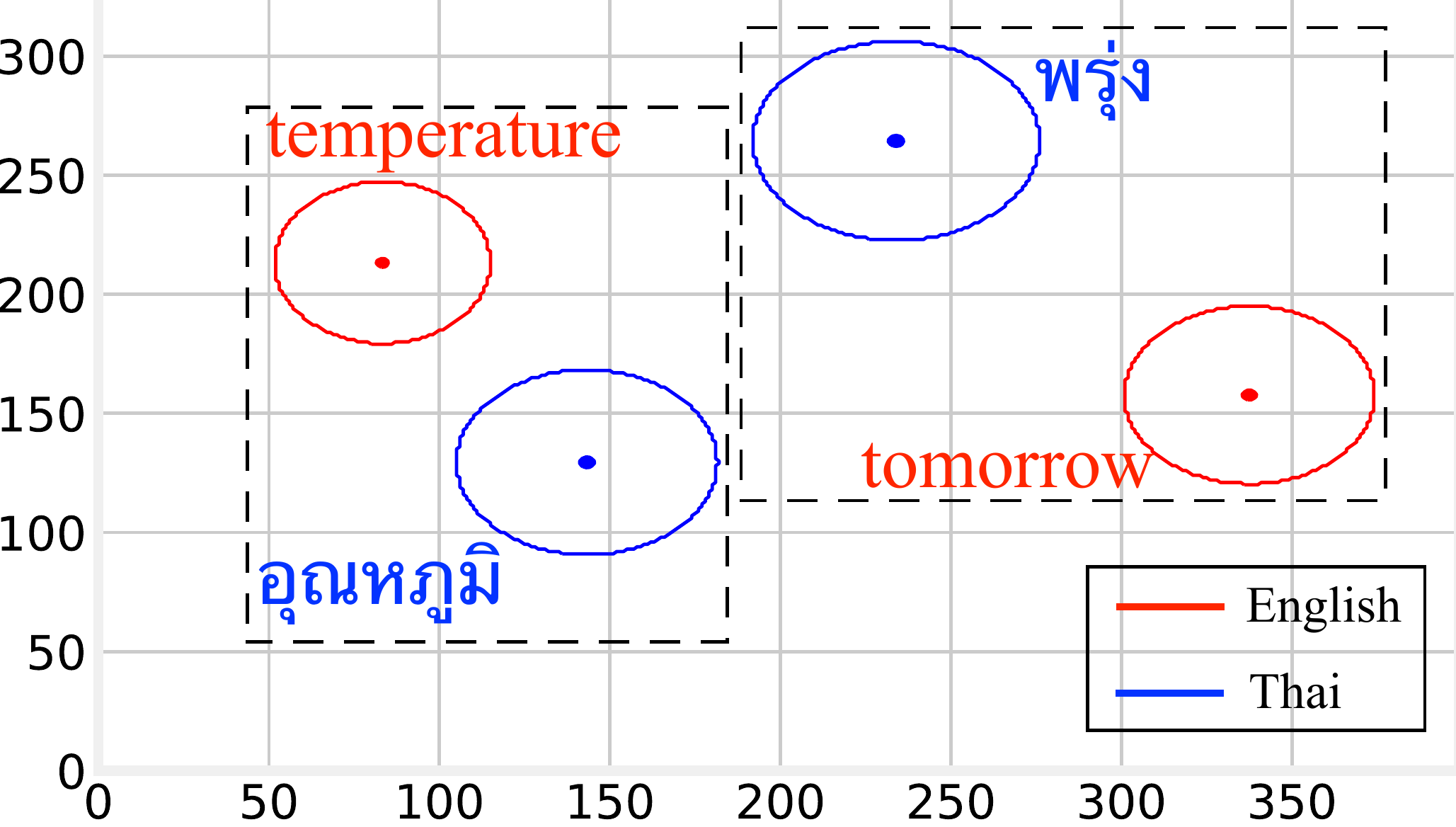}
}
\subfigure[LVM + LR]{
    \includegraphics[scale=0.35]{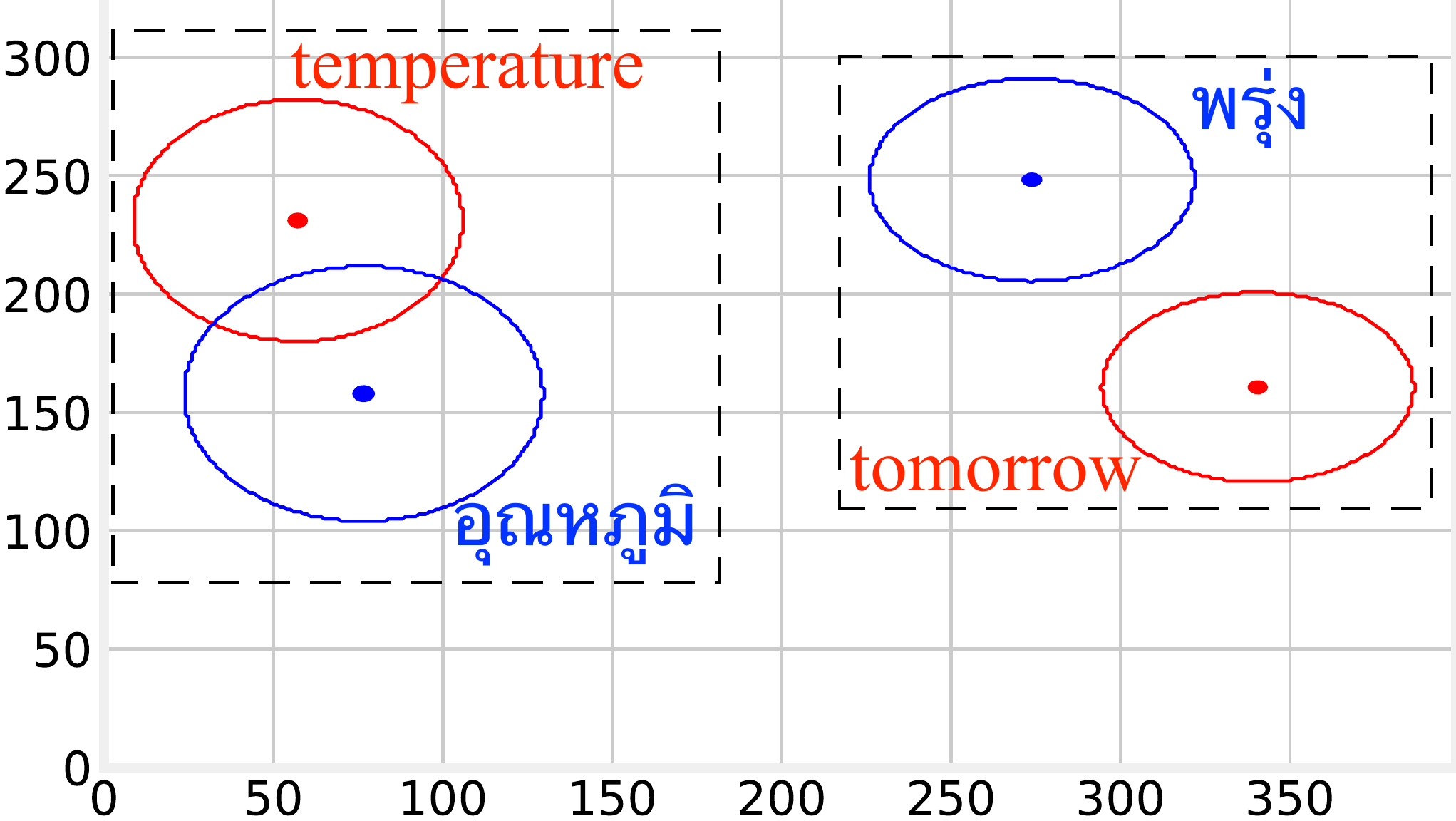}
}
\subfigure[ALVM]{
    \includegraphics[scale=0.35]{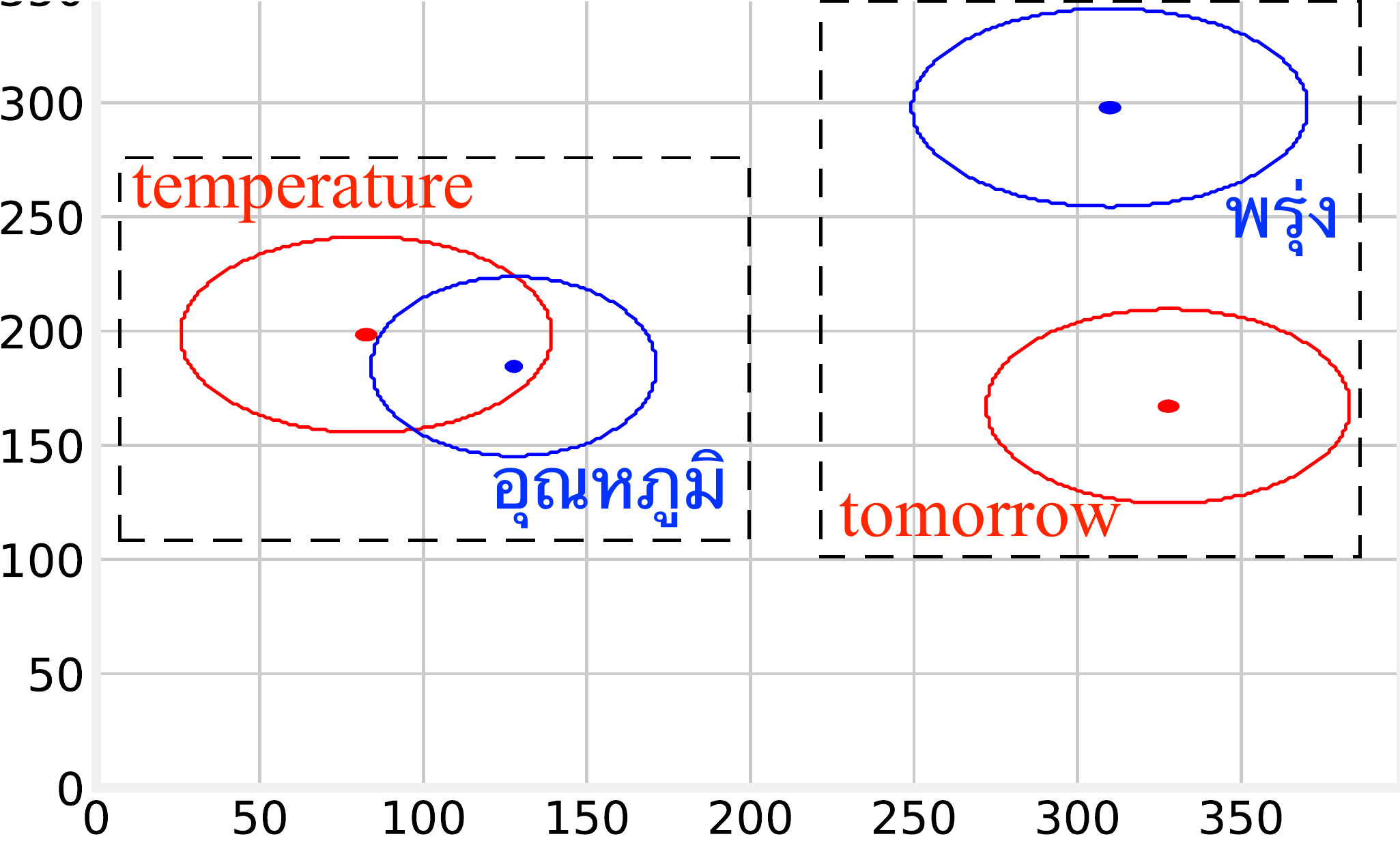}
}
\subfigure[ALVM + LR]{
    \includegraphics[scale=0.35]{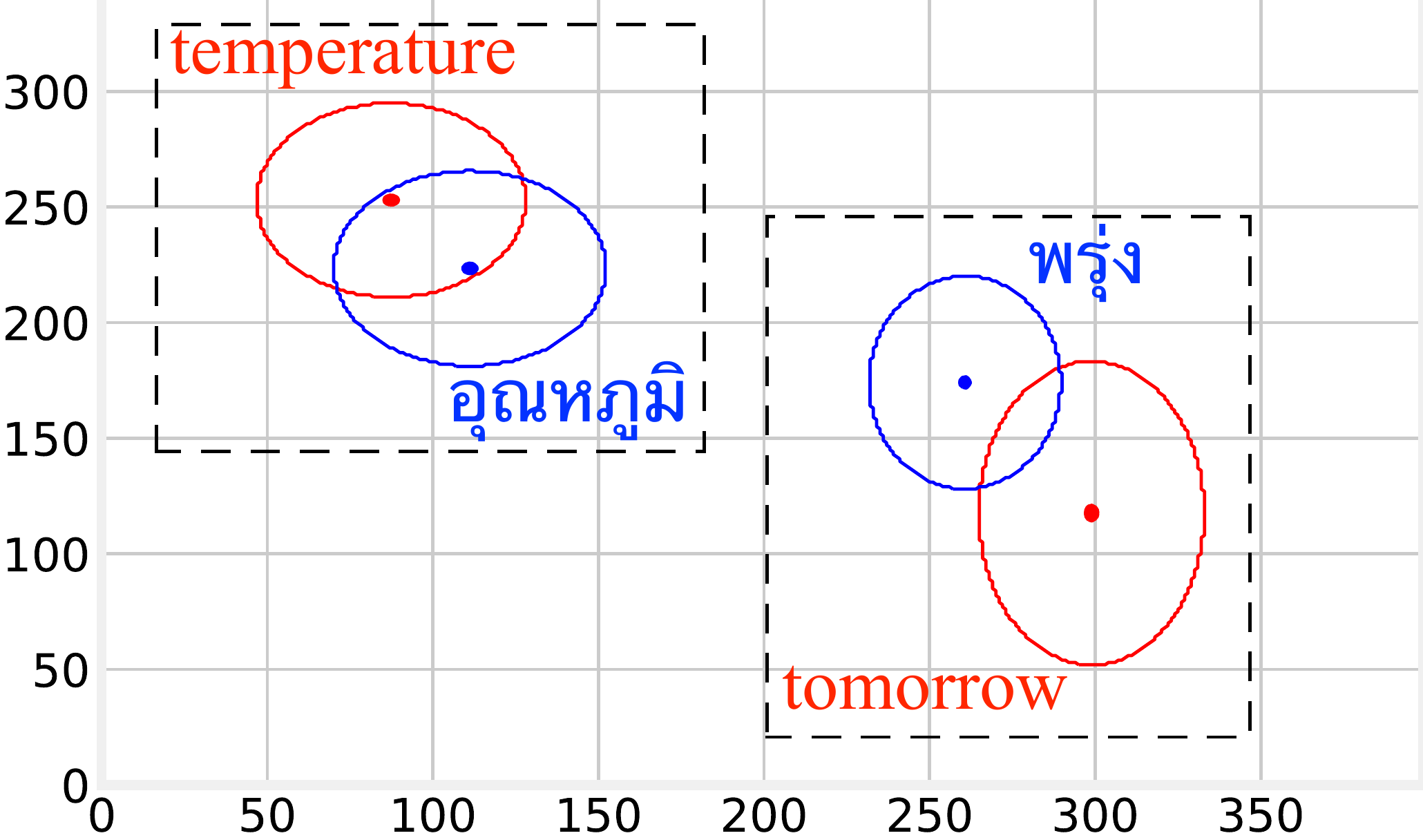}
}
\caption{Visualization for latent variables of parallel word pairs in English and Thai over different models trained on 1\% target language training set. We choose the word pairs ``temperature-
\foreignlanguage{thaicjk}{อุณหภูมิ}''
and ``tomorrow-
\foreignlanguage{thaicjk}{พรุ่ง}''
from the parallel sentences ``what will be the temperature tomorrow'' and ``\foreignlanguage{thaicjk}{อุณหภูมิ จะ อยู่ ท เท่า ไหร่ พรุ่ง}''
in English and Thai, respectively. To draw the contour plot, we sample 3000 points from the distribution of latent variables for the selected words, use PCA to project those points into 2D and calculate the mean and variance for each word.}
\label{fig:visualization-latent-variables}
\end{figure}

\paragraph{Visualization of Latent Variables}
The effectiveness of the LR and ALVM can be clearly seen from Figure~\ref{fig:visualization-latent-variables}. The former decreases the distance of latent variables for words with similar semantic meanings in different languages.
The latter makes the distributions for different slot types distinguishable.
Our model regularizes the latent variables of different slot types far from each other, and eventually it also improves the alignment of words with the same slot type. Incorporating both approaches further improves the word-level alignment across languages.
This further proves the robustness of our proposed approaches when adapting from the source language (English) to an unrelated language (Thai).

\begin{table}[!t]
\setlength{\tabcolsep}{25pt}
\centering
\resizebox{0.7\textwidth}{!}{
\begin{tabular}{lcccc}
\toprule
\multicolumn{1}{c}{\multirow{2}{*}{\textbf{Model}}}      & \multicolumn{2}{c}{\textbf{Spanish}} & \multicolumn{2}{c}{\textbf{Thai}} \\ \cmidrule(lr){2-3}  \cmidrule(lr){4-5}
\multicolumn{1}{c}{}   & \textbf{Intent}  & \multicolumn{1}{c}{\textbf{Slot}}  & \textbf{Intent} & \textbf{Slot}   \\ \midrule
\multicolumn{5}{l}{\textit{few-shot on 1\% target language training set}} \\ \midrule
\multicolumn{1}{l}{Our Model}                  & 93.82       & \multicolumn{1}{c}{78.46}          & 87.43  & 62.44     \\
\multicolumn{1}{l}{w/o Pre-training}                        & 92.75     &  \multicolumn{1}{c}{77.11}          & 86.29                     & 60.20  \\ \midrule
\multicolumn{5}{l}{\textit{few-shot on 3\% target language training set}} \\ \midrule
\multicolumn{1}{l}{Our Model}                  & 95.20       & \multicolumn{1}{c}{84.19}          & 90.97  & 70.88     \\
\multicolumn{1}{l}{w/o Pre-training}                        & 94.51   & \multicolumn{1}{c}{82.83}          & 89.72                     & 69.66  \\ \midrule
\multicolumn{5}{l}{\textit{zero-shot setting}} \\ \midrule
\multicolumn{1}{l}{Our Model}                  & 92.31       & \multicolumn{1}{c}{72.49}          & 75.77  & 33.28     \\
\multicolumn{1}{l}{w/o Pre-training}                        & 91.02       & \multicolumn{1}{c}{71.72}          & 75.18                     & 32.69  \\ \bottomrule
\end{tabular}
}
\caption{Results of the ablation study for the label sequence encoder pre-training (averaged over three runs). ``Our model'' refers to the one that combines LR, ALVM and delex. with BiLSTM-LVM.}
\label{tab:label-sequence-encoder-pretraining}
\end{table}

\paragraph{Effectiveness of Label Sequence Encoder Pre-training}
Label sequence encoder pre-training helps the label encoder to generate more expressive representations for label sequences, which ensures the effectiveness of the label regularization approach. From Table~\ref{tab:label-sequence-encoder-pretraining}, we can clearly observe the consistent performance gains made by pre-training in both few-shot and zero-shot scenarios.

\subsection{Short Summary}
In this section, we focus on improving the representation learning for low-resource languages from two perspectives: 1) improving model's robustness against imperfect cross-lingual alignments; 2) using regularization-based methods to further align cross-lingual representations.
These methods boost the word-level and sentence-level representation alignments across languages. Experimental results show that our methods achieve a remarkable performance boost compared to the strong baselines in both zero-shot and few-shot scenarios, especially for topologically distant target languages. We find that the low-resource representations can be easily and greatly enhanced by focusing on just the task-related keywords.
Visualization for latent variables further proves that our proposed methods are effective at improving the low-resource cross-lingual representations.

\newpage

\section{Order-Reduced Modeling}
Cross-lingual models trained on source language tasks possess the capability to directly transfer to target languages. However, since word order variances generally exist in different languages, cross-lingual models that overfit into the word order of the source language could have sub-optimal performance in target languages. 

In this section, we aim to address the challenge of language discrepancies in cross-lingual transfer and explore whether reducing the word order information fitted into the models can improve the task knowledge transfer to low-resource target languages. We introduce several methods to make models encode less word order information of the source language and test them based on cross-lingual word embeddings and a pre-trained multilingual model. 
Results show that: 1) our proposed Order-Reduced Transformer (ORT) is robust to word order shuffled sequences, and it consistently outperforms order-sensitive models; 2) preserving the order-agnostic property of M-BERT positional embeddings gives the model a better generalization ability to target languages; and 3) encoding excessive or insufficient word order information leads to inferior cross-lingual performance.

\subsection{Model Description}

\subsubsection{Order-Reduced Transformer}
Given that Transformer~\cite{vaswani2017attention} relies on positional embeddings to encode word order information, we propose to remove them so as to reduce the amount of word order information injected into input sentences. Note that given a linear layer as the feed-forward layer for Transformer, as introduced in~\citet{vaswani2017attention}, removing the positional embeddings module would mean getting rid of all the word order information, which would lead to a large performance drop in the source language and low performance for the cross-lingual transfer. Therefore, we utilize a one-dimensional convolutional network (Conv1d)~\cite{kim2014convolutional} as the feed-forward layer to extract n-gram features from the Multi-Head Attention features. Specifically, we formulate the encoding process as follows:
\begin{equation}
    g[1:n] = \texttt{MultiHead}(E(X[1:n])),
\end{equation}
where $X[1:n]$ represents the $n$-token input sequence; $E$ denotes the embedding layer; and $g[1:n] \in R^{n \times d} $, where $d$ is the hidden size of Transformer, represents the sequence features generated by Multi-Head Attention.

After that, a feature $c_i$ is generated from the window of features $g[i:i+h-1]$ by
\begin{equation}
  c_i = \texttt{Conv1d}(g[i:i+h-1]),
\end{equation}
where $h$ is the kernel size of Conv1d and the dimension of $c_i$ is equal to the number of output channels in Conv1d. 
We add padding for this convolution process to ensure the output feature length is the same as the length of the input tokens. Finally, the output feature sequences from Conv1d are the concatenation of $c_i$, where $i \in [1,n]$.

In this way, we fit the model with less word order information since the model only encodes the local n-gram features, and the prediction for each token is made based on the token itself and its neighbor tokens.

\subsubsection{Shuffling Word Order}
\label{sec:order-shuffle}
Instead of removing positional embeddings, we propose to permutate the word order of input sequences in the source language training samples so as to train models to be robust to different word orders.
Meanwhile, we keep the order of tokens in each entity the same and consider them as one ``word'' to ensure we don't break entities in the sequences.

We follow~\citet{lample2017unsupervised} to generate permutations similar to the noise observed with word-by-word translation (i.e., word order differences across languages). Concretely, we apply a random permutation $\sigma$ to the input sequence, verifying the condition $\forall i \in \{1,n\}, |\sigma(i) - i| \leq k$, where n is the length of the input sentence and k is a tunable parameter that controls the degree of shuffling.
We use the order-shuffled training samples to train the models and make them less sensitive to word orders.

\subsubsection{Order-Agnostic Positional Embeddings}
Another alternative method is to make the positional embeddings of Transformer order-agnostic so as to encode less order information.
In light of M-BERT's astonishing cross-lingual performance~\cite{pires2019multilingual, wu2019beto},
we speculate that the positional embeddings in M-BERT are order-agnostic.
Hence, we leverage M-BERT's positional embeddings to initialize the positional embeddings for Transformer, and we freeze them in the training phase to prevent them from fitting into the source language word order.
In the experiments, since M-BERT's positional embeddings are based on its tokenizer, we leverage it to tokenize sequences and generate cross-lingual embeddings from M-BERT. Then, a Transformer encoder with M-BERT's positional embeddings is added on top of the M-BERT embeddings. We freeze the parameters of M-BERT in the training phase to ensure the cross-lingual embeddings from M-BERT do not fit into the source language word order.

\subsubsection{Fine-tuning M-BERT}
The original fine-tuning of M-BERT to downstream cross-lingual tasks is done by adding a linear layer on top of M-BERT and fine-tuning all the parameters of the model to the source language task~\cite{pires2019multilingual, wu2019beto}. This inescapably fits the model with the source language word order.
To circumvent this issue, we freeze the positional embeddings in M-BERT in the fine-tuning stage. By doing so, the positional embeddings can still provide the word order information for M-BERT to encode input sequences, and the model avoids fitting the word order of the source language.

\begin{table}[!t]
\centering
\setlength{\tabcolsep}{20pt}
\resizebox{0.6\textwidth}{!}{
\begin{tabular}{lcccc}
\toprule
\# samples & en  & es  & de & nl \\ \midrule
Train & 14,040 & 8,319 & 12,152 & 15,802 \\
Validation & 3,249 & 1,914 & 2,867 & 2,895 \\
Test & 3,452 & 1,516 & 3,005 & 5,194 \\ \bottomrule
\end{tabular}
}
\caption{Number of samples for each language in the CoNLL 2002 and CoNLL 2003 NER datasets.}
\label{tab:ner-data-statistics}
\end{table}

\begin{table}[!t]
\centering
\setlength{\tabcolsep}{20pt}
\resizebox{0.99\textwidth}{!}{
\begin{tabular}{lcccccccc}
\toprule
\# samples & en & ar & bg & ca & zh & hr & cs & da  \\ \midrule
Train & 12,543 & 6,075 & 8,907 & 13,123 & 3,997 & 6,983 & 102,993 & 4,383 \\
Validation & 2,002 & 909 & 1,115 & 1,709 & 500 & 849 & 11,311 & 564 \\
Test & 2,077 & 680 & 1,116 & 1,846 & 500 & 1,057 & 12,203 & 565 \\ \toprule
\# samples & nl & et & fi & fr & de & he & hi & id  \\ \midrule
Train & 18,058 & 20,837 & 12,217 & 14,554 & 13,814 & 5,241 & 13,304 & 4,477 \\
Validation & 1,394 & 2,633 & 1,364 & 1,478 & 799 & 484 & 1,659 & 559 \\
Test & 1,472 & 2,737 & 1,555 & 416 & 977 & 491 & 1,684 & 557 \\ \toprule
\# samples & it & ja & ko & la & lv & no & pl & pt  \\ \midrule
Train & 13,121 & 7,164 & 27,410 & 15,906 & 5,424 & 29,870 & 19,874 & 17,993 \\
Validation & 564 & 511 & 3,016 & 1,234 & 1,051 & 4,300 & 2,772 & 1,770 \\
Test & 482 & 557 & 3,276 & 1,260 & 1,228 & 3,450 & 2,827 & 1,681 \\ \toprule
\# samples & ro & ru & sk & sl & es & sv & uk  \\ \midrule
Train & 8,043 & 48,814 & 8,483 & 8,556 & 28,492 & 4,303 & 4,513  \\
Validation & 752 & 6,584 & 1,060 & 734 & 3,054 & 504 & 577 \\
Test & 729 & 6,491 & 1,061 & 1,898 & 2,147 & 1,219 & 783 \\ \bottomrule
\end{tabular}
}
\caption{Number of samples for each language in the Universal Dependencies we use.}
\label{tab:pos-data-statistics}
\end{table}

\subsection{Experimental Setup}

\subsubsection{Datasets}
We test our methods on three natural language understanding tasks in the cross-lingual setting, namely, part-of-speech (POS) tagging, named entity recognition (NER), and semantic parsing. For the POS task, we choose the same language set as~\citet{ahmad2018difficulties} (31 languages in total) from the Universal Dependencies~\cite{nivre2017universal} to evaluate our methods. And we use the CoNLL 2002 and CoNLL 2003 datasets~\cite{tjong2002introduction,sang2003introduction}, which contain English (en), German (de), Spanish (es), and Dutch (nl), to evaluate our methods for the NER task. Finally, for the semantic parsing task, we use the multilingual semantic parsing (containing the intent detection and slot filling tasks) dataset introduced by~\citet{schuster2019crosslingual}, which contains English (en), Spanish (es), and Thai (th) across weather, alarm and reminder domains. 
The data statistics for the semantic parsing, NER, and POS tagging tasks are shown in Table~\ref{tab:semantic-parsing-data-statistics}, Table~\ref{tab:ner-data-statistics}, and Table~\ref{tab:pos-data-statistics}, respectively.

\subsubsection{Our Models and Baselines}
All our models and baseline models consist of a sequence encoder to produce features for input sequences and a conditional random field (CRF) layer~\cite{lample2016neural,ma2016end} to make predictions based on the sequence features. 
For the sequence encoder, we use Bidirectional LSTM (\textbf{BiLSTM}), Transformer (\textbf{TRS}) using sinusoidal functions as positional embeddings, Relative Positional Transformer (\textbf{RPT}) proposed in~\citet{ahmad2018difficulties}~\footnote{They utilize the relative positional embeddings proposed in~\citet{shaw2018self} to encode less word order information for cross-lingual adaptation.}, or Order-Reduced Transformer (\textbf{ORT}).
All Transformer-based encoders use Conv1d as the feed-forward layer for a fair comparison.
Word order shuffling is applied to BiLSTM and TRS baselines to make them less sensitive to word orders.
We fine-tune M-BERT by adding a linear layer on top of it, and we compare two different fine-tuning M-BERT methods (with and without freezing the positional embeddings).

\subsubsection{Training}
We evaluate our models with cross-lingual word embeddings (word-level) and M-BERT embeddings (subword-level). For the word-level embeddings, we leverage the RCSLS multilingual embeddings~\cite{joulin2018loss} for the POS and NER tasks, and we use the refined RCSLS (based on our proposed cross-lingual embedding refinement in Section~\ref{sec:cross-lingual-semantic-parsing}) for the semantic parsing task since it is specifically refined for this task. 
We set the kernel size as 3 for the feed-forward layer Conv1d in the Transformer encoder. For the word order shuffled data, we generate ten word order-shuffled samples with $k=\infty$ (can generate any permutation) for each source language training sample.
Note that the word order shuffling can not be applied for M-BERT-based models since they are pre-trained based on the correct language order, and it is not suitable to feed them with order-shuffled sequences.
For all the tasks, we use English as the source language and other languages as target languages. We follow~\citet{ahmad2018difficulties} to calculate the language distances between target languages and English.
The semantic parsing task consists of intent detection (ID) and slot filling (SF), and we mainly focus on the SF task, since it is the core of the semantic parsing.
We use the standard BIO-based F1-score for evaluating the NER and SF tasks, as in~\citet{lample2016neural}, and accuracy score for evaluating the POS task, as in~\citet{kim2017cross}. 
We use 2-layer BiLSTM with a hidden size of 256. For Transformer and ORT, the number of heads is 8, and the hidden size is 256. We use an Adam optimizer with a learning rate of 1e-3, and the batch size we use is 64. For the models using cross-lingual word embeddings, we freeze the embeddings in the training stage to ensure that the cross-lingual alignment will not be broken.
In the zero-shot setting, we do not leverage any target language data, and we use an early stop strategy based on the performance in the source language validation set to select the model. In the few-shot setting, we utilize very few target language data samples for training, and the early stop strategy is based on the performance in the target language validation set. The number of trainable parameters for the Transformer and BiLSTM based models (without considering M-BERT) is around 1 million.

\subsubsection{Applying ORT into Strong Baselines}
We apply the ORT into two strong cross-lingual models for zero-shot cross-lingual semantic parsing~\cite{liu2019zero}~\footnote{This is our previous proposed models XL-Parsing.} and NER~\cite{chen2019multi}, and both are based on BiLSTM as the sequence encoder, which is order-sensitive. To ensure a fair comparison, we keep all settings the same as the originals, except that the sequence encoder is replaced. 
For the semantic parsing model from~\citet{liu2019zero}, we replace the BiLSTM with ORT. And for the NER model from~\citet{chen2019multi}, we replace the BiLSTM in the shared feature extractor module with ORT.

\begin{table}[!t]
\renewcommand{\arraystretch}{1.1}
\centering
\setlength{\tabcolsep}{22pt}
\resizebox{0.98\textwidth}{!}{
\begin{tabular}{lccccccccc}
\toprule
\multicolumn{1}{c}{\multirow{2}{*}{}} & \multicolumn{5}{c}{\textbf{Named Entity Recognition Task}}                              & \multicolumn{4}{c}{\textbf{Slot Filling Task}}              \\ \cmidrule(lr){2-6}  \cmidrule(lr){7-10}
\multicolumn{1}{l}{}                  & \textbf{en}    & \textbf{es}    & \textbf{de}    & \multicolumn{1}{c}{\textbf{nl}}    & \multicolumn{1}{c}{\textbf{avg}}   & \textbf{en}    & \textbf{es}    & \multicolumn{1}{c}{\textbf{th}}    & \textbf{avg}   \\ \midrule
\multicolumn{1}{l}{Dist. to English}        & 0.00  & 0.12  & 0.14  & \multicolumn{1}{c}{0.14}  & \multicolumn{1}{c}{0.13}  & 0.00  & 0.12  & \multicolumn{1}{c}{0.31}  & 0.22  \\ \toprule
\multicolumn{10}{l}{\textbf{\textit{Frozen Word-level Embeddings}}} \\ \midrule
\multicolumn{1}{l}{BiLSTM}            & 87.99 & 33.71 & 25.28 & \multicolumn{1}{c}{15.28} & \multicolumn{1}{c}{24.76} & \textbf{94.87} & 59.51 & \multicolumn{1}{c}{20.63} & 40.07 \\
\multicolumn{1}{l}{\hspace{3mm} w/ shuffled data}  & 83.85 & 30.09 & 22.87 & \multicolumn{1}{c}{13.22} & \multicolumn{1}{c}{22.06} & 93.57 & 62.02 & \multicolumn{1}{c}{21.43} & 41.73 \\
\multicolumn{1}{l}{Transformer (TRS)}       & \textbf{88.67} & 30.76 & 30.54 & \multicolumn{1}{c}{18.53} & \multicolumn{1}{c}{26.61} & 94.78 & 62.67 & \multicolumn{1}{c}{22.33} & 42.50 \\
\multicolumn{1}{l}{\hspace{3mm} w/ shuffled data}  & 82.75 & 28.54 & 28.43 & \multicolumn{1}{c}{16.17} & \multicolumn{1}{c}{24.38} & 92.07 & 63.86 & \multicolumn{1}{c}{24.17} & 44.02 \\
\multicolumn{1}{l}{\citet{ahmad2018difficulties}}       & 87.86 & 32.49 & 31.83 & \multicolumn{1}{c}{19.24} & \multicolumn{1}{c}{27.85} & 94.23 & 62.07 & \multicolumn{1}{c}{23.14} & 42.61 \\ \midrule
\multicolumn{1}{l}{ORT}               & 88.41 & \textbf{34.33} & \textbf{33.54} & \multicolumn{1}{c}{\textbf{24.14}} & \multicolumn{1}{c}{\textbf{30.67}} & 94.50 & \textbf{66.84} & \multicolumn{1}{c}{\textbf{25.53}} & \textbf{46.19} \\ \toprule
\multicolumn{10}{l}{\textbf{\textit{Frozen M-BERT Embeddings}}} \\ \midrule
\multicolumn{1}{l}{Transformer (TRS)}       & 89.53 & 58.93 & 46.28 & \multicolumn{1}{c}{63.15} & \multicolumn{1}{c}{56.12} & \textbf{94.93} & 46.75 & \multicolumn{1}{c}{9.76}  & 28.26 \\
\multicolumn{1}{l}{\hspace{3mm} w/ M-BERT PE}       & 88.44 & 58.27 & \textbf{47.63} & \multicolumn{1}{c}{64.12} & \multicolumn{1}{c}{56.67} & 94.53 & 47.23 & \multicolumn{1}{c}{\textbf{10.06}} & 28.65 \\
\multicolumn{1}{l}{\citet{ahmad2018difficulties}}       & \textbf{89.96} & \textbf{60.55} & 45.43 & \multicolumn{1}{c}{61.58} & \multicolumn{1}{c}{55.85} & 94.38 & 47.80 & \multicolumn{1}{c}{8.83}  & 28.32 \\ \midrule
\multicolumn{1}{l}{ORT}               & 89.46 & 58.35 & 45.95 & \multicolumn{1}{c}{\textbf{66.31}} & \multicolumn{1}{c}{\textbf{56.87}} & 94.55 & \textbf{48.42} & \multicolumn{1}{c}{9.92}  & \textbf{29.17} \\ \toprule
\multicolumn{10}{l}{\textbf{\textit{M-BERT Fine-tuning}}} \\ \midrule
\multicolumn{1}{l}{Fine-tune M-BERT}         & \textbf{91.95} & 74.49 & 69.13 & \multicolumn{1}{c}{77.32} & \multicolumn{1}{c}{73.65} & \textbf{95.97} & 69.41 & \multicolumn{1}{c}{10.45} & 39.93 \\
\multicolumn{1}{l}{\hspace{3mm} w/ frozen PE}      & 91.87 & \textbf{74.98} & \textbf{70.22} & \multicolumn{1}{c}{\textbf{77.63}} & \multicolumn{1}{c}{\textbf{74.28}} & 95.90 & \textbf{70.30} & \multicolumn{1}{c}{\textbf{12.53}} & \textbf{41.42} \\ \midrule
\end{tabular}
}
\caption{Zero-shot cross-lingual results on NER and SF tasks (averaged over three runs) for the three settings. 
We freeze the word-level embeddings in the training stage to ensure their cross-lingual alignment is preserved.
We use ``w/ shuffled data'' to denote the models trained with the word order shuffled source language training samples. ``PE'' denotes positional embeddings, and we use ``w/ M-BERT PE'' to represent that the model initialized with the frozen M-BERT positional embeddings. ``avg'' denotes the average performance over the target languages (English is excluded).}
\label{tab:zs-cross-lingual-ner-sf}
\end{table}

\begin{table}[!t]
\renewcommand{\arraystretch}{1.09}
\centering
\resizebox{0.999\textwidth}{!}{
\begin{tabular}{cc|cccccc|cccc|cc}
\toprule
\multirow{3}{*}{\textbf{Lang}} & \multirow{3}{*}{\begin{tabular}[c]{@{}c@{}}\textbf{Dist. to} \\ \textbf{English}\end{tabular}} & \multicolumn{6}{c|}{\textbf{Frozen Word-level Embeddings}}                                                                                                                           & \multicolumn{4}{c|}{\textbf{Frozen M-BERT Embeddings}}                                     & \multicolumn{2}{c}{\textbf{M-BERT Fine-tuning}}                                    \\ \cmidrule{3-14} 
                      &                                                                              & BiLSTM & \begin{tabular}[c]{@{}c@{}}BiLSTM w/\\ shuffled data\end{tabular} & TRS   & \begin{tabular}[c]{@{}c@{}}TRS w/ \\ shuffled data\end{tabular} & RPT   & ORT   & TRS   & \begin{tabular}[c]{@{}c@{}}TRS w/ \\ M-BERT PE\end{tabular} & RPT   & ORT   & M-BERT & \begin{tabular}[c]{@{}c@{}}M-BERT w/ \\ frozen PE\end{tabular} \\ \midrule
en                    & 0.00                                                                         & \textbf{93.76}  & 87.66                                                             & 93.07 & 86.84                                                           & 92.77 & 93.74 & 92.73 & 92.36                                                       & 92.47 & \textbf{93.07} & 97.20  & \textbf{97.21}                                                          \\ \midrule
no                    & 0.06                                                                         & 34.48  & 23.50                                                             & 44.29 & 40.37                                                           & 44.45 & \textbf{55.05}$^\ddagger$ & 65.19 & 66.34                                                       & \textbf{68.44} & 65.73 & 75.72  & \textbf{76.11}                                                          \\
sv                    & 0.07                                                                         & 27.74  & 21.83                                                             & 43.83 & 29.66                                                           & 39.66 & \textbf{56.92}$^\ddagger$ & 74.84 & 76.38                                                       & \textbf{76.47} & 75.75 & 85.02  & \textbf{85.48}                                                          \\
fr                    & 0.09                                                                         & 28.92  & 22.29                                                             & 50.70 & 40.16                                                           & 47.13 & \textbf{62.72}$^\ddagger$ & 76.06 & 77.27                                                       & \textbf{79.62} & 79.24$^\ddagger$ & 88.57  & \textbf{88.82}                                                          \\
pt                    & 0.09                                                                         & 41.34  & 29.71                                                             & 59.00 & 48.78                                                           & 55.14 & \textbf{66.77} & 83.23 & 83.94                                                       & \textbf{84.94} & 84.64 & 90.66  & \textbf{91.10}                                                          \\
da                    & 0.10                                                                         & 51.77  & 37.61                                                             & 49.14 & 49.18                                                           & 56.87 & \textbf{63.21}$^\ddagger$ & 77.69 & 78.77                                                       & \textbf{79.53} & 79.02 & 87.19  & \textbf{87.61}                                                          \\
es                    & 0.12                                                                         & 41.86  & 31.49                                                             & 51.82 & 44.58                                                           & 50.82 & \textbf{60.29} & 76.58 & \textbf{79.43}$^\ddagger$                                                       & 77.93 & 77.91 & 86.56  & \textbf{86.88}                                                          \\
it                    & 0.12                                                                         & 37.14  & 24.65                                                             & 54.78 & 42.88                                                           & 49.93 & \textbf{66.55}$^\ddagger$ & 73.59 & \textbf{77.46}$^\ddagger$                                                       & 77.21 & 77.38$^\ddagger$ & 88.98  & \textbf{89.85}$^\ddagger$                                                          \\
hr                    & 0.13                                                                         & 29.35  & 22.11                                                             & 45.88 & 38.85                                                           & 45.87 & \textbf{55.40}$^\ddagger$ & 68.56 & 71.12$^\ddagger$                                                       & \textbf{71.70} & 71.08$^\ddagger$ & 82.57  & \textbf{83.45}$^\ddagger$                                                          \\
ca                    & 0.13                                                                         & 38.20  & 26.02                                                             & 50.97 & 46.24                                                           & 49.66 & \textbf{59.56} & 74.61 & \textbf{78.22}$^\ddagger$                                                       & 75.84 & 77.96$^\ddagger$ & 85.85  & \textbf{86.11}                                                          \\
pl                    & 0.13                                                                         & 43.13  & 31.41                                                             & 49.16 & 32.72                                                           & 52.02 & \textbf{62.97}$^\ddagger$ & 66.93 & \textbf{68.73}                                                       & 65.82 & 68.53 & 80.11  & \textbf{80.61}                                                          \\
uk                    & 0.13                                                                         & 29.60  & 28.99                                                             & 42.70 & 35.14                                                           & 49.40 & \textbf{56.13}$^\ddagger$ & 73.45 & \textbf{75.15}                                                       & 73.23 & 75.08 & 83.83  & \textbf{84.41}                                                          \\
sl                    & 0.13                                                                         & 33.67  & 29.08                                                             & 43.94 & 40.00                                                           & 42.80 & \textbf{60.01}$^\ddagger$ & 62.76 & 64.68                                                       & \textbf{67.42} & 64.49 & 75.71  & \textbf{76.58}                                                          \\
nl                    & 0.14                                                                         & 34.95  & 24.08                                                             & 48.69 & 34.87                                                           & 45.02 & \textbf{63.30}$^\ddagger$ & 79.28 & 79.07                                                       & 79.48 & \textbf{80.08} & 86.90  & \textbf{87.68}                                                          \\
bg                    & 0.14                                                                         & 28.42  & 24.71                                                             & 39.78 & 29.98                                                           & 43.98 & \textbf{54.49}$^\ddagger$ & 68.63 & \textbf{70.13}                                                       & 68.93 & 69.42 & \textbf{80.66}  & 80.60                                                          \\
ru                    & 0.14                                                                         & 29.65  & 26.83                                                             & 47.96 & 40.12                                                           & 51.53 & \textbf{59.20}$^\ddagger$ & 78.39 & 79.53                                                       & 77.88 & \textbf{79.76} & 88.80  & \textbf{89.05}                                                          \\
de                    & 0.14                                                                         & 31.52  & 25.12                                                             & 43.24 & 35.01                                                           & 41.11 & \textbf{50.59} & 68.41 & 67.75                                                       & \textbf{69.00} & 68.96 & \textbf{81.02}  & 80.66                                                          \\
he                    & 0.14                                                                         & 21.67  & 19.74                                                             & 34.82 & 24.38                                                           & 33.97 & \textbf{39.24} & 64.30 & 66.12                                                       & 65.22 & \textbf{66.88}$^\ddagger$ & 69.74  & \textbf{70.30}                                                          \\
cs                    & 0.14                                                                         & 34.06  & 27.53                                                             & 48.30 & 36.97                                                           & 51.00 & \textbf{63.79}$^\ddagger$ & 75.05 & \textbf{76.46}                                                       & 75.17 & 76.17 & 85.71  & \textbf{86.20}                                                          \\
ro                    & 0.15                                                                         & 35.05  & 28.36                                                             & 45.43 & 41.10                                                           & 46.29 & \textbf{61.23}$^\ddagger$ & 69.17 & \textbf{72.42}$^\ddagger$                                                       & 71.93 & 71.37$^\ddagger$ & 81.38  & \textbf{82.03}                                                          \\
sk                    & 0.17                                                                         & 39.98  & 35.62                                                             & 47.66 & 38.46                                                           & 50.28 & \textbf{61.10}$^\ddagger$ & 68.13 & \textbf{69.21}                                                       & 68.77 & 68.46 & 80.50  & \textbf{81.23}                                                          \\
id                    & 0.17                                                                         & 32.54  & 24.09                                                             & 34.11 & 26.76                                                           & \textbf{41.61} & 39.10 & 59.97 & \textbf{62.08}$^\ddagger$                                                       & 60.13 & 62.07$^\ddagger$ & 71.80  & \textbf{72.86}$^\ddagger$                                                          \\
lv                    & 0.18                                                                         & 49.05  & 35.21                                                             & 52.43 & 50.66                                                           & \textbf{58.80} & 57.87 & 64.74 & \textbf{68.46}$^\ddagger$                                                       & 66.81 & 67.94$^\ddagger$ & 79.01  & \textbf{79.64}                                                          \\
fi                    & 0.20                                                                         & 37.88  & 29.52                                                             & 44.83 & 38.53                                                           & 47.16 & \textbf{53.85} & 68.65 & \textbf{72.04}$^\ddagger$                                                       & 69.22 & 71.87$^\ddagger$ & 81.09  & \textbf{81.98}$^\ddagger$                                                          \\
et                    & 0.20                                                                         & 30.13  & 26.50                                                             & 41.86 & 25.64                                                           & 42.00 & \textbf{49.81} & 57.50 & \textbf{63.00}$^\ddagger$                                                       & 58.27 & 62.29$^\ddagger$ & 75.66  & \textbf{75.80}                                                          \\
zh                    & 0.23                                                                         & 26.66  & 23.95                                                             & 28.92 & 24.81                                                           & 27.94 & \textbf{31.20} & 53.37 & 55.58$^\ddagger$                                                       & 53.51 & \textbf{55.66}$^\ddagger$ & 63.86  & \textbf{64.06}                                                          \\
ar                    & 0.26                                                                         & 7.75   & 12.40                                                             & \textbf{25.77} & 4.32                                                            & 24.97 & 22.29 & 25.94 & \textbf{29.96}$^\ddagger$                                                       & 24.39 & 29.01$^\ddagger$ & 27.68  & \textbf{36.94}$^\ddagger$                                                          \\
la                    & 0.28                                                                         & -      & -                                                                 & -     & -                                                               & -     & -     & 42.22 & \textbf{46.11}$^\ddagger$                                                       & 43.32 & 43.34 & 45.52  & \textbf{46.96}$^\ddagger$                                                          \\
ko                    & 0.33                                                                         & 10.79  & 7.87                                                              & \textbf{21.82} & 5.16                                                            & 16.71 & 19.02 & 35.77 & 38.17$^\ddagger$                                                       & 36.08 & \textbf{39.97}$^\ddagger$ & 42.18  & \textbf{45.45}$^\ddagger$                                                          \\
hi                    & 0.40                                                                         & 20.97  & 17.68                                                             & 27.66 & 25.11                                                           & 29.00 & \textbf{35.05} & 47.95 & \textbf{52.76}$^\ddagger$                                                       & 49.45 & 51.90$^\ddagger$ & \textbf{56.45}  & 55.63                                                          \\
ja                    & 0.49                                                                         & -      & -                                                                 & -     & -                                                               & -     & -     & 40.59 & \textbf{43.17}$^\ddagger$                                                       & 41.46 & 42.82$^\ddagger$ & 44.97  & \textbf{45.08}                                                          \\ \midrule
avg                   & 0.17                                                                         & 32.44  & 25.64                                                             & 43.55 & 34.66                                                           & 44.11 & \textbf{53.10} & 64.72 & \textbf{66.98}                                                       & 65.91 & 66.83 & 75.12  & \textbf{75.97}                                                          \\ \bottomrule
\end{tabular}
}
\caption{Zero-shot cross-lingual results on the POS task (averaged over three runs). Languages are sorted by the word-ordering distance to English. Since the word-level embeddings for la and ja languages are absent, we do not report these results. $^\ddagger$ denotes that the performance improvement of the proposed models is higher than their corresponding average improvements.}
\label{tab:zs-cross-lingual-pos}
\end{table}

\subsection{Results and Discussion}

\subsubsection{Zero-shot Adaptation}
\paragraph{Order-Reduced Transformer}
As we can see from Table~\ref{tab:zs-cross-lingual-ner-sf}, removing positional embeddings from Transformer (ORT) only makes the performance in the source language (English) drop slightly by around 0.5\%. This indicates that leveraging only local order information results in a good performance in sequence labeling tasks. In other words, relying just on the information from the neighboring words (how many neighboring words depend on the kernel size in Conv1d) can ensure relatively good performance for sequence labeling tasks.
On the other hand, in terms of zero-shot adaptation to target languages (from Table~\ref{tab:zs-cross-lingual-ner-sf} and~\ref{tab:zs-cross-lingual-pos}), ORT achieves consistently better performance than the order-sensitive encoders (i.e., BiLSTM and TRS) as well as the order-reduced encoder RPT~\cite{ahmad2018difficulties}. 
For example, in the SF task, ORT outperforms BiLSTM, TRS, and RPT by 6.12\%, 3.69\%, and 3.58\%, respectively, on the F1-score, in terms of the average performance of using word-level embeddings. 

Compared to the order-sensitive models, ORT fits the word order of the source language less, which increases its adaptation robustness to target languages. We conjecture that the reason why ORT outperforms RPT is that RPT still keeps the relative word distances. Although RPT reduces the order information that the model encodes, it might not be suitable for target languages that do not have similar relative word distance patterns to English, while ORT removes all the order information in positional embeddings, which makes it more robust to the word order differences.

\paragraph{Shuffling Word Order}
From Table~\ref{tab:zs-cross-lingual-ner-sf} and~\ref{tab:zs-cross-lingual-pos}, we can see that the models trained with word order shuffled data lead to a visible performance drop in English, especially for the POS and NER tasks. For target languages, however, we observe that the performance improves in the SF task by using such data. For example, using the order shuffled data to train the Transformer improves the performance by 1.52\% on the average F1-score.
For cross-lingual adaptation, performance loss in the source language has a negative impact on the performance in target languages. In the SF task, the performance drop in English is relatively small ($\sim$2\%); hence, the benefits from being less sensitive to word orders are greater than the performance losses in English.

On the other hand, using order shuffled data makes the performance in target languages worse for the NER and POS tasks. For example, for the POS task, the average accuracy drops 8.89\% for the Transformer trained with the order shuffled data compared to the one trained without such data.
We also observe large performance drops for the NER and POS tasks in English caused by using the order shuffled data (for example, for the POS task, the drop is $\sim$6\%) since the models for these tasks are more vulnerable to the shuffled word order. In this case, the performance losses in English are larger than the benefits of being less sensitive to word orders.

\paragraph{Order-Agnostic Positional Embeddings}
As we can see from Table~\ref{tab:zs-cross-lingual-ner-sf} and~\ref{tab:zs-cross-lingual-pos}, compared to TRS alone, we observe that TRS trained with M-BERT PE (frozen) only results in a slight performance drop in English, while it generally brings better zero-shot adaptation performance to target languages. For example, in the POS task, TRS with M-BERT PE achieves 2.26\% higher average accuracy than that without M-BERT PE.
Since M-BERT is trained using 104 languages, its positional embeddings are fitted to different word orders across various languages and become order-agnostic. 
Since the pre-trained positional embeddings are frozen, their order-agnostic property is retained, which brings more robust adaptation to target languages.

In addition, we notice that ORT achieves similar performance to TRS with M-BERT PE, which further illustrates the effectiveness of encoding partial order information for zero-shot cross-lingual adaptation.

\paragraph{Fine-tuning M-BERT}
From Table~\ref{tab:zs-cross-lingual-ner-sf} and~\ref{tab:zs-cross-lingual-pos}, we observe that the results of fine-tuning M-BERT in the source language, English, are similar for both methods (less than 0.1\% difference) while freezing the positional embeddings in the fine-tuning stage generally brings better zero-shot cross-lingual performance in target languages.
Although positional embeddings are frozen, they can still provide order information for the model to encode sequences, which ensures the performance in English does not greatly drop. In the meantime, the positional embeddings are not affected by the English word order, and the order-agnostic trait of the positional embeddings is preserved, which boosts the generalization ability to target languages.

\begin{table}[t]
\centering
\setlength{\tabcolsep}{20pt}
\resizebox{0.63\textwidth}{!}{
\begin{tabular}{lcccc}
\toprule
\multicolumn{1}{c}{\multirow{2}{*}{\textbf{Models}}}  & \multicolumn{2}{c}{\textbf{Spanish}}                           & \multicolumn{2}{c}{\textbf{Thai}}              \\ \cmidrule(lr){2-3} \cmidrule(lr){4-5}
\multicolumn{1}{l}{}     & \textbf{ID}                    & \multicolumn{1}{c}{\textbf{SF}}  & \textbf{ID}                    & \textbf{SF}                      \\ \midrule
\multicolumn{1}{l}{\citet{liu2019zero}}    & 90.20                     & \multicolumn{1}{c}{65.79} & 73.43                     & 32.24  \\ 
\multicolumn{1}{l}{\quad using TRS}    & 89.71                     & \multicolumn{1}{c}{67.10} & 74.68                     & 31.20  \\
\multicolumn{1}{l}{\quad using ORT}    & \textbf{91.46}                     & \multicolumn{1}{c}{\textbf{71.36}} & \textbf{75.02}                     & \textbf{34.61}  \\ \bottomrule
\end{tabular}
}
\caption{Zero-shot results for the intent detection (ID) accuracy and slot filling (SF) F1-score on the semantic parsing task.}
\label{tab:semantic-parsing-sota-ort}
\end{table}

\begin{table}[t]
\centering
\setlength{\tabcolsep}{22pt}
\resizebox{0.65\textwidth}{!}{
\begin{tabular}{lcccc}
\toprule
\multicolumn{1}{c}{\textbf{Models}}  & \textbf{de}    & \textbf{es}                        & \multicolumn{1}{c}{\textbf{nl}}    & \textbf{avg}                   \\ \midrule
\multicolumn{1}{l}{\citet{chen2019multi}}  & 56.00                     & 73.50                     & \multicolumn{1}{c}{72.40} & 67.30                     \\ 
\multicolumn{1}{l}{\quad using TRS}  & 56.89                     & 73.72                     & \multicolumn{1}{c}{72.22} & 67.61                     \\ 
\multicolumn{1}{l}{\quad using ORT}  & \textbf{58.97}                     & \textbf{74.65}                     & \multicolumn{1}{c}{\textbf{72.56}} & \textbf{68.73}           \\ \bottomrule
\end{tabular}
}
\caption{Zero-shot results on the NER task.}
\label{tab:ner-sota-ort}
\end{table}

\paragraph{Applying ORT to Strong Models}
As shown in Table~\ref{tab:semantic-parsing-sota-ort} and~\ref{tab:ner-sota-ort}, we leverage ORT to replace the order-sensitive encoder, BiLSTM, in the strong zero-shot cross-lingual sequence labeling models proposed in~\citet{liu2019zero} and \citet{chen2019multi}. The zero-shot cross-lingual semantic parsing model proposed in~\citet{liu2019zero} is the current state-of-the-art for the multilingual semantic parsing dataset~\cite{schuster2019crosslingual}, and the model in~\citet{chen2019multi} achieves promising results in the zero-shot cross-lingual NER task. As we can see, replacing the order-sensitive encoders in their models with ORT can still boost the performance. We conjecture that since there are always cross-lingual performance drops caused by word order differences, reducing the word order of the source language fitted into the model can improve the performance.

In addition, we observe that the performance stays similar when we replace BiLSTM with TRS, which illustrates that the performance improvement made by ORT does not come from TRS, but from the model's insensitivity to word order.

\paragraph{Improvements vs. Language Distance}
As we can see from Table~\ref{tab:zs-cross-lingual-pos}, our proposed order-reduced models (e.g., ORT and TRS w/ M-BERT PE) outperform baseline models in almost all languages.
In the word-level embeddings setting, we observe that languages that are closer to English benefit more from ORT, since most numbers denoted with `$\ddagger$' come from languages that are close to English. We conjecture that ORT predicts the label of a token based on the local information of this token (the token itself and its neighbor tokens), and languages that are closer to English could have a more similar local word order to English. Interestingly, for M-BERT fine-tuning and models using M-BERT embeddings, we can see that languages that are further from English benefit more from order-reduced models.
We speculate that the alignment quality of topologically close languages in M-BERT is generally good, and the cross-lingual performance for the languages that are closer to English is also overall satisfactory, which narrows the improvement space for these languages. In contrast, the performance boost for languages that are further from English becomes larger. Surprisingly, freezing PE in the M-BERT fine-tuning significantly improves the performance of Arabic (ar). Given that the cross-lingual performance of the original M-BERT fine-tuning in Arabic is relatively low, we conjecture that one of the major reasons comes from the word order discrepancy between English and Arabic, which leads to a large performance improvement made by freezing the M-BERT PE.

\begin{table}[t!]
\centering
\setlength{\tabcolsep}{20pt}
\resizebox{0.6\textwidth}{!}{
\begin{tabular}{cccc}
\toprule
\multicolumn{1}{c}{Models}  & $k=0$ & $k=1$   & $k=2$  \\ \midrule
\multicolumn{1}{c}{BiLSTM} & \textbf{94.87} & 85.16    & 83.68  \\ 
\multicolumn{1}{c}{TRS}    &94.78 & 84.56               & 83.06  \\
\multicolumn{1}{c}{RPT} 
& 94.23 & 84.93 & 83.86  \\ \midrule
\multicolumn{1}{c}{ORT}   &94.50 & \textbf{87.87}          & \textbf{86.95}  \\ \bottomrule
\end{tabular}
}
\caption{Slot F1-scores on different noisy SF test sets in English. $k=0$ denotes the original English test set.
}
\label{tab:sf-shuffle-en}
\end{table}

\paragraph{How Order-Insensitive Is Our Model?}
To test the word order insensitivity of ORT, we follow the order shuffled methods in Section~\ref{sec:order-shuffle}, and set $k=1$ and $k=2$ to slightly shuffle the word order of the sequences and create a noisy English test set. As we can see from Table~\ref{tab:sf-shuffle-en}, ORT achieves better results than BiLSTM, TRS, and RPT on the noisy SF test set, which further illustrates that ORT is more insensible and resistant to word order differences than the baseline encoders. This property improves the generalization ability of ORT to target language word orders.

\begin{figure}[t!]
\centering
\subfigure[Performance difference on noun.]{
    \includegraphics[scale=0.53]{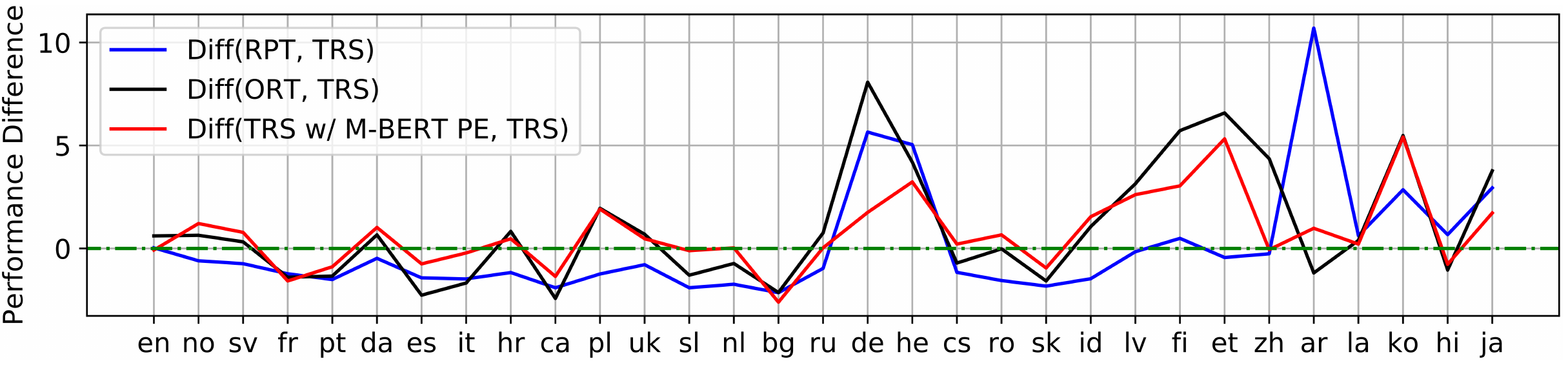}
}
\subfigure[Performance difference on verb.]{
    \includegraphics[scale=0.53]{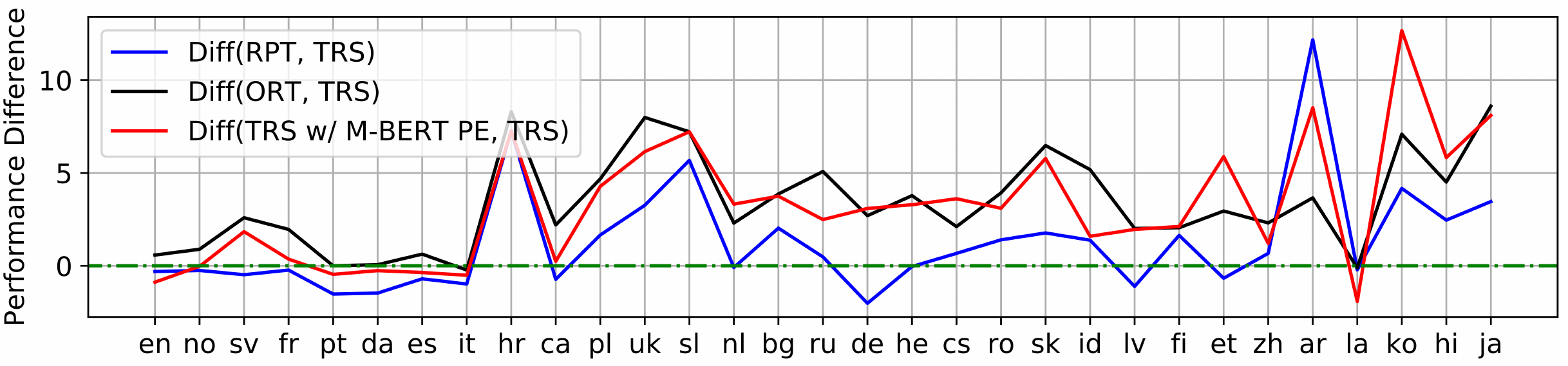}
}
\caption{Analysis on specific parts of speech. Languages are sorted by the word-ordering distance to English. We use Diff(A, B) to denote how much A outperforms B.}
\label{fig:performance_by_type}
\end{figure}

\paragraph{Performance Breakdown by Type}
We compare models (based on the frozen M-BERT embeddings) on specific POS types for the POS task. From Figure~\ref{fig:performance_by_type}, we observe that, in general, languages that are further from English benefit more from our proposed order-reduced models, which accords with the findings in Table~\ref{tab:zs-cross-lingual-pos}. Interestingly, we find that the improvements of order-reduced models (ORT and TRS w/ M-BERT PE) on the verb are larger than on the noun. 
We speculate that the order of the verb's surrounding words is different across languages, while the set of its surrounding words is more likely to remain the same or similar across languages at the semantic-level. This boosts the benefits of our proposed models, especially for the ORT which relies greatly on the extracted n-gram features from the neighboring words for the prediction.
Additionally, we find that the improvement made by RPT is also more significant on the verb than on the noun. We conjecture that the verb's relative positions to other parts of speech is more similar across languages compared to that of a noun. 

\subsubsection{Few-shot Adaptation}
\begin{figure}[t!]
\centering
\subfigure[Few-shot F1-scores for Spanish.]{
    \includegraphics[scale=0.58]{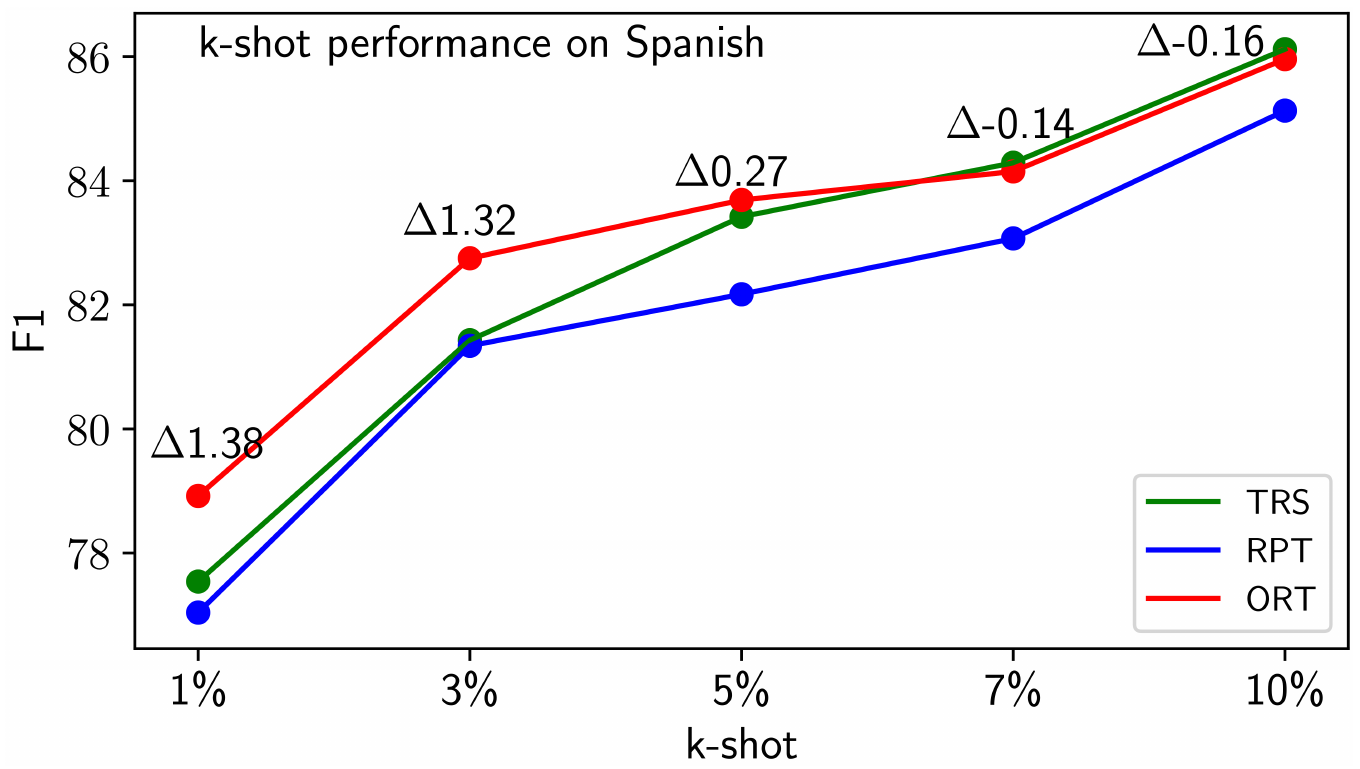}
}
\subfigure[Few-shot F1-scores for Thai.]{
    \includegraphics[scale=0.58]{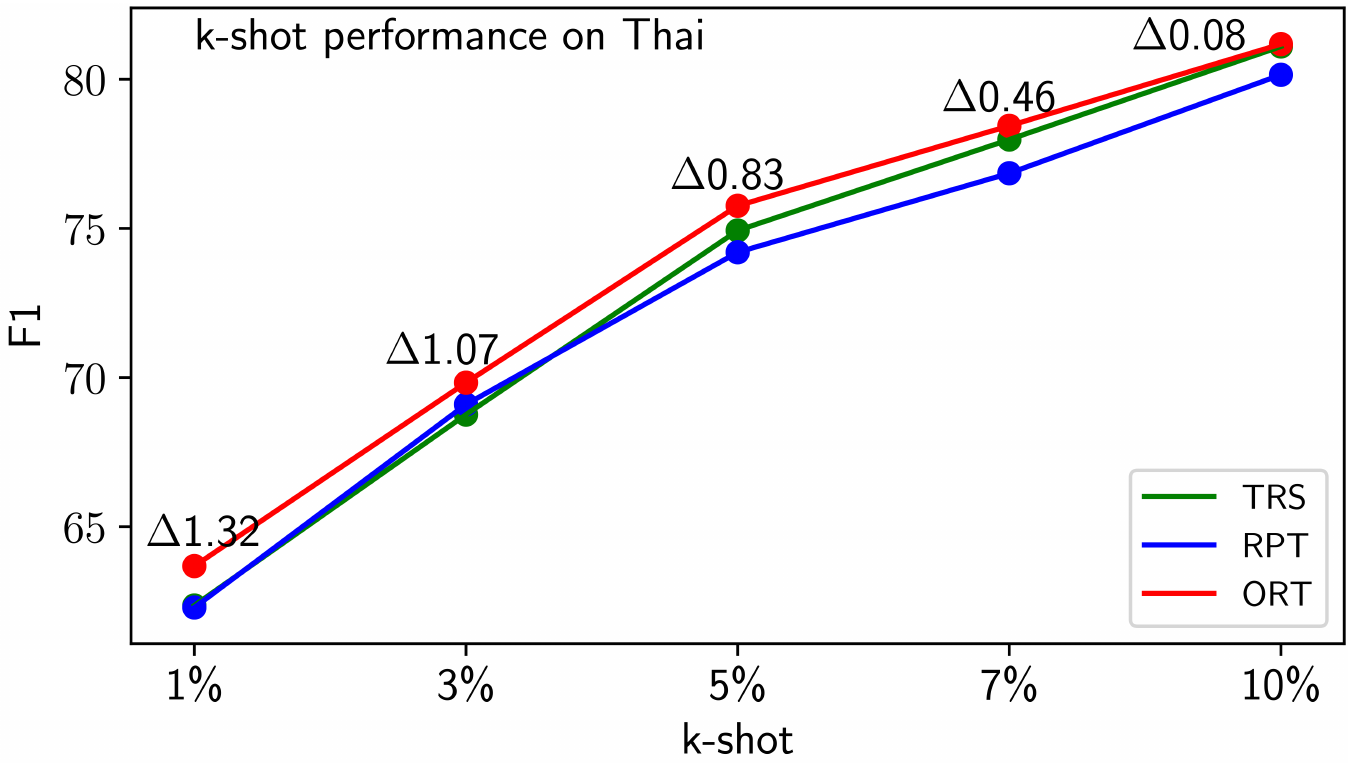}
}
\caption{Few-shot F1-scores for the SF task for Spanish and Thai. The x-axis represents the proportion of target language training samples in the training set. The numbers with $\Delta$ denote how much ORT outperforms TRS.}
\label{fig:fewshot-sf}
\end{figure}

Since we do not observe the order information for target languages in the zero-shot scenario, the order-reduced models will have a more robust adaptation ability. Then, the question we want to ask is whether order-reduced models can still improve the performance if a few training samples in target languages are available. 
We test with different numbers of target language training samples for the SF task, and the results are shown in Figure~\ref{fig:fewshot-sf}. 
We observe that the improvements in the few-shot scenarios are lower than in the zero-shot scenario, where ORT improves over TRS by 4.17\% and 3.22\% on Spanish and Thai, respectively (from Table~\ref{tab:zs-cross-lingual-ner-sf}), and as the proportion of target language training samples goes up, the improvement made by ORT goes down. This is because the model is able to learn the target language word order based on the target language training samples, which decreases the advantages of the order-reduced models. 
We also observe that RPT generally achieves worse performance than TRS, and we conjecture that RPT requires more training samples to learn the relative word order information than TRS, which lowers its generalization ability to the target language in the few-shot scenario.

\subsubsection{Ablation Study}
In this part, we explore the model variations in terms of positional embeddings, the feed-forward layer for TRS and ORT, adding different permutations to the shuffled word order, and whether to use the CRF layer.
We test the models' zero-shot performance on the SF task for this ablation study, and results are illustrated in Table~\ref{tab:ablation-study-sf}. 

\paragraph{Positional Embeddings}
We observe that TRS+CRF using trainable positional embeddings achieves similar performance to using sinusoidal positional embeddings. 

\paragraph{Feed-forward Layer}
We can see that the performance of TRS+CRF using linear layers as the feed-forward layer is on par with using Conv1d. We conjecture that the reason is that the positional embeddings in TRS have already encoded the word order information of the whole input sequence which makes the type of feed-forward layer less important.
However, when we replace Conv1d with linear layers for the feed-forward layer in ORT+CRF, the performance greatly drops ($\sim$8.5\% F1-score drops for Spanish and $\sim$5\% F1-score drops for Thai), and the performance continues to significantly drop when the CRF layer is removed (ORT+Linear).
This is because ORT can not encode any order information when the feed-forward layer Conv1d is replaced with the linear layer, and not any word order information is injected into the ORT+Linear model in which the CRF layer is removed. This makes the model perform badly in the source language and then weakens its adaptation ability to target languages.
In addition, we observe that the Conv1d feed-forward layer is also important for TRS trained with order-shuffled data. This is because Conv1d encodes the order of tokens in the entity (we do not shuffle the tokens in an entity), which is essential for detecting entities.

\begin{table}[t!]
\centering
\setlength{\tabcolsep}{20pt}
\resizebox{.9\textwidth}{!}{
\begin{tabular}{lccccc}
\toprule
\multicolumn{1}{l}{\textbf{Models}} & \textbf{PE}   & \textbf{Feed-forward}  & \textbf{k} &  \textbf{es}    & \textbf{th}    \\ \midrule
TRS + CRF           & Trainable & Linear    & -  & 62.13 & 22.68 \\ 
TRS + CRF           & Sinusoid  & Linear & -     & 62.55 & 21.82 \\
\quad w/ shuffled data          & Sinusoid & Linear  &  $\infty$   & 58.89 & 19.27 \\ \midrule
TRS + Linear           & Sinusoid  & Conv1d   & -    & 55.40 & 19.33 \\
TRS + CRF           & Sinusoid  & Conv1d   & -    & 62.67 & 22.33 \\
\quad w/ shuffled data  & Sinusoid & Conv1d   & 2   & 61.12 & 21.24 \\ 
\quad w/ shuffled data & Sinusoid & Conv1d   & 3   & 63.20 & 23.34 \\ 
\quad w/ shuffled data & Sinusoid & Conv1d   & 4   & 63.54 & 23.59 \\
\quad w/ shuffled data & Sinusoid & Conv1d   & $\infty$   & \textbf{63.86} & \textbf{24.17} \\ \midrule
ORT + Linear         & -         & Linear & -     & 39.65 & 13.52 \\
ORT + CRF         & -         & Linear & -     & 58.27 & 20.35 \\ 
ORT + Linear         & -         & Conv1d & -     & 61.76 & 22.44 \\
ORT + CRF     & -         & Conv1d  & -    & \textbf{66.84} & \textbf{25.53} \\ \bottomrule
\end{tabular}
}
\caption{Ablation study on positional embeddings, feed-forward layer, word order shuffling, and CRF layer. Results are the F1-scores for the zero-shot SF task.
``-'' denotes that the model does not have this module. ``+CRF'' and ``+Linear'' denotes using and not using the CRF layer, respectively.}
\label{tab:ablation-study-sf}
\end{table}

\paragraph{Different Permutations of Word Order}
We try using different permutations (changing the value of k) of the word order to generate order-shuffled data.
As we can see, when we slightly shuffle the word order ($k=2$), the performance becomes worse than not using order-shuffled data. This is because the model fits the slightly shuffled word order, which is not similar to that of the target languages. After more perturbations are added to the word order, TRS becomes more robust to order differences.

\paragraph{Effectiveness of the CRF Layer}
For sequence labeling tasks, the CRF layer, which models the conditional probability of label sequences, could also implicitly model the source language word order in training. Therefore, we conduct an ablation study to test the effectiveness of the CRF layer for the cross-lingual models.
From Table~\ref{tab:ablation-study-sf}, we can see that removing the CRF layer makes the performance worse. We conjecture that although the CRF layer might contain some information on the word order pattern in the source language, 
It also models the conditional probability for tokens that belong to the same entity so that it learns when the start or the end of an entity is. This is important for sequence labeling tasks, and models that have the CRF layer removed might not have this ability.
For example, in the SF task, when the user says ``set an alarm for 9 pm'', ``for 9 pm'' belongs to the ``DateTime'' entity, and the CRF layer learns to model ``for'' and ``pm'' as the start and end of the ``DateTime'' entity, respectively. Without the CRF layer, models treat the features of these tokens independently.

\begin{figure}[t!]
\centering
\includegraphics[scale=0.7]{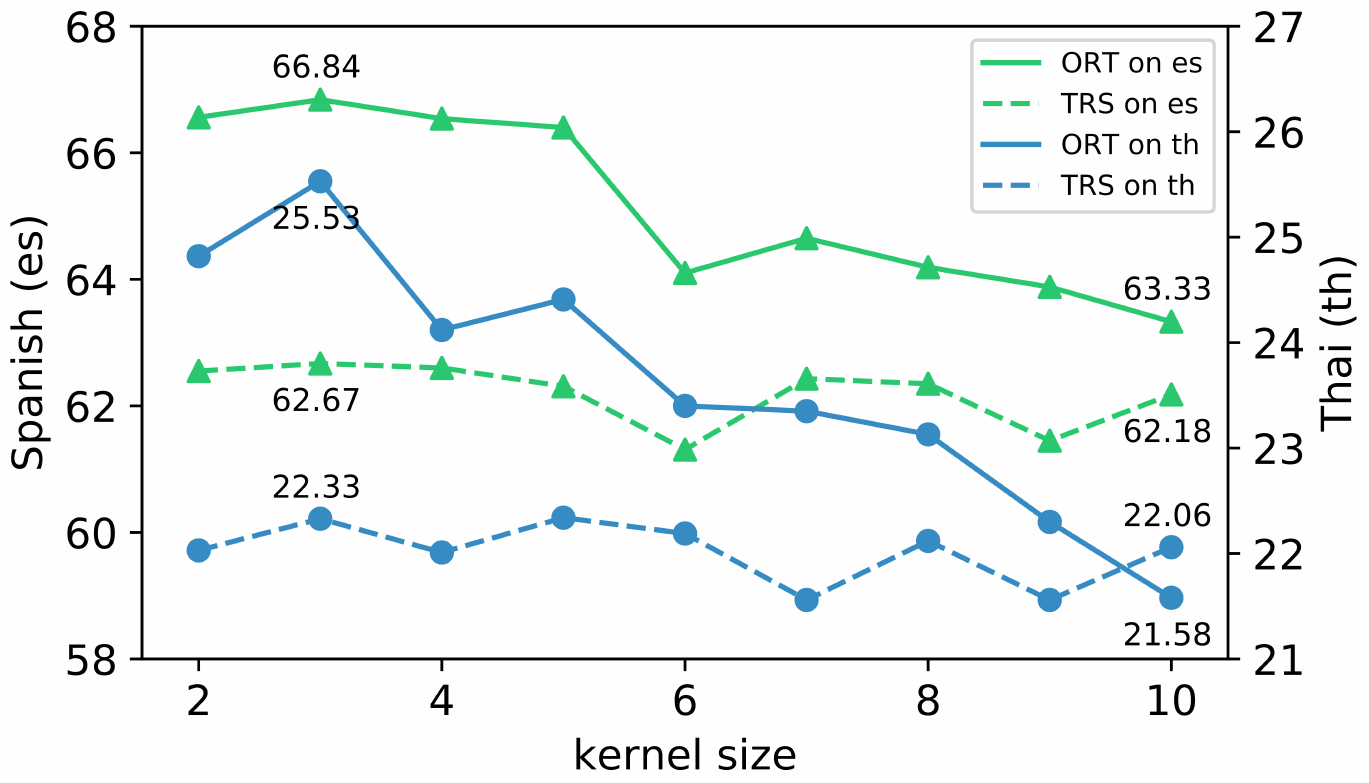}
\caption{Zero-shot results on the SF task with different kernel sizes for ORT and TRS.}
\label{fig:kerne_size_sf}
\end{figure}

\paragraph{Kernel Size vs. Performance}
Since the kernel size of Conv1d represents the amounts of local word order information that ORT encodes, we explore how the kernel size affects the performance.
As shown in Figure~\ref{fig:kerne_size_sf}, with the increase of kernel size, the zero-shot performance of ORT decreases, and the performance of ORT becomes similar to TRS's when the kernel size is 10. This is because the larger the kernel size, the more order information the model will encode. Hence, the model's generalization ability to target languages decreases when the kernel size is too large.

\subsection{Short Summary}
In this section, we focus on tackling the challenge of language discrepancies in the cross-lingual transfer.
We investigate whether reducing the word order of the source language fitted into the models can improve cross-lingual sequence labeling performance. 
We propose several methods to build order-reduced models, and then compare them with order-sensitive baselines. 
Extensive experimental results show that our proposed order-reduced Transformer (ORT) is robust to the word order shuffled sequences, and it consistently outperforms the order-sensitive models as well as relative positional Transformer (RPT). This confirms that modeling partial word orders instead of the whole sequence can improve the robustness of the model against the word order differences between languages and the task knowledge transfer to low-resource languages.
Additionally, preserving the order-agnostic property for the M-BERT positional embeddings gives the model a better generalization ability to target languages. Furthermore, we observe that encoding excessive or insufficient word order information leads to inferior cross-lingual performance, and models that do not encode any word order information perform badly in both source and target languages.

\newpage

%% file: chapter/sec-4-crossdomain.tex
\chapter{Adaptation to Low-Resource Target Domains} \label{sec-crossdomain}
Deep learning methods generally suffer from the inferior performance in low-resource domains due to the data scarcity issue. 
Therefore, enabling models to quickly adapt to low-resource target domains where only a few or even zero training samples are available has become an essential yet challenging task. 
The major challenges of low-resource domain adaptation are two-fold: first, how to effectively learn the low-resource representations and quickly adapt to the unseen label types in diverse target domains where the training samples are insufficient; second, how to cope with the performance drop caused by large domain discrepancies and enable an effective task knowledge transfer from source to target domains. 
In this chapter, we focus on tackling these two challenges, and we will present the two key contributions in the following sections:
\begin{itemize}
    \item We introduce a coarse-to-fine framework, Coach, to improve the effectiveness of low-resource representation learning. Coach decomposes the representation learning process into a coarse-grained and a fine-grained feature learning, and we find that simplifying task structures makes the representation learning more effective for low-resource domains. Results show that our framework achieves better domain adaptation performance than previous state-of-the-art models, and it is especially effective for unseen label types.
    
    \item We collect a multi-domain NER dataset spanning over five diverse domains with specialized entity categories to catalyze the cross-domain NER research. We propose to leverage different levels of the domain corpus and a more challenging pre-training strategy to do domain-adaptive pre-training (DAPT). We find that focusing on the corpus with domain-specialized entities and utilizing a more challenging pre-training strategy can better address the domain discrepancy issue in the task knowledge transfer and bring a performance boost for distant target domains.
    
\end{itemize}

\newpage

\begin{figure}[t!]
\centering
\includegraphics[width=.8\linewidth]{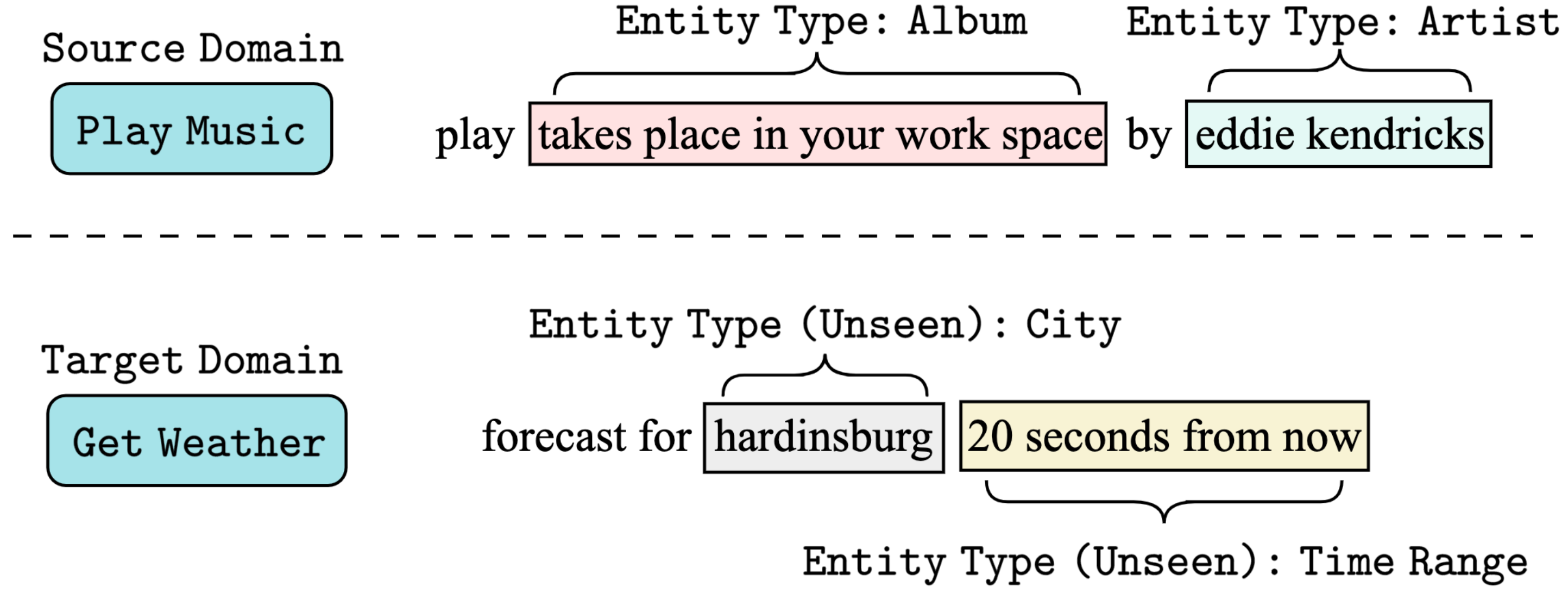}
\caption{An example of categorizing unseen entity types in target domains.}
\label{fig:unseen-entity-example}
\end{figure}

\section{Coarse-to-Fine Cross-Domain Framework}
In this section, we propose a \textbf{Coa}rse-to-fine approa\textbf{ch} (\textbf{Coach}) framework which focuses on learning effective low-resource representations and enabling a fast adaptation to unseen label types in low-resource target domains. 
Coach decomposes the representation learning process into a coarse-grained and a fine-grained feature learning. It first learns the general pattern of entities by detecting whether the tokens are entities or not. Then, it predicts the specific types for the detected entities. In addition, we propose a \textbf{template regularization} method to improve the adaptation robustness by regularizing the representation of input sentences based on templates. 
Figure~\ref{fig:unseen-entity-example} illustrates the challenges of categorizing unseen entity types in the cross-domain adaptation.

\subsection{Motivation}
\begin{figure}[!t]
\centering
\subfigure[Framework proposed by~\citet{bapna2017towards}.]{
    \includegraphics[scale=0.825]{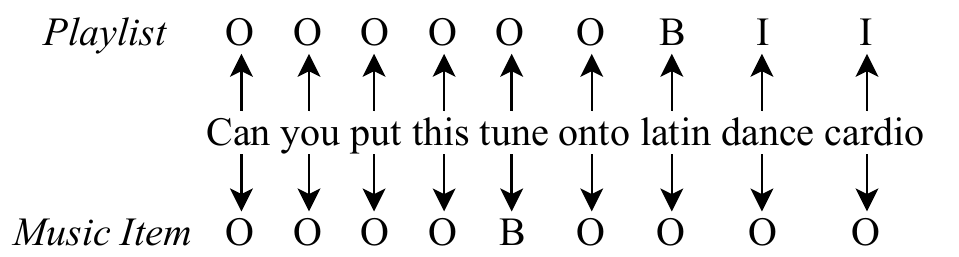}
}
\subfigure[Our proposed framework, Coach.]{
    \includegraphics[scale=0.825]{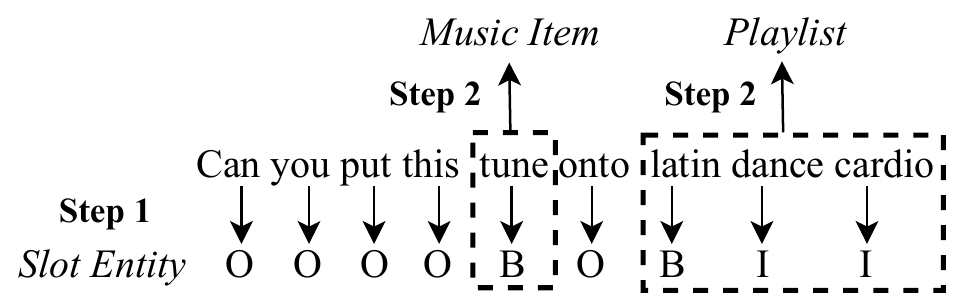}
}
\caption{Cross-domain sequence tagging frameworks.}
\label{fig:cross-domain-sequence-tagging-frameworks}
\end{figure}

Recently, \citet{bapna2017towards} proposed a cross-domain entity tagging framework, which, as shown in Figure~\ref{fig:cross-domain-sequence-tagging-frameworks}, conducts entity tagging individually for each slot entity type using the standard Begin-Inside-Outside (BIO) structure.
Their framework first generates word-level representations of input sequences, which are then concatenated with the representation of each entity type description, and the predictions are based on the concatenated features for each entity type. Due to the inherent variance of slot entities across domains, it is difficult for this framework to capture the whole slot entity (e.g., ``latin dance cardio'' in Figure~\ref{fig:cross-domain-sequence-tagging-frameworks}) in the target domain. There also exists a multiple prediction problem. For example, ``tune'' in Figure~\ref{fig:cross-domain-sequence-tagging-frameworks} could be predicted as ``B'' for both ``music item'' and ``playlist'', which would cause additional trouble for the final prediction.

We emphasize that in order to capture the whole slot entity, it is pivotal for the model to share its parameters for all slot types in the source domains and learn the general pattern of slot entities.
Therefore, as depicted in Figure~\ref{fig:cross-domain-sequence-tagging-frameworks}, we propose a new cross-domain tagging framework called Coach, a coarse-to-fine approach. It first \textbf{coarsely} learns the slot entity pattern by predicting whether the tokens are slot entities or not. Then, it combines the features for each slot entity and predicts the specific (\textbf{fine}) entity type based on the similarity with the representation of each entity type description. Moreover, in this way, our framework is able to avoid the multiple predictions problem.

\subsection{Model Description}
\begin{figure}[!t]
    \centering
    \includegraphics[scale=0.8]{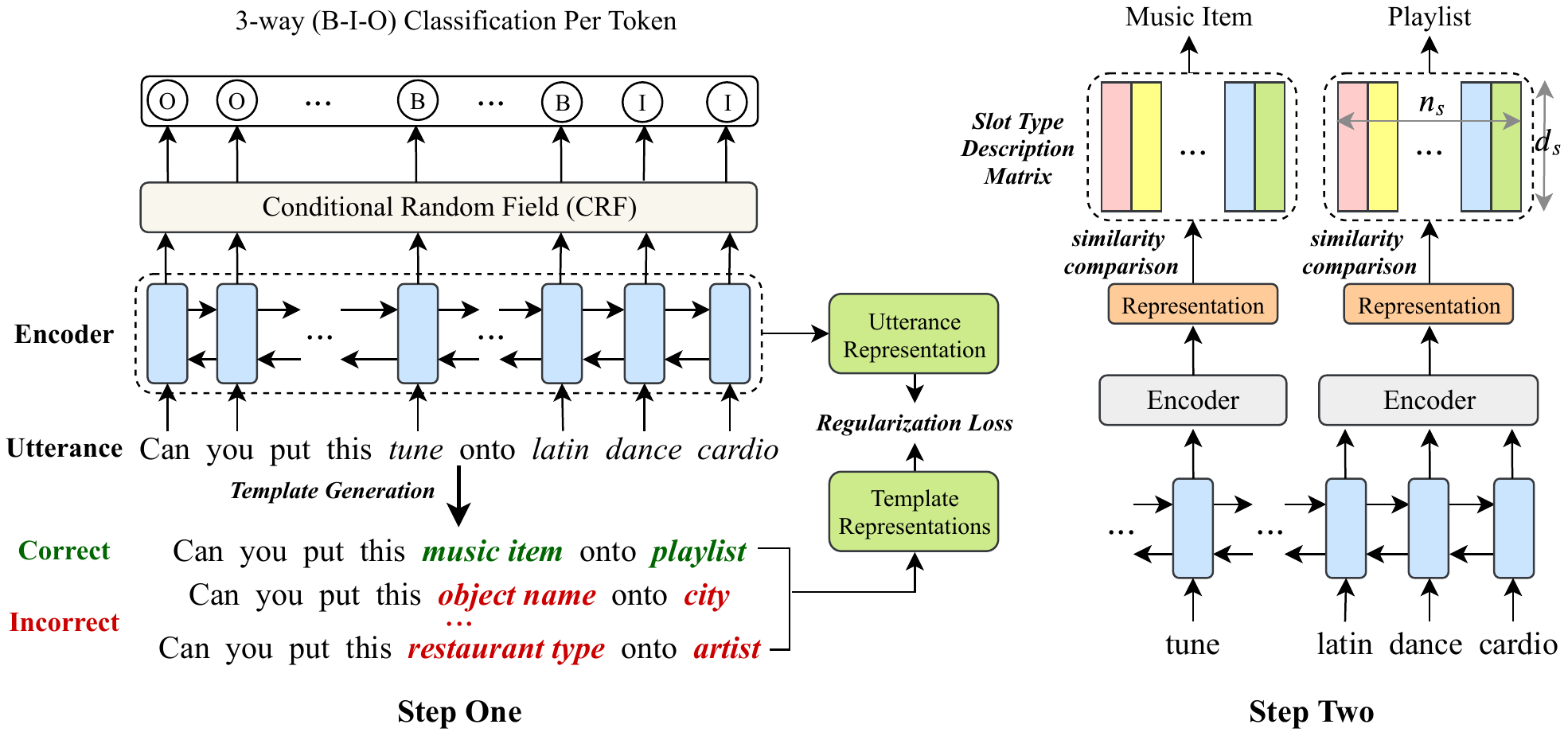}
    \caption{Illustration of our framework, Coach, and the template regularization approach.}
    \label{fig:coach-framework}
\end{figure}

Figure~\ref{fig:coach-framework} provides an illustration of our Coach framework, and the template regularization approach. 
The entity tagging process in our Coach framework consists of two steps. In the first step, we utilize a BiLSTM-CRF structure~\cite{lample2016neural} to learn the general pattern of slot entities by having our model predict whether tokens are slot entities or not (i.e., 3-way classification for each token).
In the second step, our model further predicts a specific type for each slot entity based on the similarities with the description representations of all possible entity types. To generate representations of slot entities, we leverage another encoder, BiLSTM~\cite{hochreiter1997long}, to encode the hidden states of slot entity tokens and produce representations for each slot entity.

We represent the input sequence (utterance) with $n$ tokens as $\textbf{w} = [w_1, w_2, ..., w_{n}]$, and $\textbf{E}$ denotes the embedding layer for utterances. The whole process can be formulated as follows:
\begin{equation}
    [h_1, h_2, ..., h_n] = \textnormal{BiLSTM} (\textbf{E}(\textbf{w})),
\end{equation}
\begin{equation}
    [p_1, p_2, ..., p_n] = \textnormal{CRF}([h_1, h_2, ..., h_n]),
\end{equation}
where $[p_1, p_2, ..., p_n]$ are the logits for the 3-way classification. Then, for each slot entity, we take its hidden states to calculate its representation:
\begin{equation}
    r_k = \textnormal{BiLSTM}([h_i, h_{i+1},... h_j]), \label{eq:encode-entities}
\end{equation}
\begin{equation}
    s_k = M_{desc} \cdot r_k,  \label{eq:measure}
\end{equation}
where $r_k$ denotes the representation of the $k^{th}$ slot entity, $[h_i, h_{i+1}, ..., h_j]$ denotes the BiLSTM hidden states for the $k^{th}$ slot entity, $M_{desc} \in R^{n_s \times d_s} $ is the representation matrix of the entity description ($n_s$ is the number of possible entity types and $d_s$ is the dimension of entity descriptions), and $s_k$ is the specific entity type prediction for this $k^{th}$ slot entity. We obtain the slot description representation $ r^{desc} \in R^{d_s} $ by summing the embeddings of the N slot description tokens (similar to~\citet{shah2019robust}):
\begin{equation}
    r^{desc} = \sum_{i=1}^{N} \textbf{E}(t_i),
\end{equation}
where $t_i$ is the $i^{th}$ token and \textbf{E} is the same embedding layer as that for utterances.

\subsection{Template Regularization}
In some cases, similar or the same entity types in the target domain can also be found in the source domains.
Nevertheless, it is still challenging for the model to recognize the slot types in the target domain owing to the variance between the source domains and the target domain. To improve the adaptation ability, we introduce a template regularization method.

As shown in Figure~\ref{fig:coach-framework}, we first replace the slot entity tokens in the utterance with different entity labels to generate correct and incorrect utterance templates. Then, we use BiLSTM and an attention layer~\cite{felbo2017using} to generate the utterance and template representations:
\begin{equation}
    e_t = h_t w_a,~~ \alpha_t = \frac{exp(e_t)}{\sum_{j=1}^n exp(e_j)},~~ R=\sum_{t=1}^{n} \alpha_t h_t,
\end{equation}
where $h_t$ is the BiLSTM hidden state in the $t^{th}$ step, $w_a$ is the weight vector in the attention layer and $R$ is the representation for the input utterance or template. 

We minimize the regularization loss functions for the right and wrong templates, which can be formulated as follows:
\begin{equation}
    L^{r} = \textnormal{MSE} (R^{u}, R^{r}),
\end{equation}
\begin{equation}
    L^{w} = -\beta \times \textnormal{MSE} (R^{u}, R^{w}),
\end{equation}
where $R^{u}$ is the representation for the user utterance, $R^{r}$ and $R^{w}$ are the representations of right and wrong templates, we set $\beta$ as one, and \textnormal{MSE} denotes mean square error. Hence, in the training phase, we minimize the distance between $R^{u}$ and $R^{r}$ and maximize the distance between $R^{u}$ and $R^{w}$.
To generate a wrong template, we replace the correct entity with another random entity, and we generate two wrong templates for each utterance.
To ensure the representations of the templates are meaningful (i.e., similar templates have similar representations) for training $R^{u}$, in the first several epochs, the regularization loss is only to optimize the template representations, and in the following epochs, we optimize both template representations and utterance representations. 


By doing so, the model learns to cluster the representations in the same or similar templates into a similar vector space. Hence, the hidden states of tokens that belong to the same entity type tend to be similar, which boosts the robustness of these entity types in the target domain.

\subsection{Experimental Setup}
\subsubsection{Dataset}
We evaluate our framework on SNIPS~\cite{coucke2018snips}, a public semantic parsing dataset which contains 39 slot entity types across seven domains (intents) and around 2000 training samples per domain. To test our framework, each time, we choose one domain as the target domain and the other six domains as the source domains.

Moreover, we also study another adaptation case where there is no unseen label in the target domain. We utilize the CoNLL-2003 English named entity recognition (NER) dataset as the source domain~\cite{sang2003introduction}, and the CBS SciTech News NER dataset from~\citet{jia2019cross} as the target domain. These two datasets have the same four types of entities, namely, PER (person), LOC (location), ORG (organization), and MISC (miscellaneous).

\subsubsection{Baselines}

We use the following baselines to compare with our model:

\paragraph{Concept Tagger (CT)}
\citet{bapna2017towards} proposed a slot filling framework that utilizes entity descriptions to cope with the unseen slot types in the target domain.

\paragraph{Robust Zero-shot Tagger (RZT)}
Based on CT, \citet{shah2019robust} leveraged example values of entities to improve robustness of cross-domain adaptation. This is the previous state-of-the-art model in the cross-domain slot filling task.

\paragraph{BiLSTM-CRF}
This baseline is only for the cross-domain NER. Since there are no unseen labels in the NER target domain, the BiLSTM-CRF~\cite{lample2016neural} uses the same label set for the source and target domains and casts the task as an entity classification for each token. This makes the model applicable in both zero-shot and few-shot scenarios.

\subsubsection{Training}
We use a 2-layer BiLSTM with a hidden size of 200 and a dropout rate of 0.3 for both the template encoder and utterance encoder. Note that the parameters in these two encoders are not shared.
The BiLSTM for encoding the hidden states of slot entity tokens has one layer with a hidden size of 200, and the output is of the same dimension as the concatenated word-level and char-level embeddings. We use Adam optimizer with a learning rate of 0.0005. Cross-entropy loss is leveraged to train the 3-way classification in the first step, and the specific entity type predictions are used in the second step. We split 500 data samples in the target domain as the validation set for choosing the best model and the remainder are used for the test set. In the experiments, we use standard BIO structure to calculate the F1 scores.
We use word-level~\cite{bojanowski2017enriching} and character-level~\cite{hashimoto2017joint} embeddings for our model as well as all the baselines. The baseline models also use the same entity descriptions as our model for a fair comparison.

\begin{table}[!t]
\setlength{\tabcolsep}{8pt}
\centering
\resizebox{0.999\textwidth}{!}{
\begin{tabular}{lcccccccccccc}
\toprule
 \multicolumn{1}{l}{\multirow{2}{*}{\textbf{Domain}}}  & \multicolumn{4}{c}{\textbf{Zero-shot}} & \multicolumn{4}{c}{\textbf{Few-shot on 20 (1\%) samples}} & \multicolumn{4}{c}{\textbf{Few-shot on 50 (2.5\%) samples}}         \\ \cmidrule(lr){2-5} \cmidrule(lr){6-9} \cmidrule(lr){10-13}
\multicolumn{1}{l}{}  & \textbf{CT}     & \multicolumn{1}{c|}{\textbf{RZT}}   & \textbf{Coach}  & \textbf{+TR}   & \textbf{CT}        & \multicolumn{1}{c|}{\textbf{RZT}}    & \textbf{Coach}     & \textbf{+TR}      & \textbf{CT}    & \multicolumn{1}{c|}{\textbf{RZT}}   & \textbf{Coach}  & \textbf{+TR}   \\ \midrule
AddToPlaylist      & 38.82  & \multicolumn{1}{c|}{42.77} & 45.23 & \textbf{50.90} & 58.36     & \multicolumn{1}{c|}{\textbf{63.18}}     & 58.29    & 62.76    & 68.69 & \multicolumn{1}{c|}{\textbf{74.89}} & 71.63 & 74.68 \\
BookRestaurant     & 27.54  & \multicolumn{1}{c|}{30.68} & 33.45 & \textbf{34.01} & 45.65     & \multicolumn{1}{c|}{50.54}     & 61.08    & \textbf{65.97}    & 54.22 & \multicolumn{1}{c|}{54.49} & 72.19 & \textbf{74.82} \\
GetWeather         & 46.45  & \multicolumn{1}{c|}{50.28} & 47.93 & \textbf{50.47} & 54.22     & \multicolumn{1}{c|}{58.86}     & 67.61    & \textbf{67.89}    & 63.23 & \multicolumn{1}{c|}{58.87} & \textbf{81.55} & 79.64 \\
PlayMusic          & 32.86  & \multicolumn{1}{c|}{\textbf{33.12}} & 28.89 & 32.01 & 46.35     & \multicolumn{1}{c|}{47.20}     & 53.82    & \textbf{54.04}    & 54.32 & \multicolumn{1}{c|}{59.20} & 62.41 & \textbf{66.38} \\
RateBook           & 14.54  & \multicolumn{1}{c|}{16.43} & \textbf{25.67} & 22.06 & 64.37     & \multicolumn{1}{c|}{63.33}     & \textbf{74.87}    & 74.68    & 76.45 & \multicolumn{1}{c|}{76.87} & \textbf{86.88} & 84.62 \\
SearchCreativeWork & 39.79  & \multicolumn{1}{c|}{44.45} & 43.91 & \textbf{46.65} & 57.83     & \multicolumn{1}{c|}{\textbf{63.39}}     & 60.32    & 57.19    & 66.38 & \multicolumn{1}{c|}{\textbf{67.81}} & 65.38 & 64.56 \\
FindScreeningEvent & 13.83  & \multicolumn{1}{c|}{12.25} & \textbf{25.64} & 25.63 & 48.59     & \multicolumn{1}{c|}{49.18}     & 66.18    & \textbf{67.38}    & 70.67 & \multicolumn{1}{c|}{74.58} & 78.10 & \textbf{83.85} \\ \midrule
\textbf{Average F1}            & 30.55  & \multicolumn{1}{c|}{32.85} & 35.82 & \textbf{37.39} & 53.62     & \multicolumn{1}{c|}{56.53}     & 63.17    & \textbf{64.27}    & 64.85 & \multicolumn{1}{c|}{66.67} & 74.02 & \textbf{75.51} \\ \bottomrule
\end{tabular}
}
\caption{F1-scores on SNIPS. Scores in each row represents the performance of the leftmost target domain, and TR denotes template regularization.}
\label{tab:cross-domain-semantic-parsing}
\end{table}

\subsection{Results and Discussion}

\subsubsection{Quantitative Analysis}
As illustrated in Table~\ref{tab:cross-domain-semantic-parsing}, we can clearly see that our models are able to achieve significantly better performance than the previous state-of-the-art approach, RZT.
The CT framework suffers from the difficulty of capturing the whole slot entity, while our framework is able to recognize the slot entity tokens by sharing its parameters across all slot types. Based on the CT framework, the performance of RZT is still limited, and Coach outperforms RZT by a $\sim$3\% F1-score in the zero-shot setting. Additionally, template regularization further improves the adaptation robustness by helping the model cluster the utterance representations into a similar vector space based on their corresponding template representations.

Interestingly, our models achieve impressive performance in the few-shot scenario. In terms of the averaged performance, our best model (Coach+TR) outperforms RZT by $\sim$8\% and $\sim$9\% F1-scores on the 20-shot and 50-shot settings, respectively.
We conjecture that our model is able to better recognize the whole slot entity in the target domain and map the representation of the slot entity belonging to the same slot type into a similar vector space to the representation of this slot type based on Eq (\ref{eq:measure}). This enables the model to quickly adapt to the target domain slots.

\begin{table}[!t]
\setlength{\tabcolsep}{25pt}
\centering
\resizebox{0.7\textwidth}{!}{
\begin{tabular}{lcc}
\toprule
\textbf{Models}    & \textbf{0 samples}    & \textbf{50 samples}   \\ \midrule
CT~(\citet{bapna2017towards})  & 61.43 & 65.85 \\
RZT~(\citet{shah2019robust})   & 61.94    & 65.21     \\ 
BiLSTM-CRF        & 61.77    & 66.57     \\ \midrule
Coach & 64.08   & \textbf{68.35}     \\
Coach + TR   & \textbf{64.54}   & 67.45     \\ \bottomrule
\end{tabular}
}
\caption{F1-scores on the NER target domain (CBS SciTech News).}
\label{tab:cross-domain-named-entity-recognition}
\end{table}

We further evaluate our model on the scenario where there are no unseen entity types in the target domain. From Table~\ref{tab:cross-domain-named-entity-recognition}, we see that our Coach framework is also suitable for the case where all entity types in the target domain exist in the source domain, in terms of both zero-shot and few-shot scenarios, while CT and RZT are not as effective as BiLSTM-CRF. 
However, we observe that template regularization loses its effectiveness in this task, since the text in NER is relatively more open, which makes it hard to capture the templates for each entity label type.

\subsubsection{Analysis on Seen and Unseen Entities}
We take a further step to test the models on seen and unseen entities in target domains to analyze the effectiveness of our approaches. To test the performance, we split the test set into ``unseen'' and ``seen'' parts. An utterance is categorized into the ``unseen'' part as long as there is an unseen entity type (i.e., the entity type does not exist in the remaining six source domains) in it. Otherwise we categorize it into the ``seen'' part. The results for the ``seen'' and ``unseen'' categories are shown in Table~\ref{tab:cross-domain-seen-unseen}.
We observe that our approaches generally improve on both unseen and seen entity types compared to the baseline models. For the improvements in the unseen slots, our models are better able to capture the unseen slots since they explicitly learn the general pattern of slot entities.
Interestingly, our models also bring large improvements in the seen entity types. We conjecture that it is also challenging to adapt models to seen entities due to the large variance between the source and target domains. For example, slot entities belonging to the ``object type'' in the ``RateBook'' domain are different from those in the ``SearchCreativeWork'' domain. Hence, the baseline models might fail to recognize these seen entities in the target domain, while our approaches can adapt to the seen entity types more quickly in comparison.

In addition, we observe that our template regularization method improves performance in both seen and unseen entities, which illustrates that clustering representations based on templates can boost the adaptation ability.

\begin{table}[!t]
\setlength{\tabcolsep}{20pt}
\centering
\resizebox{0.9\textwidth}{!}{
\begin{tabular}{lcccccc}
\toprule
\multirow{2}{*}{\textbf{Models}} & \multicolumn{2}{c}{\textbf{0 samples}} & \multicolumn{2}{c}{\textbf{20 samples}} & \multicolumn{2}{c}{\textbf{50 samples}} \\ \cmidrule(lr){2-3} \cmidrule(lr){4-5} \cmidrule(lr){6-7}
& \textbf{unseen}         & \textbf{seen}          & \textbf{unseen}          & \textbf{seen}          & \textbf{unseen}         & \textbf{seen}          \\ \midrule
CT        & 27.10           & 44.18         & 50.13           & 61.21         & 62.05          & 69.64         \\
RZT        & 28.28          & 47.15         & 52.56           & 63.26         & 63.96          & 73.10          \\ \midrule
Coach      & 32.89          & 50.78         & 61.96           & 73.78         & 74.65          & 76.95         \\
Coach+TR    & \textbf{34.09}          & \textbf{51.93}         & \textbf{64.16}           & \textbf{73.85}         & \textbf{76.49}          & \textbf{80.16}         \\ \bottomrule
\end{tabular}
}
\caption{Averaged F1-scores for seen and unseen slot entities over all target domains. The number of samples denotes the training samples used for the target domain.}
\label{tab:cross-domain-seen-unseen}
\end{table}

\begin{table}[!t]
\setlength{\tabcolsep}{20pt}
\centering
\resizebox{0.99\textwidth}{!}{
\begin{tabular}{lcccccc}
\toprule
\multirow{2}{*}{\textbf{Task}} & \multicolumn{3}{c}{\textbf{zero-shot}} & \multicolumn{3}{c}{\textbf{few-shot on 50 samples}} \\ \cmidrule(lr){2-4}  \cmidrule(lr){5-7}
& \textbf{sum}      & \textbf{trs}      & \textbf{bilstm}     & \textbf{sum}          & \textbf{trs}          & \textbf{bilstm}         \\ \midrule
SNIPS          & 33.89    & 34.33    & \textbf{35.82}    & 73.80         & 72.66        & \textbf{74.02}        \\ 
NER (CBS SciTech News)        & 63.04    & 63.29    & \textbf{64.47}    & 66.98        & 68.04        & \textbf{68.35}        \\ \bottomrule
\end{tabular}
}
\caption{Ablation study in terms of the methods to encode the entity tokens on Coach. The results for SNIPS are averaged over all domains.}
\label{tab:ablation-study-methods-of-encode-entity-tokens}
\end{table}

\subsubsection{Ablation Study}
We conduct an ablation study in terms of the methods to encode the entity tokens (described in Eq. (\ref{eq:encode-entities})) to investigate how they affect the performance. Instead of using BiLSTM, we try two alternatives. One is to use the encoder of Transformer (TRS)~\cite{vaswani2017attention}, and the other is to simply sum the hidden states of slot entity tokens. From Table~\ref{tab:ablation-study-methods-of-encode-entity-tokens}, we can see that there is no significant performance difference among different methods, and we observe that using BiLSTM to encode the entity tokens generally achieves better results.

\subsection{Short Summary}
We introduce a novel coarse-to-fine framework, Coach, for the cross-domain slot filling, which aims to improve the low-resource representation learning and address the issue of categorizing unseen entity types in target domains. 
Coach shares its parameters across all entity types and learns to predict whether input tokens are entities or not. Then, it detects concrete entity types for these entity tokens based on the entity type descriptions. Moreover, we propose template regularization to further improve the adaptation robustness.
We find that simplifying task structures makes the representation learning more effective for low-resource domains.
Experimental results show that our model significantly outperforms existing cross-domain slot filling approaches, and it also achieves better performance for the cross-domain NER task.

\newpage

\section{Domain-Adaptive Pre-Training}
In this section, we first present a cross-domain named entity recognition (NER) dataset (CrossNER), a fully-labeled collection of NER data spanning over five diverse domains with specialized entity categories for different domains, to catalyze the cross-domain NER research. 
Additionally, we also collect an unlabeled domain-related corpus for each domain. After that, we conduct comprehensive experiments to explore the effectiveness of leveraging different levels of the domain corpus and pre-training strategies to perform domain-adaptive pre-training (DAPT) for the cross-domain task.

\subsection{The CrossNER Dataset}

\subsubsection{Motivation}
Previous cross-domain NER studies~\cite{jia2020multi,jia2019cross,yang2017transfer} consider the CoNLL2003 English NER dataset~\cite{sang2003introduction} from Reuters News as the source domain, and utilize NER datasets from Twitter~\cite{derczynski2016broad,lu2018visual}, biomedicine~\cite{nedellec2013overview} and CBS SciTech News~\cite{jia2019cross} as target domains. 
However, we find two drawbacks in utilizing these datasets for cross-domain NER evaluation.
First, most target domains are either close to the source domain or not narrowed down to a specific topic or domain. Specifically, the CBS SciTech News domain is close to the Reuters News domain (both are related to news) and the content in the Twitter domain is generally broad since diverse topics will be tweeted about and discussed on social media. 
Second, the entity categories for the target domains are limited. Except the biomedical domain, which has specialized entities in the biomedical field, the other domains (i.e., Twitter and CBS SciTech News) only have general categories, such as person and location. However, we expect NER models to recognize certain entities related to target domains.

To address the drawbacks, we introduce a new human-annotated cross-domain NER dataset, dubbed CrossNER, which contains five diverse domains, namely, politics, natural science, music, literature and artificial intelligence (AI). Each domain has particular entity categories; for example, there are ``politician'', ``election'' and ``political party'' categories specialized for the politics domain. As in previous works, we consider the CoNLL2003 English NER dataset as the source domain, and the five domains in CrossNER as the target domains.
We collect $\sim$1000 development and test examples for each domain and a small number of data samples (100 or 200) in the training set for each domain since we consider a low-resource scenario for target domains.
In addition, we collect five corresponding unlabeled domain-related corpora for the DAPT, given its effectiveness for domain adaptation~\cite{beltagy2019scibert,donahue2019lakhnes,lee2020biobert,gururangan2020don}.

\subsubsection{Unlabeled Corpora Collection}
To collect CrossNER, we first construct five unlabeled domain-specific (politics, natural science, music, literature and AI) corpora from Wikipedia.  
Wikipedia contains various categories and each category has further subcategories. It serves as a valuable source for us to collect a large corpus related to a certain domain. For example, to construct the corpus in the politics domain, we gather Wikipedia pages that are in the politics category as well as its subcategories, such as political organizations and political cultures. We utilize these collected corpora to investigate DAPT.

\subsubsection{NER Data Collection}
After the unlabeled corpora construction, we extract sentences from these corpora for annotating named entities. The annotation consists of pre-annotation and annotation processes, which we describe as follows.

\paragraph{Pre-Annotation Process}
For each domain, we sample sentences from our collected unlabeled corpus, which are then given named entity labels. Before annotating the sampled sentences, we leverage the DBpedia Ontology~\cite{mendes2012dbpedia}\footnote{The DBpedia Ontology contains 320 entity classes and categorizes 3.64 million entities.} to automatically detect entities and pre-annotate the selected samples. By doing so, we can alleviate the workload of annotators and potentially avoid annotation mistakes. However, the quality of the pre-annotated NER samples is not satisfactory since some entities will be incorrectly labeled and many entities are not in the DBpedia Ontology. In addition, we utilize the hyperlinks in Wikipedia and mark tokens that have hyperlinks to facilitate the annotation process and assist annotators in noticing entities. This is because tokens having hyperlinks are highly likely to be the named entity.

\paragraph{Annotation Process}
Each data sample requires two well-trained NER annotators to annotate it and one NER expert to double check it and give final labels. The data collection proceeds in three steps. First, one annotator needs to detect and categorize the entities in the given sentences. Second, the other annotator checks the annotations made by the first annotator, makes markings if he/she thinks that the annotations \textbf{could} be wrong and gives another annotation. Finally, the expert first goes through the annotations again and checks for possible mistakes, and then makes the final decision for disagreements between the first two annotators. 
In order to ensure the quality of annotations, the second annotator concentrates on looking for possible mistakes made by the first annotator instead of labeling from scratch. In addition, the expert will give a second round check and confer with the first two annotators when he/she is unsure about the annotations. A total 63.64\% of entities (the number of pre-annotated entities divided by the number of entities with hyperlinks) are pre-annotated based on the DBpedia Ontology, 73.33\% of entities (the number of corrected entities divided by the number of entities with hyperlinks) are corrected in the first annotation stage, 8.59\% of entities are annotated as possibly incorrect in the second checking stage, and finally, 8.57\% of annotations (out of all annotations) are modified by the experts. More annotation details are reported in the Appendix~\ref{appendix:annotation-details-crossner}.


\begin{table*}[]
\setlength{\tabcolsep}{10pt}
\renewcommand{\arraystretch}{1.15}
\centering
\resizebox{0.999\textwidth}{!}{
\begin{tabular}{cccccccc}
\toprule
\multirow{2}{*}{\textbf{Domain}}                                  & \multicolumn{3}{c}{\textbf{Unlabeled Corpus}} & \multicolumn{3}{c}{\textbf{Labeled NER}} & \multirow{2}{*}{\textbf{Entity Categories}}    \\ \cmidrule(lr){2-4}  \cmidrule(lr){5-7}
  & \textbf{\# paragraph}    & \textbf{\# sentence}    & \textbf{\# tokens}   & \textbf{\# Train}         & \textbf{\# Dev}         & \textbf{\# Test}        & \\ \midrule
Reuters & - & - & - & 14,987 & 3,466 & 3,684 & person, organization, location, miscellaneous \\ \midrule
Politics        & 2.76M           & 9.07M           & 176.56M      & 200           & 541         & 651         & \begin{tabular}[c]{@{}c@{}}politician, person, organization, political party, event, \\ election, country, location, miscellaneous\end{tabular}    \\ \midrule
\begin{tabular}[c]{@{}c@{}}Natural\\ Science\end{tabular}         & 1.72M           & 5.32M           & 98.50M       & 200           & 450         & 543         & \begin{tabular}[c]{@{}c@{}}scientist, person, university, organization, country, location, discipline, \\ enzyme, protein, chemical compound, chemical element, event,\\  astronomical object, academic journal,  award, theory, miscellaneous\end{tabular} \\ \midrule
Music      & 3.49M           & 9.82M           & 194.62M       & 100           & 380         & 456         & \begin{tabular}[c]{@{}c@{}}music genre, song, band, album, musical artist, musical instrument,\\ award, event, country, location, organization, person, miscellaneous\end{tabular}     \\ \midrule
Literature    & 2.69M           & 9.17M           & 177.33M       & 100           & 400         & 416         & \begin{tabular}[c]{@{}c@{}}book, writer, award, poem, event, magazine, person, location,\\  organization, country, miscellaneous\end{tabular}    \\ \midrule
\begin{tabular}[c]{@{}c@{}}Artificial\\ Intelligence\end{tabular} & 97.04K           & 287.62K           & 5.20M       & 100           & 350         & 431         & \begin{tabular}[c]{@{}c@{}}field, task, product, algorithm, researcher, metrics, university, \\ country, person, organization, location, miscellaneous\end{tabular}     \\ \bottomrule
\end{tabular}
}
\caption{Data statistics of unlabeled domain corpora, labeled NER samples and entity categories for each domain.}
\label{tab:crossner-data-statistics}
\end{table*}

\subsubsection{Data Statistics}
The data statistics of the Reuters News domain~\cite{sang2003introduction} and the five collected domains are illustrated in Table~\ref{tab:crossner-data-statistics}. In general, it is easy to collect a large unlabeled corpus for one domain, while for some low-resource domains, the corpus size could be small. As we can see from the statistics of the unlabeled corpus, the size is large for all domains except the AI domain (only a few AI-related pages exist in Wikipedia). Since DAPT experiments usually require a large amount of unlabeled sentences~\cite{wu2020tod,gururangan2020don}, this data scarcity issue introduces a new challenge for the DAPT.

We make the size of the training set (from Table~\ref{tab:crossner-data-statistics}) relatively small since cross-domain NER models are expected to perform fast adaptation with a small-scale of target domain data samples. In addition, there are domain-specialized entity types for each domain, resulting in a hierarchical category structure. For example, there are ``politician'' and ``person'' classes, but if a person is a politician, that person should be annotated as a ``politician'' entity, and not, a ``person'' entity.
Similar cases can be found for ``scientist'' and ``person'', ``political party'' and ``organization'', etc. We believe this hierarchical category structure will bring a challenge to this task since the model needs to better understand the context of inputs and be more robust in recognizing entities.

\begin{figure}
    \centering
    \subfigure[Overlaps of the NER datasets between domains]{
        \includegraphics[scale=0.52]{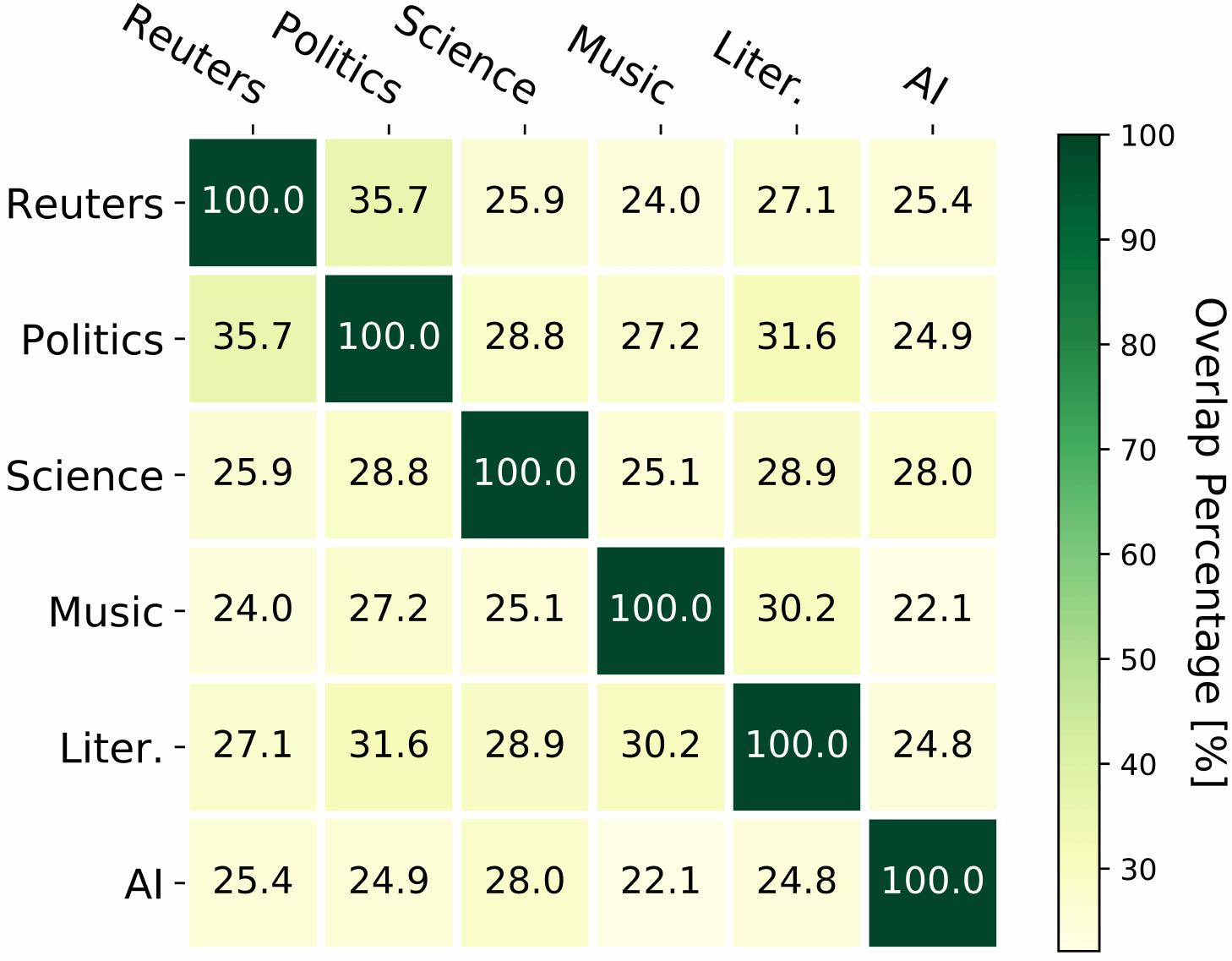}
    }
    \subfigure[Overlaps of the unlabeled corpora between domains]{
        \includegraphics[scale=0.52]{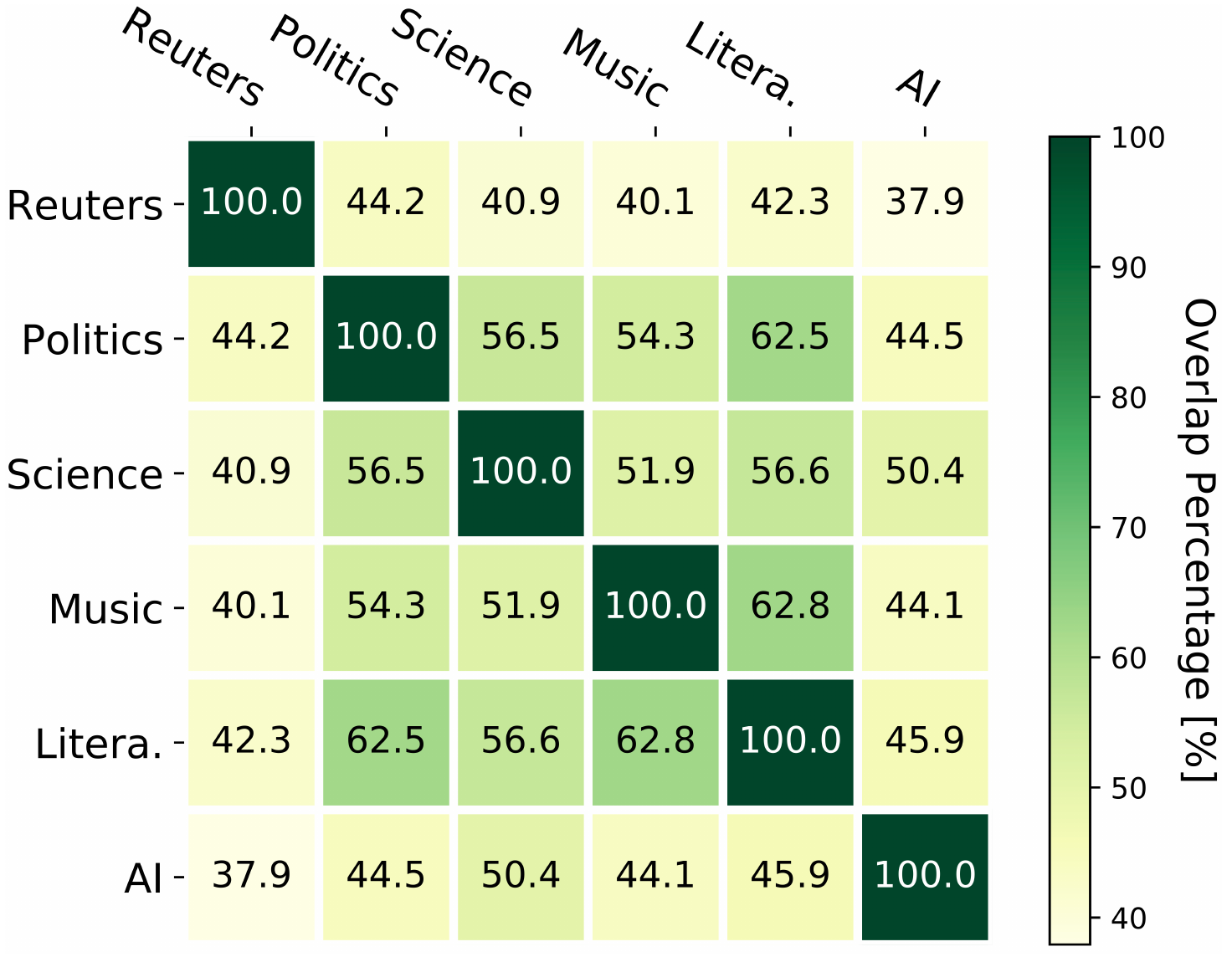}
    }
    \caption{Vocabulary overlaps between domains (\%). ``Reuters'' denotes the Reuters News domain, ``Science'' denotes the natural science domain and ``Litera.'' denotes the literature domain.}
    \label{fig:domain-overlap}
\end{figure}

\subsubsection{Domain Overlap}
The vocabulary overlaps of the NER datasets between domains (including the source domain (Reuters News domain) and the five collected target domains) are shown in Figure~\ref{fig:domain-overlap} (a). Vocabularies for each domain are created by considering the top 5K most frequent words (excluding stopwords). We observe that the vocabulary overlaps between domains are generally small, which further illustrates that the overlaps between domains are comparable and the domains of our collected datasets are diverse. 

The vocabulary overlaps of the unlabeled corpora between domains are shown in Figure~\ref{fig:domain-overlap} (b). The Reuters News corpus is taken from~\citet{jia2019cross}. For each domain, we sample 150K paragraphs from the domain-related corpus and create the vocabulary by considering the top 50K most frequent words (excluding stopwords). We can see that the overlaps in the unlabeled corpus are generally larger than those in the NER datasets. Since the corpora are large, they will contain more frequent words that overlap with those in other domains. In addition, the overlaps for other domain pairs are still comparably small except the vocabulary overlaps between the politics and literature domains, and music and literature domains (above 60\%).

\subsection{Model Description}
We continue pre-training the language model BERT~\cite{devlin2019bert} on the unlabeled corpus (i.e., DAPT) for the domain adaptation. 
The DAPT is explored in two directions. First, we investigate how different levels of corpora influence the pre-training.
Second, we explore the effectiveness between token-level and span-level masking in the DAPT.

\subsubsection{Pre-training Corpus}
When the size of a domain-related corpus is enormous, continuing to pre-train language models on it would be time-consuming. In addition, there would be noisy and domain-unrelated sentences in the collected corpus which could weaken the effectiveness of the DAPT. Therefore, we investigate whether extracting more indispensable content from the large corpus for pre-training can achieve comparable or even better cross-domain performance.

We consider three levels of the corpora for pre-training. The first is the \textbf{domain-level} corpus, which is the largest corpus we can collect related to a certain domain. 
The second is the \textbf{entity-level} corpus, which is a subset of the domain-level corpus and is made up of sentences having plentiful entities. Practically, it can be extracted from the domain-level corpus based on an entity list. We leverage the entity list in DBpedia Ontology and extract sentences that contain multiple entities to construct the entity-level corpus. 
The third is the \textbf{task-level} corpus, which is explicitly related to the NER task in the target domain. 
To construct this corpus, we select sentences having domain-specialized entities existing in the DBpedia Ontology. The size of the task-level corpus is expected to be much smaller than the entity-level corpus. However, its content should be more beneficial than that of the entity-level corpus.

Taking this further, we propose to \textbf{integrate} the entity-level and the task-level corpus. Instead of simply merging these two corpora, we first upsample the task-level corpus (double the size in practice) and then combine it with the entity-level corpus. Hence, models will tend to focus more on the task-level sentences in the DAPT.

\subsubsection{Span-level Pre-training}
Inspired by~\citet{joshi2020spanbert}, we propose to change the token-level masking (MLM) in BERT~\cite{devlin2019bert} into span-level masking for the DAPT. In BERT, MLM first randomly masks 15\% of the tokens in total, and then replaces 80\% of the masked tokens with special tokens (\texttt{[MASK]}), 10\% with random tokens and 10\% with the original tokens. 
We follow the same masking strategy as BERT except the first masking step. In the first step, after the random masking, we move the individual masked index position into its adjacent position that is next to another masked index position in order to produce more masked spans, while we do not touch the continuous masked indices (i.e., masked spans). For example, the randomly masked sentence:

\texttt{Western music's effect would [MASK] to grow within the country} \\ \texttt{[MASK] sphere} \\
\noindent would become 

\texttt{Western music's effect would continue to grow within the [MASK]} \\
\texttt{[MASK] sphere}.

Intuitively, span-level masking provides a more challenging task for pre-trained language models. For example, predicting ``San Francisco'' is much harder than predicting ``San'' given ``Francisco'' as the next word. Hence, the span-level masking can facilitate BERT to better understand the domain text so as to complete the more challenging task.

\subsection{Experimental Setup}

\subsubsection{Overall Setup}
We consider the CoNLL2003 English NER dataset~\cite{sang2003introduction} from Reuters News, which contains person, location, organization and miscellaneous entity categories, as the source domain and five domains in CrossNER as target domains. Our model is based on BERT~\cite{devlin2019bert} in order to have a fair comparison with the current state-of-the-art model~\cite{jia2020multi}, and we follow~\citet{devlin2019bert} to fine-tune BERT on the NER task.

Before training on the source or target domains, we conduct the DAPT on BERT when the unlabeled domain-related corpus is leveraged. Moreover, in the DAPT, different types of unlabeled corpora are investigated (i.e., domain-level, entity-level, task-level and integrated corpora), and different masking strategies are inspected (i.e., token-level and span-level masking).
Then, we carry out three different settings for the domain adaptation, which are described as follows:
\begin{itemize}
    \item \textbf{Directly Fine-tune:} We ignore the source domain training samples, and fine-tune BERT directly on the target domain data.
    \item \textbf{Pre-train then Fine-tune:} We first pre-train BERT on the source domain data, and then fine-tune it to the target domain samples.
    \item \textbf{Jointly Train:} We jointly fine-tune BERT on both source and target domain data samples. Since the size of the data samples in the target domains is smaller than in the source domain, we upsample the target domain data to balance the source and target domain data samples.
\end{itemize}

\subsubsection{Baselines}
We compare our methods to the following baselines:
\begin{itemize}
    \item BiLSTM-CRF~\cite{lample2016neural}, which incorporates bi-directional LSTM~\cite{hochreiter1997long} and conditional random fields for named entity recognition. We combine source domain data samples and the upsampled target domain data samples to jointly train this model (i.e., the joint training setting mentioned in the experimental settings). We use the word-level embeddings from~\citet{pennington2014glove} and the char-level embeddings from~\citet{hashimoto2017joint}.
    \item Our previously proposed framework, Coach, for entity tagging domain adaptation. It splits the task into two stages by first detecting the entities and then categorizing the detected entities.
    \item \citet{jia2019cross}, which integrates language modeling tasks and NER tasks in both source and target domains to perform cross-domain knowledge transfer. We follow their settings and provide the domain-level corpus for the language modeling tasks.
    \item \citet{jia2020multi}, which is a multi-cell compositional LSTM structure based on BERT representations~\cite{devlin2019bert} for domain adaptation. This is the current state-of-the-art cross-domain NER model. 
\end{itemize}

\subsubsection{Training}
We perform the DAPT for 15 epochs on the pre-trained corpus. We add a linear layer on top of BERT~\cite{devlin2019bert} and then fine-tune the whole model on the NER task. We select the best hyper-parameters by searching a combination of batch size and learning rate with the following range: batch size \{16, 32\} and learning rate \{$1\times10^{-5}$, $3\times10^{-5}$, $5\times10^{-5}$\}.
In the Directly Fine-tune and Pre-train then Fine-tune settings, we use batch size 16 and learning rate $5\times10^{-5}$, while in the Jointly Train setting, we use batch size 16 and learning rate $1\times10^{-5}$.
Additionally, we upsample the target domain data in the Jointly Train setting to balance the data samples between the source and target domains. Given that the number of training samples in the source domain is around 100 times larger than the target domain samples, the number of times we multiply the target domain data is searched within the range \{10, 50, 100, 150, 200\}, and we find that 100 is generally suitable for all domains. We use F1-score to evaluate the models as it is a commonly used evaluation metrics for the NER models~\cite{ma2016end,lample2016neural,chiu2016named}. We use the early stop strategy and select the model based on the performance on the development set of the target domain.

\begin{table}[!t]
\setlength{\tabcolsep}{10pt}
\renewcommand{\arraystretch}{1.15}
\centering
\resizebox{0.99\textwidth}{!}{
\begin{tabular}{ccccccccc}
\toprule
\textbf{Models}        & \textbf{Masking}                      & \textbf{Corpus}                 & \textbf{Politics} & \textbf{Science} & \textbf{Music} & \textbf{Litera.} & \textbf{AI}    &      \textbf{Average}         \\ \toprule
\multicolumn{9}{l}{\textbf{\textit{Fine-tune Directly on Target Domains (Directly Fine-tune)}}} \\ \midrule
\multirow{8}{*}{BERT-based}     & \multicolumn{2}{c}{w/o DAPT}               & 66.56    & 63.73   & 66.59 & 59.95      & 50.37   & 61.44              \\ \cmidrule(lr){2-3}  \cmidrule(lr){4-9}
    & \multirow{4}{*}{Token-level} & Domain-level           & 67.21    & 64.63   & 70.56 & 62.54      & 53.66  & 63.72                     \\ 
&       & Entity-level           & 67.59    & 65.97   & 70.64 & 63.77      & 53.94      & 64.38                 \\ 
&        & Task-level             & 67.30     & 65.04   & 70.37 & 62.10       & 53.19      & 63.60                 \\ 
 & & Integrated             & \textbf{68.83}    & \textbf{66.55}   & \textbf{72.42} & \textbf{63.95}      & \textbf{55.44}      & \textbf{65.44}                 \\ \cmidrule(lr){2-3}  \cmidrule(lr){4-9}
 & \multirow{3}{*}{Span-level}  & Entity-level           & 68.58    & 66.70    & 71.62 & 64.67      & 55.65  & 65.44                     \\
  &         & Task-level             & 68.37    & 65.84   & 70.66 & 63.85      & 54.48    & 64.64                   \\
 &      & Integrated             & \textbf{70.45}    & \textbf{67.59}   & \textbf{73.39} & \textbf{64.96}      & \textbf{56.36}     & \textbf{66.55}                  \\ \toprule
\multicolumn{9}{l}{\textbf{\textit{Pre-train on the Source Domain then Fine-tune on Target Domains (Pre-train then Fine-tune)}}} \\ \midrule
       \multirow{8}{*}{BERT-based}  & \multicolumn{2}{c}{w/o DAPT}     & 68.71    & 64.94   & 68.30  & 63.63      & 58.88     & 64.89                  \\ \cmidrule(lr){2-3}  \cmidrule(lr){4-9}
& \multirow{4}{*}{Token-level} & Domain-level           & 69.37    & 66.68   & 72.05 & 65.15      & 61.48   & 66.95                    \\
  &    & Entity-level           & 70.32    & 67.03   & 71.55 & 65.76      & 61.52    & 67.24                   \\       &                                              &            Task-level             & 70.21    & 65.99   & 71.74 & 65.32      & \multicolumn{1}{l}{60.29} & 66.71  \\   &                                              &      Integrated             & \textbf{71.44}    & \textbf{67.53}   & \textbf{74.02} & \textbf{66.57}      & \multicolumn{1}{l}{\textbf{61.90}} & \textbf{68.29}  \\ \cmidrule(lr){2-3}  \cmidrule(lr){4-9}                                              & \multirow{3}{*}{Span-level}  & Entity-level           & 71.85    & 68.04   & 73.34 & 66.28      & \multicolumn{1}{l}{61.66} & 68.23  \\   &     & Task-level             & 70.77    & 67.41   & 73.01 & 66.58      & \multicolumn{1}{l}{61.68} & 67.89  \\   &       & Integrated             & \textbf{72.05}    & \textbf{68.78}   & \textbf{75.71} & \textbf{69.04}      & \multicolumn{1}{l}{\textbf{62.56}} & \textbf{69.63}  \\ \toprule
\multicolumn{9}{l}{\textbf{\textit{Jointly Train on Both Source and Target Domains (Jointly Train)}}} \\ \midrule
     \multirow{8}{*}{BERT-based}     & \multicolumn{2}{c}{w/o DAPT}               & 68.85    & 65.03   & 67.59 & 62.57      & 58.57       & 64.52                \\ \cmidrule(lr){2-3}  \cmidrule(lr){4-9}
 & \multirow{4}{*}{Token-level} & Domain-level           & 69.49    & 66.37   & 71.94 & 63.74      & 60.53     & 66.41                  \\ 
& \multicolumn{1}{l}{} &       Entity-level           & 70.01    & 66.55   & 71.51 & 63.35      & 61.29     & 66.54                  \\  
& \multicolumn{1}{l}{} &     Task-level             & 70.14    & 66.06   & 70.70  & 62.68      & 60.14  & 65.94                     \\  
& \multicolumn{1}{l}{} &     Integrated             & \textbf{71.09}    & \textbf{67.58}   & \textbf{72.57} & \textbf{64.27}      & \textbf{62.55}       & \textbf{67.61}                \\ \cmidrule(lr){2-3}  \cmidrule(lr){4-9}
& \multirow{3}{*}{Span-level}  & Entity-level           & 71.90     & 68.04   & 71.98 & 64.23      & 61.63    & 67.55                   \\ 
& \multicolumn{1}{l}{} &    Task-level             & 71.31    & 67.75   & 71.17 & 63.24      & 60.83      & 66.86                 \\ 
& \multicolumn{1}{l}{}    &    Integrated             & \textbf{72.76}    & \textbf{68.28}   & \textbf{74.30}  & \textbf{65.18}      & \textbf{63.07}        & \textbf{68.72}               \\ \toprule
\multicolumn{9}{l}{\textbf{\textit{Baseline Models}}} \\ \midrule
\multicolumn{1}{l}{BiLSTM-CRF (word)}  &    \multicolumn{1}{c}{-}       & \multicolumn{1}{c}{-} & 52.52    & 44.60    & 40.77 & 35.69      & 38.24     & 42.36           \\ 
\multicolumn{1}{l}{BiLSTM-CRF (word + char)}   & \multicolumn{1}{c}{-}       & \multicolumn{1}{c}{-} & 56.60    & 49.97   & 44.79 & 43.03      & 43.56    & 47.59                   \\ \midrule
\multicolumn{1}{l}{Coach (word)}      & \multicolumn{1}{c}{-}       & \multicolumn{1}{c}{-} & 54.01    & 44.88   & 45.58 & 36.18      & 40.41      & 44.21                 \\
\multicolumn{1}{l}{Coach (word + char)}  & \multicolumn{1}{c}{-}       & \multicolumn{1}{c}{-} & 61.50     & 52.09   & 51.66 & 48.35      & 45.15    & 51.75                   \\ \midrule
\multicolumn{1}{l}{\citet{jia2019cross}}          & \multicolumn{1}{c}{-}       & \multicolumn{1}{c}{-} & 68.44    & 64.31   & 63.56 & 59.59      & 53.70    & 61.92                   \\ \midrule
\multicolumn{1}{l}{\citet{jia2020multi}}          & \multicolumn{1}{c}{-}       & \multicolumn{1}{c}{-} & 70.56    & 66.42   & 70.52 & 66.96      & 58.28       & 66.55                \\
\multicolumn{1}{l}{+ DAPT (Span-level \& Integrated)}          & \multicolumn{1}{c}{-}       & \multicolumn{1}{c}{-} & \textbf{71.45}    & \textbf{67.68}   & \textbf{74.19} & \textbf{68.63}      & \textbf{61.64}       & \textbf{68.71}                \\ \bottomrule
\end{tabular}
}
\caption{F1-scores of our proposed methods in three settings and baseline models. Results are averaged over three runs.}
\label{tab:cross-domain-results-on-crossner}
\end{table}

\begin{table}[!t]
\setlength{\tabcolsep}{20pt}
\renewcommand{\arraystretch}{1.35}
\centering
\resizebox{0.98\textwidth}{!}{
\begin{tabular}{cccccc}
\toprule
& \textbf{Politics} & \textbf{Science} & \textbf{Music}   & \textbf{Litera.} & \textbf{AI}     \\ \midrule
Domain-level & 177M (1x)  & 99M (1x)  & 195M (1x) & 177M (1x) & 5.2M (1x)  \\
Entity-level & 67M (0.37x)   & 36M (0.36x)  & 96M (0.49x)  & 87M (0.49x)  & 2.7M (0.52x)  \\ 
Task-level   & 16M (0.09x)   & 3.9M (0.04x)  & 26M (0.13x)  & 14M (0.08x)  & 0.2M (0.04x) \\
Integrated   & 99M (0.56x)  & 44M (0.44x)  & 148M (0.76x) & 115M (0.65x) & 3.1M (0.60x)  \\ \bottomrule
\end{tabular}
}
\caption{Number of tokens of different corpus types. The number in the brackets represents the size ratio between the corresponding corpus and the domain-level corpus.}
\label{corpus_statistics}
\end{table}

\subsection{Results and Discussion}

\subsubsection{Corpus Types \& Masking Strategies}
From Table~\ref{tab:cross-domain-results-on-crossner}, we can see that DAPT using the entity-level or the task-level corpus achieves better or on par results with using the domain-level corpus, while according to the corpus statistics illustrated in Table~\ref{corpus_statistics}, the size of the entity-level corpus is generally around half or less than half that of the domain-level corpus, and the size of the task-level corpus is much smaller than the domain-level corpus. 
We conjecture that the content of a corpus with plentiful entities is more suitable for DAPT in the NER task. In addition, selecting sentences with plentiful entities is able to filter numerous noisy sentences and partial domain-unrelated sentences from the domain corpus. Picking sentences having domain-specialized entities also filters a great many sentences that are not explicitly related to the domain and makes the DAPT more effective and efficient. 
In general, DAPT using the task-level corpus performs slightly worse than using the entity-level corpus. This can be attributed to the large corpus size differences. Furthermore, integrating the entity-level and task-level corpora is able to consistently boost the adaptation performance compared to utilizing other corpus types, although the size of the integrated corpus is still smaller than the domain-level corpus. This is because the integrated corpus ensures the pre-training corpus is relatively large and, in the meantime, focuses on the content that is explicitly related to the NER task in the target domain. 
The results suggest that the corpus content is essential for the DAPT. Surprisingly, the DAPT is still effective for the AI domain even though the corpus size in this domain is relatively small, which illustrates that the DAPT is also practically useful in a small corpus setting.

As we can see from Table~\ref{tab:cross-domain-results-on-crossner}, when leveraging the same corpus, the span-level masking consistently outperforms the token-level masking. For example, in the Pre-train then Fine-tune setting, DAPT on the integrated corpus and using span-level masking outperforms that using token-level masking by a 1.34\% F1-score on average. This is because predicting spans is a more challenging task than predicting tokens, forcing the model to better comprehend the domain text and providing it a more powerful capability to do the downstream tasks.
Moreover, adding the integrated corpus and DAPT using the span-level masking~\citet{jia2020multi} further improves on the F1-score by 2.16\% on average.

From Table~\ref{tab:cross-domain-results-on-crossner}, we can clearly observe the improvements when the source domain data samples are leveraged. For example, compared to the Directly Fine-tune, Pre-train then Fine-tune (w/o DAPT) improves on the F1-score by 3.45\% on average, and Jointly Train (w/o DAPT) improves on the F1-score by 3.08\% on average.
We notice that Pre-train then Fine-tune generally leads to better performance than Jointly Train. We speculate that jointly training on both the source and target domains makes it difficult for the model to concentrate on the target domain task, leading to a sub-optimal result, while for the Pre-train then Fine-tune, the model learns the NER task knowledge from the source domain data in the pre-training step and then focuses on the target domain task in the fine-tuning step.
Finally, we can see that our best model can outperform the existing state-of-the-art model in all five domains.
However, the averaged F1-score of the best model is not yet perfect (lower than 70\%), which highlights the need for more advanced cross-domain models.

\begin{figure}[!t]
\centering
\subfigure[Directly Fine-tune.]{
    \includegraphics[scale=0.435]{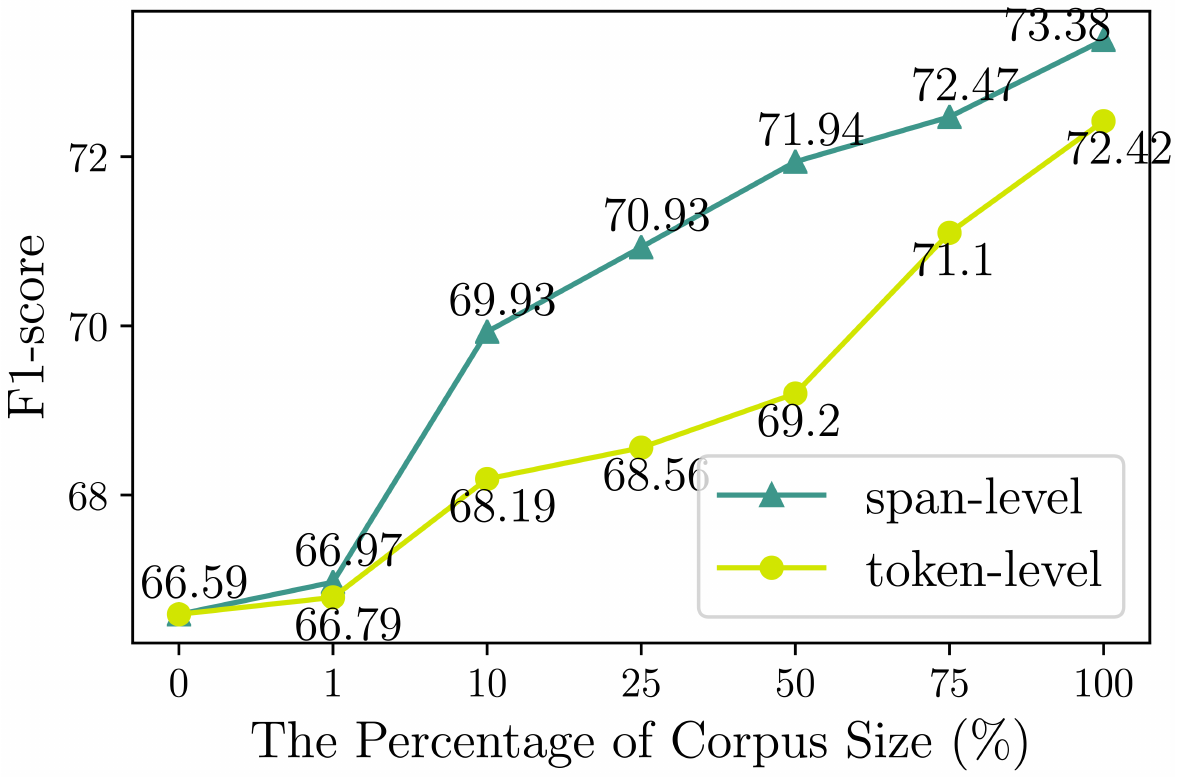}
}
\subfigure[Pre-train then Fine-tune.]{
    \includegraphics[scale=0.435]{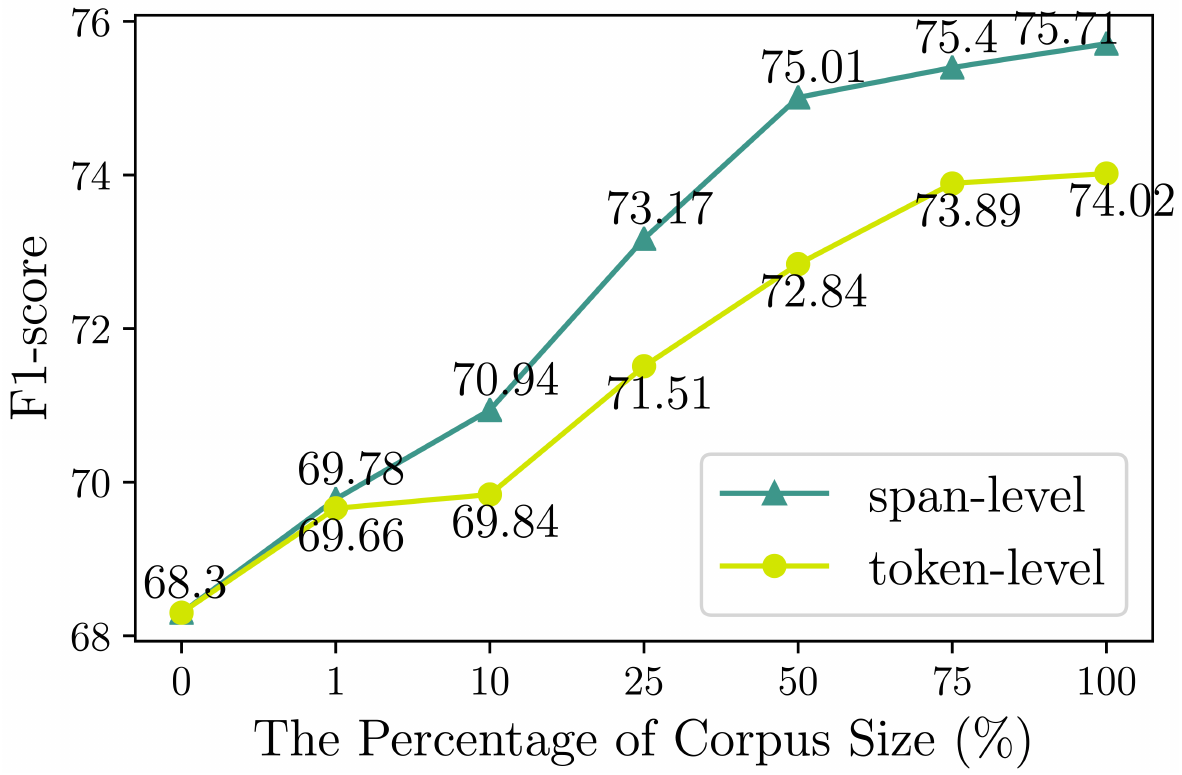}
}
\subfigure[Jointly Train.]{
    \includegraphics[scale=0.435]{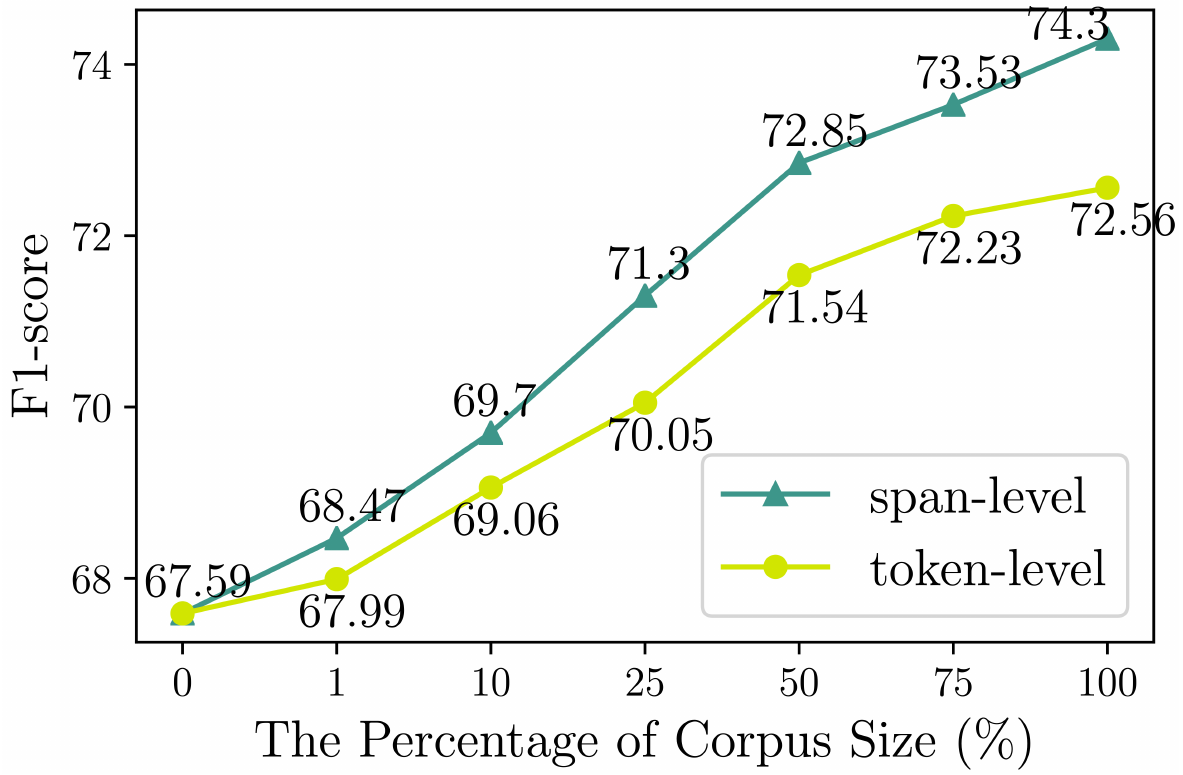}
}
\caption{Comparisons among utilizing different percentages of the music domain's integrated corpus and masking strategies in the DAPT for the three settings.}
\label{fig:performance_vs_corpus_size}
\end{figure}

\subsubsection{Performance vs. Unlabeled Corpus Size}
Given that a large-scale domain-related corpus might sometimes be unavailable, we investigate the effectiveness of different corpus sizes for the DAPT and explore how the masking strategies will influence the adaptation performance.
As shown in Figure~\ref{fig:performance_vs_corpus_size}, as the size of unlabeled corpus increases, the performance generally keeps improving. 
This implies that the corpus size is generally essential for the DAPT, and within a certain corpus size, the larger the corpus is, the better the domain adaptation performance the DAPT will produce.
Additionally, we notice that in the Pre-train then Fine-tune setting, the improvement becomes comparably lower when the percentage reaches 75\% or higher. We conjecture that it is relatively difficult to improve the performance once it reached a certain level.

Furthermore, little performance improvement is observed for both the token-level and span-level masking strategies when only a small-scale corpus (1\% of the music integrated corpus, 1.48M) is available. As the corpus size increases, the span-level masking starts to outperform the token-level masking. We notice that in the Directly Fine-tune setting, the performance discrepancy between the token-level and span-level is first increasing and then decreasing, while the performance discrepancies are generally increasing in the other two settings. We hypothesize that the span-level masking can learn the domain text more efficiently since it is a more challenging task, while the token-level masking requires a larger corpus to better understand the domain text.

\begin{figure}[t!]
    \centering
    \includegraphics[scale=0.55]{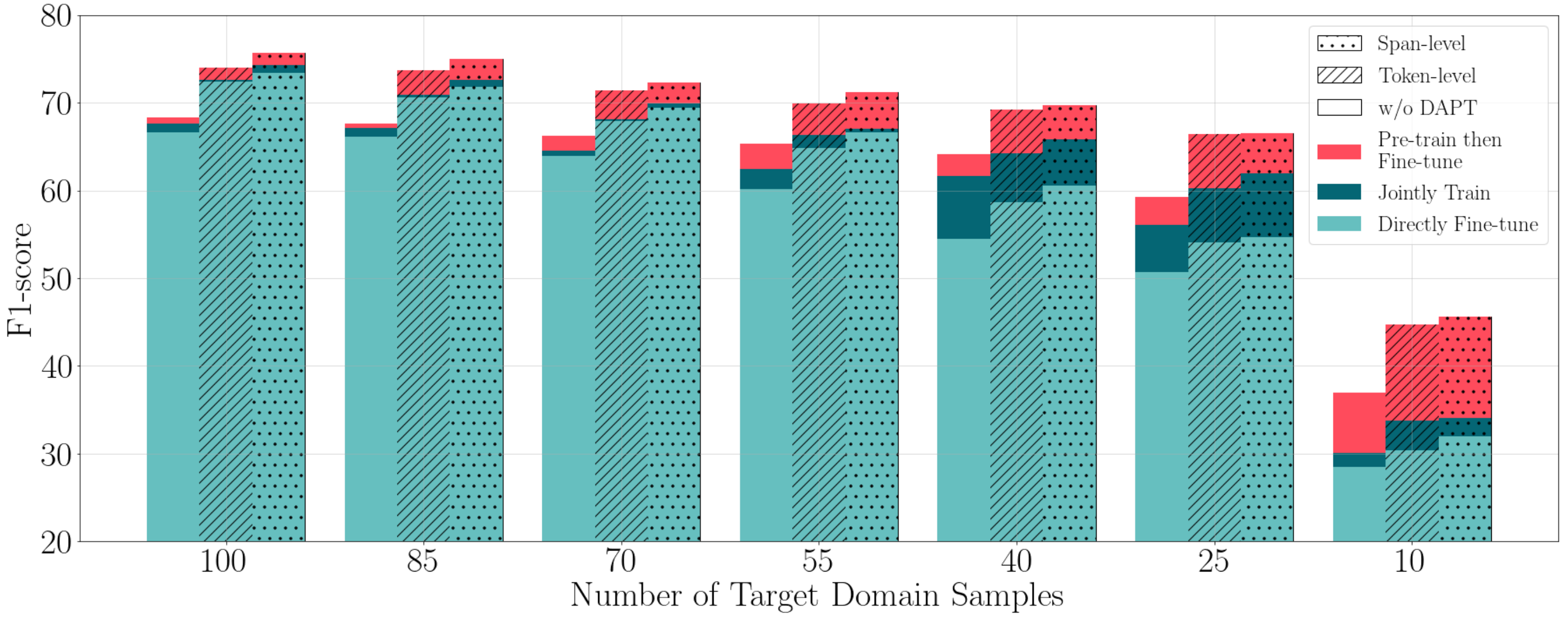}
    \caption{Few-shot F1-scores (averaged over three runs) in the music domain. We use the integrated corpus for the DAPT.}
    \label{fig:performance-fewshot}
\end{figure}

\subsubsection{Performance vs. Target Domain Sample Size}
From Figure~\ref{fig:performance-fewshot}, we can see that the performance drops when the number of target domain samples is reduced, the span-level pre-training generally outperforms token-level pre-training, and the task becomes extremely difficult when only a few data samples (e.g., 10 samples) are available in the target domain. Interestingly, as the target domain sample size decreases, the advantage of using source domain training samples becomes more significant, for example, Pre-train then Fine-tune outperforms Directly Fine-tune by $\sim$10\% F1 when the sample size is reduced to 40 or lower. This is because these models are able to gain the NER task knowledge from a large amount of source domain examples and then possess the ability to quickly adapt to the target domain.
Additionally, using DAPT significantly improves the performance in the Pre-train then Fine-tune setting when target domain samples are scarce (e.g., 10 samples). This can be attributed to the boost in the domain adaptation ability made by the DAPT, which allows the model to quickly learn the NER task in the target domain. Furthermore, we notice that with the decreasing of the sample size, the performance discrepancy between Pre-train then Fine-tune and Jointly Train becomes larger. We speculate that in the Jointly Train setting, the models focus on the NER task in both the source and target domains. This makes the models tend to ignore the target domain when the sample size is too small, while for the Pre-train then Fine-tune setting, the models can focus on the target domain in the fine-tuning stage to ensure the good performance in the target domain.

\begin{table}[t!]
\setlength{\tabcolsep}{10pt}
\centering
\resizebox{0.99\textwidth}{!}{
\begin{tabular}{lcccccccccc}
\toprule
\multicolumn{1}{c}{\textbf{Models}} & \textbf{Genre} & \textbf{Song}  & \textbf{Band}  & \textbf{Album} & \textbf{Artist} & \textbf{Country} & \textbf{Loc.}  & \textbf{Org.}  & \textbf{Per.}  & \textbf{Misc.}  \\ \midrule
\multicolumn{11}{l}{\textit{\textbf{Directly Fine-tune}}} \\ \midrule
w/o DAPT                     & 77.35      & 26.57 & 71.65 & 60.61 & 80.71       & 84.93   & 67.82  & 59.36 & 7.97   & 17.20 \\
Span-level + Integrated      & 78.89      & 42.63 & \textbf{83.03} & 65.93 & \textbf{85.70}       & 83.86   & 77.17 & 69.43 & 10.09 & 19.20 \\ \midrule
\multicolumn{11}{l}{\textit{\textbf{Pre-train then Fine-tune}}}                                  \\ \midrule
w/o DAPT                     & 79.12      & 42.04 & 68.45  & 61.79  & 76.75        & \textbf{88.85}   & 78.47 & 69.70 & 10.15  & \textbf{29.80} \\
Span-level + Integrated      & \textbf{80.82}      & \textbf{58.67} & 82.72 & \textbf{69.28} & 84.58       & 85.61   & \textbf{80.54} & \textbf{75.16} & \textbf{12.59}  & 28.67 \\ \bottomrule
\end{tabular}
}
\caption{F1-scores (averaged over three runs) for the categories in the music domain over the Directly Fine-tune and Pre-train then Fine-tune settings. Span-level+Integrated denotes that the span-level masking and integrated corpus are utilized for the DAPT. ``Loc.'', ``Org.'', ``Per.'' and ``Misc.'' denote ``Location'', ``Organization'', ``Person'' and ``Miscellaneous'', respectively.}
\label{fig:fine-grained}
\end{table}

\subsubsection{Fine-grained Comparison}
In this section, we further explore the effectiveness of the DAPT and leveraging NER samples in the source domain. As shown in Table~\ref{fig:fine-grained}, the performance is improved on almost all categories when the DAPT or the source domain NER samples are utilized. We observe that using source domain NER data might hurt the performance in some domain-specialized entity categories, such as ``artist'' (``musical artist'') and ``band''. This is because that since ``artist'' is a subcategory of ``person'' and models pre-trained on the source domain tend to classify artists as ``person'' entities. Similarly, ``band'' is a subcategory of ``organization'', which leads to the same misclassification issue after the source domain pre-training.
When the DAPT is used, the performance on some domain-specialized entity categories is greatly improved (e.g., ``song'', ``band'' and ``album'').
We notice that the performance on the ``person'' entity is relatively low compared to other categories. It is because that the hierarchical category structure could cause models to be confused between ``artist'' and ``person'' entities, and we find out that 84.81\% of ``person'' entities are misclassified as ``artist'' in the best model we have.

\subsection{Short Summary}
In this section, we introduce CrossNER, a humanly-annotated NER dataset spanning over five diverse domains with specialized entity categories for each domain, to facilitate further research in the NER domain adaptation field.
In addition, we collect the corresponding domain-related corpora for the study of DAPT.
Moreover, we conduct comprehensive experiments and analyses in terms of the size of the domain-related corpus and different pre-training strategies in the DAPT for the cross-domain NER task.
We find that a more challenging pre-training fashion can better address the domain discrepancy issue in the task knowledge transfer.
We compare our proposed method with a set of existing strong baselines and show that our method consistently brings better performance on target domains that are topologically distant from the source domain.


\newpage

%% file: chapter/sec-5-x2parsing.tex
\chapter{Adaptation to Target Languages in Target Domains}\label{sec-x2parsing}

\begin{figure}[!t]
    \centering
    \resizebox{0.9\textwidth}{!}{  
    \includegraphics{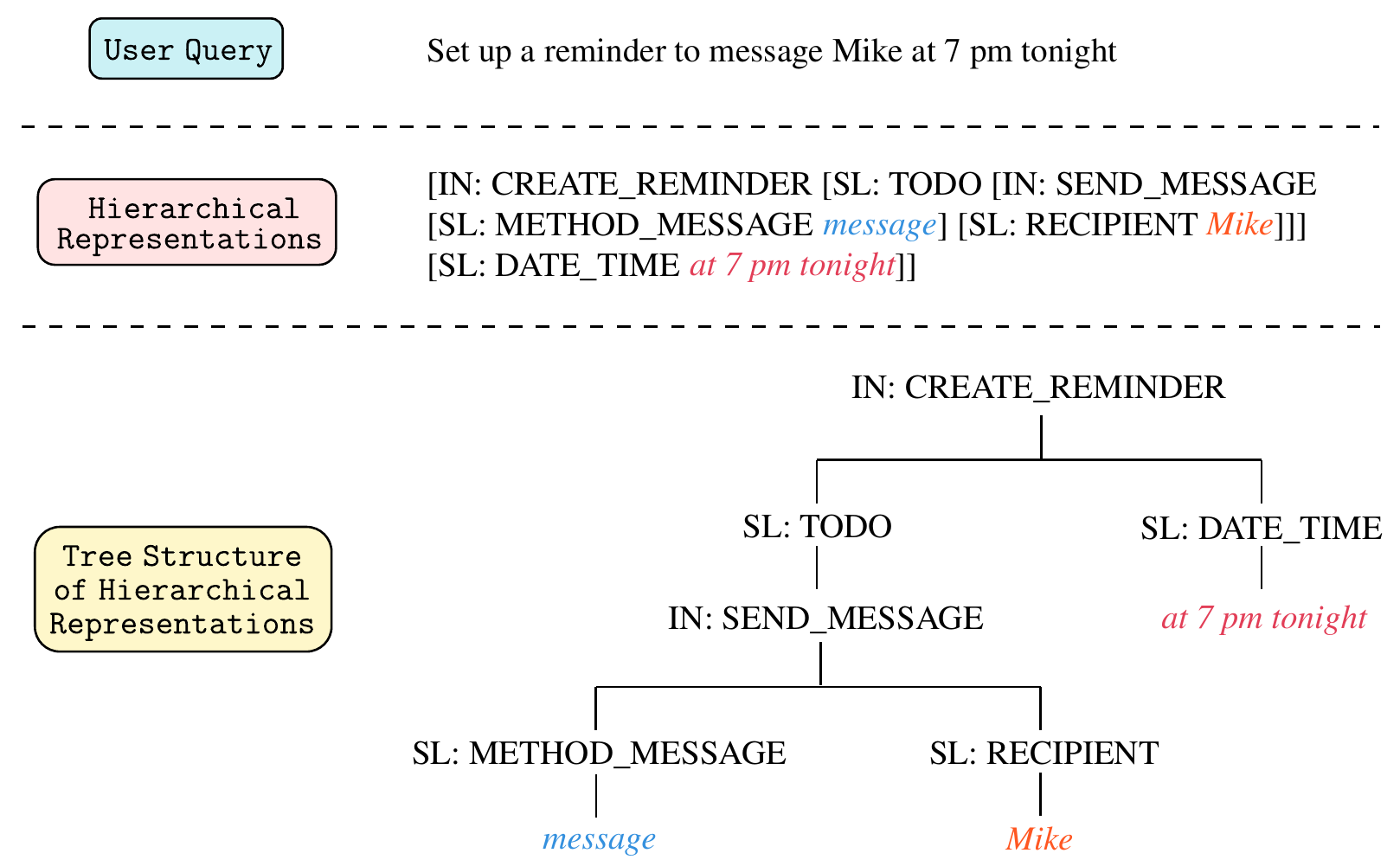}
    }
    \caption{Illustration of the compositional semantic parsing task, where IN denotes the intent and SL denotes the slot entity.}
    \label{fig:compositional-semantic-parsing}
\end{figure}

In the previous two chapters, we explored our models' adaptation ability to low-resource target languages in the same domains, and to low-resource target domains in the same language. 
Taking this further, we tackle the challenges of adapting models to low-resource target languages in low-resource target domains, which extends the models' scalability in data scarcity scenarios.
We focus on the compositional semantic parsing task, where the model needs to cope with complex user queries, and detect and categorize nested intents and slot entities. Figure~\ref{fig:compositional-semantic-parsing} provides an illustration of this task, which allows an in-depth understanding of complex user queries, and serves as an essential component of virtual assistants.

In this chapter, we propose X2Parser, a cross-lingual and cross-domain compositional semantic parser, that is transferable to low-resource target languages and target domains.
We provide a new perspective to simplify this task by flattening the hierarchical representations and casting the problem into several sequence labeling tasks. We introduce a fertility-based slot predictor that first learns to dynamically detect the number of labels for each token, and then predicts the slot entity types.
We conduct extensive experiments in different few-shot settings and explore the combination of cross-lingual and cross-domain scenarios. 
We find that simplifying task structures makes the representation learning more effective for low-resource languages and domains.
Experimental results show that our model significantly outperforms existing strong baselines in different low-resource settings and notably reduces the latency compared to a generation-based model. Our model can also be applied for general nested entity recognition tasks.

\section{X2Parser}

\subsection{Task Decomposition}

\subsubsection{Motivation}
As illustrated in Figure~\ref{fig:compositional-semantic-parsing}, hierarchical representations are used in the compositional semantic parsing task to express the output structures. Previous state-of-the-art approaches~\cite{rongali2020don,li2021mtop} to this task are generation-based models that learn to directly generate the hierarchical representations which contain nested intents and slot labels. 
However, hierarchical representations containing nested annotations for intents and slot entities are relatively complex, and the models~\cite{rongali2020don,li2021mtop} need to learn when to generate the starting intent or slot entity, when to copy tokens from the input sequence, and when to generate the end of the label.
Hence, we need large enough training data to train a good model to produce such representations, and the model's performance will be greatly limited in low-resource scenarios. 
Therefore, instead of incorporating intents and slot entities into one representation, we propose to predict them separately so that we can simplify the parsing problem and enable the model to easily learn the skills for each decomposed task, and finally, our model can achieve a better adaptation ability in low-resource scenarios.

\begin{figure*}[!t]
    \centering
    \resizebox{0.999\textwidth}{!}{  
    \includegraphics{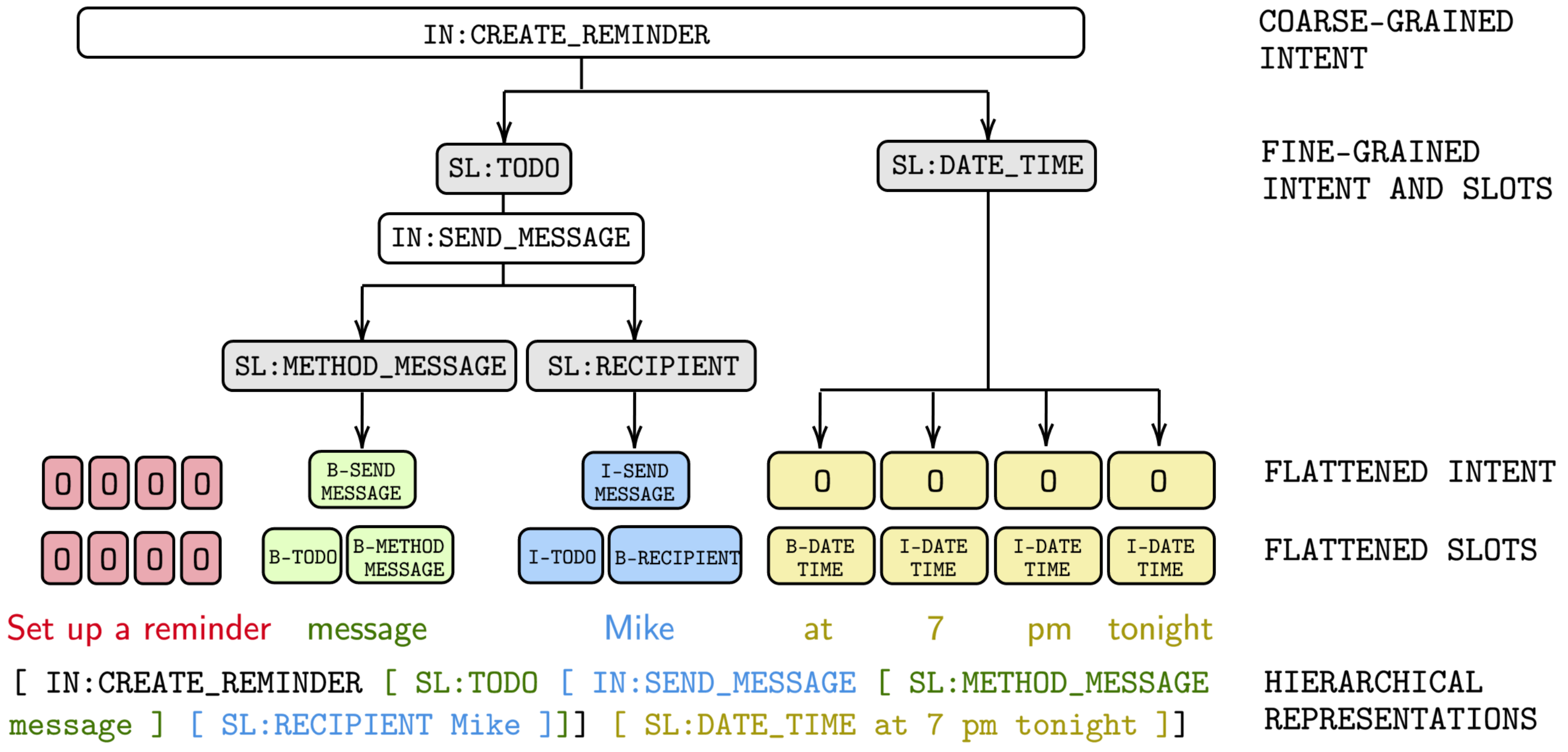}
    }
    \caption{One data example with the illustration of our proposed flattened intents and slots representations, as well as the hierarchical representations used in~\citet{li2021mtop}.}
    \label{fig:flatten_vs_hierarchical}
\end{figure*}

As illustrated in Figure~\ref{fig:flatten_vs_hierarchical}, we obtain the coarse-grained intent, flattened fine-grained intents and flattened slot entity labels from the hierarchical representations, and train the model based on these three categories in a multi-task fashion. Note that we can always reconstruct the hierarchical representations based on the labels in these three categories, which means that the decomposed labels and the hierarchical labels are equivalent.

\subsubsection{Label Constructions}
\paragraph{Slot Labels} We extract nested slot labels from the hierarchical representations and assign the labels to corresponding tokens based on the BIO structure. As we can see from Figure~\ref{fig:flatten_vs_hierarchical}, there could exist multiple slot labels for one token, and we consider the order of the labels so as to reconstruct the hierarchical representations. Specifically, we put the more fine-grained slot label at the later position. For example, ``message'' (in Figure~\ref{fig:flatten_vs_hierarchical}) has \texttt{B-TODO} and \texttt{B-METHOD-MESSAGE} labels, and \texttt{B-METHOD-MESSAGE} comes after \texttt{B-TODO} since it is a more fine-grained slot label.

\paragraph{Intent Labels}
Each data sample has one intent label for the whole user utterance, and we extract it as an individual coarse-grained intent label. For the intents expressed by partial tokens (i.e., fine-grained intents), we use the BIO structure to label the corresponding tokens. We notice that we only need to assign one intent label to each token since the nested cases for intents are relatively simple.\footnote{More details about for nested intents are placed in the Appendix~\ref{appendix:nested_intent_label_construction}.} Therefore, the fine-grained intent classification becomes a  sequence labeling task.

\begin{figure}[!t]
    \centering
    \resizebox{0.99\textwidth}{!}{  
    \includegraphics{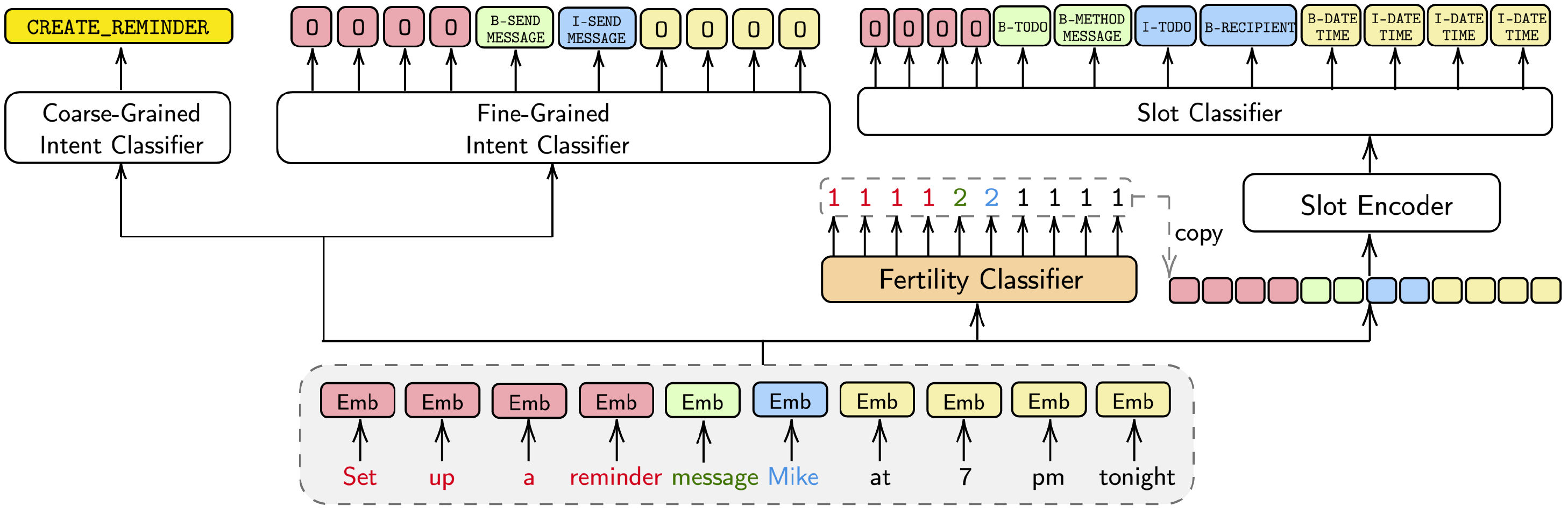}
    }
    \caption{The architecture of X2Parser. We consider the compositional semantic parsing task as a combination of the coarse-grained intent classification, fine-grained intent prediction, and slot filling tasks.}
    \label{fig:x2parser-architecture}
\end{figure}

\subsection{Architecture}
Figure~\ref{fig:x2parser-architecture} illustrates our proposed transferable cross-lingual and cross-domain parser (X2Parser).
To enable the cross-lingual ability of our model, we leverage the multilingual pre-trained model XLM-R~\cite{conneau2020unsupervised} as the sequence encoder.
Let us define $ X = \{x_1, x_2, ..., x_n\} $ as the user utterance and $ H = \{h_1, h_2, ..., h_n\}$ as the hidden states (denoted as Emb in Figure~\ref{fig:x2parser-architecture}) from XLM-R.

\subsubsection{Slot Predictor}
The slot predictor consists of a fertility classifier, a slot encoder, and a slot classifier.
Inspired by~\citet{gu2018nonautoregressive}, the fertility classifier learns to predict the number of slot labels for each token, and then it copies the corresponding number of hidden states. Finally, the slot classifier is trained to conduct the sequence labeling based on the slot labels we constructed. 
The fertility classifier not only helps the model identify the number of labels for each token but also guides the model to implicitly learn the nested slot information in user queries. It relieves the burden of the slot classifier, which needs to predict multiple slot entities for certain tokens.

\paragraph{Fertility Classifier (FC)}
We add a linear layer (FC) on top of the hidden states from XLM-R to predict the number of labels (fertility), which we formulate as follows:
\begin{equation}
    F = \{f_1, f_2, ..., f_n\} = \text{FC}(\{h_1, h_2, ..., h_n\}),
\end{equation}
where FC is an n-way classifier (n is the maximum label number) and $f_i (i \in [1,n])$ is a positive integer representing the number of labels for $x_i$. 

\paragraph{Slot Filling}
After obtaining the fertility predictions, we copy the corresponding number of hidden states from XLM-R:
\begin{equation}
    H' = \text{CopyHiddens}(H, F).
\end{equation}
Then, we add a Transformer encoder~\cite{vaswani2017attention} (slot encoder (SE)) on top of $H'$ to incorporate the sequential information into the hidden states, followed by adding a linear layer (slot classifier (SC)) to predict the slots, which we formulate as follows:
\begin{equation}
    P_{\text{slot}} = \text{SC}(\text{SE}(H')),
\end{equation}
where $P_{\text{slot}}$ is a sequence of slots that has the same length as the sum of the fertility numbers.

\subsubsection{Intent Predictor}
\paragraph{Coarse-Grained Intent} 
The coarse-grained intent is predicted based on the hidden state of the ``\texttt{[CLS]}'' token from XLM-R since it can be the representation for the whole sequence, and then we add a linear layer (coarse-grained intent classifier (CGIC)) on top of the hidden state to predict the coarse-grained intent:
\begin{equation}
    p_{\text{cg}} = \text{CGIC}(h_{\text{cls}}),
\end{equation}
where $ p_{\text{cg}} $ is a single intent prediction.

\paragraph{Fine-Grained Intent}
We add a linear layer (fine-grained intent classifier (FGIC)) on top of the hidden states H to produce the fine-grained intents:
\begin{equation}
    P_{\text{fg}} = \text{FGIC}(\{h_1, h_2, ..., h_n\}),
\end{equation}
where $P_{\text{fg}}$ is a sequence of intent labels that has the same length as the input sequence.

\begin{table}[!t]
\setlength{\tabcolsep}{15pt}
\renewcommand{\arraystretch}{1.2}
\centering
\resizebox{0.99\textwidth}{!}{
\begin{tabular}{ccccccccc}
\toprule
\multirow{2}{*}{\textbf{Domain}} & \multicolumn{6}{c}{\textbf{Number of Utterances}}              & \multirow{2}{*}{\begin{tabular}[c]{@{}c@{}}\textbf{Intent}\\ \textbf{Types}\end{tabular}} & \multirow{2}{*}{\begin{tabular}[c]{@{}c@{}}\textbf{Slot}\\ \textbf{Types}\end{tabular}} \\ \cmidrule(lr){2-7}
& \textbf{English} & \textbf{German} & \textbf{French} & \textbf{Spanish} & \textbf{Hindi}  & \textbf{Thai}   & & \\ \midrule
\textbf{Alarm}                   & 1,783   & 1,581  & 1,706  & 1,377   & 1,510  & 1,783  & 6                                                                       & 5                                                                     \\
\textbf{Calling}                 & 2,872   & 2,797  & 2,057  & 2,515   & 2,490  & 2,872  & 19                                                                      & 14                                                                    \\
\textbf{Event}                   & 1,081   & 1,051  & 1,115  & 911     & 988    & 1,081  & 12                                                                      & 12                                                                    \\
\textbf{Messaging}               & 1,053   & 1,239  & 1,335  & 1,164   & 1,082  & 1,053  & 7                                                                       & 15                                                                    \\
\textbf{Music}                   & 1,648   & 1,499  & 1,312  & 1,509   & 1,418  & 1,648  & 27                                                                      & 12                                                                    \\
\textbf{News}                    & 1,393   & 905    & 1,052  & 1,130   & 930    & 1,393  & 3                                                                       & 6                                                                     \\
\textbf{People}                  & 1,449   & 1,392  & 763    & 1,408   & 1,168  & 1,449  & 17                                                                      & 16                                                                    \\
\textbf{Recipes}                 & 1,586   & 1,002  & 762    & 1,382   & 929    & 1,586  & 3                                                                       & 18                                                                    \\
\textbf{Reminder}                & 2,439   & 2,321  & 2,202  & 1,811   & 1,833  & 2,439  & 19                                                                      & 17                                                                    \\
\textbf{Timer}                   & 1,358   & 1,014  & 1,165  & 1,159   & 1,047  & 1,358  & 9                                                                       & 5                                                                     \\
\textbf{Weather}                 & 2,126   & 1,785  & 1,990  & 1,816   & 1,800  & 2,126  & 4                                                                       & 4                                                                     \\ \midrule
\textbf{Total}                   & 18,788  & 16,585 & 15,459 & 16,182  & 15,195 & 18,788 & 117   & 78   \\ \bottomrule
\end{tabular}
}
\caption{Data statistics of the MTOP dataset. The data are roughly divided into a 70:10:20 percent split for train, eval and test}
\label{tab:mtop_data_statistics}
\end{table}

\section{Experimental Setup}
\subsection{Datasets}
\begin{figure}[t!]
    \centering
    \resizebox{0.55\textwidth}{!}{  
    \includegraphics{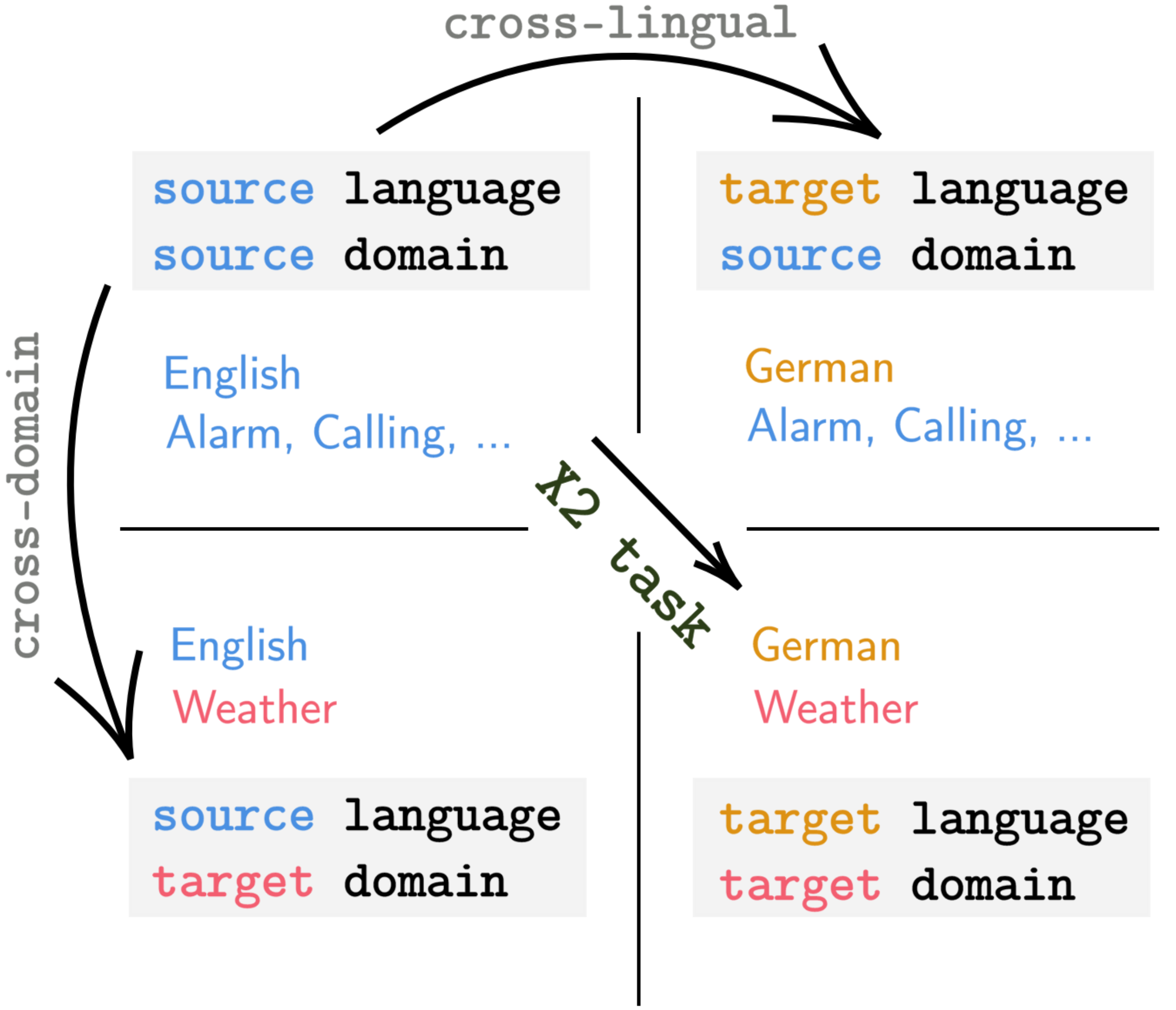}
    }
    \caption{Illustration of the cross-lingual task, cross-domain task, and the combination of both (X2 task).}
    \label{fig:task_settings}
\end{figure}

We conduct experiments on the MTOP dataset proposed by~\citet{li2021mtop}, which contains six languages, English (en), German (de), French (fr), Spanish (es), and Thai (th), and 11 domains, alarm, calling, event, messaging, music, news, people, recipes, reminder, timer, and weather.
The data statistics are shown in Table~\ref{tab:mtop_data_statistics}.
As illustrated in Figure~\ref{fig:task_settings}, we explore the cross-lingual setting, cross-domain setting, and a combination of both.

\subsubsection{Cross-Lingual Setting}
In the cross-lingual setting, we use English as the source language and the other languages as target languages. In addition, we consider a zero-shot scenario where we only use English data for training.

\subsubsection{Cross-Domain Setting}
In the cross-domain setting, we only consider training and evaluation in English.
We choose ten domains as source domains and the other domain as the target domain. Different from the cross-lingual setting, we consider a few-shot scenario where we first train the model using the data from the ten source domains, and then we fine-tune the model using a few data samples (e.g., 10\% of the data) from the target domain.
We consider the few-shot scenario because zero-shot adapting the model to the target domain is extremely difficult due to the unseen intent and slot entity types, while zero-shot adaptation to target languages is easier because multilingual pre-trained models are used.

\subsubsection{Cross-Lingual Cross-Domain Setting}
This setting combines the cross-lingual and cross-domain settings. Specifically, we first train the model on the English data from the ten source domains, and then fine-tune it on a few English data samples from the other (target) domain. Finally, we conduct the zero-shot evaluation on all the target languages of the target domain.

\subsection{Baselines}

We compare our model with the following baselines:

\paragraph{Seq2Seq w/ XLM-R} 
\citet{rongali2020don} proposed a sequence-to-sequence (Seq2Seq) model using a pointer-generator network~\cite{see2017get} to handle nested queries, and achieved new state-of-the-art results in English. \citet{li2021mtop} adopted this architecture for zero-shot cross-lingual adaptation. They replaced the encoder with the XLM-R~\cite{conneau2020unsupervised} and used a customized decoder to learn to generate intent and label types and copy tokens from the inputs.\footnote{In order to compare the performance in the cross-domain and cross-lingual cross-domain settings, we follow~\citet{li2021mtop} to reimplement this baseline since the source code is not publicly available.}

\paragraph{Seq2Seq w/ CRISS}
This is the same architecture as \textit{Seq2Seq w/ XLM-R}, except that \citet{li2021mtop} replaced XLM-R with the multilingual pre-trained model, CRISS~\cite{Chau2020cross}, as the encoder for the zero-shot cross-lingual adaptation.

\paragraph{Neural Layered Model (NLM)}
This baseline conducts the multi-task training based on the same task decomposition as X2Parser, but it replaces the slot predictor module in X2Parser with a neural layered model~\cite{ju2018neural},\footnote{This model was originally proposed to tackle the nested named entity recognition task.} while keeping the other modules the same.
Unlike our fertility-based slot predictor, NLM uses several stacked layers to predict entities of different levels. We use this baseline to verify the effectiveness of our fertility-based slot predictor.

\subsection{Training}
We use XLM-R Large~\cite{conneau2020unsupervised} as the sequence encoder. For a word (in an utterance) with multiple subword tokens, we take the representations from the first subword token to predict the labels for this word. The Transformer encoder (slot encoder) has one layer with a head number of 4, a hidden dimension of 400, and a filter size of 64. We set the fertility classifier as a 3-way classifier since the maximum label number for each token in the dataset is 3. We train X2Parser using the Adam optimizer~\cite{kingma2015adam} with a learning rate of 2e-5 and a batch size of 32.
We follow~\citet{li2021mtop} and use the exact match accuracy to evaluate the models. For our model, the prediction is considered correct only when the predictions for the coarse-grained intent, fine-grained intents, and the slots are all correct. To ensure a fair comparison, we use the same three random seeds to run each model and calculate the averaged score for each target language and domain.

\begin{table}[!t]
\setlength{\tabcolsep}{15pt}
\renewcommand{\arraystretch}{1.3}
\centering
\resizebox{0.99\textwidth}{!}{
\begin{tabular}{lccccccc}
\toprule
\textbf{Model}   & \textbf{en}     & \textbf{es} & \textbf{fr} & \textbf{de} & \textbf{hi} & \textbf{th} & \textbf{Avg.} \\ \midrule
Seq2Seq w/ CRISS~\cite{li2021mtop}  & \textbf{84.20}        & 48.60       & 46.60       & 36.10       & 31.20       & 0.00        & 32.50         \\
Seq2Seq w/ XLM-R~\cite{li2021mtop}    & 83.90       & 50.30       & 43.90       & 42.30       & 30.90       & 26.70       & 38.82         \\
Neural Layered Model (NLM) & 82.40 & 59.99       & 58.16       & 54.91       & 29.31       & 28.78       & 46.23         \\ \midrule
X2Parser        & 83.39           & \textbf{60.30}       & \textbf{58.34}       & \textbf{56.16}       & \textbf{37.06}       & \textbf{29.35}       & \textbf{48.24}         \\ \bottomrule
\end{tabular}
}
\caption{Exact match accuracies for the zero-shot \textbf{cross-lingual setting}. ``Avg.'' denotes the averaged performance over all target languages (English excluded). The results of X2Parser and NLM are averaged over five runs.}
\label{tab:cross_lingual_results}
\end{table}

\section{Results and Discussion}
\subsection{Quantitative Results}
\subsubsection{Cross-Lingual Setting}
As we can see from Table~\ref{tab:cross_lingual_results}, X2Parser achieves similar performance in English compared to Seq2Seq-based models, while it significantly outperforms them in the zero-shot cross-lingual setting, with $\sim$10\% accuracy improvement on average. In the English training process, the Seq2Seq-based models can well learn the specific scope of tokens that need to be copied and assigned to a specific label type based on numerous training data.
However, these models will easily lose effectiveness when the input sequences are in target languages due to the inherent variances across languages and the difficulty of generating hierarchical representations. X2Parser separates the compositional semantic parsing task into predicting intents and slots individually, which lowers the task difficulty and boosts its zero-shot adaptation ability to target languages.
Interestingly, we find that compared to \textit{Seq2Seq w/ XLM-R}, X2Parser greatly boosts the performance on target languages that are topologically close to English (e.g., French (fr)) with more than 10\% scores, while the improvements for languages that are topologically distant from English (e.g., Thai (th) and Hindi (hi)) are relatively limited. 
We argue that the large discrepancies between English and Thai make the representation alignment quality between English and Thai (Hindi) in XLM-R relatively low, and their different language patterns lead to unstable slot and intent predictions. These factors limit the improvement X2Parser can achieve on the adaptation to topologically distant languages. 

From Table~\ref{tab:cross_lingual_results}, although NLM achieves marginally lower performance in English compared to \textit{Seq2Seq w/ XLM-R}, it produces significant improvements in target languages. This can be attributed to the fact that NLM leverages the same task decomposition as X2Parser, which further indicates the effectiveness of decomposing the compositional semantic parsing task into intent and slot predictions for low-resource scenarios.
Additionally, X2Parser surpasses NLM by $\sim$2\% exact match accuracy on average in target languages. We conjecture that the stacked layers in NLM could make the model confused about which layer needs to generate which entity types, and this confusion is aggravated in the zero-shot cross-lingual setting where no training data are available. However, our fertility-based method helps the model implicitly learn the structure of hierarchical slots by predicting the number of labels for each token, which allows the slot classifier to predict the slot types more easily in the cross-lingual setting.

\begin{table}[!t]
\setlength{\tabcolsep}{10pt}
\renewcommand{\arraystretch}{1.3}
\centering
\resizebox{0.99\textwidth}{!}{
\begin{tabular}{ccccccccccccc}
\toprule
\multicolumn{1}{c}{\textbf{Model}}    & \textbf{Alarm} & \textbf{Call.} & \textbf{Event} & \textbf{Msg.}  & \textbf{Music} & \textbf{News}  & \textbf{People} & \textbf{Recipe} & \textbf{Remind} & \textbf{Timer} & \multicolumn{1}{c}{\textbf{Weather}} & \textbf{Avg.}  \\ \midrule
\multicolumn{1}{c}{Seq2Seq}  & 67.94   & 64.25      & 61.93      & 50.11      & 32.20   & 43.20      & 52.54       & 34.21       & 46.32       & 44.83      & \multicolumn{1}{c}{73.58}        &  51.92     \\
\multicolumn{1}{c}{NLM}      & 76.32 & 70.02 & 73.60 & 70.58 & \textbf{56.52} & 58.01 & 67.33  & 50.01  & 57.28  & 64.37 & \multicolumn{1}{c}{80.15}   & 65.83 \\ \midrule
\multicolumn{1}{c}{X2Parser} & \textbf{76.72} & \textbf{73.16} & \textbf{77.33} & \textbf{71.45} & 55.19 & \textbf{64.43} & \textbf{69.77}  & \textbf{51.78}  & \textbf{58.86}  & \textbf{65.98} & \multicolumn{1}{c}{\textbf{81.17}}   & \textbf{67.80} \\ \bottomrule
\end{tabular}
}
\caption{Exact match accuracies (averaged over three runs) for the \textbf{cross-domain setting} in English. The scores represent the performance for the corresponding target domains. We use 10\% of training samples in the target domain. ``Seq2Seq'' denotes the ``Seq2Seq w/ XLM-R'' baseline (same for the following tables and figures).}
\label{tab:cross_domain_results}
\end{table}

\subsubsection{Cross-Domain Setting}
As shown in Table~\ref{tab:cross_domain_results}, X2Parser and NLM notably surpass the Seq2Seq model, with $\sim$15\% improvements on the averaged scores. This can be largely attributed to the effectiveness of our proposed task decomposition for low-resource scenarios. 
Seq2Seq models need to learn when to generate the label, when to copy tokens from the inputs, and when to produce the end of the label to generate hierarchical representations. This generation process requires a relatively large number of data samples to learn, which leads to the weak few-shot cross-domain performance for the Seq2Seq model. Furthermore, X2Parser outperforms NLM, with a $\sim$2\% averaged score.
We conjecture that our fertility classifier guides the model to learn the inherent hierarchical information from the user queries, making it easier for the slot classifier to predict slot types for each token. However, the NLM's slot classifier, which consists of multiple stacked layers, needs to capture the hierarchical information and correctly assign slot labels of different levels to the corresponding stacked layer, which requires relatively larger data to learn.

\begin{table}[!t]
\setlength{\tabcolsep}{10pt}
\renewcommand{\arraystretch}{1.3}
\centering
\resizebox{0.99\textwidth}{!}{
\begin{tabular}{ccccccccccccc}
\toprule
\multicolumn{1}{c}{\textbf{Model}}    & \textbf{Alarm} & \textbf{Call.} & \textbf{Event} & \textbf{Msg.}  & \textbf{Music} & \textbf{News}  & \textbf{People} & \textbf{Recipe} & \textbf{Remind} & \textbf{Timer} & \multicolumn{1}{c}{\textbf{Weather}} & \textbf{Avg.}  \\ \midrule
\multicolumn{1}{c}{Seq2Seq}  & 34.29      &  47.00     & 41.81      & 25.86      & 19.21      & 25.39      &  22.13      & 16.12       & 9.80       &  20.01     & \multicolumn{1}{c}{36.90}        &  22.25     \\
\multicolumn{1}{c}{NLM}      & 48.53 & 43.30 & 44.62 & 43.32 & 36.25 & 28.60 & 43.29  & 28.54  & 20.50  & 34.16 & \multicolumn{1}{c}{59.57}   & 39.15 \\ \midrule
\multicolumn{1}{c}{X2Parser} & \textbf{48.72} & \textbf{51.30} & \textbf{53.22} & \textbf{43.99} & \textbf{37.25} & \textbf{34.85} & \textbf{45.97}  & \textbf{32.99}  & \textbf{27.87}  & \textbf{36.61} & \multicolumn{1}{c}{\textbf{60.05}}   & \textbf{42.98} \\ \bottomrule
\end{tabular}
}
\caption{Exact match accuracies (averaged over three runs) for the \textbf{cross-lingual cross-domain setting}. The result for each domain is the averaged performance over all target languages. We use 10\% of training samples in the English target domain, and do not use any data in the target languages.}
\label{tab:cross_lingual_cross_domain_results}
\end{table}

\begin{figure*}[!t]
    \centering
    \resizebox{\textwidth}{!}{
    \includegraphics{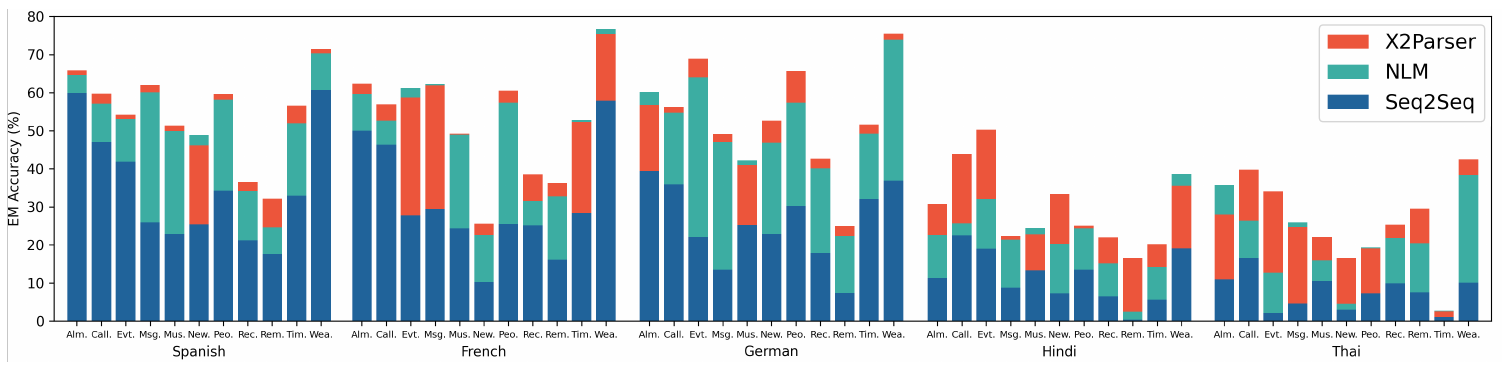}
    }
    \caption{Full \textbf{cross-lingual cross-domain} results (across all target languages of target domains) for Table~\ref{tab:cross_lingual_cross_domain_results}.}
    \label{fig:x2-combined-results}
\end{figure*}

\paragraph{Cross-Lingual Cross-Domain Setting}
From Table~\ref{tab:cross_lingual_cross_domain_results} and Figure~\ref{fig:x2-combined-results}, we can further observe the effectiveness of our proposed task decomposition and X2Parser in the cross-lingual cross-domain setting. X2Parser and NLM consistently outperform the Seq2Seq model in all target languages of the target domains and boost the averaged exact match accuracy by $\sim$20\%. Additionally, from Table~\ref{tab:cross_lingual_cross_domain_results}, X2Parser also consistently outperforms NLM on all 11 domains and surpasses it by 3.84\% accuracy on average. From Figure~\ref{fig:x2-combined-results}, X2Parser greatly improves on NLM in topologically distant languages (i.e., Hindi and Thai). It illustrates the powerful transferability and robustness of the fertility-based slot prediction that enables X2Parser to have a good zero-shot cross-lingual performance after it is fine-tuned to the target domain.

\begin{figure}[!t]
\begin{minipage}{.33\textwidth}
  \centering
  \includegraphics[width=1\linewidth]{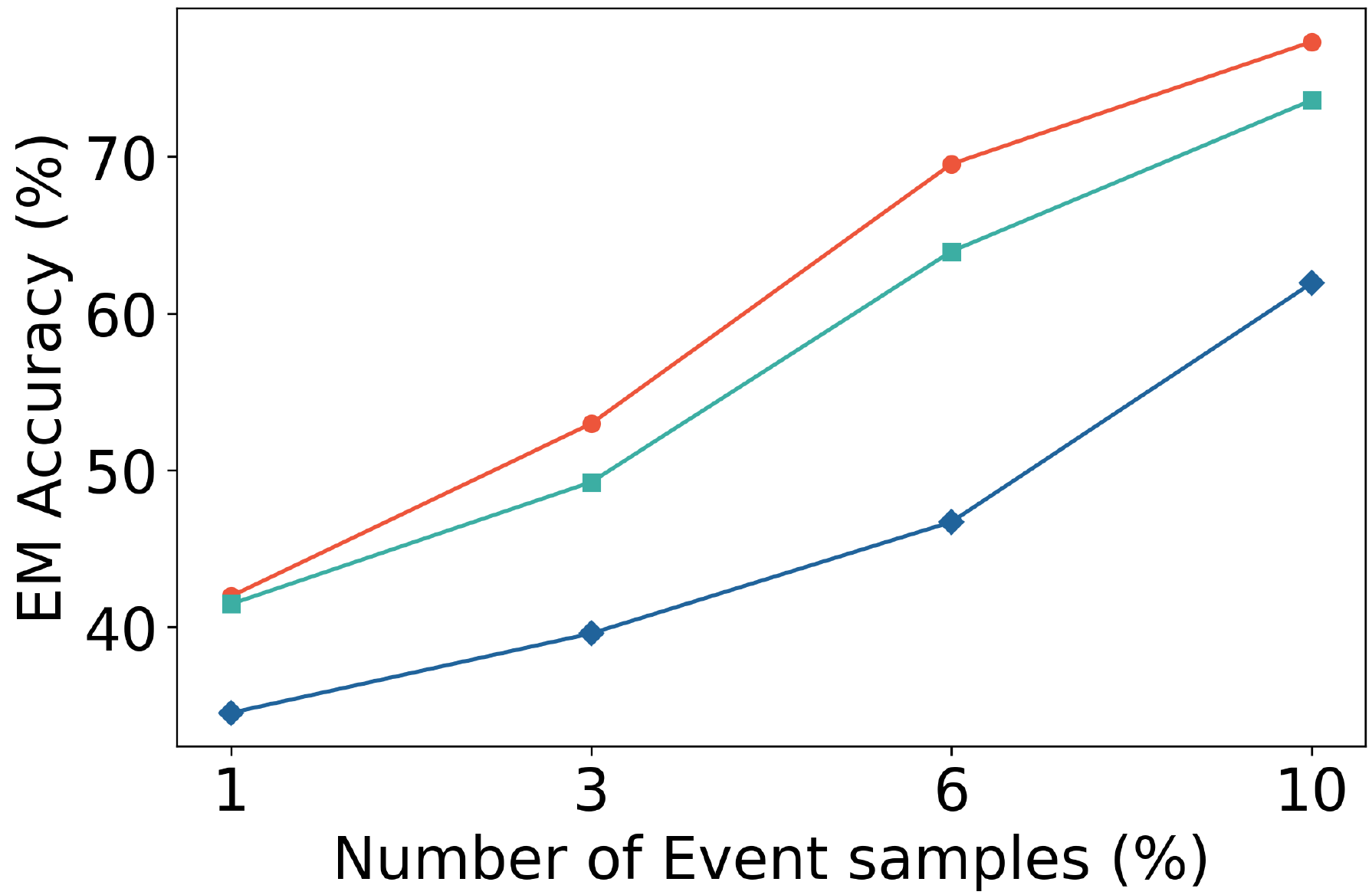}  
\end{minipage}
\begin{minipage}{.33\textwidth}
  \centering
  \includegraphics[width=\linewidth]{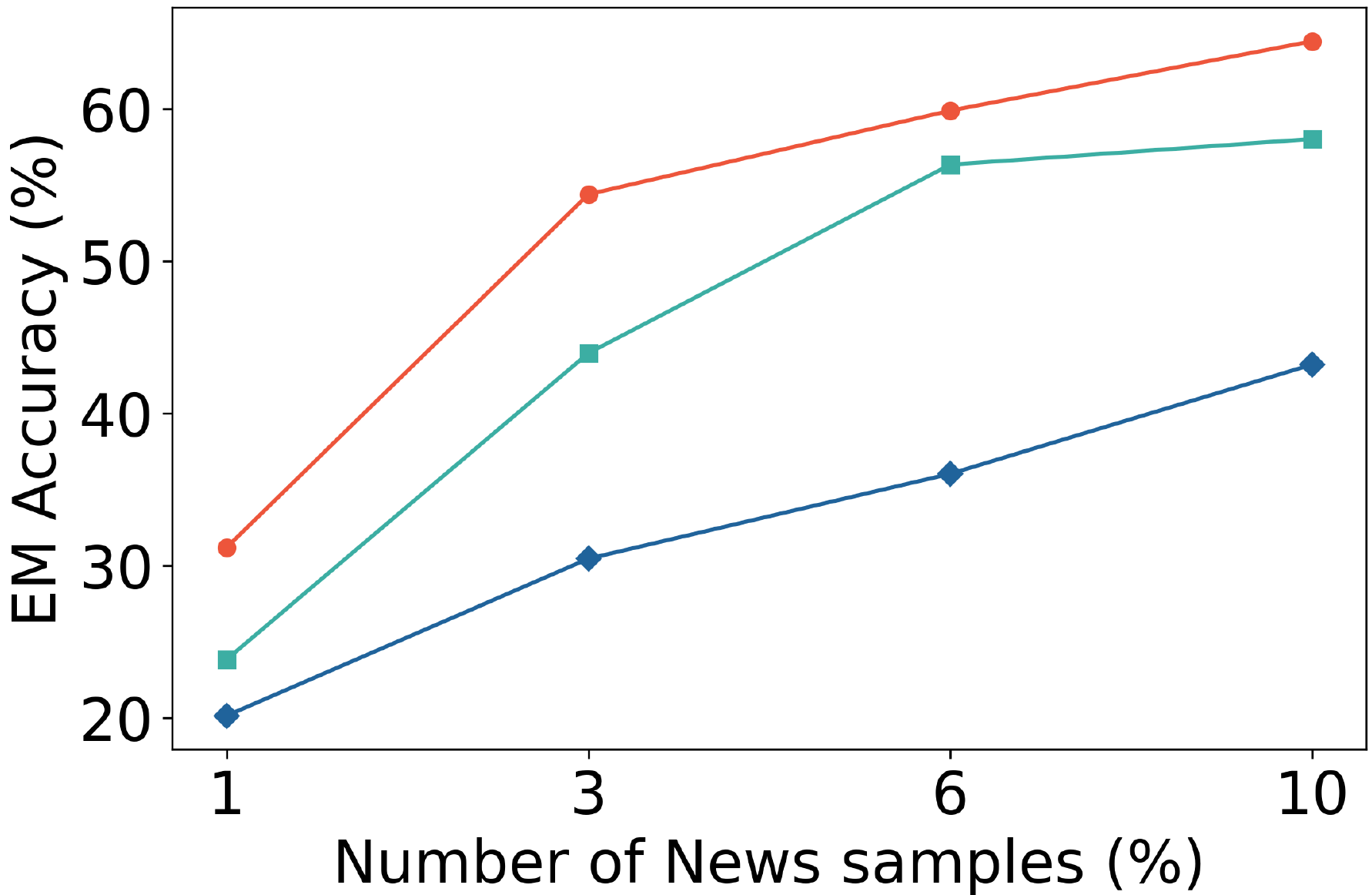}  
\end{minipage}
\begin{minipage}{.33\textwidth}
  \centering
  \includegraphics[width=\linewidth]{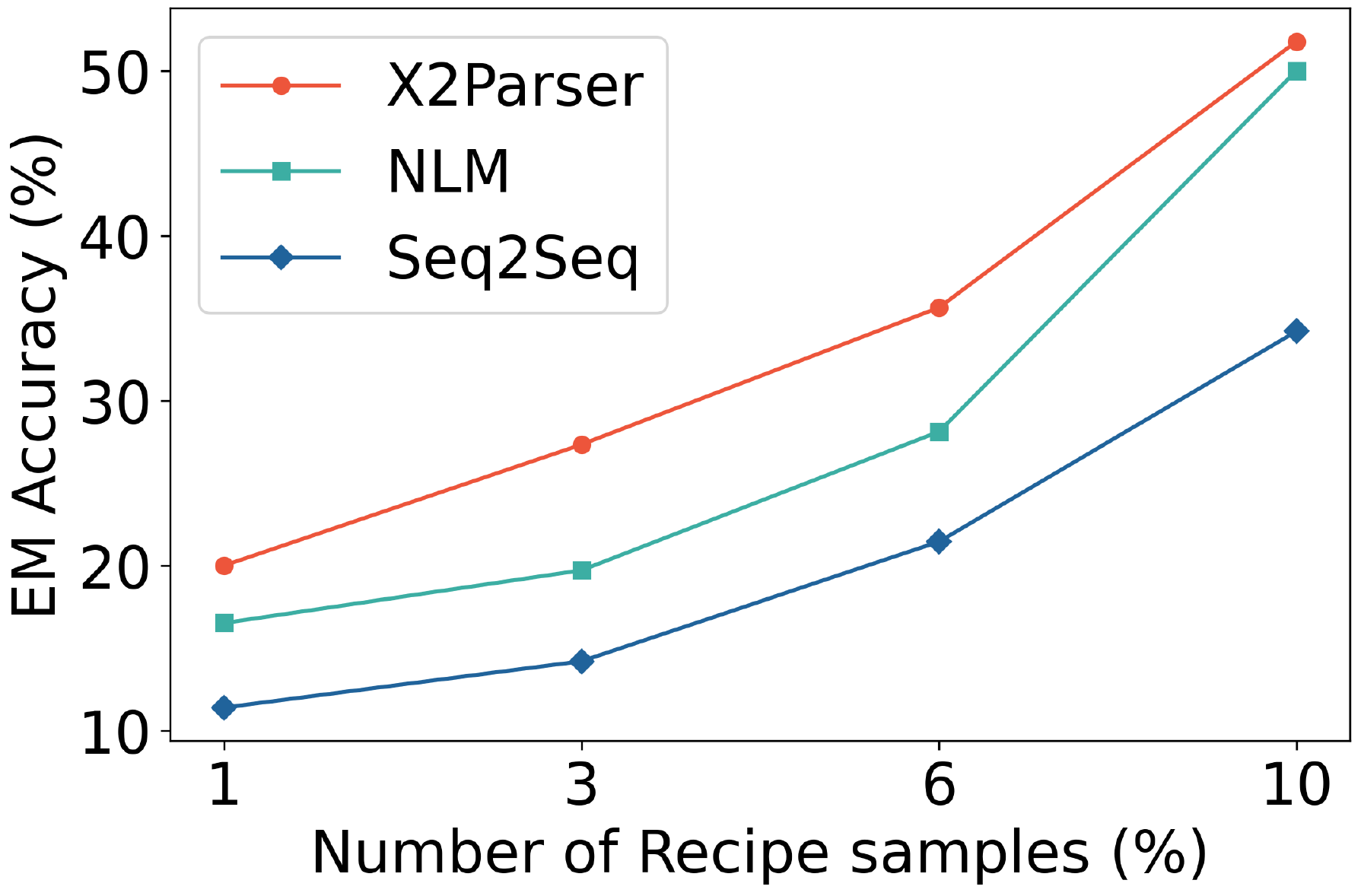}
\end{minipage}
\caption{Few-shot exact match results on the \textbf{cross-domain setting} for Event, News and Recipe target domains.}
\label{fig:results-cross-domain}
\end{figure}

\begin{figure}[!t]
\begin{minipage}{.33\textwidth}
  \centering
  \includegraphics[width=1\linewidth]{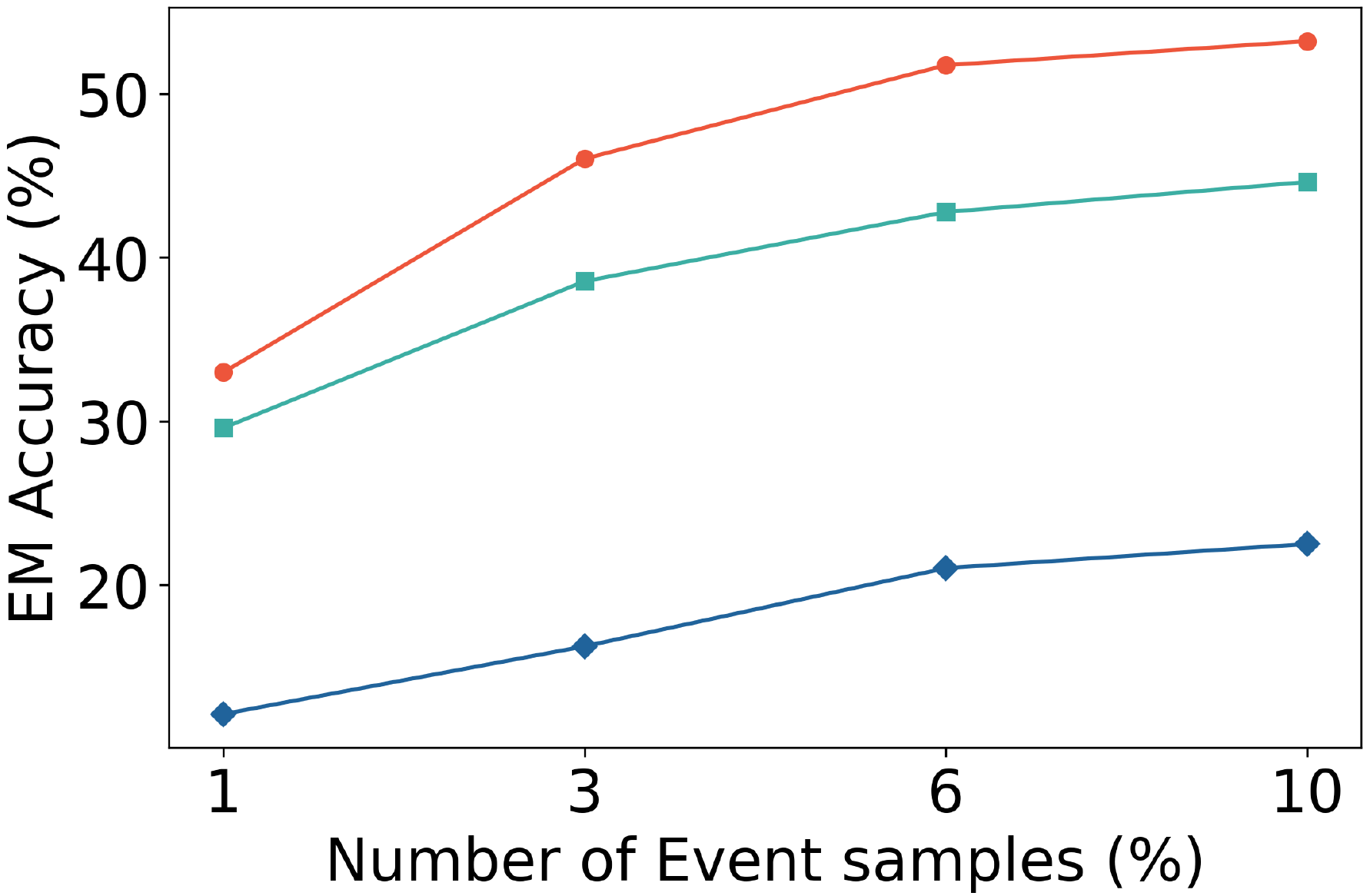}
\end{minipage}
\begin{minipage}{.33\textwidth}
  \centering
  \includegraphics[width=\linewidth]{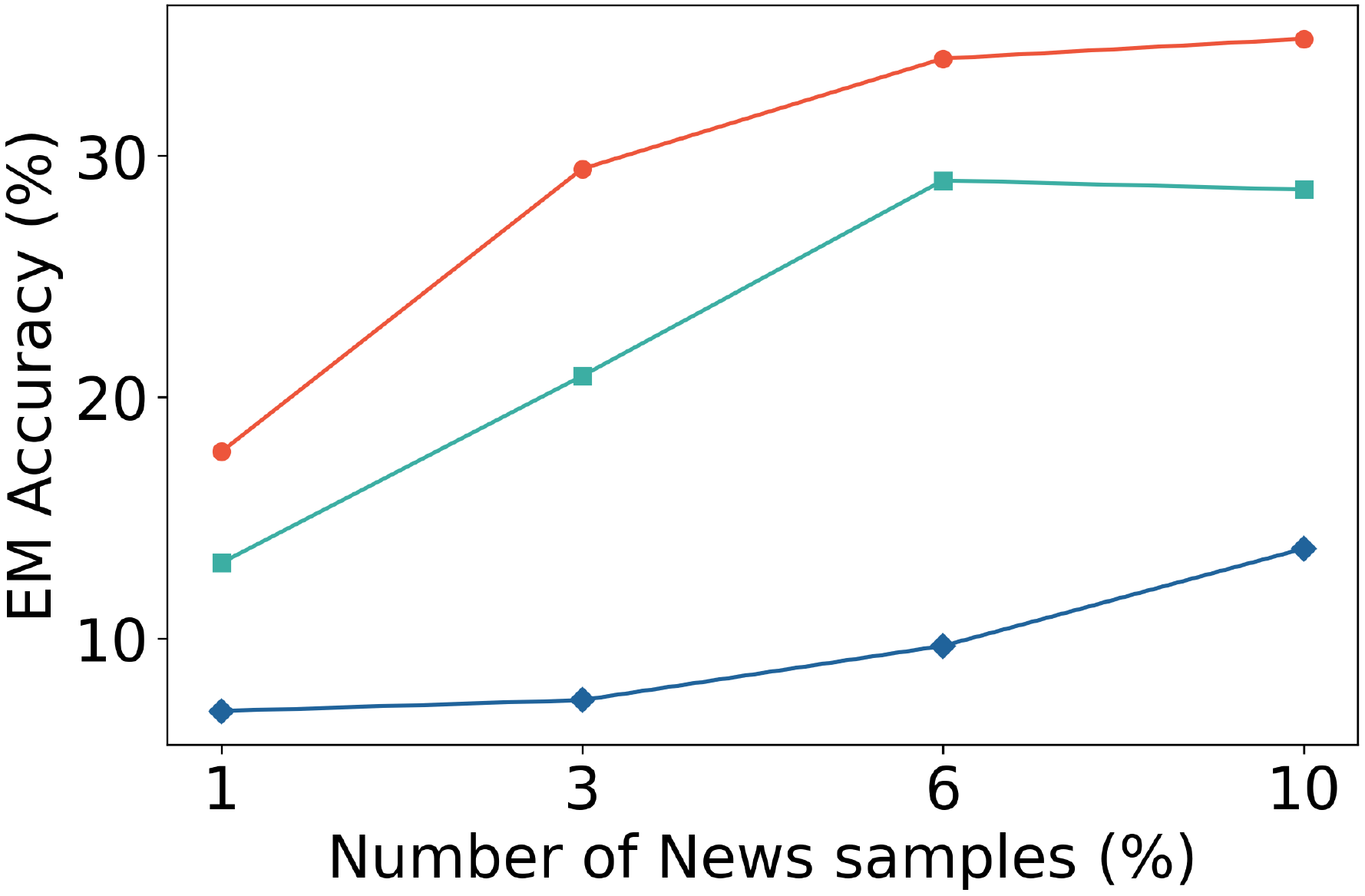}
\end{minipage}
\begin{minipage}{.33\textwidth}
  \centering
  \includegraphics[width=\linewidth]{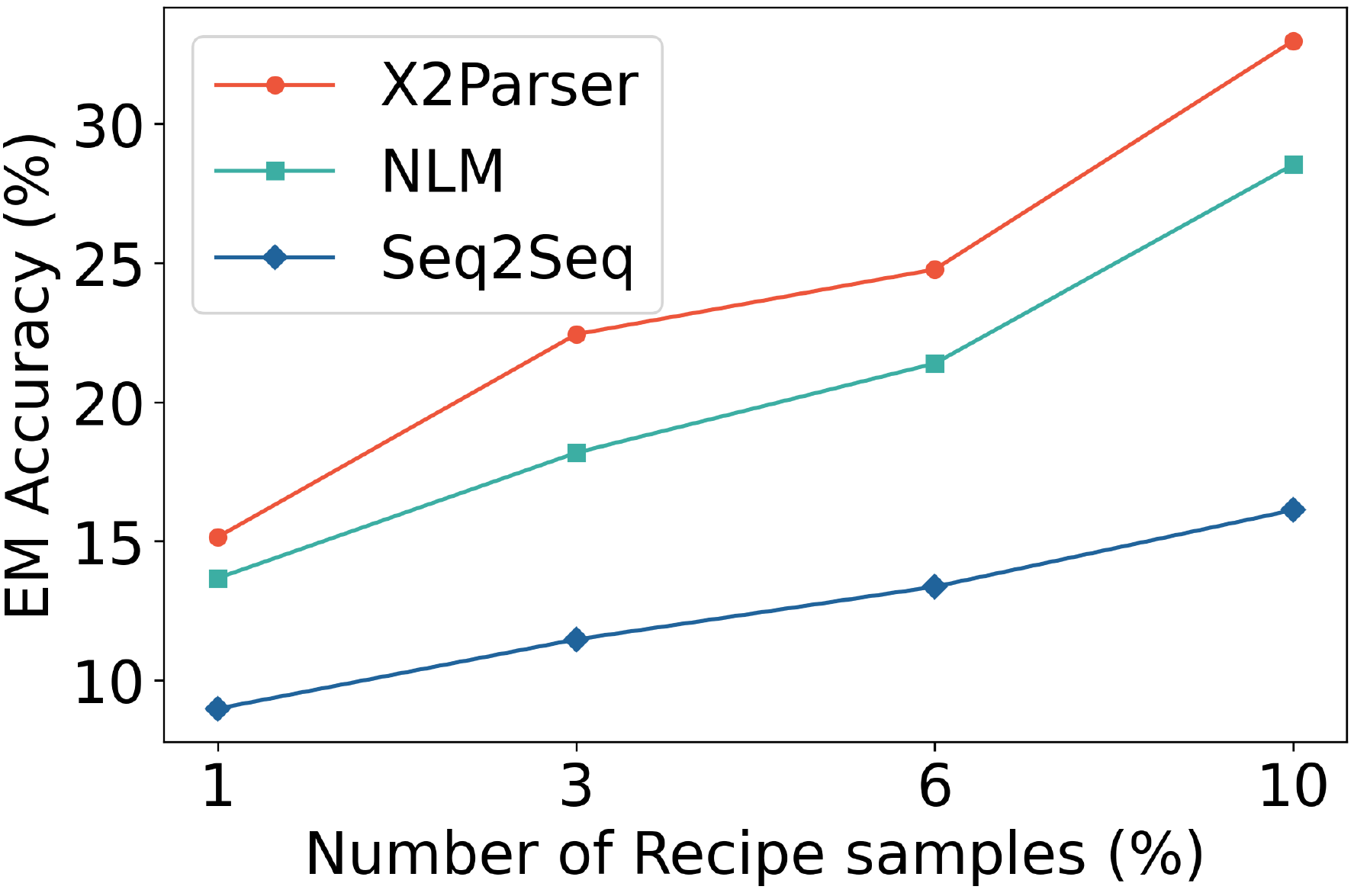}
\end{minipage}
\caption{Few-shot exact match results on the \textbf{cross-lingual cross-domain setting} for Event, News and Recipe target domains. The results are averaged over all target languages.}
\label{fig:results-x2}
\end{figure}

\subsection{Few-shot Analysis}
We conduct few-shot experiments using different sample sizes from the target domain for the cross-domain and cross-lingual cross-domain settings. The few-shot results on the Event, News, and Recipe target domains for both settings\footnote{We only report three domains for brevity, and place the full results for all 11 target domains in the Appendix~\ref{appendix:few-shot-results}.} are shown in Figure~\ref{fig:results-cross-domain} and Figure~\ref{fig:results-x2}. We find that the performance of the Seq2Seq model is generally poor in both settings, especially when only 1\% of data samples are available.
With the help of the task decomposition, NLM and X2Parser remarkably outperform the Seq2Seq model in various target domains for both the cross-domain and cross-lingual cross-domain settings across different few-shot scenarios (from 1\% to 10\%).
Moreover, X2Parser consistently surpasses NLM for both the cross-domain and cross-lingual cross-domain settings in different few-shot scenarios, which further verifies the strong adaptation ability of our model. 

Interestingly, we observe that the improvement of X2Parser over Seq2Seq grows as the number of training samples increases. For example, in the cross-lingual cross-domain setting of the event domain, the improvement goes from 20\% to 30\% as the training data increases from 1\% to 10\%.
We conjecture that in the low-resource scenario, the effectiveness of X2Parser will be greatly boosted when a relatively large data samples are available, while the Seq2Seq model needs much more training data to achieve good performance.

\begin{table}[!t]
\setlength{\tabcolsep}{10pt}
\renewcommand{\arraystretch}{1.3}
\centering
\resizebox{0.99\textwidth}{!}{
\begin{tabular}{ccccccccccccc}
\toprule
\multirow{2}{*}{\textbf{Model}} & \multicolumn{2}{c}{\textbf{Spanish}} & \multicolumn{2}{c}{\textbf{French}} & \multicolumn{2}{c}{\textbf{German}} & \multicolumn{2}{c}{\textbf{Hindi}} & \multicolumn{2}{c}{\textbf{Thai}} & \multicolumn{2}{c}{\textbf{Average}} \\ \cmidrule(lr){2-3} \cmidrule(lr){4-5} \cmidrule(lr){6-7} \cmidrule(lr){8-9} \cmidrule(lr){10-11} \cmidrule(lr){12-13}
                       & \textbf{NN}           & \textbf{Nested}       & \textbf{NN}          & \textbf{Nested}       & \textbf{NN}          & \textbf{Nested}       & \textbf{NN}          & \textbf{Nested}      & \textbf{NN}         & \textbf{Nested}      & \textbf{NN}           & \textbf{Nested}       \\ \midrule
Seq2Seq                & 56.21        & 29.38        & 48.11       & 32.83        & 46.02       & 20.25        & 37.84       & 22.30       & 33.27      & 13.56       & 44.29        & 23.66        \\
NLM                    & 65.65        & \textbf{41.95}        & 61.02       & 42.91        & 56.90       & 37.94        & 36.48       & 24.36       & 34.15      & 15.70       & 50.84        & 32.57        \\ \midrule
X2Parser               & \textbf{66.69}        & 39.19        & \textbf{63.45}       & \textbf{44.28}        & \textbf{58.43}       & \textbf{39.71}        & \textbf{42.64}       & \textbf{28.55}       & \textbf{35.96}      & \textbf{16.67}       & \textbf{53.43}        & \textbf{33.68}        \\ \bottomrule
\end{tabular}
}
\caption{Zero-shot cross-lingual exact match accuracies for nested and non-nested (NN) cases.}
\label{tab:nested_nonnested}
\end{table}

\subsection{Analysis on Nested \& Non-Nested Data}
To further understand how our model improves the performance, we split the test data in the MTOP dataset~\cite{li2021mtop} into nested and non-nested samples. We consider the user utterances that do not have fine-grained intents and nested slots as the non-nested data sample and the rest of the data as the nested data sample. As we can see from Table~\ref{tab:nested_nonnested}, X2Parser significantly outperforms the Seq2Seq model on both nested and non-nested user queries with an average of $\sim$10\% accuracy improvement in both cases. X2Parser also consistently surpasses NLM on all target languages in both the nested and non-nested scenarios, except for the Spanish nested case, which further illustrates the stable and robust adaptation ability of X2Parser.

\begin{figure}[t!]
    \centering
    \resizebox{0.55\textwidth}{!}{  
    \includegraphics{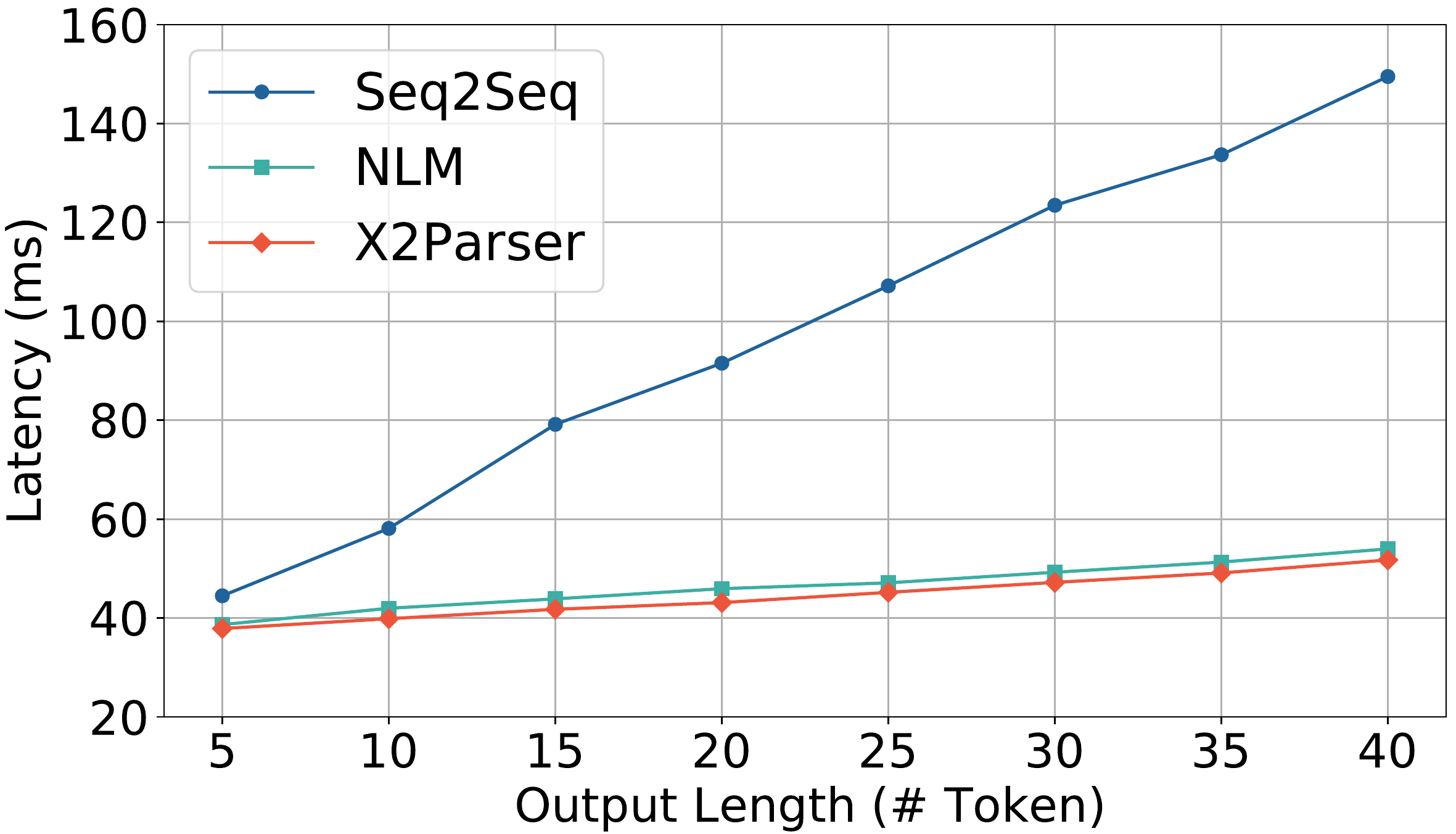}
    }
    \caption{Averaged latencies for our model and baselines on different output lengths of the MTOP dataset.}
    \label{fig:latency}
\end{figure}

\subsection{Latency Analysis}
We can see from Figure~\ref{fig:latency} that, as the output length increases, the latency discrepancy between the Seq2Seq-based model (Seq2Seq) and sequence labeling-based models (NLM and X2Parser) becomes larger, and when the output length reaches 40 tokens (around the maximum length in MTOP), X2Parser can achieve an up to 66\% reduction in latency compared to the Seq2Seq model. This can be attributed to the fact that the Seq2Seq model has to generate the outputs token by token, while X2Parser and NLM can directly generate all the outputs. In addition, the inference speed of X2Parser is slightly faster than that of NLM. This is because NLM uses several stacked layers to predict slot entities of different levels, and the higher-level layers have to wait for the predictions from lower-level layers, which slightly decreases the inference speed.

\section{Short Summary}
In this section, we develop a transferable and non-autoregressive model (X2Parser) for compositional semantic parsing task that can better learn the low-resource representations and adapt to target languages and domains with a faster inference speed. Unlike previous models that learn to generate hierarchical representations, we propose to decompose the task into intent and slot predictions so as to lower the difficulty of the task, and then we cast both prediction tasks into sequence labeling problems. After that, we further propose a fertility-based method to cope with the slot prediction task where each token could have multiple labels.
We find that simplifying task structures makes the representation learning more effective for low-resource languages and domains.
Results illustrate that X2Parser significantly outperforms strong baselines in all low-resource settings. Furthermore, our model is able to reduce the latency by up to 66\% compared to a generation-based model.

\newpage

%% file: chapter/sec-6-conclusion.tex
\chapter{Conclusion}\label{sec-conclusion}

This thesis focuses on leveraging transfer learning (i.e., cross-lingual, cross-domain) methods to address data scarcity issues for low-resource languages and domains.
We discuss the challenges in cross-lingual and cross-domain adaptations in natural language understanding (NLU), and introduce transferable models for the semantic parsing, named entity recognition, and part-of-speech tagging tasks. 
We propose language and domain adaptation approaches which improve low-resource representations and tackle language and domain discrepancies in the task knowledge transfer. In this chapter, we conclude the thesis, summarize our contributions, and discuss possible future work.

We propose a cross-lingual embedding refinement, a transferable latent variable model, and a label regularization method for cross-lingual adaptation. These methods further refine the representations of task-related keywords across languages and regularize the model training to learn robust representations for low-resource languages. We find that the representations for low-resource languages can be easily and greatly improved by focusing on just the keywords, and experiments show that our methods significantly outperform previous state-of-the-art models.

We present an Order-Reduced Transformer (ORT), which removes the dependence on the global word order information and focuses on only the local word orders, to improve the task knowledge transfer across languages. We find that modeling partial word orders instead of the whole sequence can improve the robustness of the model against the word order differences between languages and the task knowledge transfer to low-resource languages. Experiments show that ORT consistently outperforms the order-sensitive models on multiple cross-lingual NLU tasks. We further study fitting models with different amounts of word order information, and find that encoding excessive or insufficient word orders into the model results in inferior quality of the task knowledge transfer.

We collect CrossNER, a fully-labeled collection of named entity recognition (NER) data spanning five diverse domains with specialized entity categories, to catalyze the cross-domain NER research. Then, we study domain-adaptive pre-training (DAPT) on different levels of the domain corpus and pre-training strategies for the domain adaptation. We propose to integrate different levels of domain-related corpora, and conduct DAPT using a span-level masking method. We discover that focusing on a factional corpus containing domain-specialized entities and utilizing a more challenging manner of pre-training can better address the domain discrepancy issue in the task knowledge transfer.

We introduce a coarse-to-fine framework called Coach, and a cross-lingual and cross-domain parsing framework called X2Parser. Coach decomposes the semantic parsing task into predicting coarse-grained and fine-grained labels, while X2Parser simplifies the hierarchical task structures into flattened ones. We observe that simplifying task structures makes the representation learning more effective for low-resource languages and domains. Results show that our proposed frameworks significantly outperform previous state-of-the-art models in low-resource scenarios, and learning flattened representations (X2Parser) is also more efficient for long and complex queries.

Lastly, the main contribution of this thesis is that it introduces effective transfer learning approaches to greatly improve the generalization ability of NLU models to a much wider range of areas and substantially reduce the need for training samples in low-resource languages and domains. 
One of the ultimate goals of AI is to allow machines to quickly learn new tasks and adapt to new domains just like humans. This thesis successfully pushes the research in the NLU area one step towards this goal.

In future work, we can develop more advanced order-reduced modeling methods and incorporate them into the multilingual language model pre-training to improve the low-resource representation learning and boost the model's adaptation robustness to topologically distant languages. Additionally, inspired from our findings on continual pre-training, we can study how to select more effective domain-related corpora and develop more powerful pre-training strategies for domain-adaptive pre-training. 
Moreover, we can continue exploring the effectiveness of simplifying model structures to other low-resource NLU as well as natural language generation (NLG) tasks.
Furthermore, given that it is resource-consuming to train gigantic language models (GLMs) on the source language/domain data, we can investigate efficient training methods on these GLMs (e.g., prompt learning, adapters) for cross-lingual and cross-domain adaptation.



\newpage

%% file: chapter/sec-publications.tex

\chapter*{\underline{List of Publications}}
\vspace{12mm}

(* denotes equal contribution)

\begin{itemize}
    \item \textbf{Zihan Liu}, Yan Xu, Tiezheng Yu, Wenliang Dai, Ziwei Ji, Samuel Cahyawijaya, Andrea Madotto, Pascale Fung. "CrossNER: Evaluating Cross-Domain Named Entity Recognition." In Proceedings of 35th AAAI Conference on Artificial Intelligence, 2021.
    \item \textbf{Zihan Liu}, Genta Indra Winata, Samuel Cahyawijaya, Andrea Madotto, Zhaojiang Lin, Pascale Fung. "On the Importance of Word Order Information in Cross-lingual Sequence Labeling." In Proceedings of 35th AAAI Conference on Artificial Intelligence, 2021.
    \item \textbf{Zihan Liu}, Genta Indra Winata, Peng Xu, Zhaojiang Lin, Pascale Fung. "Cross-lingual Spoken Language Understanding with Regularized Representation Alignment." In Proceedings of the 2020 Conference on Empirical Methods in Natural Language Processing (EMNLP), 2020.
    \item \textbf{Zihan Liu}, Genta Indra Winata, Peng Xu, Pascale Fung. "Coach: A Coarse-to-Fine Approach for Cross-domain Slot Filling" In Proceedings of the 58th Annual Meeting of the Association for Computational Linguistics, 2020.
    \item \textbf{Zihan Liu}*, Genta Indra Winata*, Zhaojiang Lin, Peng Xu, Pascale Fung. "Attention-Informed Mixed-Language Training for Zero-shot Cross-lingual Task-oriented Dialogue Systems." In Proceedings of 34th AAAI Conference on Artificial Intelligence, 2020.
    \item \textbf{Zihan Liu}, Jamin Shin, Yan Xu, Genta Indra Winata, Peng Xu, Andrea Madotto, and Pascale Fung. "Zero-shot Cross-lingual Dialogue Systems with Transferable Latent Variables." In Proceedings of the 2019 Conference on Empirical Methods in Natural Language Processing and the 9th International Joint Conference on Natural Language Processing (EMNLP-IJCNLP), 2019.
    \item \textbf{Zihan Liu}, Genta Indra Winata, Pascale Fung. "Continual Mixed-Language Pre-Training for Extremely Low-Resource Neural Machine Translation." In Findings of the Association for Computational Linguistics: ACL-IJCNLP 2021, 2021.
    \item Tiezheng Yu*, \textbf{Zihan Liu}*, Pascale Fung. "AdaptSum: Towards Low-Resource Domain Adaptation for Abstractive Summarization." In Proceedings of the 2021 Conference of the North American Chapter of the Association for Computational Linguistics: Human Language Technologies, 2021.
    \item Genta Indra Winata*, Samuel Cahyawijaya*, \textbf{Zihan Liu}*, Zhaojiang Lin, Andrea Madotto, Peng Xu, Pascale Fung. "Learning Fast Adaptation on Cross-Accented Speech Recognition." In Proceedings of INTERSPEECH, 2020.
    
    \item Wenliang Dai, Samuel Cahyawijaya, \textbf{Zihan Liu}, Pascale Fung. "Multimodal End-to-End Sparse Model for Emotion Recognition." In Proceedings of the 2021 Conference of the North American Chapter of the Association for Computational Linguistics: Human Language Technologies, 2021.
    \item Tiezheng Yu*, Wenliang Dai*, \textbf{Zihan Liu}, Pascale Fung. "Vision Guided Generative Pre-trained Language Models for Multimodal Abstractive Summarization." In Proceedings of the 2021 Conference on Empirical Methods in Natural Language Processing (EMNLP), 2021.
    \item Wenliang Dai, \textbf{Zihan Liu}, Tiezheng Yu, Pascale Fung. "Modality-Transferable Emotion Embeddings for Low-Resource Multimodal Emotion Recognition." In Proceedings of the 1st Conference of the Asia-Pacific Chapter of the Association for Computational Linguistics and the 10th International Joint Conference on Natural Language Processing, 2020.
    \item Andrea Madotto, Samuel Cahyawijaya, Genta Indra Winata, Yan Xu, \textbf{Zihan Liu}, Zhaojiang Lin, Pascale Fung. "Learning Knowledge Bases with Parameters for Task-Oriented Dialogue Systems." In Proceedings of the 2020 Conference on Empirical Methods in Natural Language Processing: Findings, 2020.
    \item Genta Indra Winata*, Samuel Cahyawijaya*, Zhaojiang Lin, \textbf{Zihan Liu}, Peng Xu, Pascale Fung. "Meta-transfer learning for code-switched speech recognition." In Proceedings of the 58th Annual Meeting of the Association for Computational Linguistics, 2020.
    \item Genta Indra Winata*, Samuel Cahyawijaya*, Zhaojiang Lin, \textbf{Zihan Liu}, and Pascale Fung. "Lightweight and Efficient End-to-End Speech Recognition Using Low-Rank Transformer." In ICASSP 2020-2020 IEEE International Conference on Acoustics, Speech and Signal Processing (ICASSP), 2020.
    \item Zhaojiang Lin, Peng Xu, Genta Indra Winata, Farhad Bin Siddique, \textbf{Zihan Liu}, Jamin Shin, and Pascale Fung. "CAiRE: An End-to-End Empathetic Chatbot." In Proceedings of 34th AAAI Conference on Artificial Intelligence, 2020.
    \item Genta Indra Winata, Zhaojiang Lin, Jamin Shin, \textbf{Zihan Liu}, and Pascale Fung. "Hierarchical Meta-Embeddings for Code-Switching Named Entity Recognition." In Proceedings of the 2019 Conference on Empirical Methods in Natural Language Processing and the 9th International Joint Conference on Natural Language Processing (EMNLP-IJCNLP), 2019.
    
    \item \textbf{Zihan Liu}, Genta Indra Winata, Peng Xu, Pascale Fung. "X2Parser: Cross-Lingual and Cross-Domain Framework for Task-Oriented Compositional Semantic Parsing." In Proceedings of the 6th Workshop on Representation Learning for NLP (RepL4NLP), 2021.
    \item \textbf{Zihan Liu}, Genta Indra Winata, Andrea Madotto, Pascale Fung. "Exploring fine-tuning techniques for pre-trained cross-lingual models via continual learning." In Proceedings of the 5th Workshop on Representation Learning for NLP (RepL4NLP), 2020.
    \item \textbf{Zihan Liu}, Genta Indra Winata, Pascale Fung. "Zero-Resource Cross-Domain Named Entity Recognition." In Proceedings of the 5th Workshop on Representation Learning for NLP (RepL4NLP), 2020.
    \item Yan Xu, Etsuko Ishii, Genta Indra Winata, Zhaojiang Lin, Andrea Madotto, \textbf{Zihan Liu}, Peng Xu, Pascale Fung. "CAiRE in DialDoc21: Data Augmentation for Information Seeking Dialogue System." In Proceedings of the 1st Workshop on Document-grounded Dialogue and Conversational Question Answering (DialDoc), 2021.
    \item Wenliang Dai*, Tiezheng Yu*, \textbf{Zihan Liu}, Pascale Fung. "Kungfupanda at SemEval-2020 Task 12: BERT-Based Multi-Task Learning for Offensive Language Detection." In Proceedings of the Fourteenth Workshop on Semantic Evaluation, 2020.
    \item Nayeon Lee*, \textbf{Zihan Liu}*, Pascale Fung. "Team yeon-zi at semeval-2019 task 4: Hyperpartisan news detection by de-noising weakly-labeled data." In Proceedings of the 13th International Workshop on Semantic Evaluation, 2019.
    \item Dan Su, Yan Xu, Genta Indra Winata, Peng Xu, Hyeondey Kim, \textbf{Zihan Liu}, and Pascale Fung. "Generalizing Question Answering System with Pre-trained Language Model Fine-tuning." In EMNLP 2019 MRQA Workshop, 2019.
    \item \textbf{Zihan Liu}, Yan Xu, Genta Indra Winata, and Pascale Fung. "Incorporating Word and Subword Units in Unsupervised Machine Translation Using Language Model Rescoring." In Proceedings of the Fourth Conference on Machine Translation (Volume 2: Shared Task Papers, Day 1), 2019.
    \item Zhaojiang Lin, Andrea Madotto, Genta Indra Winata, \textbf{Zihan Liu}, Yan Xu, Cong Gao, and Pascale Fung. "Learning to learn sales prediction with social media sentiment." In Proceedings of the First Workshop on Financial Technology and Natural Language Processing, 2019.
    \item Genta Indra Winata, Samuel Cahyawijaya, \textbf{Zihan Liu}, Zhaojiang Lin, Andrea Madotto, Pascale Fung. "Are Multilingual Models Effective in Code-Switching?" Computational Approaches to Linguistic Code-Switching (CALCS), 2021.
    \item Zhaojiang Lin*, \textbf{Zihan Liu}*, Genta Indra Winata*, Samuel Cahyawijaya*, Andrea Madotto*, Yejin Bang, Etsuko Ishii, Pascale Fung. "XPersona: Evaluating Multilingual Personalized Chatbot." NLP for Conversational AI (NLP4ConvAI), 2021.
\end{itemize}

%% file: chapter/sec-appendix.tex
\appendix


\chapter{Annotation Details of CrossNER}
\label{appendix:annotation-details-crossner}

\section{Annotator Training}
We gather the NER experts who are familiar with the NER task and the annotation rules. We train the annotators before he/she starts annotating. For each domain, we first give the annotator the annotation instruction and 100 annotated examples (annotated by the NER experts) and ask them to check the possible annotation errors.
After the checking process, the NER experts will inspect the results and tell the annotators the mistakes they made in the checking stage. Hence, in this process, the annotators are able to learn how to annotate the NER samples for specific domains.

\section{Annotation Instructions}
We split the annotation instructions into two parts, namely, general instructions, as well as domain-specific instructions. We describe the instructions as follows:

\subsection{General Instructions}
Each data sample requires two well-trained NER annotators to annotate it and one NER expert to double check it and give final labels. The annotation process consists of three steps. First, one annotator needs to detect and categorize the entities in the given sentences. Second, the other annotator checks the annotations made by the first annotator, makes markings if he/she thinks that the annotations \textbf{could} be wrong and gives another annotation. Finally, the expert first goes through the annotations again and checks for possible mistakes, and then makes the final decision for disagreements between the first two annotators. Notice that we mark the tokens with hyperlinks in Wikipedia, and it is highly likely that these tokens are named entities. When one entity contains another entity, we should give labels to the entity with a larger span. For example, ``Fellow of the Royal Society'' is an entity (a award entity), while ``Royal Society'' is another entity (an organization entity), we should annotate ``Fellow of the Royal Society'' instead of ``Royal Society''.
The requirements for the annotators in different annotation stages are as follows:
\begin{itemize}
    \item If you are in the first annotation stage, you need to detect and categorize the entities carefully, and check the pre-labeled entities (annotated by DBpedia Ontology described in the main paper) are correct or not, and give the correct annotations if you think the labels are wrong. Notice that the pre-label entity might not be an entity, and some tokens not labeled as entities could be entities.
    \item If you are in the second annotation stage, you need to focus on looking for the possible annotations mistakes made by the first annotator, and give another annotation if you think the labels \textbf{could} be wrong. Additionally, you need to detect and categorize the entities that are missed by the first annotator.
    \item If you are in the third annotation stage (only applicable for NER experts), you need to carefully go through the annotations and correct the possible mistakes, and in the meantime, you need to check the corrected annotations made by the second annotator and then makes final decisions for the disagreements between the two annotators. If you are unsure about the annotations, you need to confer with the two annotators.
\end{itemize}

\subsection{Domain-Specific Instructions}
We list the annotation details for the five domains, namely, politics, natural science, music, literature, and artificial intelligence.
\begin{itemize}
    \item \textbf{Politics:} The entity category list for this domain is \{person, organization, location, politician, political party, election, event, country, miscellaneous\}. The annotation rules for the abovementioned entity categories are as follows:
    \begin{itemize}
        \item \textit{Person:} The name of a person should be annotated as a person entity.
        \item \textit{Politician:} The politician entity. If a person entity is a politician, you should label this person as a politician entity instead of a person entity.
        \item \textit{Location:} The location entity, including place, bridge, city, county and etc.
        \item \textit{Country:} The country entity.
        \item \textit{Event:} The event entity, which includes festival, war, summit, Campaign, and etc.
        \item \textit{Election:} The election entity. If an event entity is an election event, you should label it as an election entity instead of an event entity.
        \item \textit{Organization:} The organization entity.
        \item \textit{political party:} The political party entity. If an organization entity is a political party, you should label it as a political entity instead of an organization entity.
        \item \textit{Miscellaneous:} An entity needs to be classified as the miscellaneous type if it does not belong to any other category.
    \end{itemize}
    Note that the annotation rules for some general entity categories (i.e., person, location, organization, event, country, miscellaneous) are the same as those in the other domains, and we do not put the annotation rules for these entity categories for the other domains.
    \item \textbf{Natural Science:} This domain contains the area of biology, chemistry, and astrophysics. The entity category list for this domain is \{scientist, person, university, organization, country, location, discipline, enzyme, protein, chemical element, chemical compound, astronomical object, academic journal, event, theory, award, miscellaneous\}.\footnote{Since data samples come from Wikipedia instead of academic papers from the natural science field, the entities are generally not difficult to categorize by annotators that are not working on these areas.} The annotation rules are as follows:
    \begin{itemize}
        \item \textit{University:} The university entity.
        \item \textit{Discipline:} The discipline entity. It contains the areas and subareas of biology, chemistry and astrophysics, such as quantum chemistry.
        \item \textit{Theory:} The theory entity. It includes law and theory entities, such as ptolemaic planetary theories.
        \item \textit{Award:} The award entity.
        \item \textit{Scientist:} If a person entity is a scientist, you should label this person as a scientist entity instead of a person entity.
        \item \textit{Protein:} The protein entity.
        \item \textit{Enzyme:} Notice that enzyme is a special type of protein. Hence, if a protein entity is an enzyme, you should label this protein as an enzyme entity instead of a protein entity.
        \item \textit{Chemical element:} The chemical element entity. Basically, this category contains the chemical elements from the periodic table.
        \item \textit{Chemical compound:} The chemical compound entity. If a chemical compound entity do not belong to protein or enzyme, you should label it as a chemical compound entity.
        \item \textit{Astronomical object:} The astronomical object entity. 
        \item \textit{Academic journal:} The academic journal entity.
    \end{itemize}
    \item \textbf{Music:} The entity category list for this domain is \{music genre, song, band, album, musical artist, musical instrument, award, event, country, location, organization, person, miscellaneous\}. The annotation rules are as follows:
    \begin{itemize}
        \item \textit{Music genre:} The music genre entity, such as country music, folk music and jazz.
        \item \textit{Song:} The song entity.
        \item \textit{Band:} The band entity. If an organization belongs to a band, you should label it as a band entity instead of an organization entity.
        \item \textit{Album:} The album entity.
        \item \textit{Musical artist:} The musical artist entity. It a person is working on the music area (e.g., he/she is a singer, composer, or songwriter), you should label it as a musical artist entity instead of a person entity.
        \item \textit{Musical instrument:} The musical instrument entity, such as piano.
    \end{itemize}
    \item \textbf{Literature:} The entity category list for this domain is \{book, writer, award, poem, event, magazine, literary genre, person, location, organization, country, miscellaneous\}. The annotation rules are as follows:
    \begin{itemize}
        \item \textit{Book:} The book entity.
        \item \textit{Poem:} The poem entity.
        \item \textit{Writer:} The writer entity. If a person is working on literature (including writer, novelist, scriptwriter, poet, and etc), you should label it as a writer entity instead of a person entity.
        \item \textit{Magazine:} The magazine that publishes articles as well as any other literature work.
        \item \textit{Literary genre:} The literary genre entity, such as novel and science fiction.
    \end{itemize}
    \item \textbf{Artificial Intelligence:} The entity category list for this domain is \{field, task, product, algorithm, researcher, metrics, university, country, person, organization, location, programming language, conference, miscellaneous\}. The annotators for this domain are all working on the AI area. The annotation rules are as follows:
    \begin{itemize}
        \item \textit{Researcher:} The researcher entity. If a person is working on research (including professor, Ph.D. student, researcher in companies, and etc), you should label it as a researcher entity instead of a person entity.
        \item \textit{Field:} The research field entity, such as machine learning, deep learning, and natural language processing.
        \item \textit{Task:} The specific task entity in the research field, such as machine translation and object detection.
        \item \textit{Product:} The product entity that includes the product (e.g., a certain kind of robot like Pepper), system (e.g., facial recognition system) and toolkit (e.g., Tensorflow and PyTorch)
        \item \textit{Algorithm:} The algorithm entity. It contains algorithms (e.g., decision trees) and models (e.g, CNN and LSTM).
        \item \textit{Metrics:} The evaluation metrics, such as F1-score.
        \item \textit{Programming Language:} The programming language, such as Java and Python.
        \item \textit{Conference:} The conference entity. It contains conference and journal entities.
    \end{itemize}
\end{itemize}

\chapter{X2Parser}

\section{Nested Intent Label Construction}
\label{appendix:nested_intent_label_construction}

In this section, we further describe how we convert the fine-grained intent prediction into a sequence labeling task (each token has only one label). We use a few examples to illustrate our intent label construction method.

\begin{figure}[!ht]
    \centering
    \subfigure[A labeling example for non-nested intent.]{
        \includegraphics[scale=0.78]{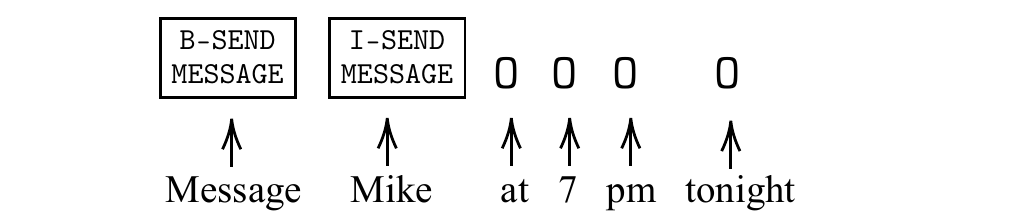}
        \label{fig:nonnested_intent}
    }
    \subfigure[A labeling example for nested intent.]{
        \includegraphics[scale=0.78]{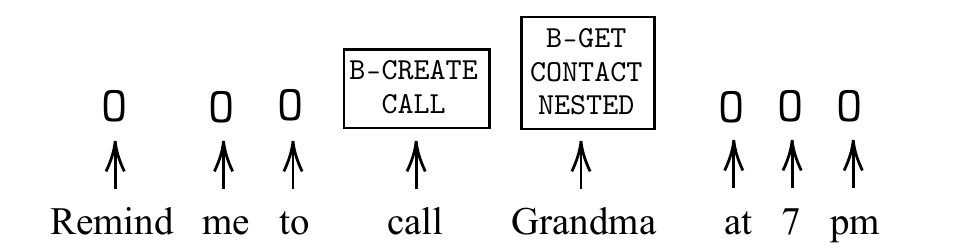}
        \label{fig:nested_intent}
    }
    \caption{Labeling examples for intents.}
    \label{fig:nested_and_nonnested_intent}
\end{figure}

As illustrated in Figure~\ref{fig:nonnested_intent}, when there are no nested intents in the input utterance, we follow the BIO structure to give intent labels. 
We can see from Figure~\ref{fig:nested_intent} that ``call Grandma'' is a \texttt{CREATE-CALL} intent and ``Grandma'' is a \texttt{GET-CONTACT} intent. Hence, the \texttt{GET-CONTACT} intent is nested in the \texttt{CREATE-CALL} intent. We use a special intent label (with ``NESTED'') for the ``\texttt{GET-CONTACT}'' intent (\texttt{B-GET-CONTACT-NESTED}) to represent that this intent is nested in another intent, and hence, the scope of the \texttt{CREATE-CALL} intent is automatically expanded from ``call'' to ``call Grandma''.~\footnote{We notice that if two intents have overlaps, one intent either fully covers the other intent or is fully covered by the other intent.}

Note that we cannot apply this labeling method to the slot prediction since one token in the user utterance could be the starting token for more than one slot entity. If that is the case, we have to use more than one slot label for this token to denote the starting position for each slot entity. 
Given that in the MTOP dataset, one token will not be the starting token of more than one intent, we can apply this method for the intent label construction.
In the future, when more complex and sophisticated datasets are collected for the task-oriented compositional semantic parsing task, where there could exist more than one intent label for each token, we can always use the fertility-based method (currently applied for the slot prediction) for the intent prediction.

\section{Full Few-shot Results}
\label{appendix:few-shot-results}

\begin{figure}[!ht]
\begin{minipage}{.33\textwidth}
  \centering
  \includegraphics[width=1\linewidth]{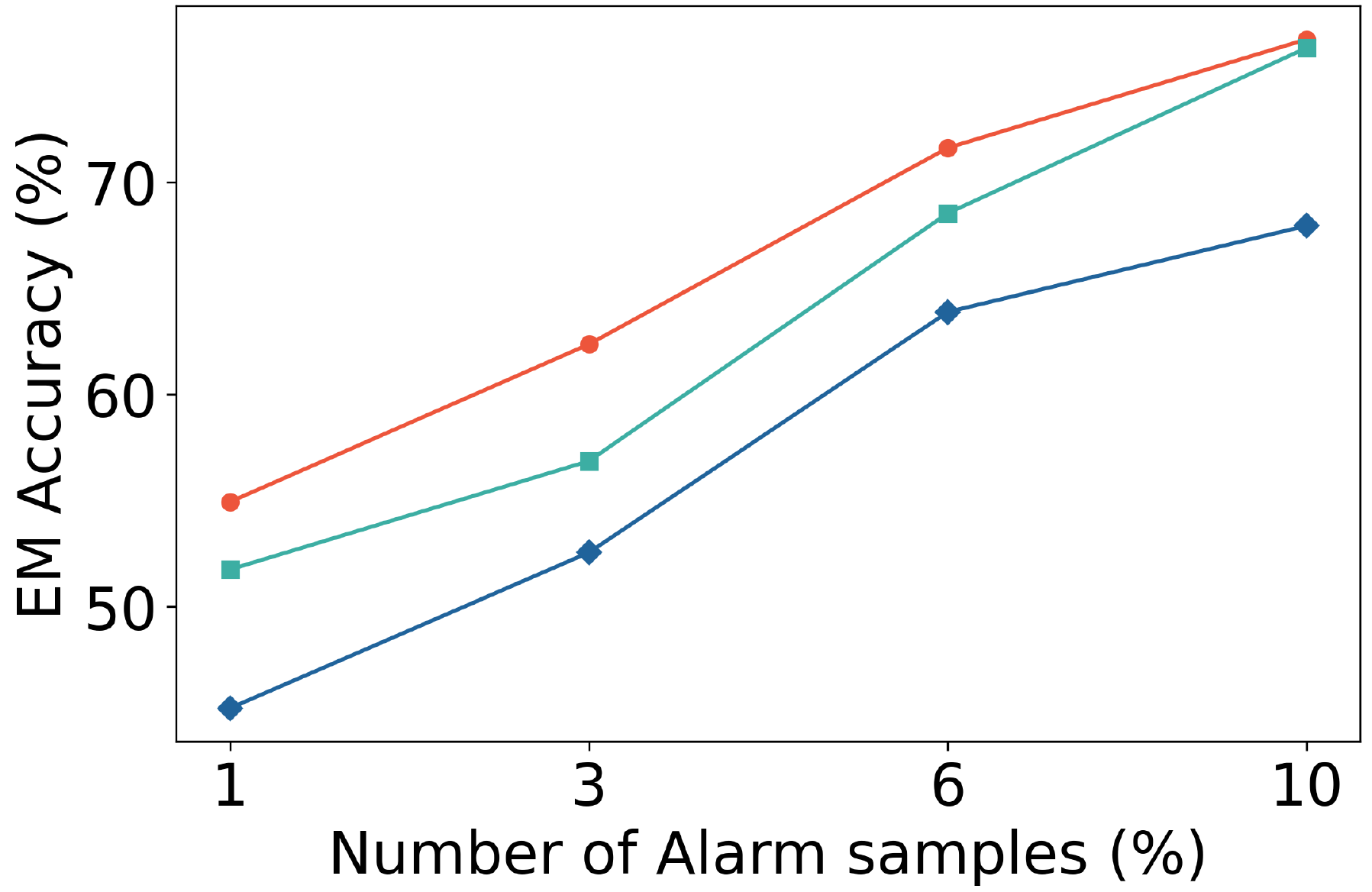}  
\end{minipage}
\begin{minipage}{.33\textwidth}
  \centering
  \includegraphics[width=\linewidth]{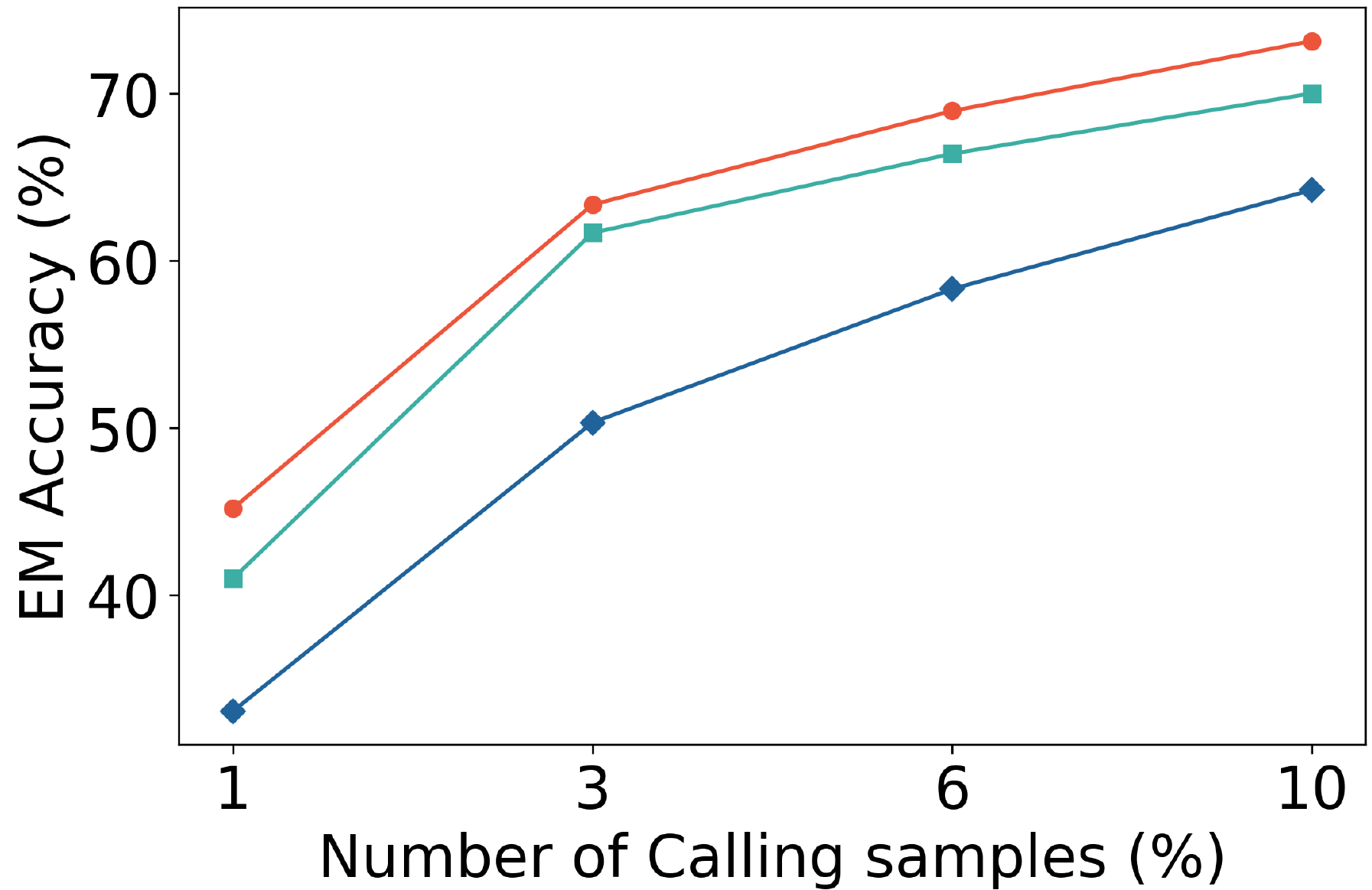}  
\end{minipage}
\begin{minipage}{.33\textwidth}
  \centering
  \includegraphics[width=\linewidth]{images/chapter-5/cross-domain/Event.pdf}
\end{minipage}
\begin{minipage}{.33\textwidth}
  \centering
  \includegraphics[width=1\linewidth]{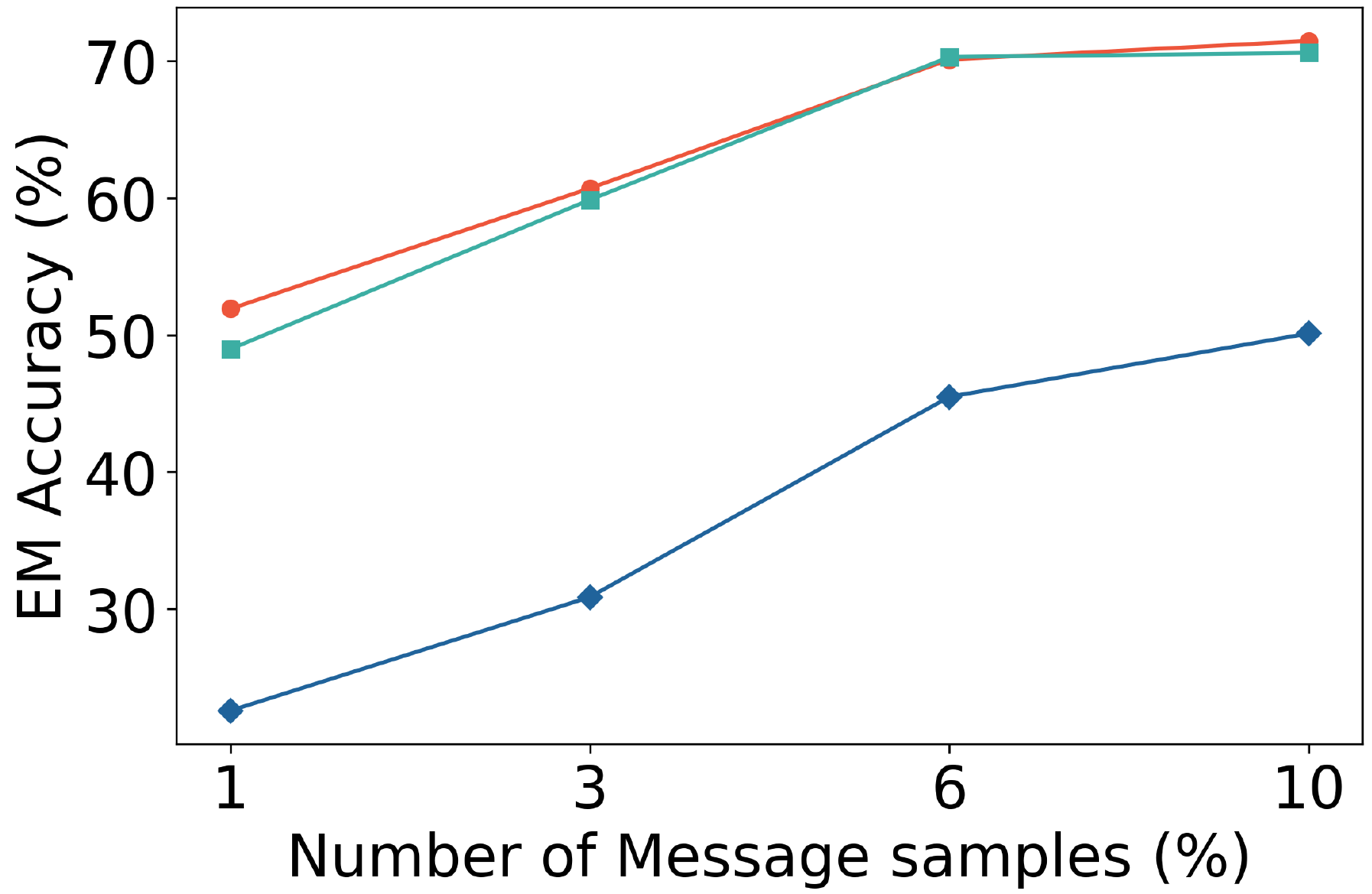}  
\end{minipage}
\begin{minipage}{.33\textwidth}
  \centering
  \includegraphics[width=\linewidth]{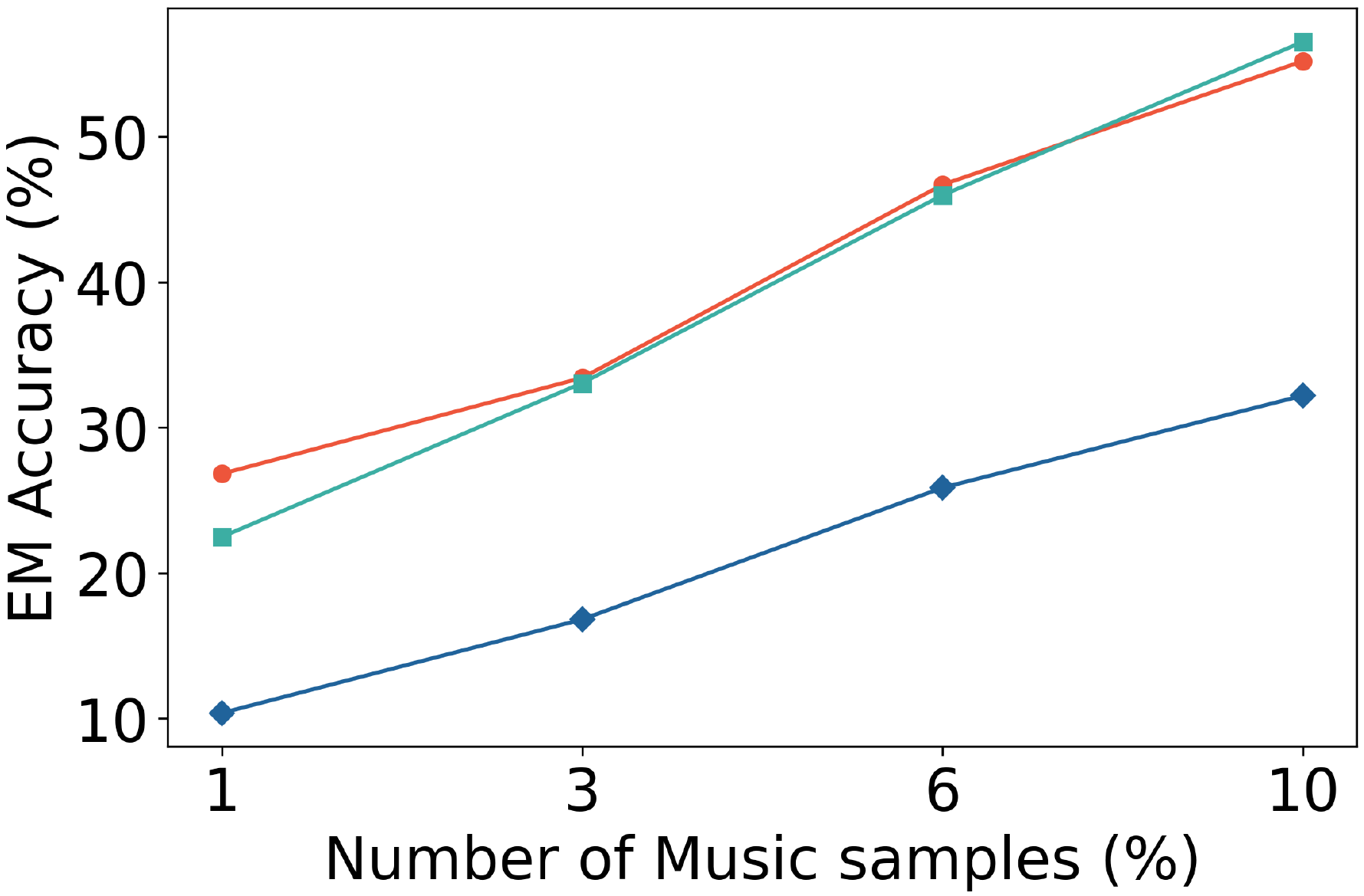}  
\end{minipage}
\begin{minipage}{.33\textwidth}
  \centering
  \includegraphics[width=\linewidth]{images/chapter-5/cross-domain/News.pdf}
\end{minipage}
\begin{minipage}{.33\textwidth}
  \centering
  \includegraphics[width=1\linewidth]{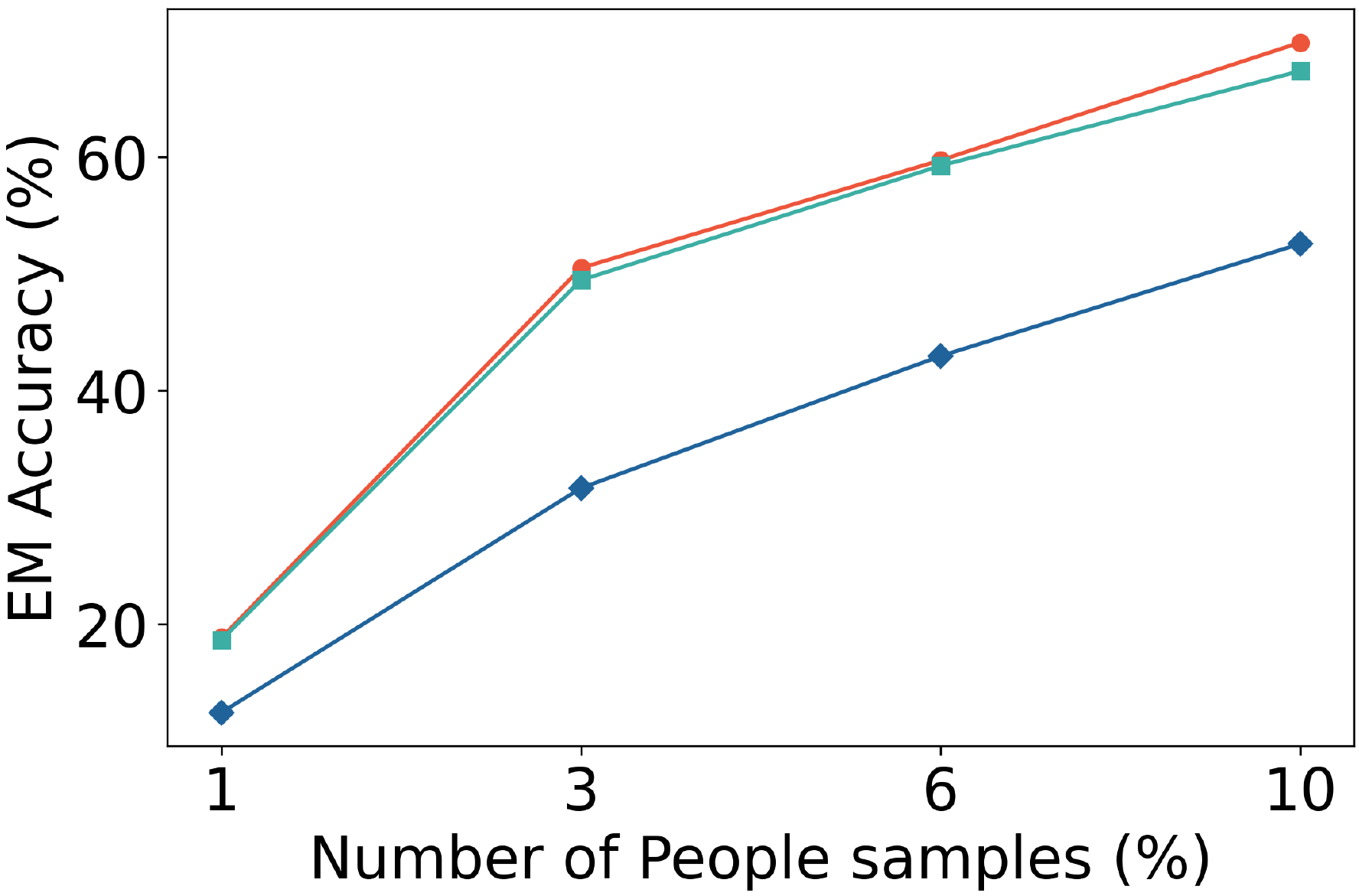}  
\end{minipage}
\begin{minipage}{.33\textwidth}
  \centering
  \includegraphics[width=\linewidth]{images/chapter-5/cross-domain/Recipe.pdf}  
\end{minipage}
\begin{minipage}{.33\textwidth}
  \centering
  \includegraphics[width=\linewidth]{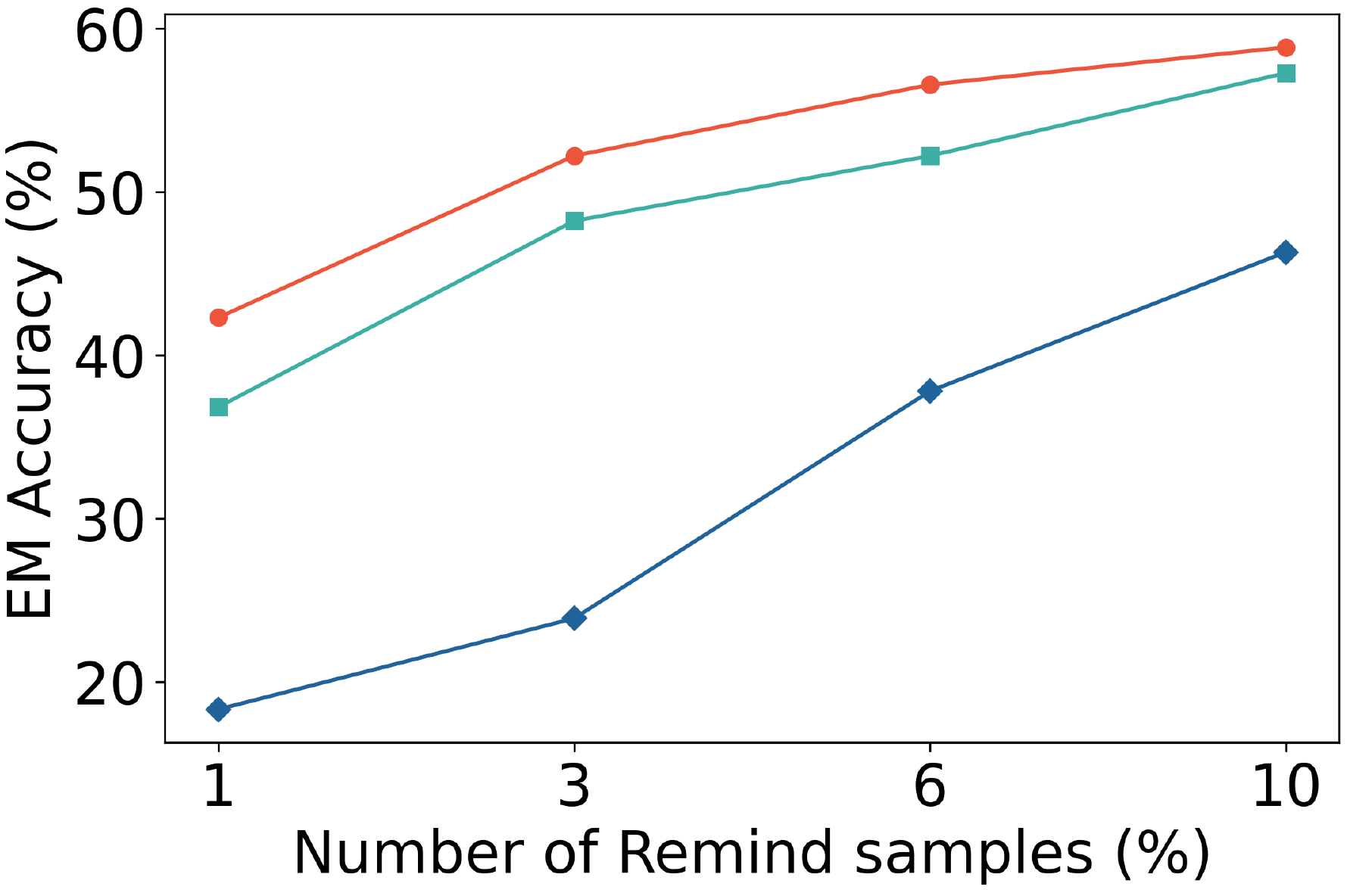}
\end{minipage}
\begin{minipage}{.33\textwidth}
  \centering
  \includegraphics[width=1\linewidth]{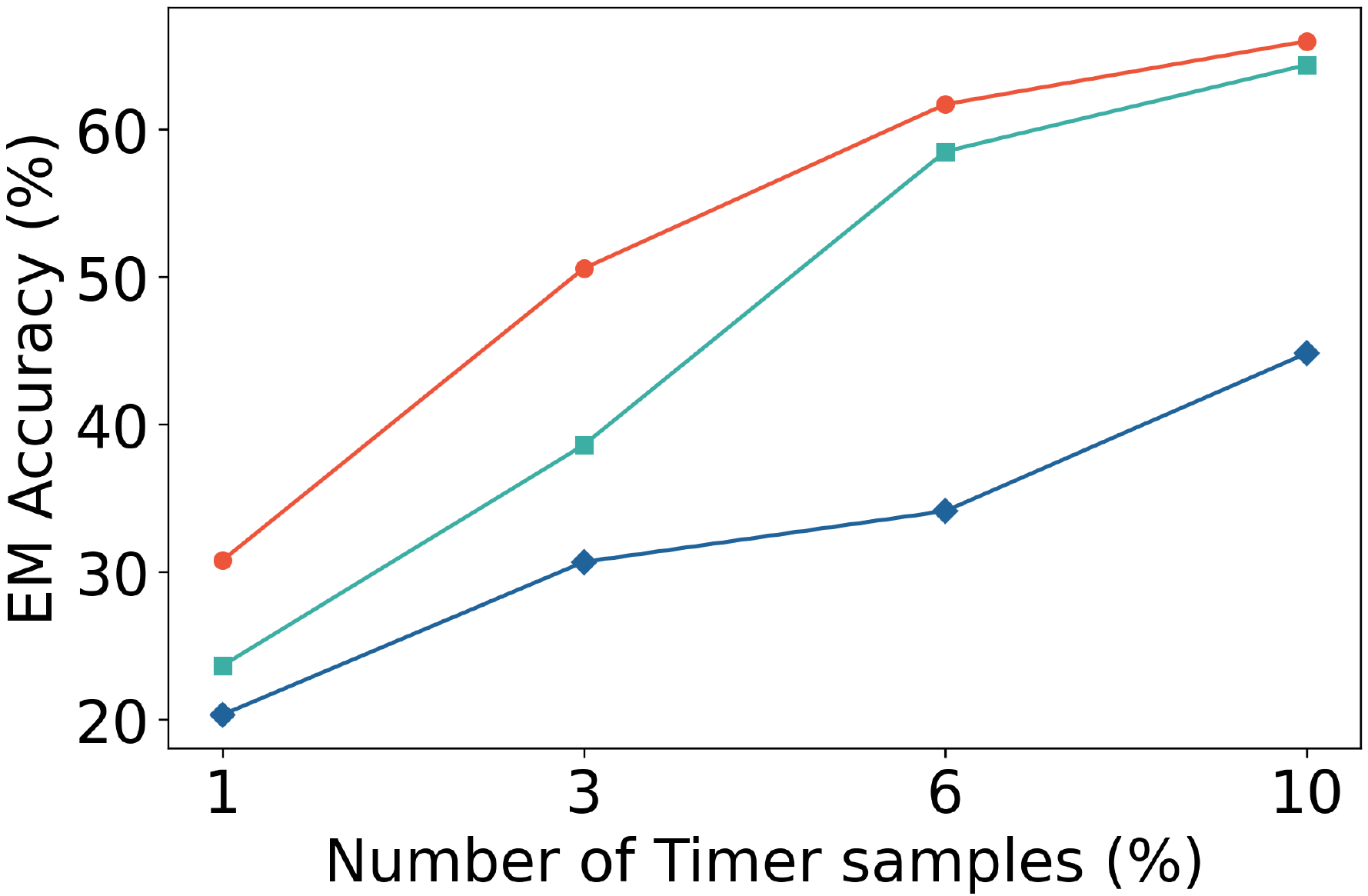}  
\end{minipage}
\begin{minipage}{.33\textwidth}
  \centering
  \includegraphics[width=\linewidth]{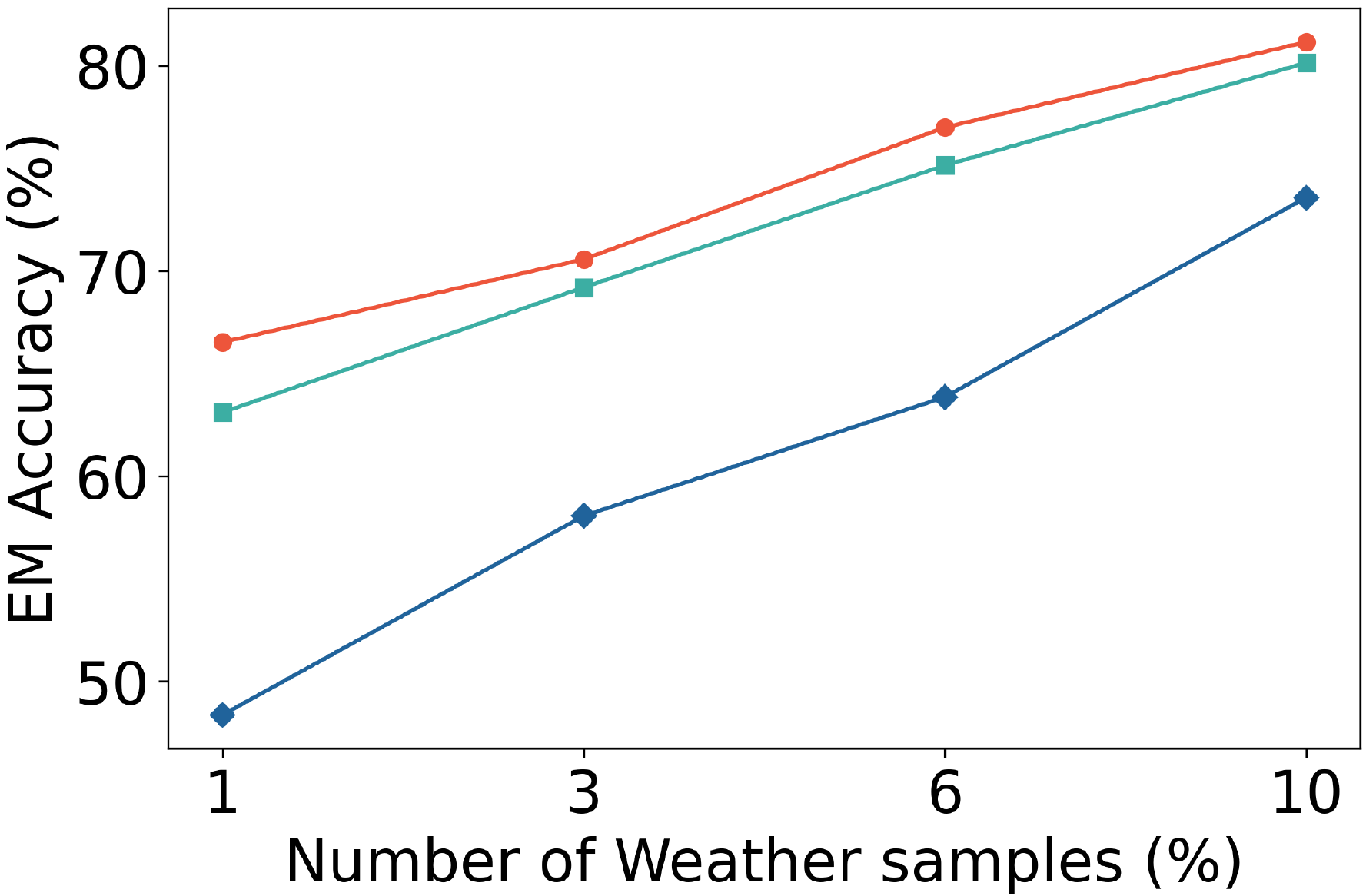}  
\end{minipage}
\caption{Few-shot exact match accuracies for the \textbf{cross-domain setting} across all 11 target domains.}
\label{fig:results-all-cross-domain}
\end{figure}

\begin{figure}[!ht]
\begin{minipage}{.33\textwidth}
  \centering
  \includegraphics[width=1\linewidth]{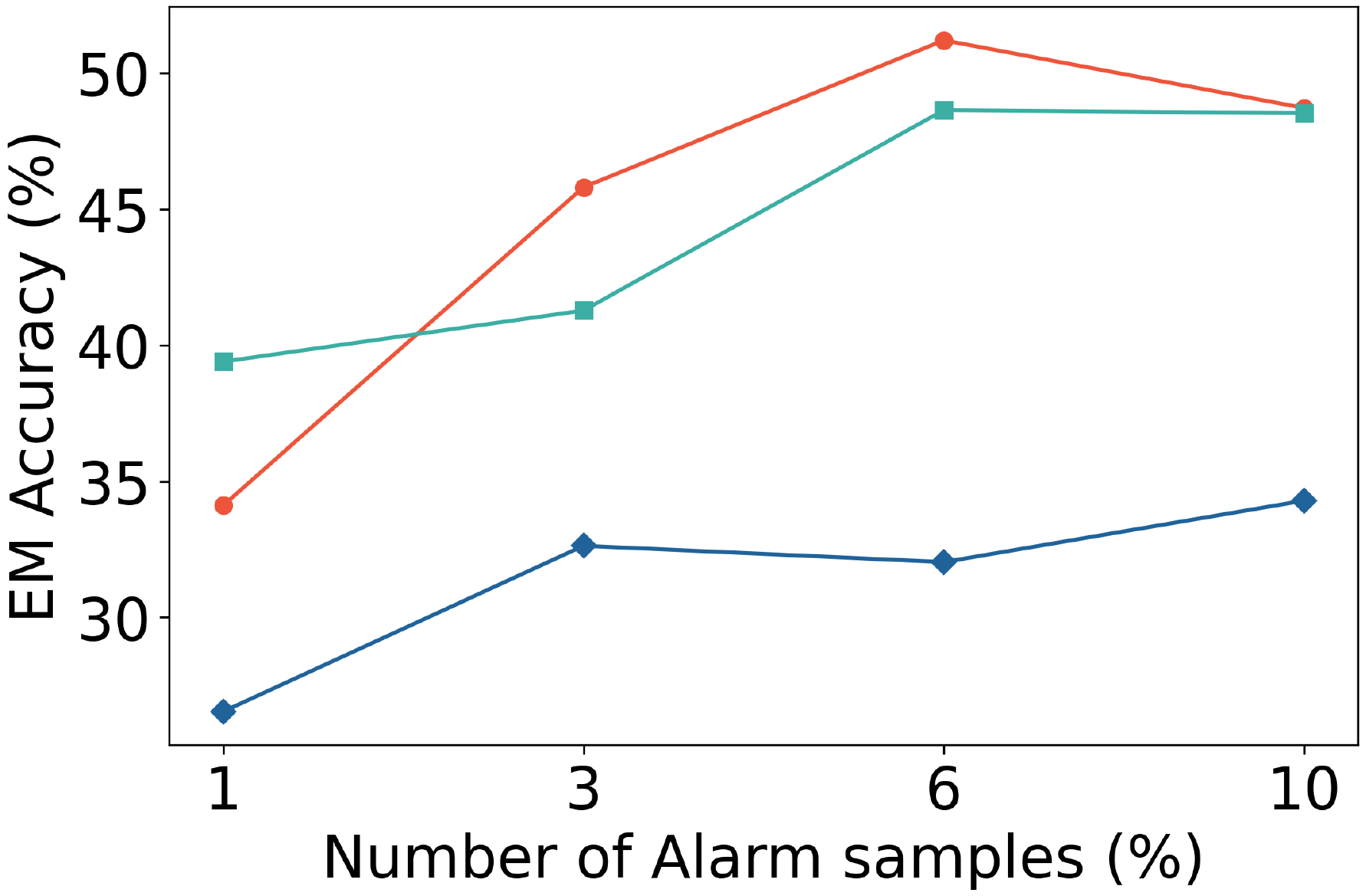}  
\end{minipage}
\begin{minipage}{.33\textwidth}
  \centering
  \includegraphics[width=\linewidth]{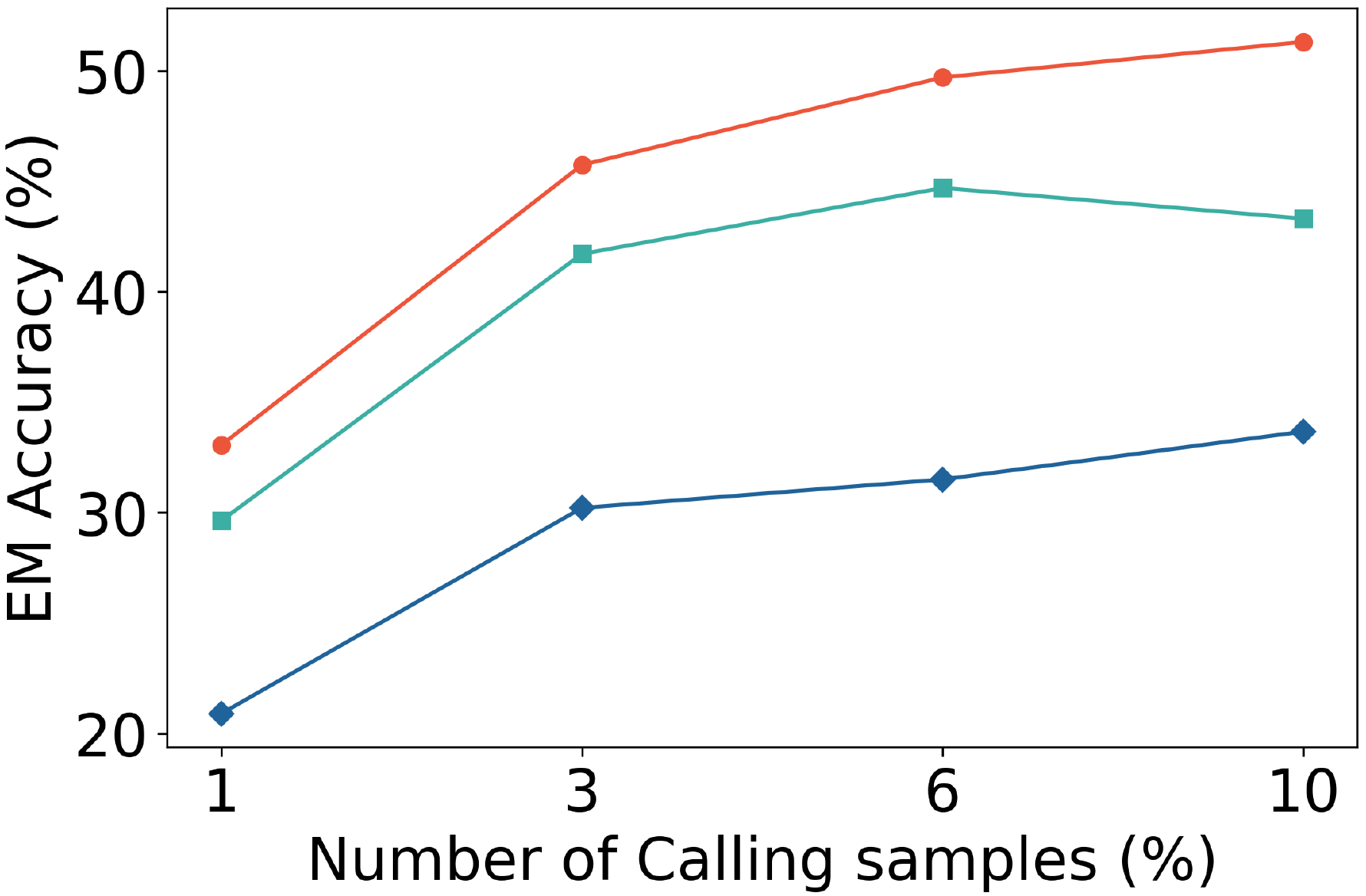}  
\end{minipage}
\begin{minipage}{.33\textwidth}
  \centering
  \includegraphics[width=\linewidth]{images/chapter-5/X2/Event.pdf}
\end{minipage}
\begin{minipage}{.33\textwidth}
  \centering
  \includegraphics[width=1\linewidth]{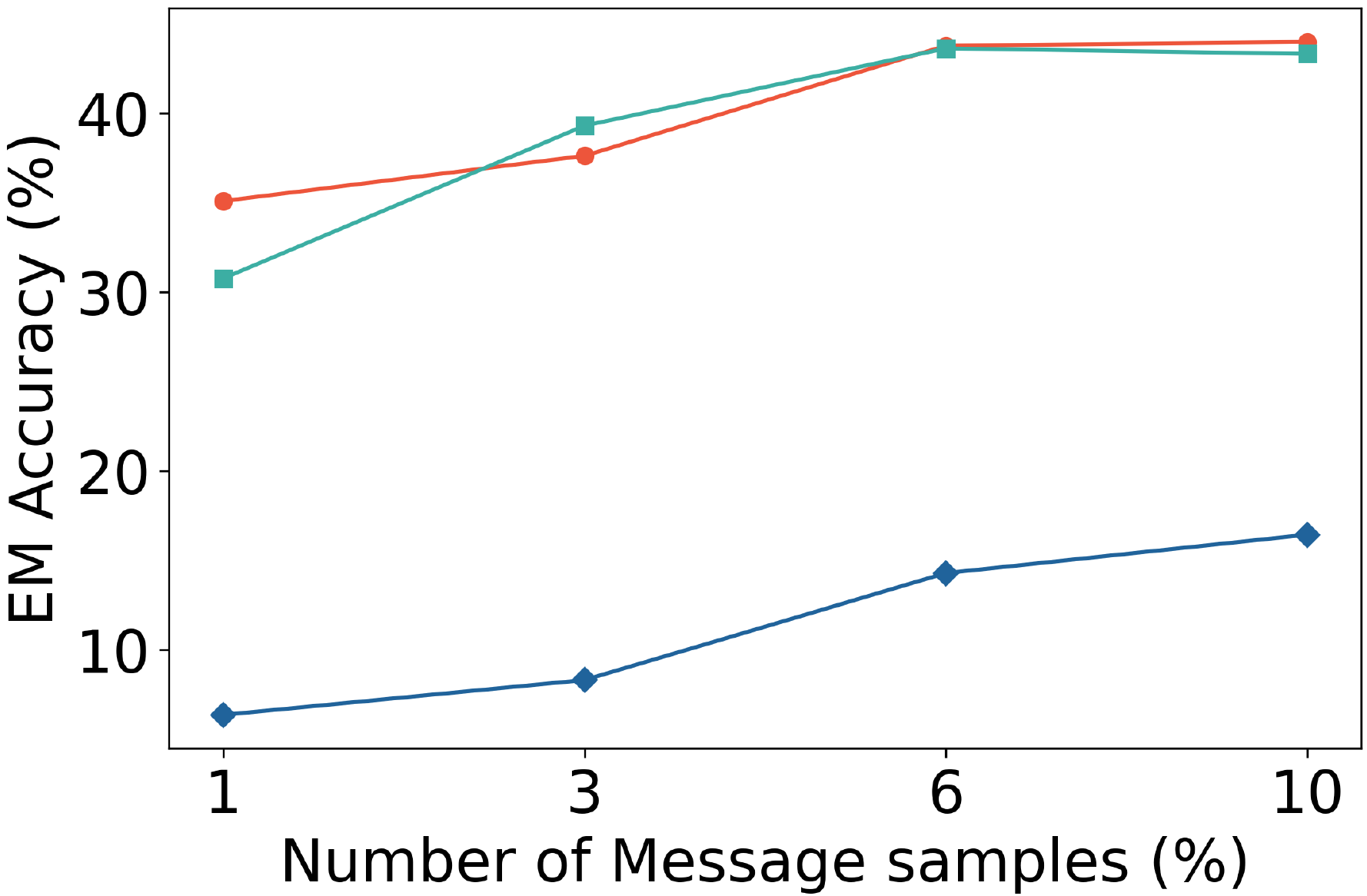}  
\end{minipage}
\begin{minipage}{.33\textwidth}
  \centering
  \includegraphics[width=\linewidth]{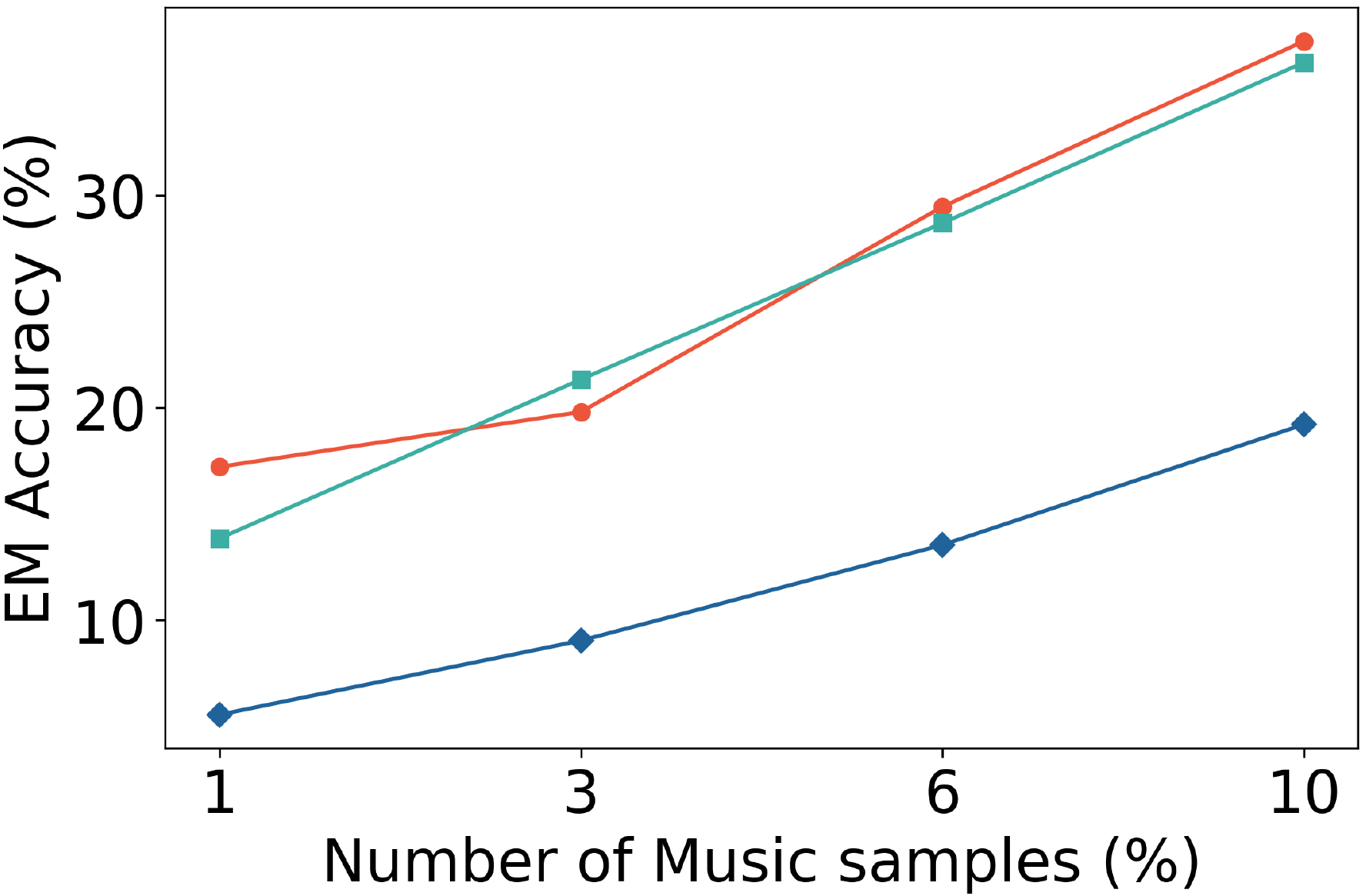}  
\end{minipage}
\begin{minipage}{.33\textwidth}
  \centering
  \includegraphics[width=\linewidth]{images/chapter-5/X2/News.pdf}
\end{minipage}
\begin{minipage}{.33\textwidth}
  \centering
  \includegraphics[width=1\linewidth]{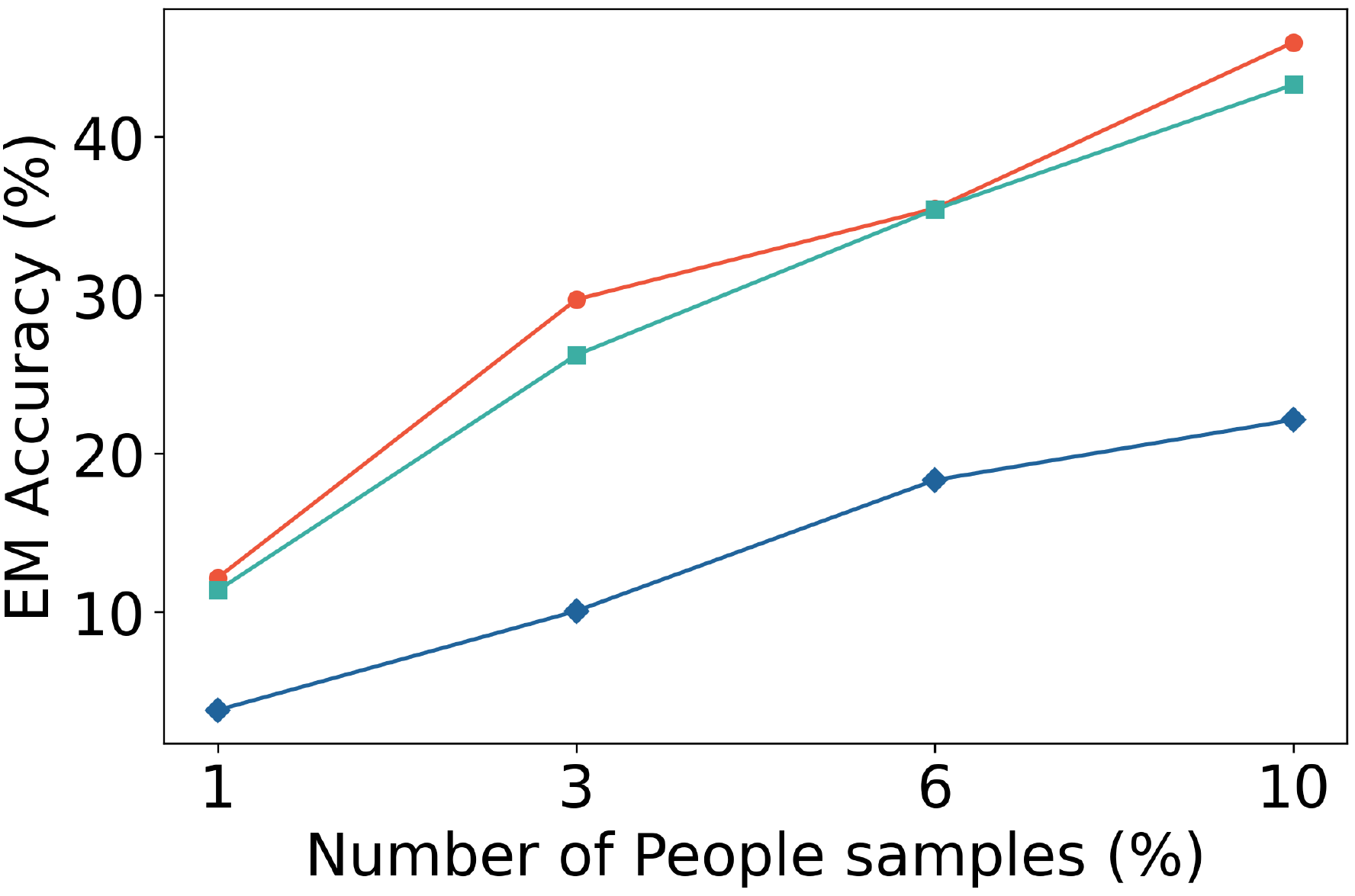}  
\end{minipage}
\begin{minipage}{.33\textwidth}
  \centering
  \includegraphics[width=\linewidth]{images/chapter-5/X2/Recipe.pdf}  
\end{minipage}
\begin{minipage}{.33\textwidth}
  \centering
  \includegraphics[width=\linewidth]{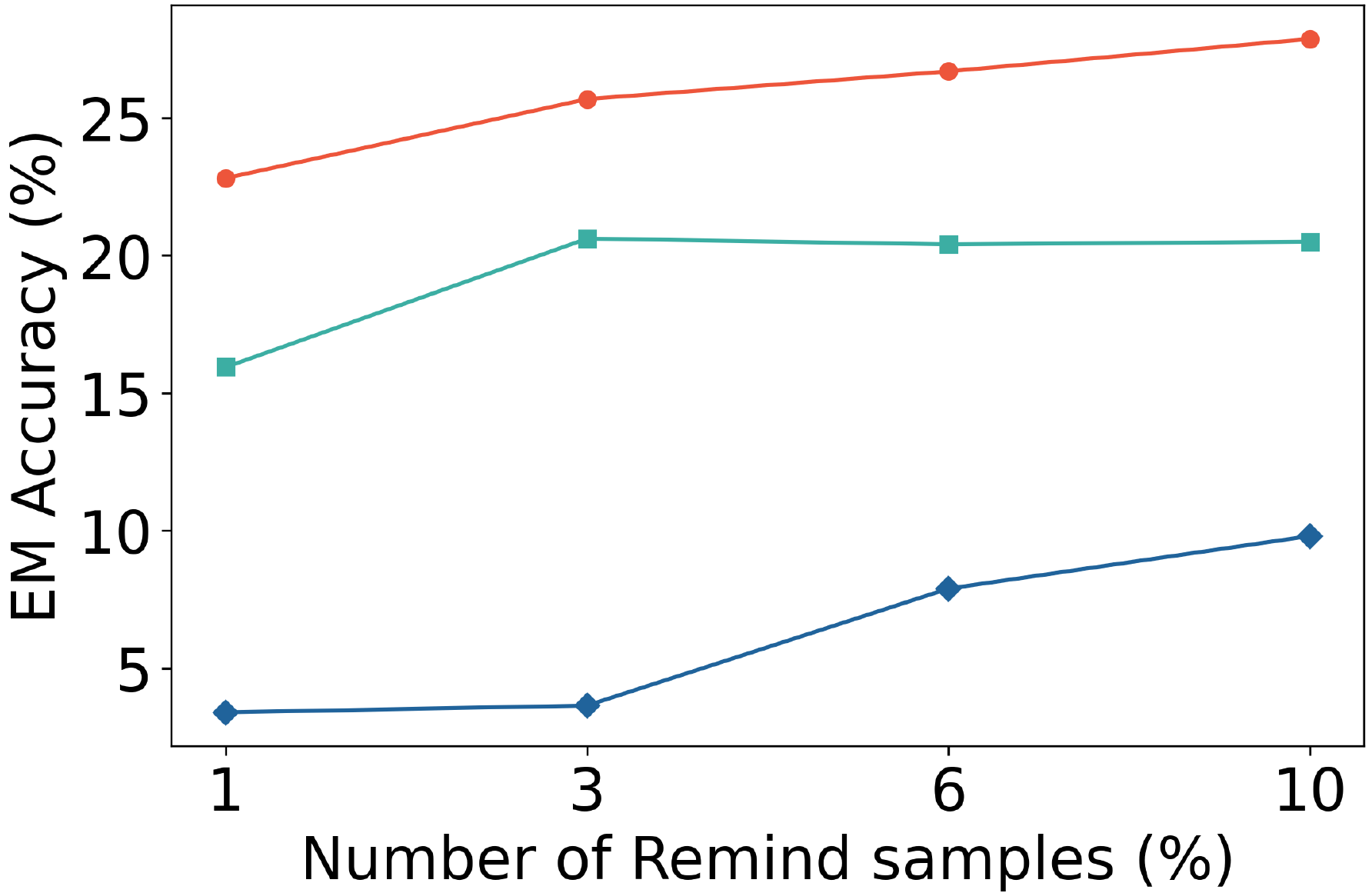}
\end{minipage}
\begin{minipage}{.33\textwidth}
  \centering
  \includegraphics[width=1\linewidth]{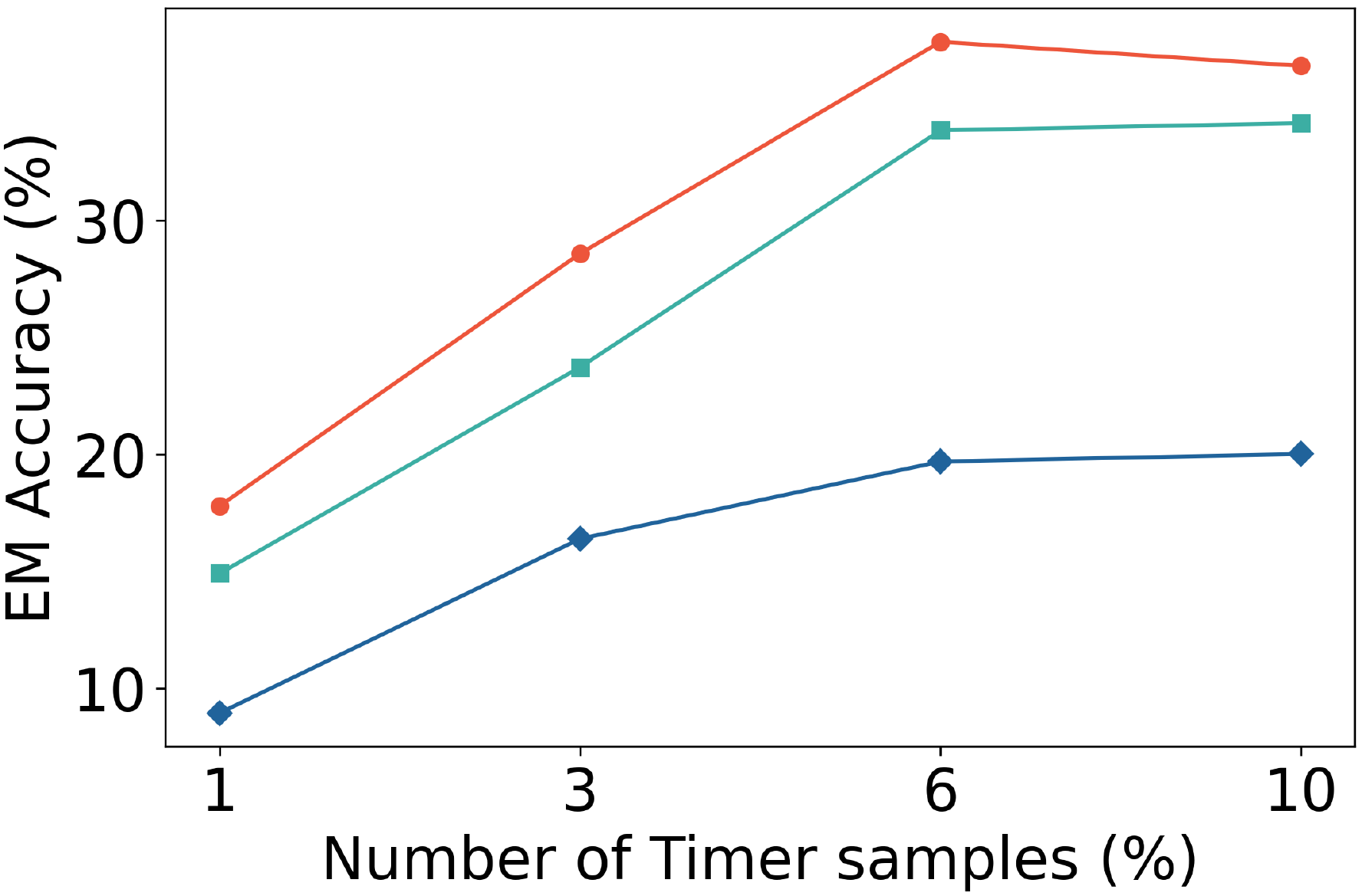}  
\end{minipage}
\begin{minipage}{.33\textwidth}
  \centering
  \includegraphics[width=\linewidth]{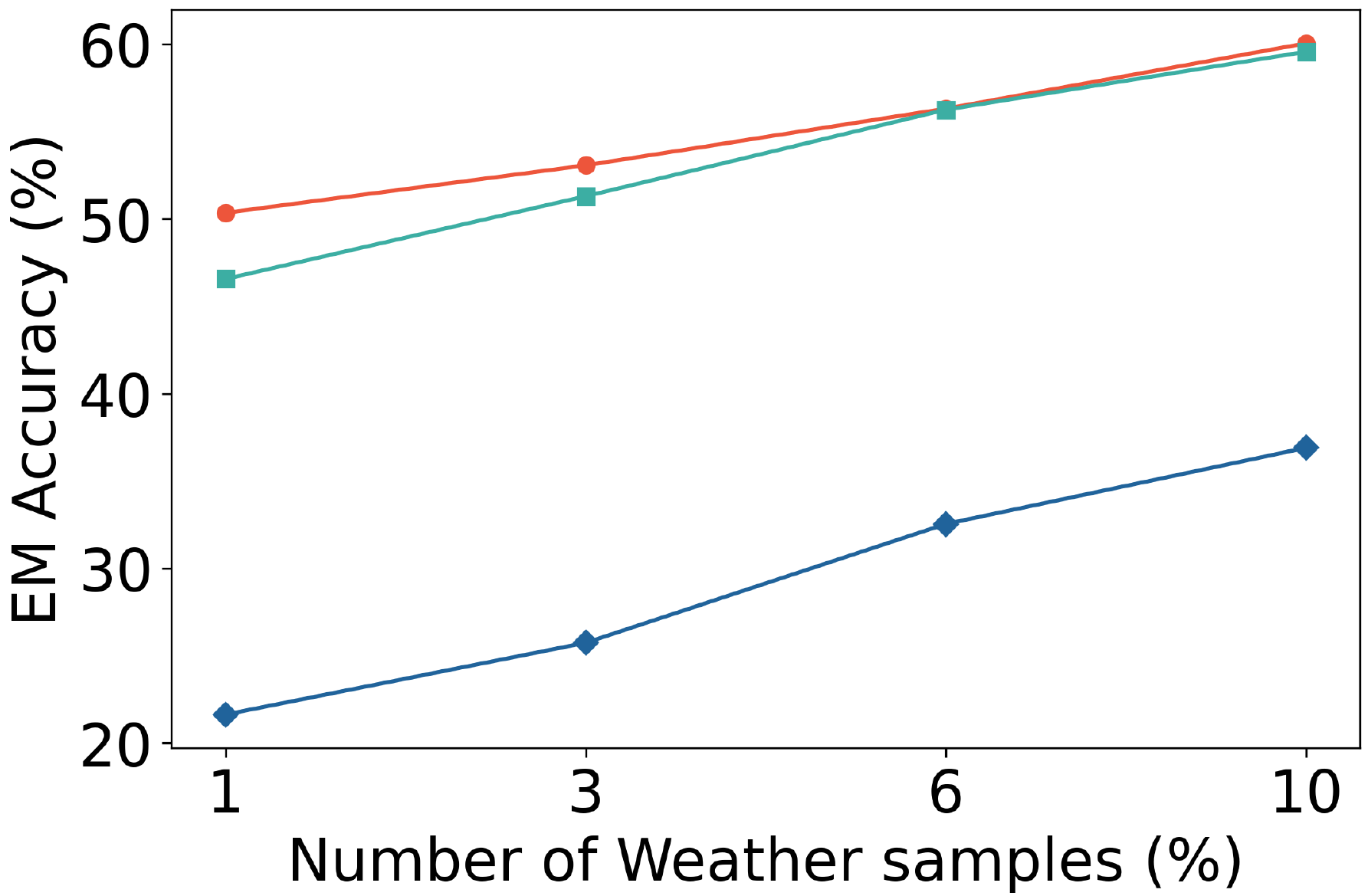}  
\end{minipage}
\caption{Few-shot Exact match accuracies for the \textbf{cross-lingual cross-domain setting} across all 11 target domains. The results are averaged over all target languages.}
\label{fig:results-all-x2}
\end{figure}

Full few-shot cross-domain results across all 11 target domains are shown in Figure~\ref{fig:results-all-cross-domain}. And full few-shot cross-lingual cross-domain results across all 11 target domains are shown in Figure~\ref{fig:results-all-x2}.

